# EXPLORATORY ARABIC OFFENSIVE LANGUAGE DATASET ANALYSIS

Fatemah Husain
f.husain@ku.edu.kw
Ozlem Uzuner
ouzuner@gmu.edu

## Abstract

This paper adding more insights towards resources and datasets used in Arabic offensive language research. The main goal of this paper is to guide researchers in Arabic offensive language in selecting appropriate datasets based on their content, and in creating new Arabic offensive language resources to support and complement the available ones.

## Introduction

Annotated offensive language datasets are used to categorize texts according to their offensive content automatically. As it is mentioned previously, some examples of offensive content are hate speech, obscene language, or vulgar language. The automated categorization process is called text classification, which depends heavily on the availability and the quality of the dataset used in building the classification model.

The offensive language datasets are a critical factor to the growth and success of the online offensive language detection systems. Multiple attributes effect the quality of datasets, such as the size, the annotation process, and the source. High quality datasets provide valuable data insights and support the classification model to learn effectively.

To pursue the goal of this paper, several available open-source datasets are surveyed from across the Arabic offensive language datasets to provide a comprehensive overview by conducting in-depth Exploratory Data Analysis (EDA). The EDA includes a statistical analysis, a textual analysis, and a contextual analysis for all datasets to investigate the content from multiple dimensions. Some visualization tools are used to better understand the content and context of the data used. The study ends-up with a summary of the results to synthesis the main findings.

The scope of this paper covers the following research questions:

- What are the content of the available Arabic offensive language datasets?
- What are the limitations of the available Arabic offensive language datasets?
- How can we complement the available Arabic offensive language datasets to contribute to text classification systems?

The paper is organized in four main sections. The methodology is described in detail in the first section. The second section presents the results and the third section builds on top of the second one by discussing and synthesizing the results. In the last section, conclusions and design considerations are presented.

**Methodology**

Four main phases are followed during the survey process. Starting by selecting datasets, formatting datasets, analyzing datasets, and ending by summarizing and synthesizing the results. The following paragraphs describe each phase of the methodology in detail.

1) Selecting Datasets:

A set of criteria are defined to select the datasets: searching, formatting, and accessibility. These criteria ensure the quality of the study.

a. Defining Searching Criteria

Datasets related to offensive language are included, such as hate speech, vulgar, or abusive. Only Arabic language datasets are considered, including dialectic Arabic.

b. Defining Formatting Criteria

Datasets from multiple formats were included. Most datasets are in Comma-Separated Values (CSV) file format, few of them are in Excel, Tab-Separated Values (TSV), and JavaScript Object Notation (JSON).

c. Defining Accessibility Criteria

Datasets that have been released freely online with open-source option are considered only.

2) Formatting Datasets:

The selected datasets are in heterogeneous formats and some of them include multiple descriptive attributes, such as publishing date, user profile, or number of annotators. Thus, we process them to be in a minimal and consistent format.

a. Filtering Attributes

We remove all unnecessary attributes that do not serve the goal of the study. Only textual messages and labels were included. The content of textual messages was intentionally kept without cleaning because all content is considered for analysis purposes, however, some datasets were provided in preprocessed format only.

b. Creating CSV Files

For each dataset, we create a CSV file to save the textual messages and labels only. This file is used for cross labels analysis and for overall dataset analysis.

c. Creating Textual Files

For each label within the datasets, we create a text file that contains only the textual content. This file is used for textual analysis and contextual analysis purposes.

3) Analyzing Datasets:

This is the most important phase of the study. The analysis phase adds value and insight about the content of the datasets. We present detailed investigations for the content of each dataset by conducting statistical, textual, and contextual analysis, in addition to generating multiple graphs to visualize the content.

a. Statistical analysis

The statistical analysis includes finding frequencies of words, frequencies of stop words, statistical measurements for the lengths of the text based on the number of tokens, and statistical measurements for the lengths of the tokens based on the number of characters to analyze their relationships with offensive content. To extract the most frequently used words for each class accurately, we remove a list of stop words from the text. The stop words list includes the NLTK Arabic stop words list, and Albadi, Kurdi, and Mishra (2018)'s stop words list. Then, we search for the words that have the prefix 'ال' to remove the prefix. We do not remove the prefix 'ال' when it is used as a part of the word and not as a prefix, such as in the word "الله". Simple count of token frequencies is useful to compare among multiple classes; however, it does not provide rich information about each class separately. We use the web-based tool Voyant[1] to further analyze the text and identify the top five most distinctive words of each class. Stop words could help in defining the context of the posts. We conduct simple frequency analysis to generate the top stop words per class, as stop words that appear only in a particular class might be better to consider in analysis as a regular word rather than as a stop word. We investigate the complexity of the text used in each class to check if there is any pattern or relationship between the complexity of the text used and the type of the offensive content. We use two measures to peruse the goal of this analysis; the number of characters per token and the number of tokens per post.

[1] https://voyant-tools.org/

b. Textual analysis

Before conducting any cleaning or filtering techniques to the data, we generate word cloud graphs for each label from each dataset using the textual files to give some intuition about the raw content of each class. Data in all datasets are extracted from user-generated content platforms that is usually written in unstructured format and using dialectic Arabic, which is not supported by most of the available textual analysis tools. Thus, we were unable to perform POS Tagging to analyze the text based on their functional roles, and investigate whether that could influence the offensive content.

c. Contextual analysis

We study the impact of context to offensive content. Context is defined in terms of text sentiment, the use of emojis, and the use of punctuations. To better understand the context of the samples, We use the Mazajak online tool[2] for Arabic sentiment analysis to predict the sentiments of tweets. Thus, each sample is classified to positive, negative, or neutral depends on its content. Emoji is often used in online communication to reflect emotion and express personality, thus, considering emojis adds value to understanding text. Punctuations provide clue for the meaning of unfamiliar phrases and context of the sentence. As a result of that we analyze the use of punctuations and their effects on offensive content.

[2] mazajak.inf.ed.ac.uk

4) Summarizing and Synthesizing Results:

After reviewing the analysis section, we connect results across the datasets and summarize the overall findings. We add more insight into the findings by synthesizing the result with findings from previous studies, and provide valuable design considerations for other researchers in the same domain of research.

**Datasets Analysis Results:**

This section contains the results from dataset analysis in chronological order based on the publication date of each dataset. A total of nine datasets satisfy the selection criteria as the following: Aljazeera.net Deleted Comments (Mubarak, Darwish, and Magdy, 2017), Egyptian Tweets (Mubarak, Darwish, and Magdy, 2017), YouTube Comments (Alakrot, Murray, and Nikolov, 2018), Religious Hate Speech (Albadi, Kurdi, and Mishra, 2018), Levantine Hate Speech and Abusive Language (Mulki et al., 2019), Tunisian Hate Speech and Abusive Language (Haddad, Mulki, and Oueslati, 2019), Multi-Platform Offensive Language Dataset (Chowdhury et al., 2020), the Fourth Workshop on Open-Source Arabic Corpora and Corpora Processing Tools (Mubarak et al., 2020), and the Multi-Platform Hate Speech Dataset (Omar, Mahmoud, & Abd El-Hafeez, 2020).

1) The Aljazeera.net Deleted Comments Dataset:

The Aljazeera.net deleted comments datasets is developed by Mubarak, Darwish, and Magdy (2017). It includes a total of 31,692 comments. Three classes are used to label the comments as the following: 5,653 clean comments, 533 obscene comments, and 25,506 offensive comments. Figure 1 shows classes distribution.

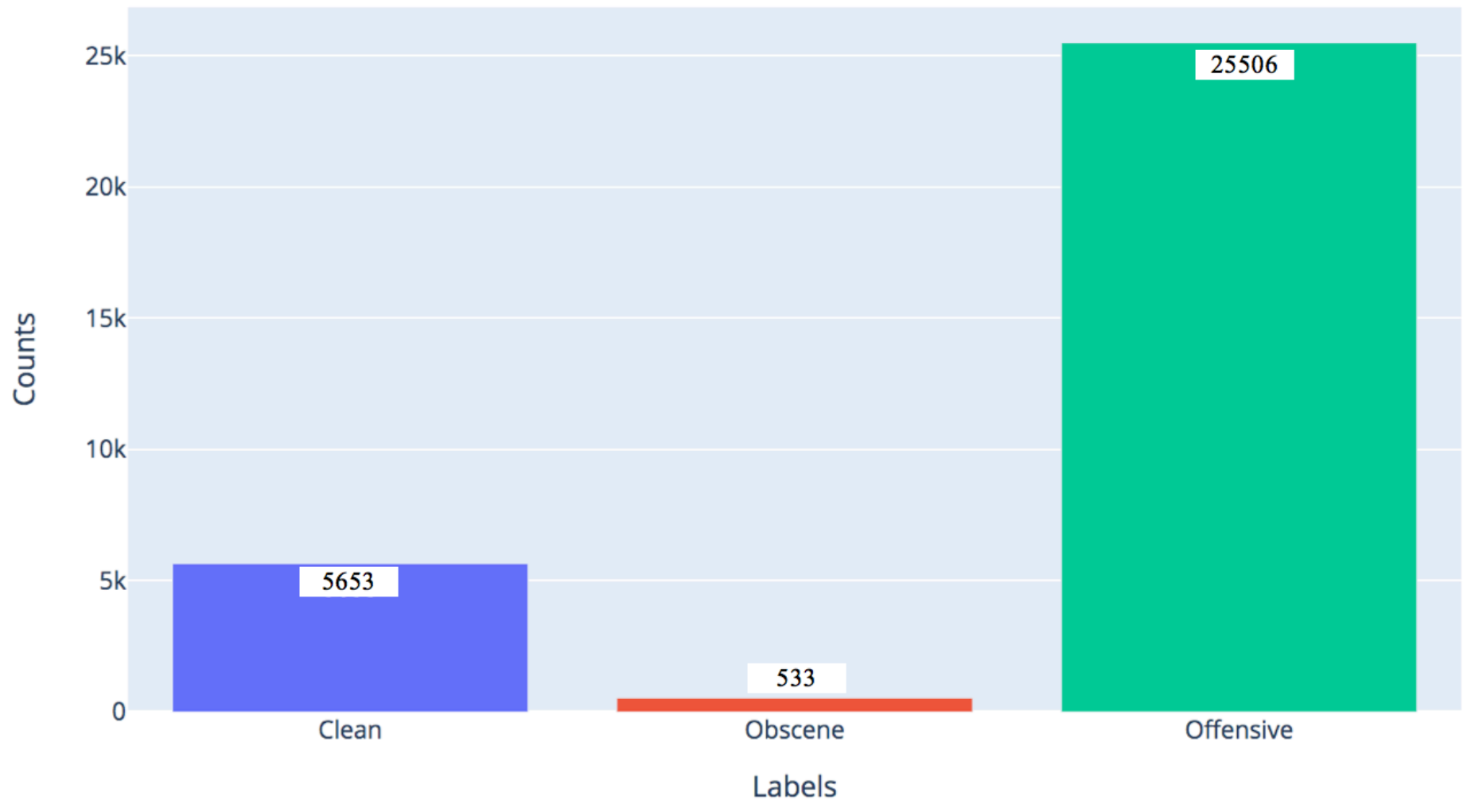

**Figure 1 Class distribution for the Aljazeera dataset**

The total number of duplicate comments is 8; 2 clean comments, 1 obscene comments, and 5 offensive comments. The following is an example from the duplicated offensive comments:

أناديكم أناديكم أشد على أياديكم وأبوس الأرض تحت نعالكم .. تجار الدين من الروفض والايرانين يجب تطهيرهم كما فعل بهم صلاح الدين . بعدها سوف تتحرر الاراضي العربيه من الاحتلال الفارسي والاسرائيلي

Translation: I am calling you, I am calling you and hold your hands and kiss the land beneath your shoes.. The land need to be cleaned from the Iranian and Shia as Salah Al-Deen did before, after that the Arabic land will get free from the Persian and Israeli invasion.

Investigating text through the word cloud from Figure 2, it can be noticed that the most common particles differ among the three classes. For example, in clean comments, “هو / he” and “في /in” are the most frequent ones, in obscene comments, “يا /you” and “لا /no” are the most used ones, and in offensive comments, “كان /was” and “من / from” are more likely to be seen than other particles. From the word cloud figure, some distinguishable words among the classes are “فيصل / Faisal” from clean comments, “كس / pussy” from obscene comments, and “سيف / sword” from offensive comments.

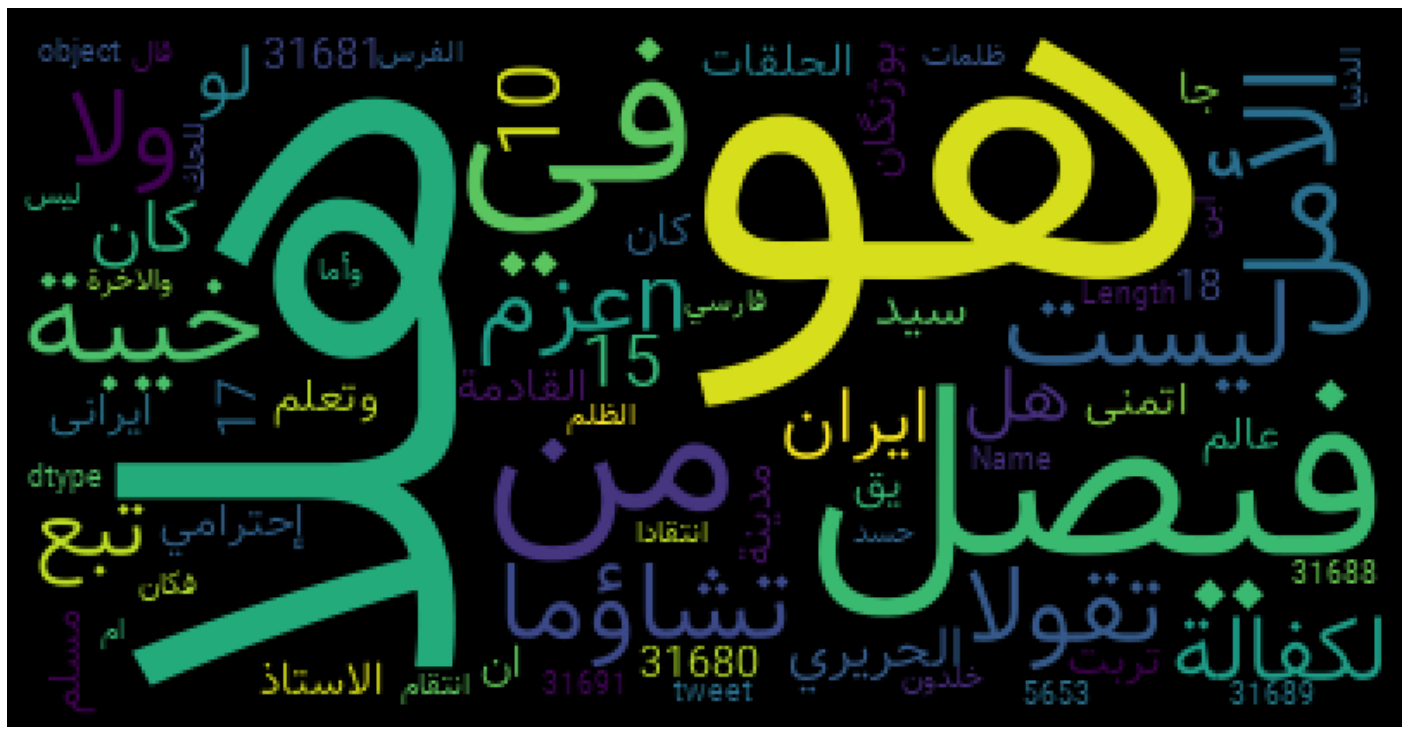

(a) Clean

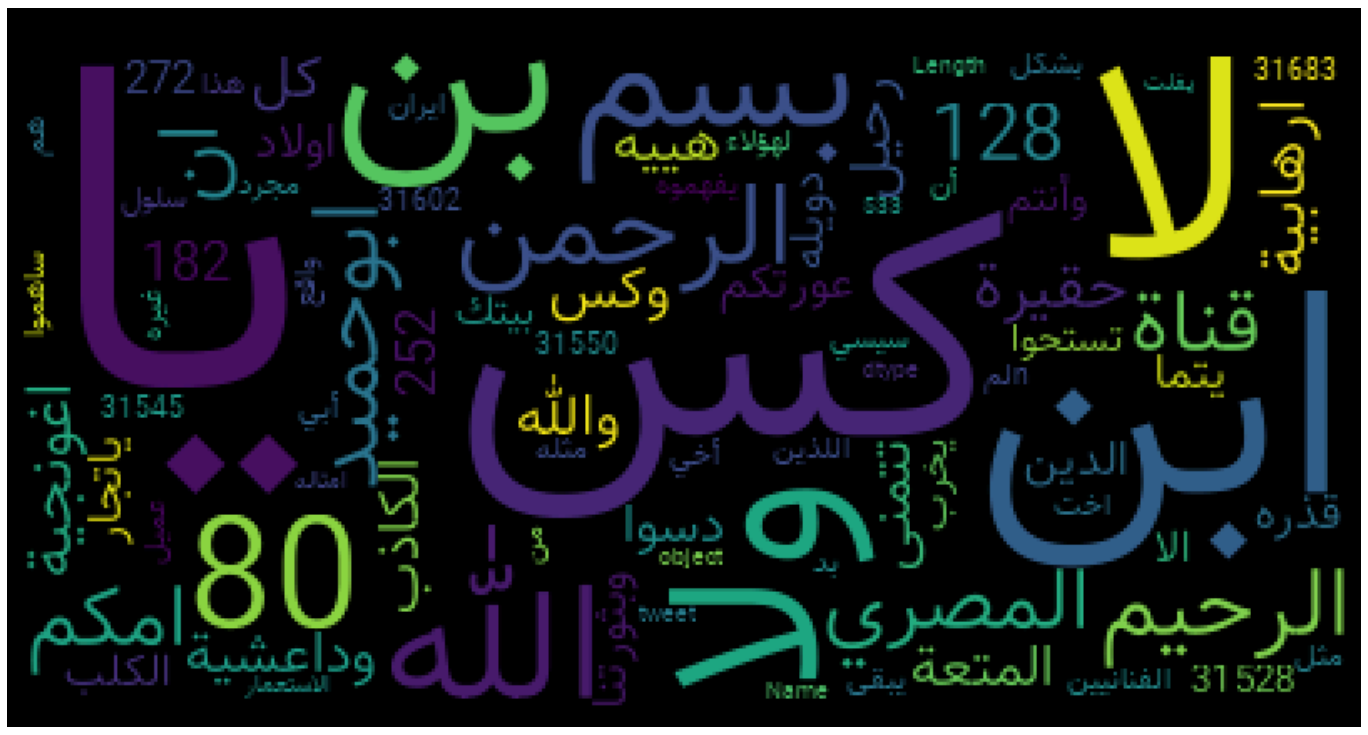

(b) Obscene

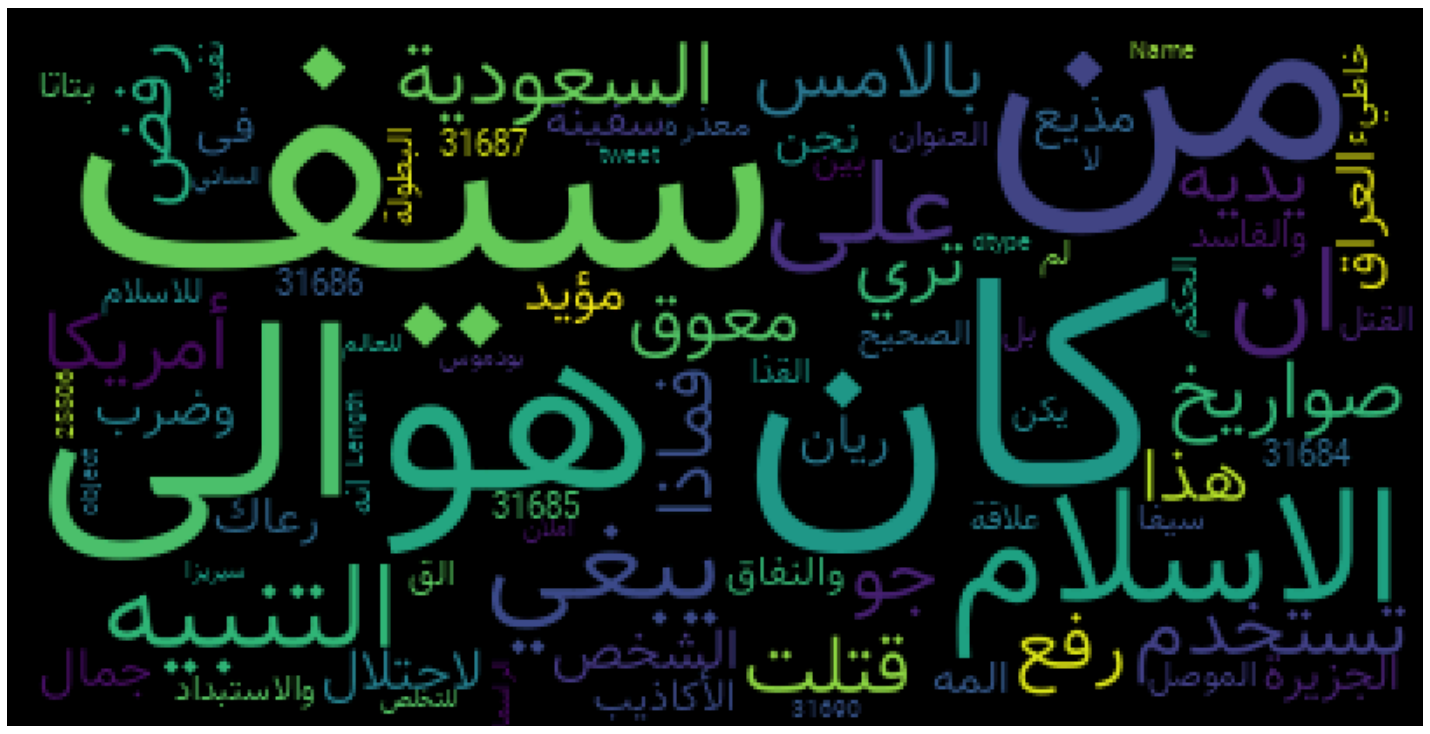

(d) Offensive

**Figure 2 The word cloud of the Aljazeera dataset (a. clean, b. obscene, c. offensive)**

In Figures 3 to 5, the top frequent 10 tokens are shown for each class separately. In all classes, the words “الله / God” is the top frequent one. For both clean and offensive comments, the second top frequent word is “دوله / state” while for obscene comments, the second top word is “ابن / son”. The third top frequent word in clean comments is “جزيره / Jazeera”, “كس / pussy” in obscene comments, and “مصر / Egypt” in offensive comments.

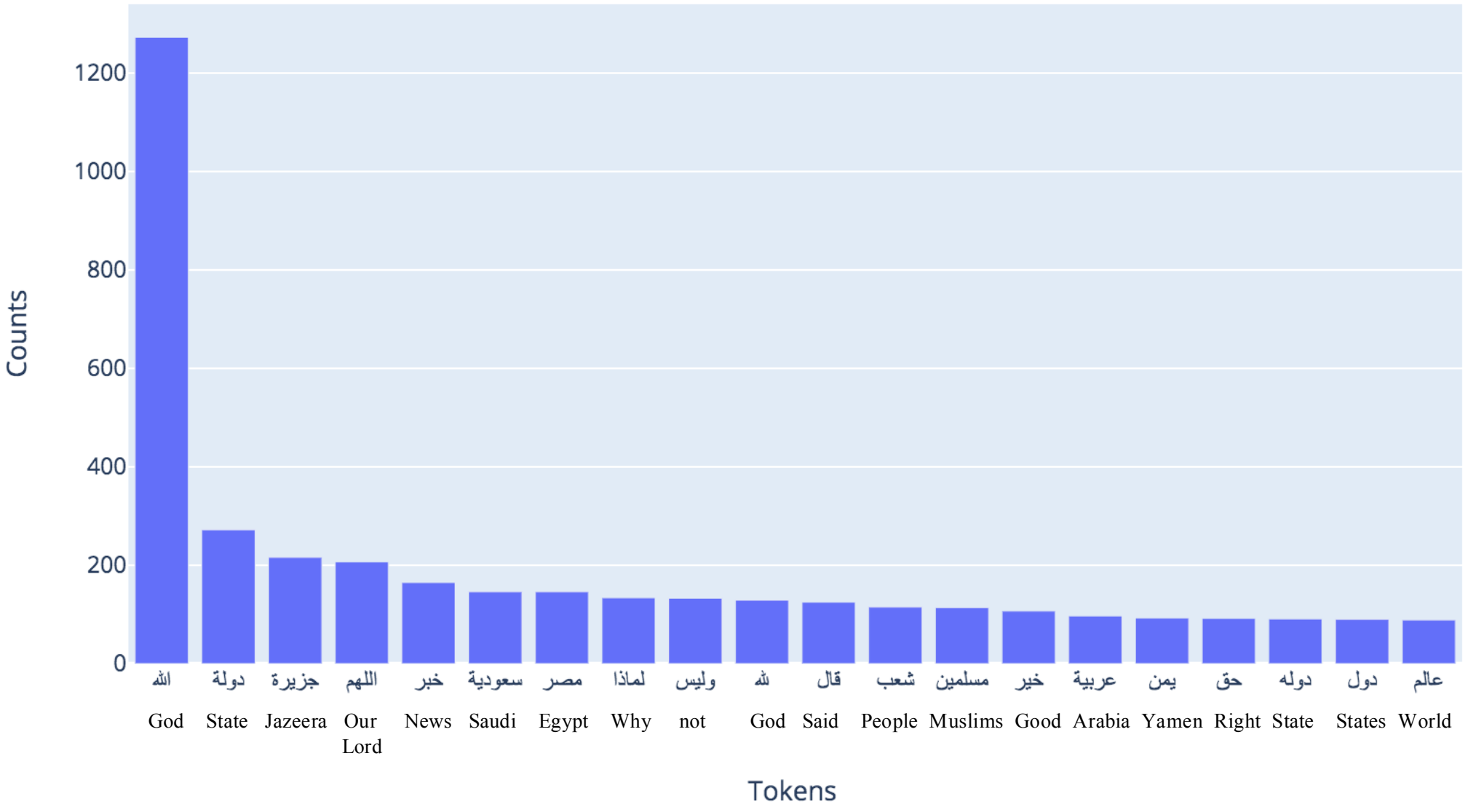

**Figure 3 Most common clean tokens in the Aljazeera dataset**

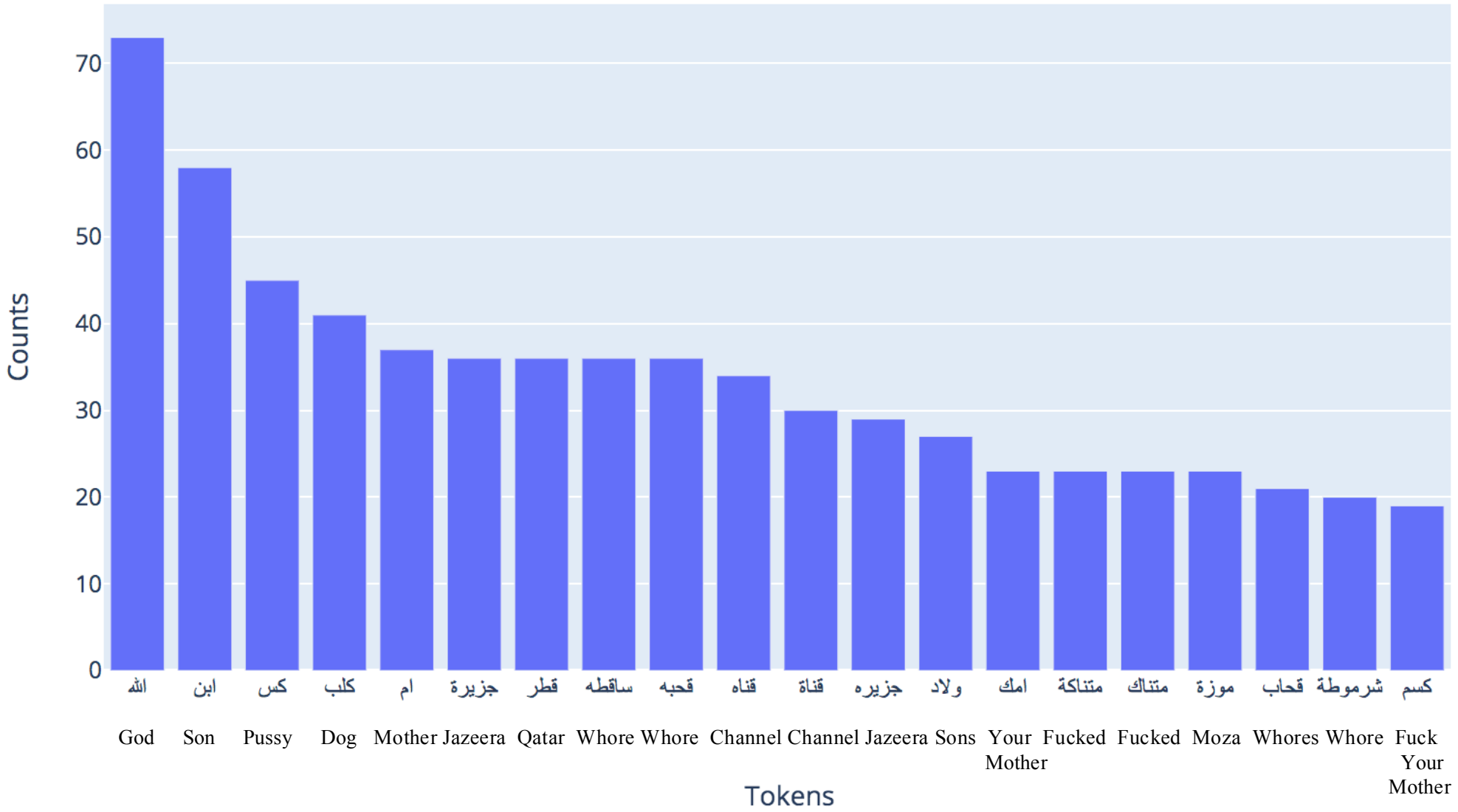

**Figure 4 Most common obscene tokens in the Aljazeera dataset**

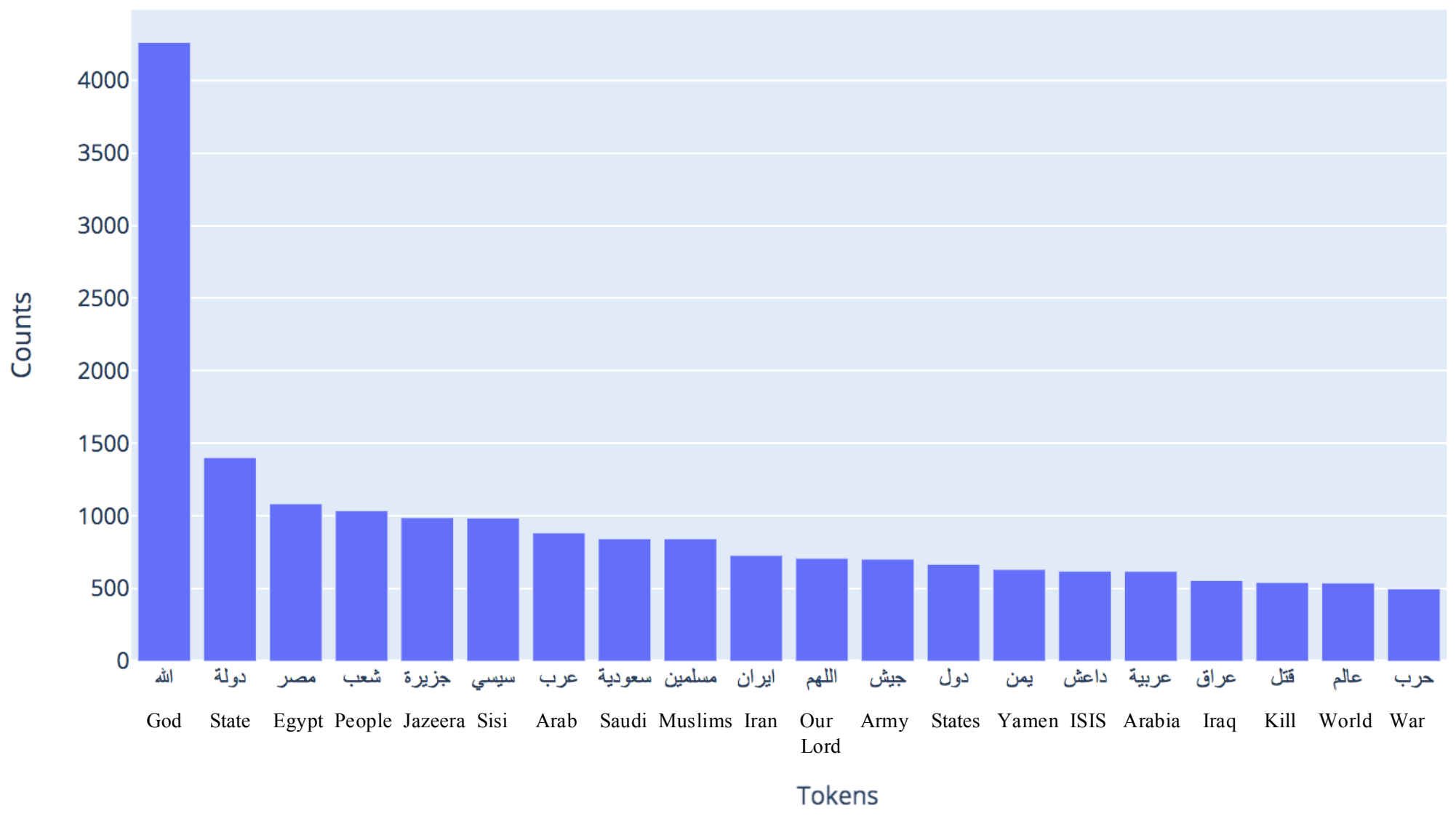

**Figure 5 Most common offensive tokens in the Aljazeera dataset**

Applying the web-based tool Voyant shows the following five distinctive words for each class:

1. Clean: الخبر / the news (130), قال / said (128), ووفقكم / bless you (38), عافاكم / make you healthy (36), خير / good (92).
2. Obscene: المتناك / fucked (20), كسم / fuck your mother (19), كس / pussy (44), المتناكة / fucked (16), القحبة (13).
3. Offensive: قال / said (445), الاسد / Al-Asad (275), السوري / Syrian (271), تركيا Turkey (244), فلسطين / Palestine (227).

Figures 6 and 7 compare the results of the statistical analysis for the length of comments and the length of tokens based on the classes. On all cases, offensive comments are the longest followed by obscene comments and clean comments are at the end. Clean Comments are the only category that has some outliers for comments length.

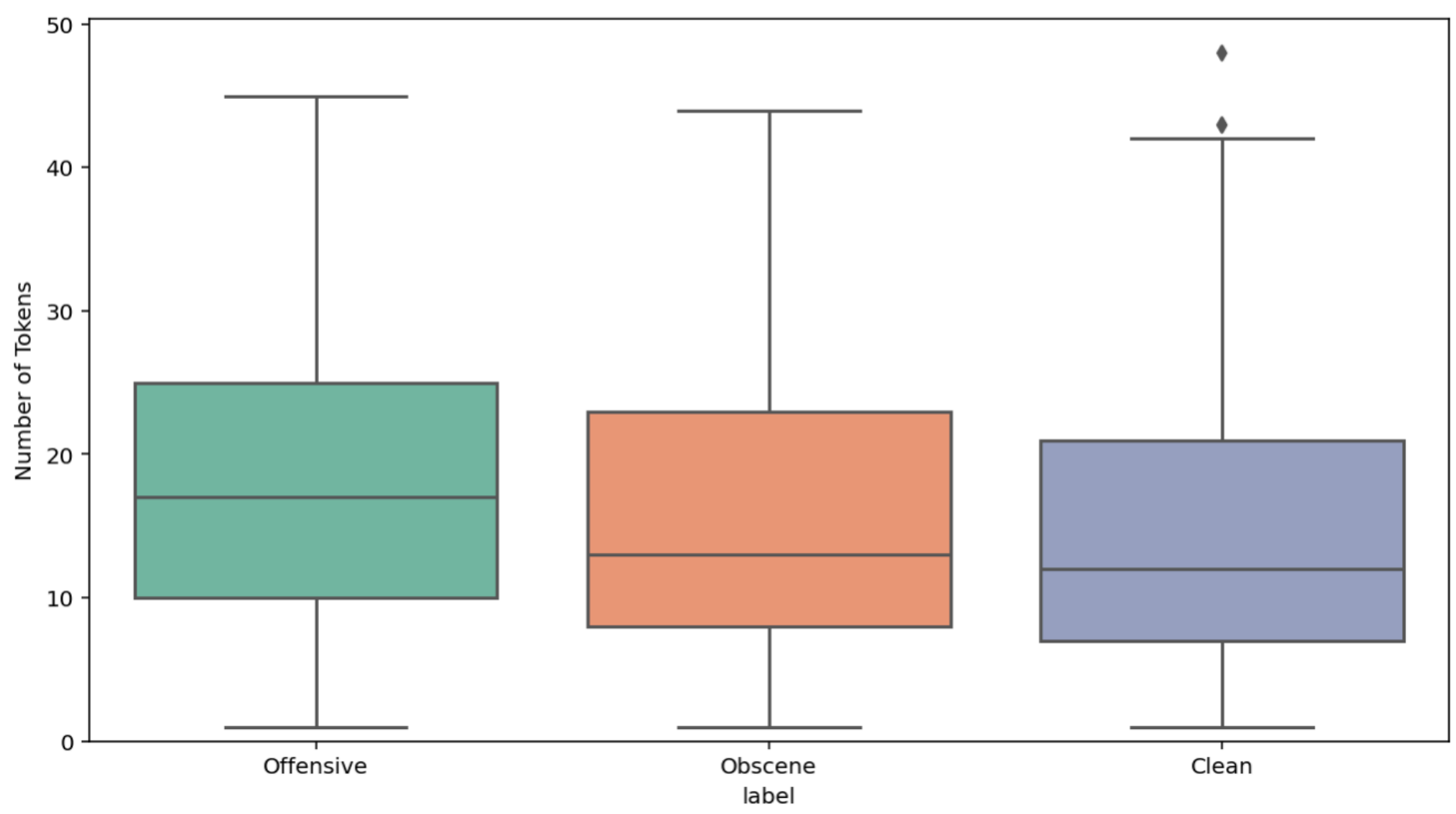

**Figure 6 Statistics of each label in the Aljazeera dataset based on the number of tokens per comment**

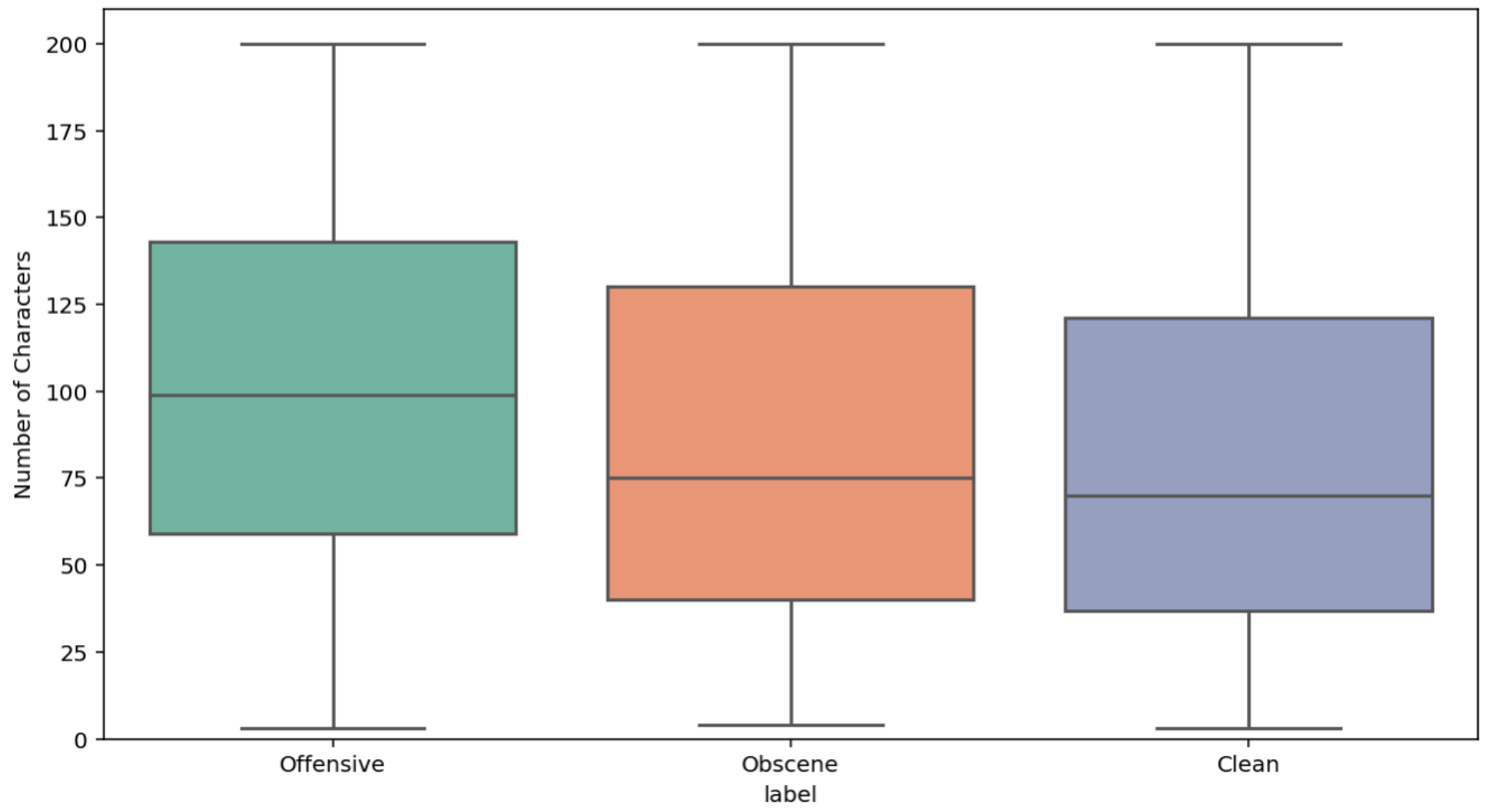

**Figure 7 Statistics of each label in the Aljazeera dataset based on the number of characters per token**

Investigating the use of stop words among the classes from Figure 8 to 10, shows a very similar pattern among all classes. For example, "من / from" is the top one among all, followed by "و / and" in both clean and offensive classes and "يا / you" in obscene class, and "في / in" is the third top frequent stop word among all classes.

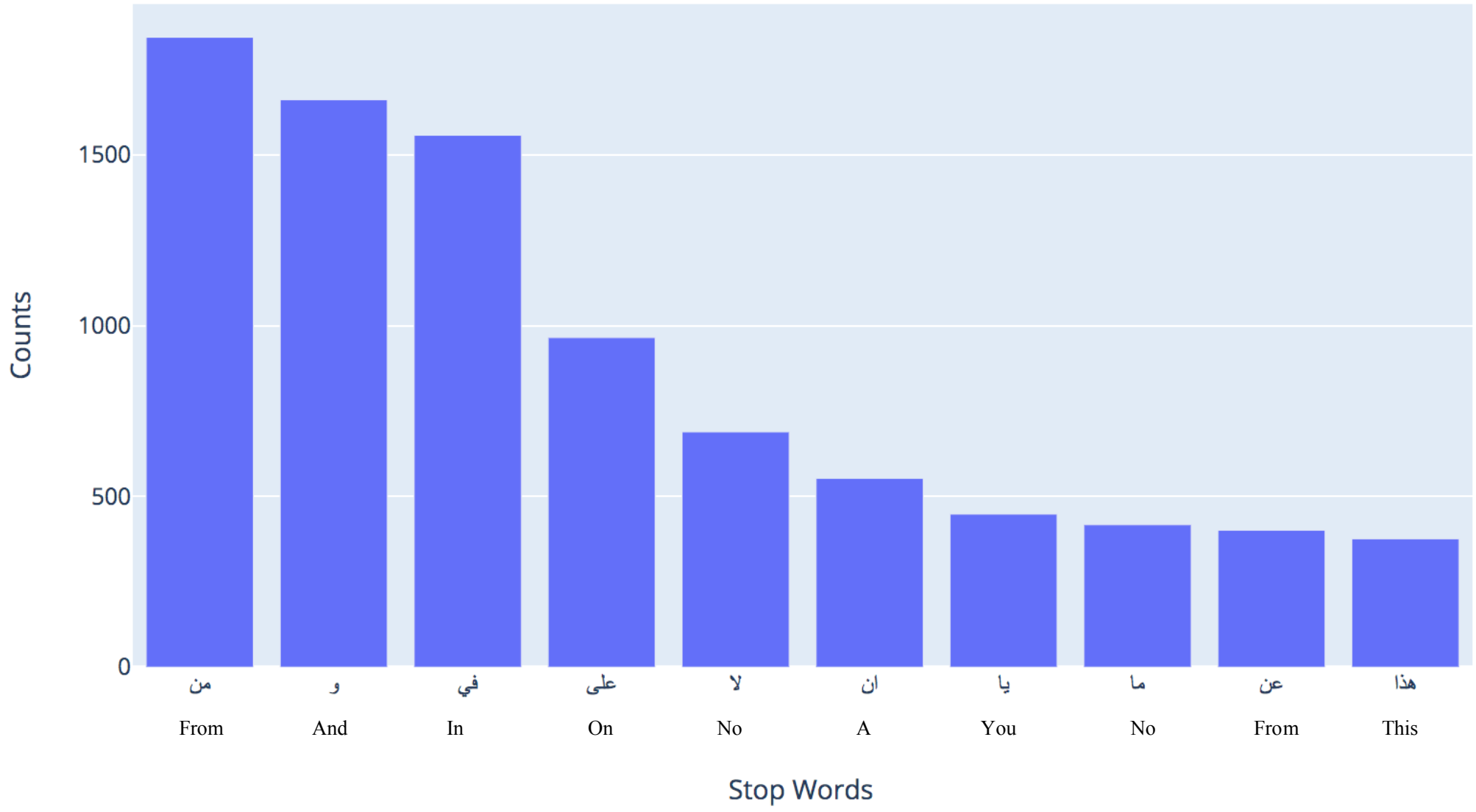

**Figure 8 Most common stop words in clean class from the Aljazeera dataset**

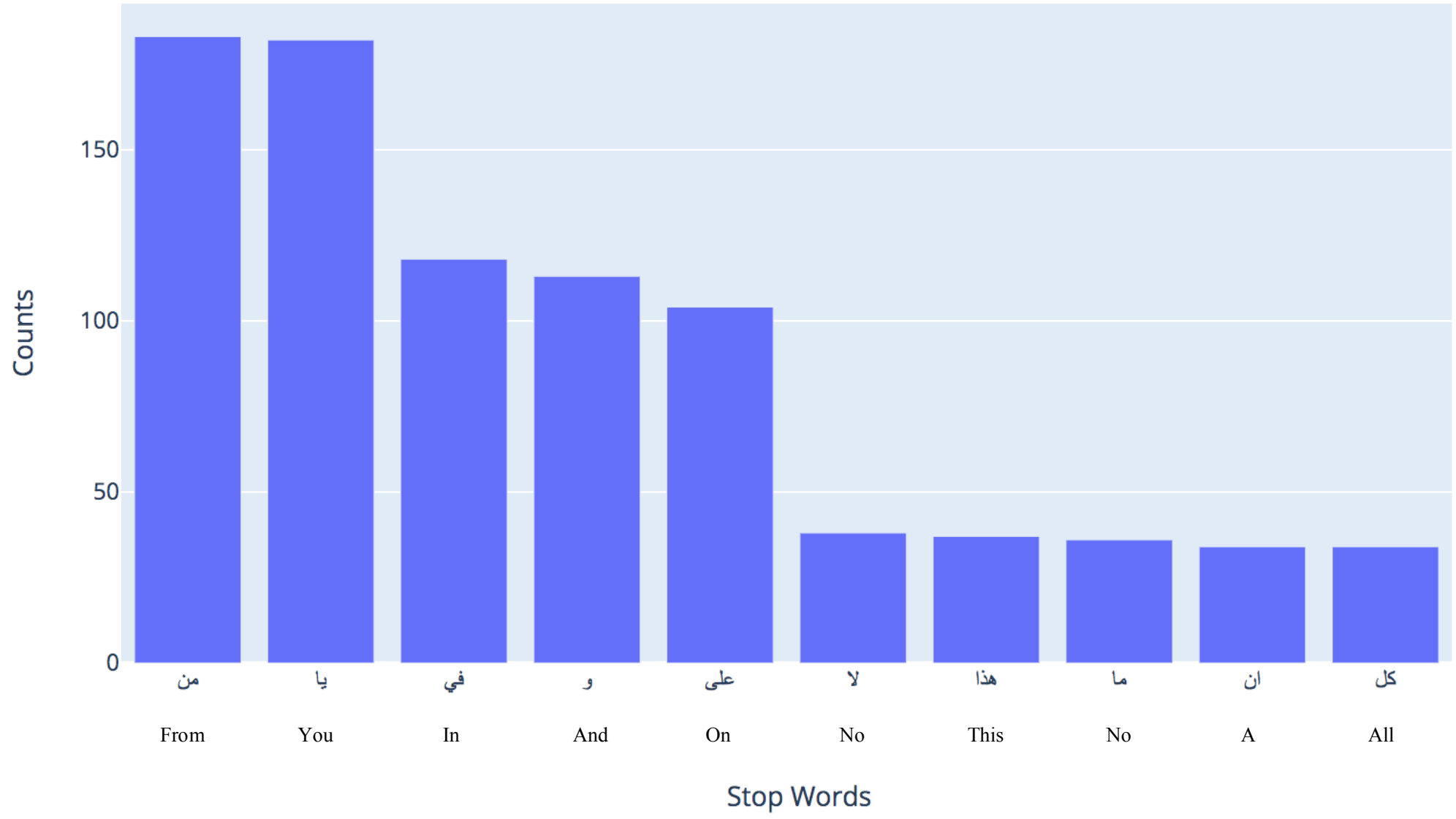

**Figure 9 Most common stop words in obscene class from the Aljazeera dataset**

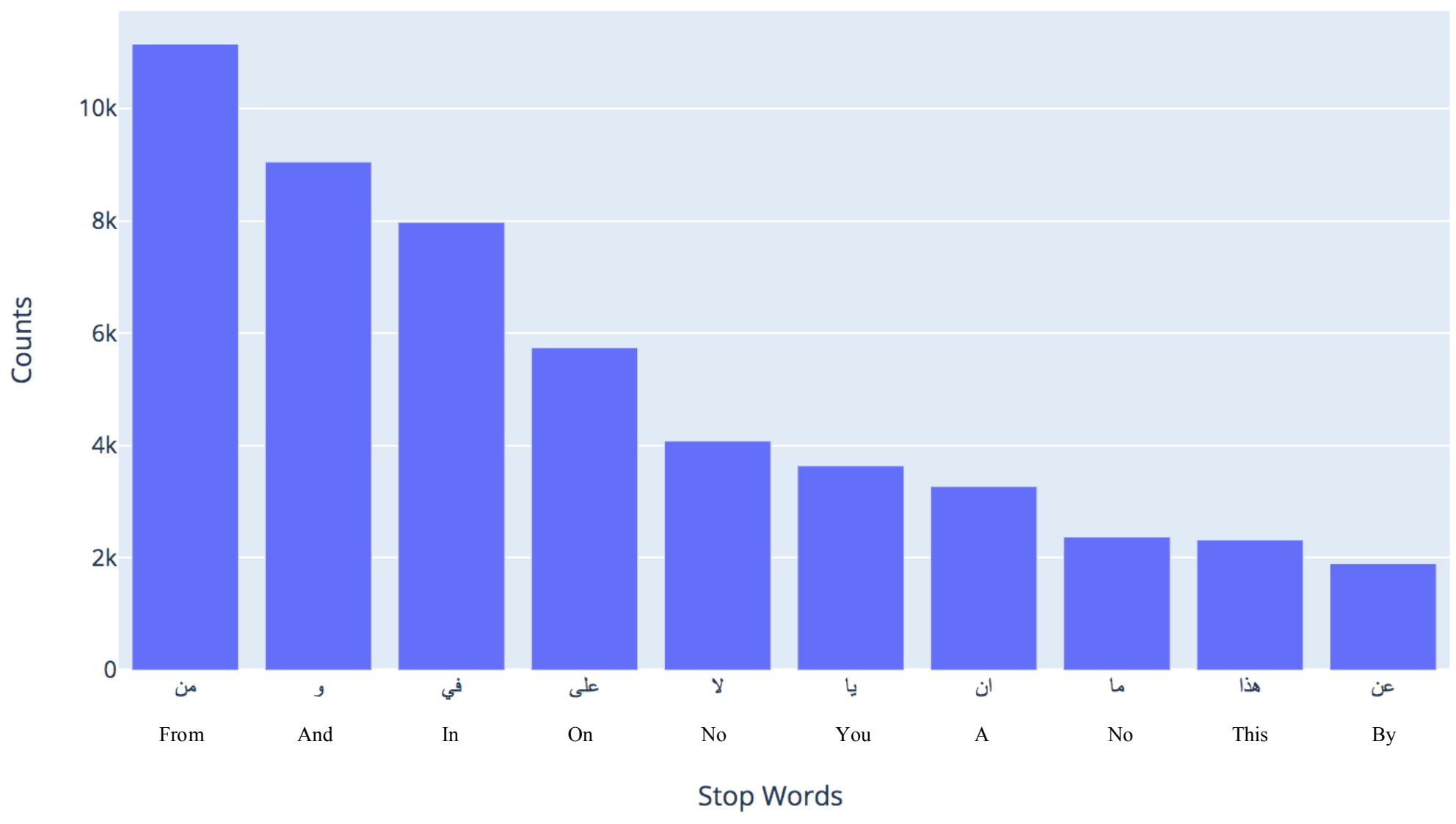

**Figure 10 Most common stop words in offensive class from the Aljazeera dataset**

Figure 11 is a bar chart for the sentiment analysis results. Obscene comments are all labeled negatively by the Mazajak online tool, while clean and offensive comments have mixed sentiments; mostly negative followed by neutral then positive.

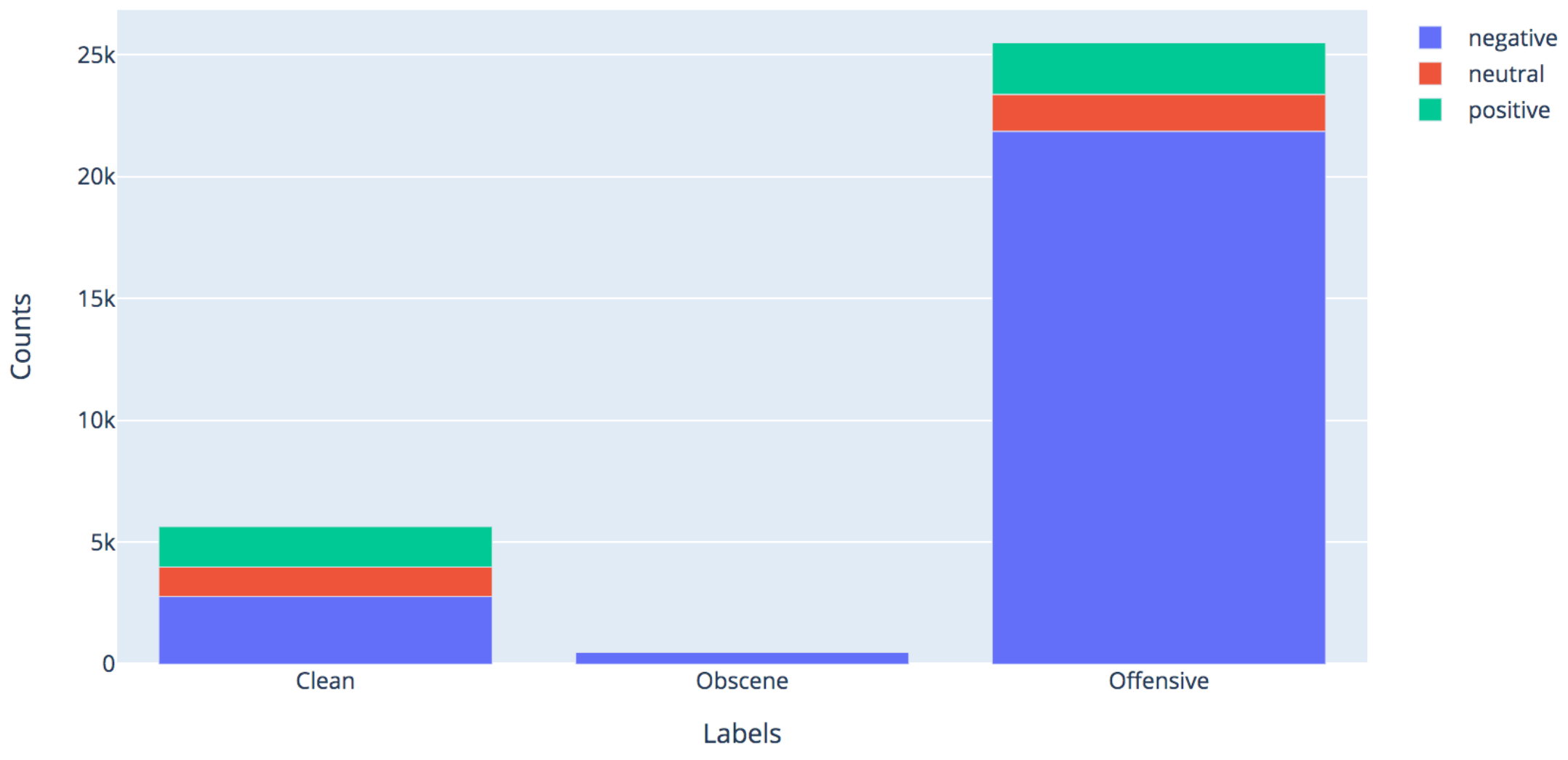

**Figure 11 Sentiment analysis based on labels for the Aljazeera dataset**

The provided comments do not have any emoji, so we couldn't analyze the use of emojis among the classes. Figures 12 to 14 show count frequencies of the top ten punctuation for each class. As can be seen from the figures, the top first three punctuation are exactly the same for all classes; “.”, “!”, and “؟” respectively.

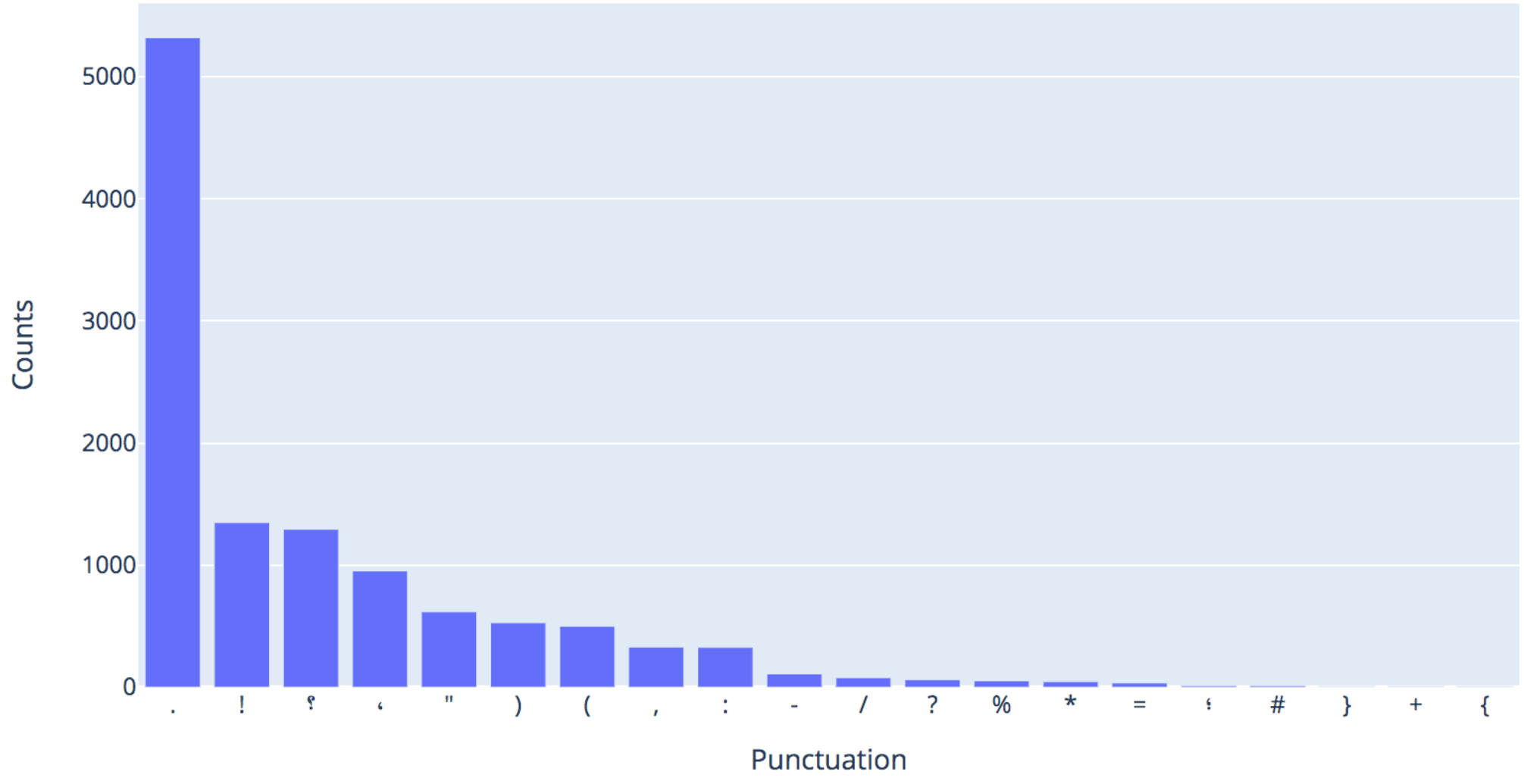

**Figure 12 Most common clean punctuation in the Aljazeera dataset**

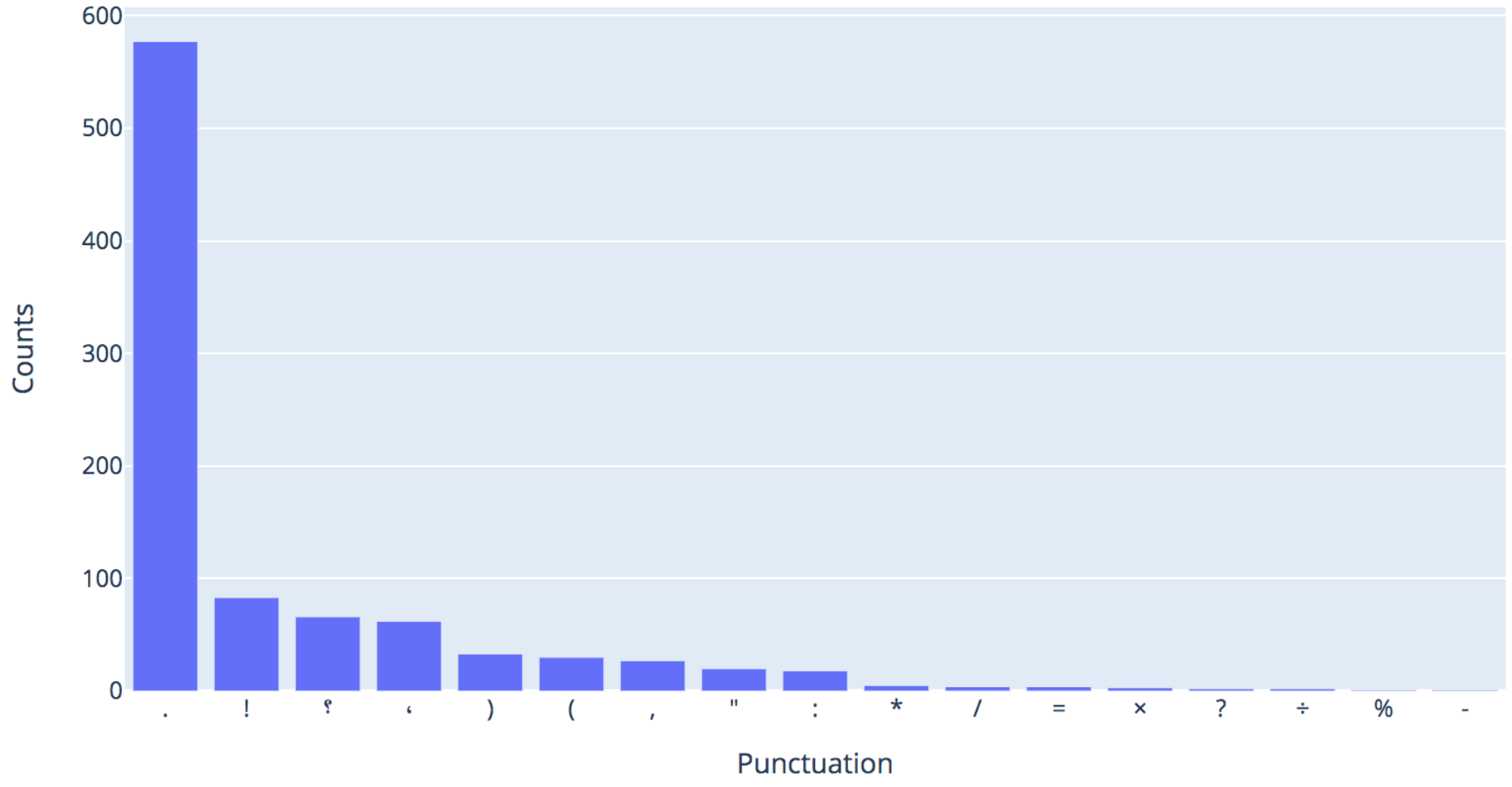

**Figure 13 Most common obscene punctuation in the Aljazeera dataset**

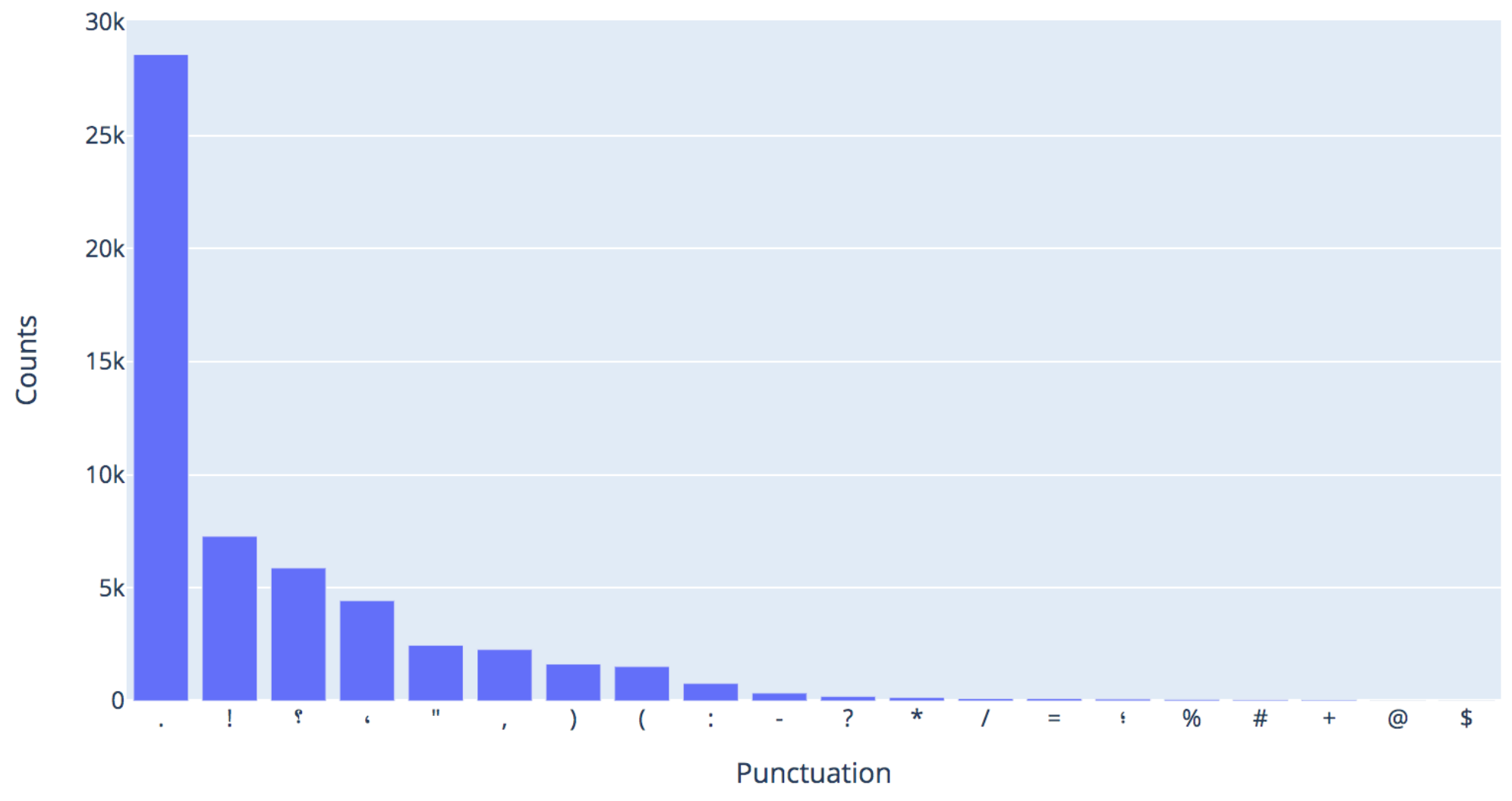

**Figure 14 Most common offensive punctuation in the Aljazeera dataset**

2) Egyptian Tweets Dataset:

The Egyptian tweets dataset developed by the same researchers of the Aljazeera deleted comments dataset, and it has the same labeling structure (Mubarak, Darwish, and Magdy, 2017). The total number of tweets is 1,100 tweets; 453 clean, 203 obscene, and 444 offensive (see Figure 15). The total number of duplicate tweets were two; the following is an example from a duplicate offensive tweet:

يا عم انت شارب أيه؟ يعني دي سحنه تجيب سياحه يا ناس ؟ دي سحنه تجيب مرض بالكتير

Translation: Sir what did you drink? she is ugly how can she brings tourists? She is ugly bringing sickness to most people

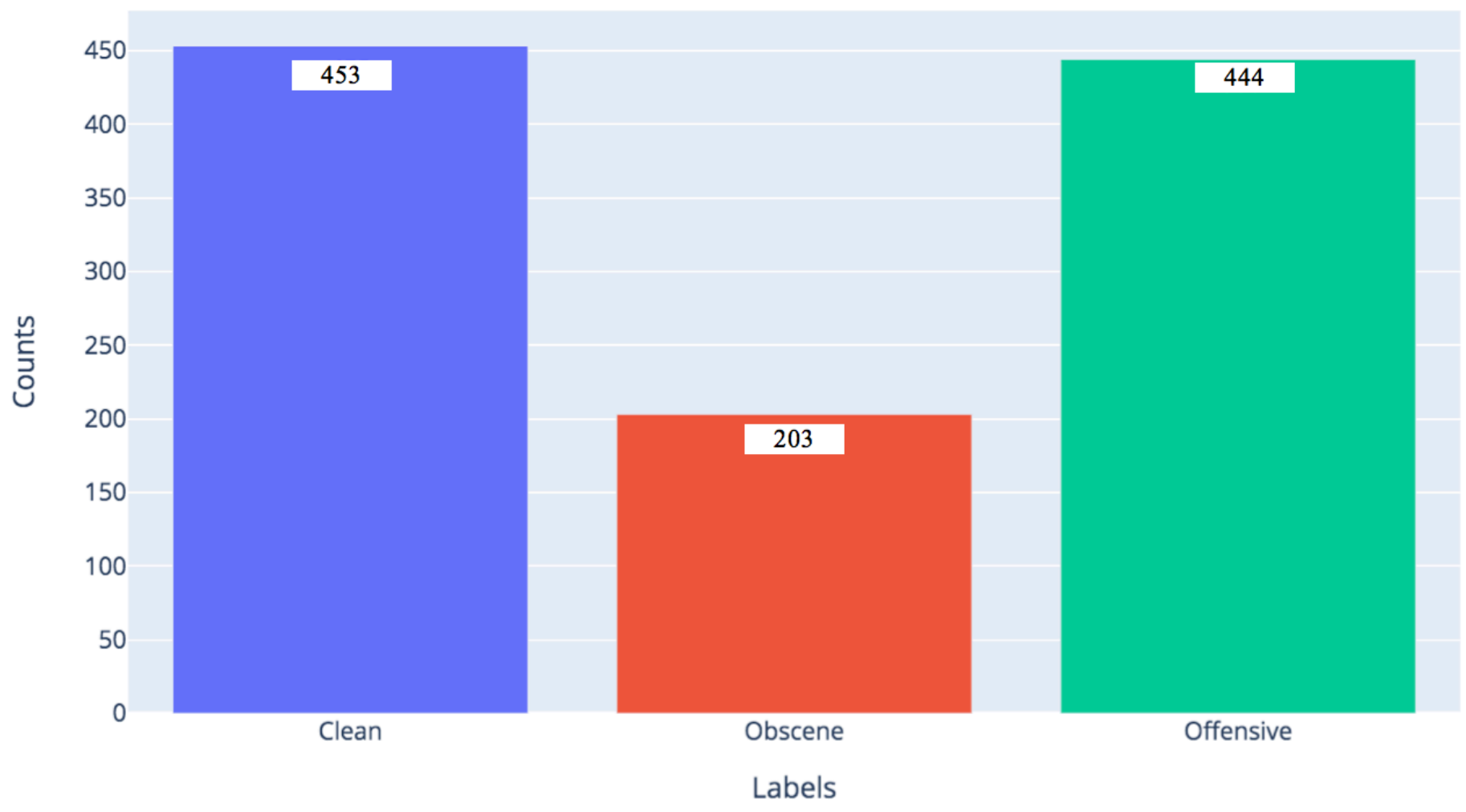

**Figure 15 Class distribution for the Egyptian Tweets dataset**

Figure 16 shows the word cloud for each class. Clean tweets show “انت / you” as the largest word, obscene tweets show “امك / your mother”, and offensive tweets show “امثالك / similar you”.

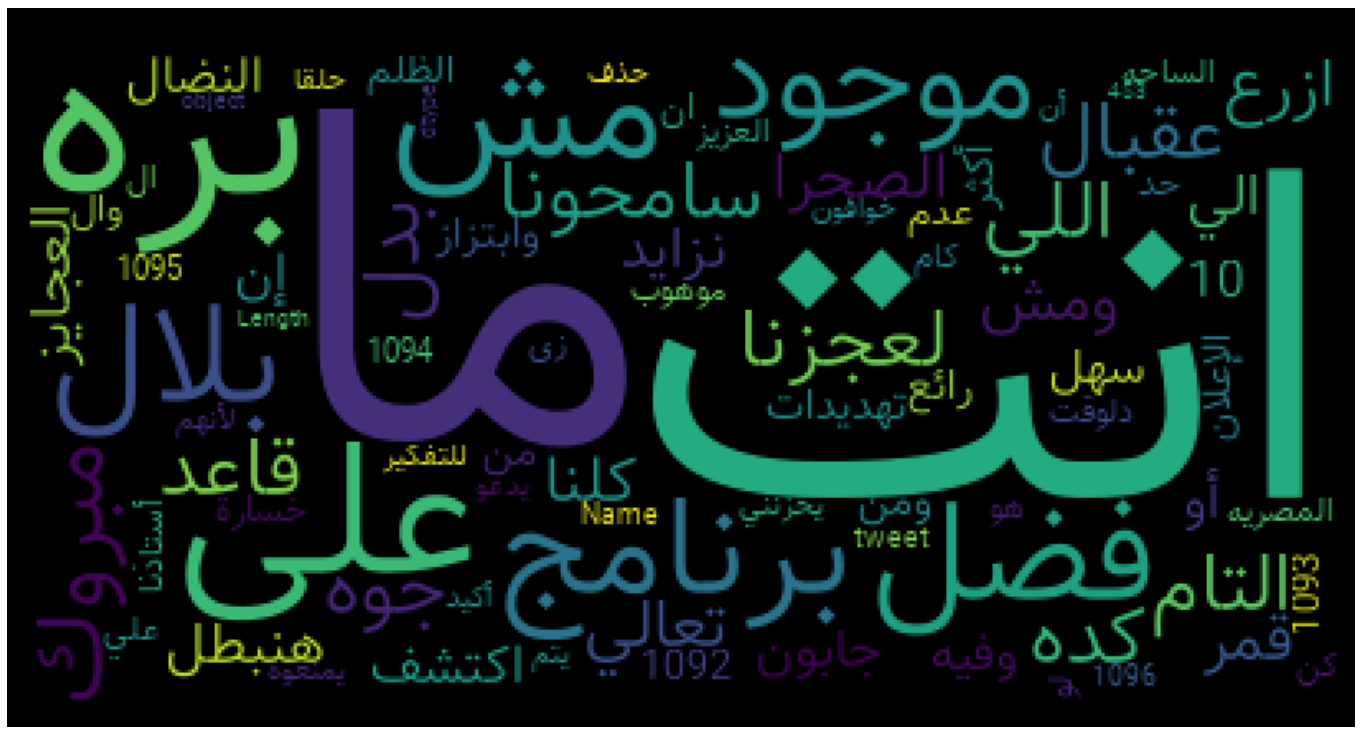

(a) Clean

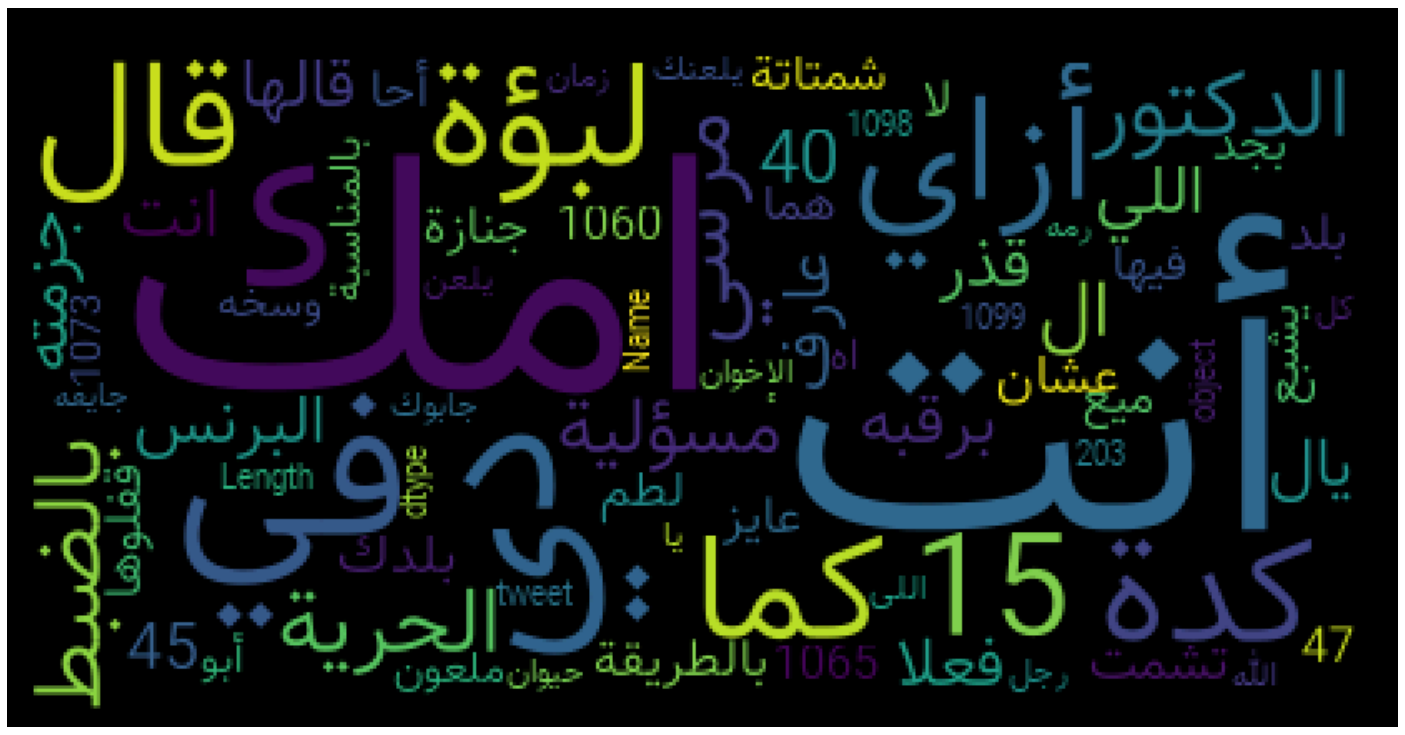

(b) Obscene

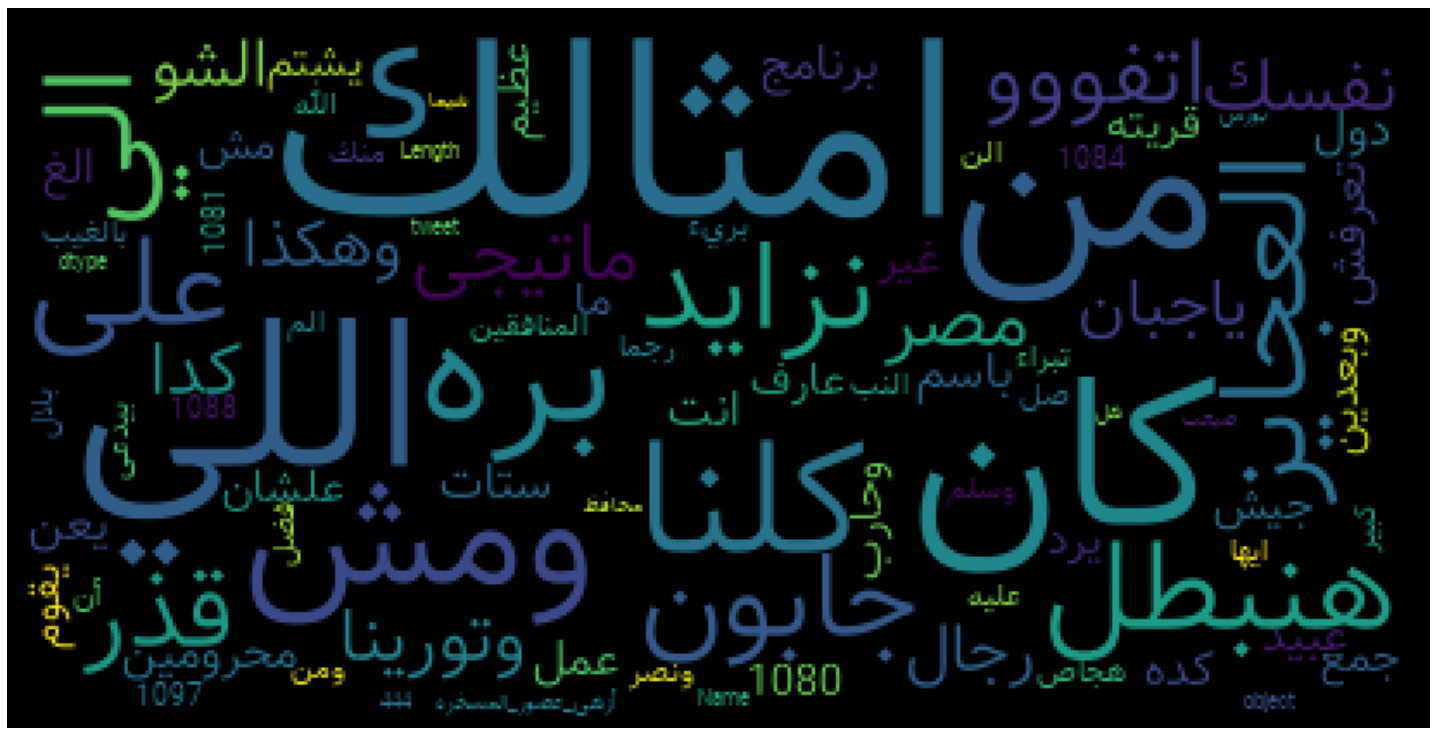

(c) Offensive

**Figure 16 The word cloud of the Egyptian Tweets dataset (a. clean, b. obscene, c. offensive)**

The top frequent tokens differ among the classes (see Figures 17 to 19). Clean tweets report “الله / God”, followed by “مصر / Egypt”, and “شعب / people”. Obscene tweets firstly report “امك / your mother”, followed by “عرص / bad behaved” and “كلب / dog”. Offensive tweets show similar counts for the top two frequent tokens; “مصر / Egypt” and “الله /God”, second in count after them is “منك / from you”, and third is “سيسي / Sisi”.

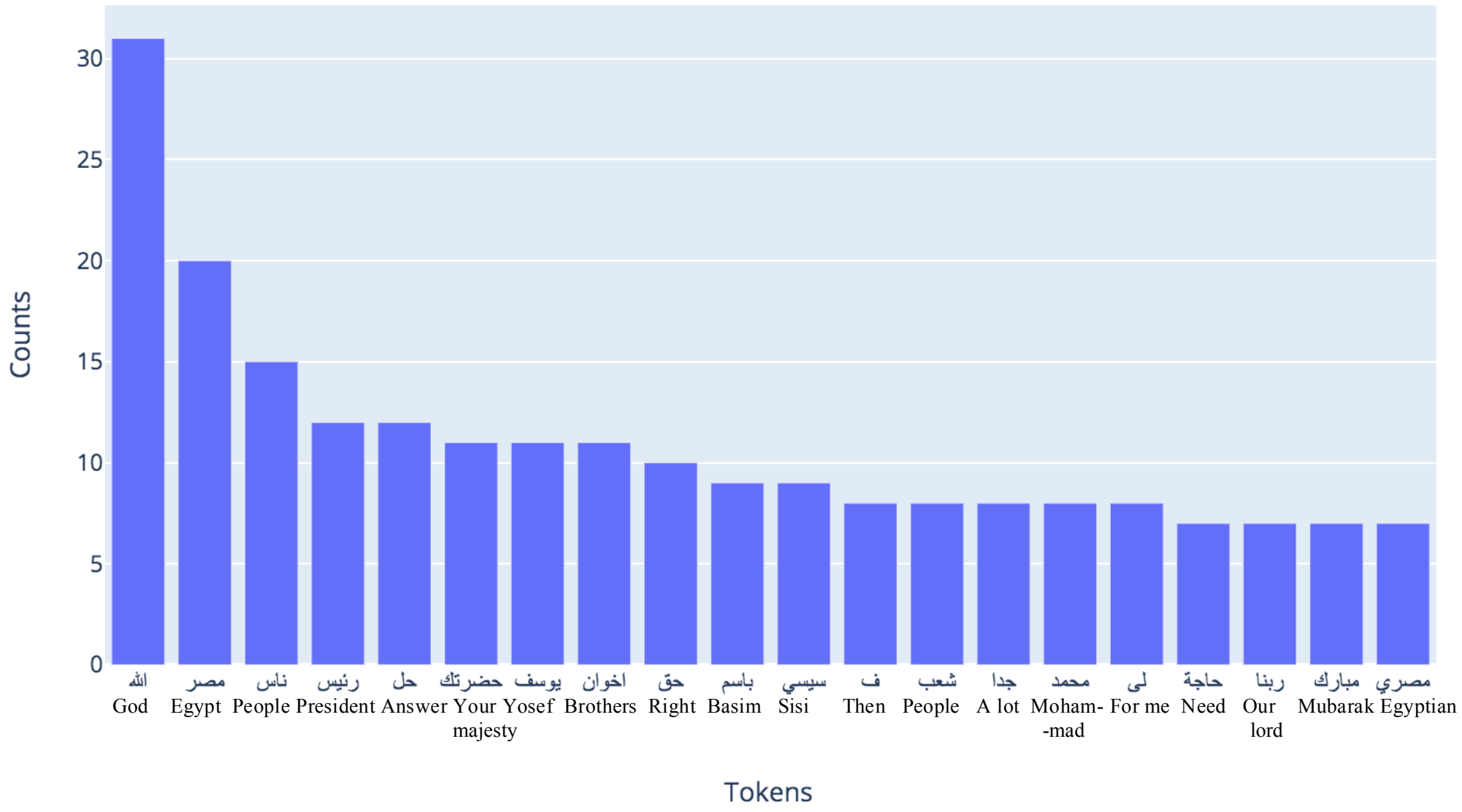

**Figure 17 Most common clean tokens in the Egyptian Tweets dataset**

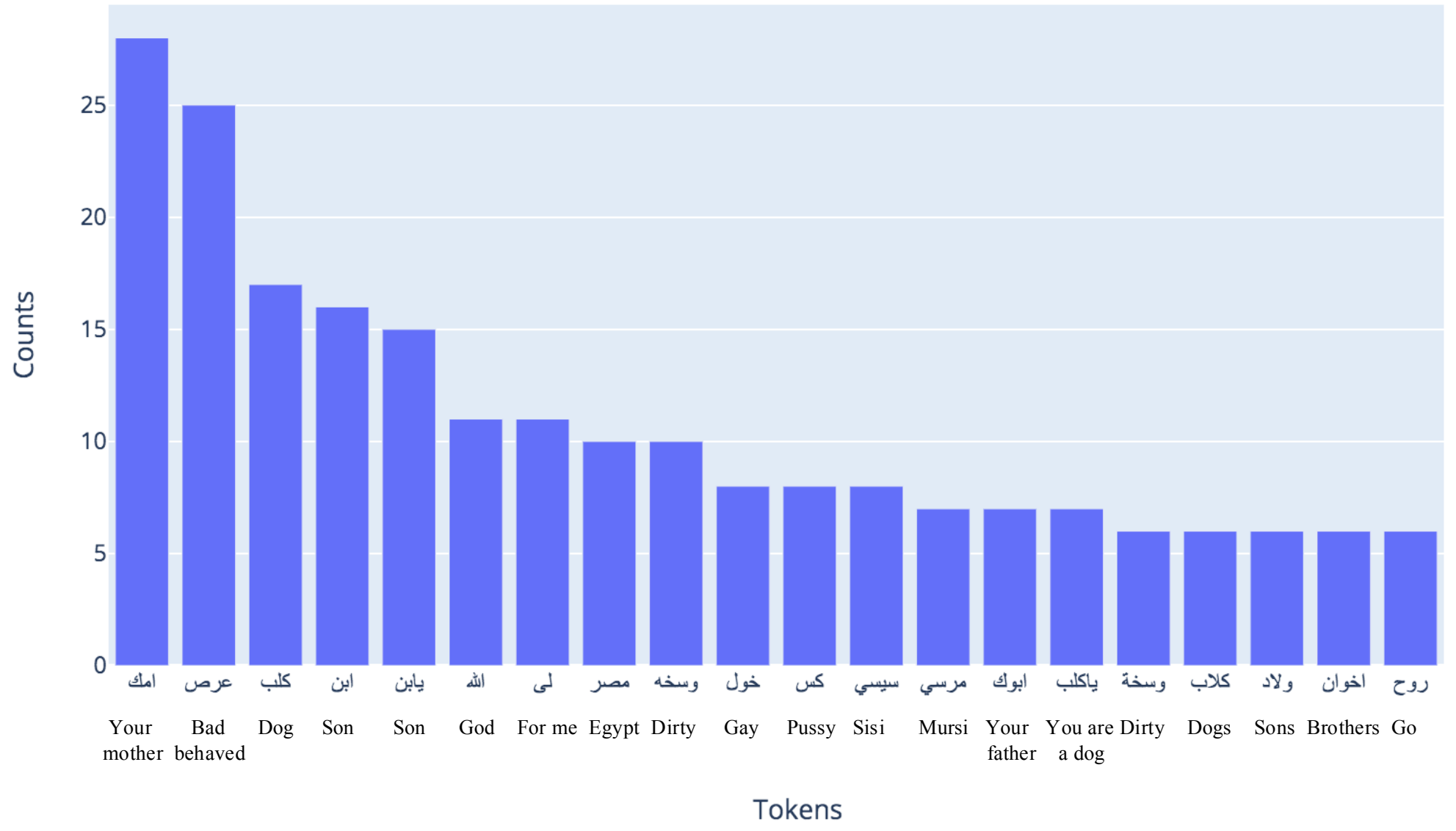

**Figure 18 Most common clean obscene in the Egyptian Tweets dataset**

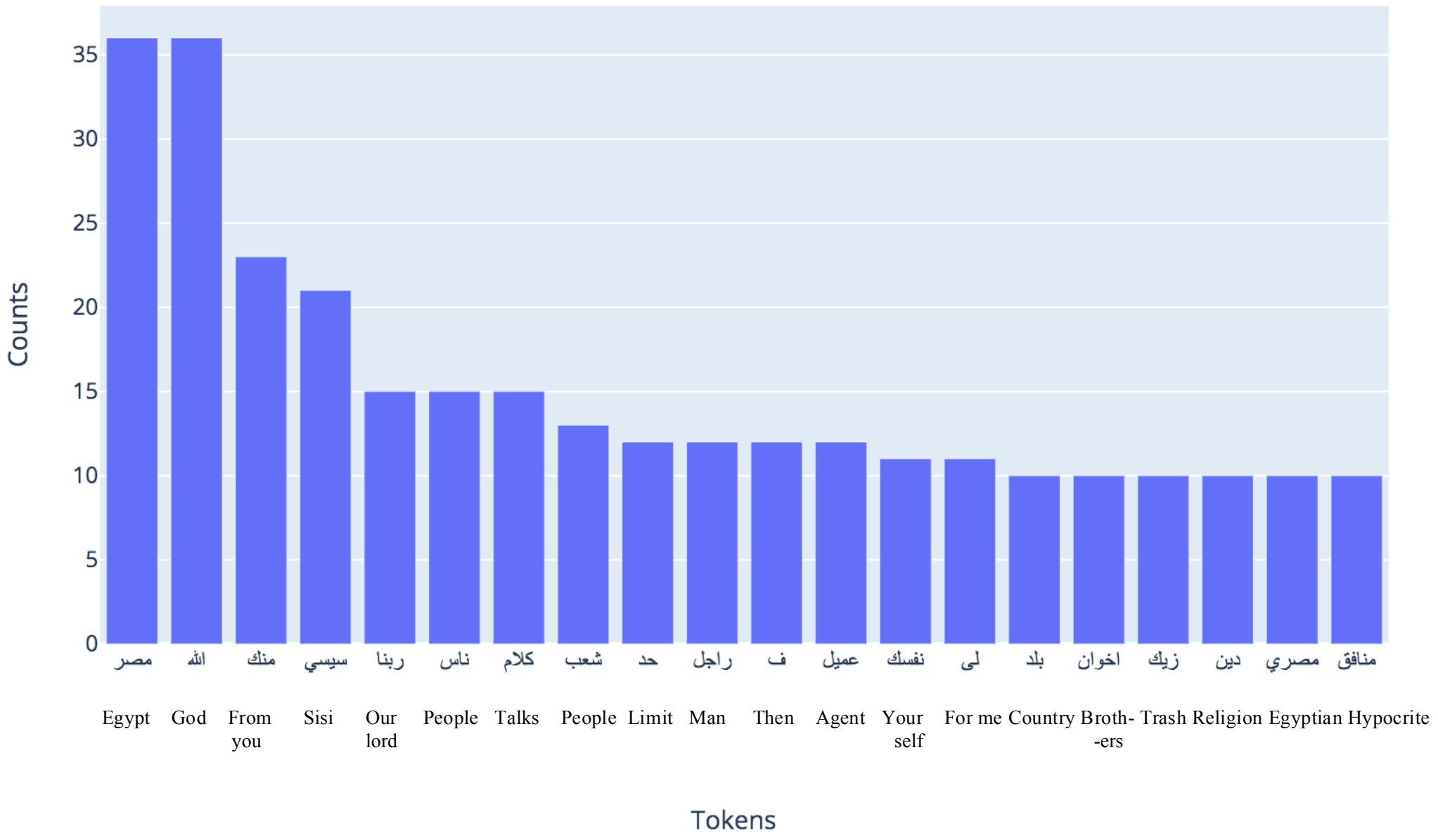

**Figure 19 Most common offensive obscene in the Egyptian Tweets dataset**

Results from the web-based tool Voyant show the following distinctive words:

1. Clean: الحل / the solution (10), الرئيس / the president (14), علاقة / relationship (5), خير / good (5), الظلم / injustice (5).
2. Obscene: العرص / the bastard (16), امك / your mother (29), عرص / bastard (10), كس / pussy (8), خول / gay (7).
3. Offensive: البرادعى / El-Baradei (6), بطل / hero (5), اهبل / stupid (5), القرف / disgusting (5), محترم / respected (4).

The two Figures; 20 and 21; demonstrate the variations among classes in terms of tweets length and tokens length. As can be noticed from both Figures, offensive tweets are the longest one in terms of tweets length and tokens length, followed by clean tweets and then by obscene tweets. In addition, maximum and minimum number of tokens and characters for clean tweets are larger than those of the other categories.

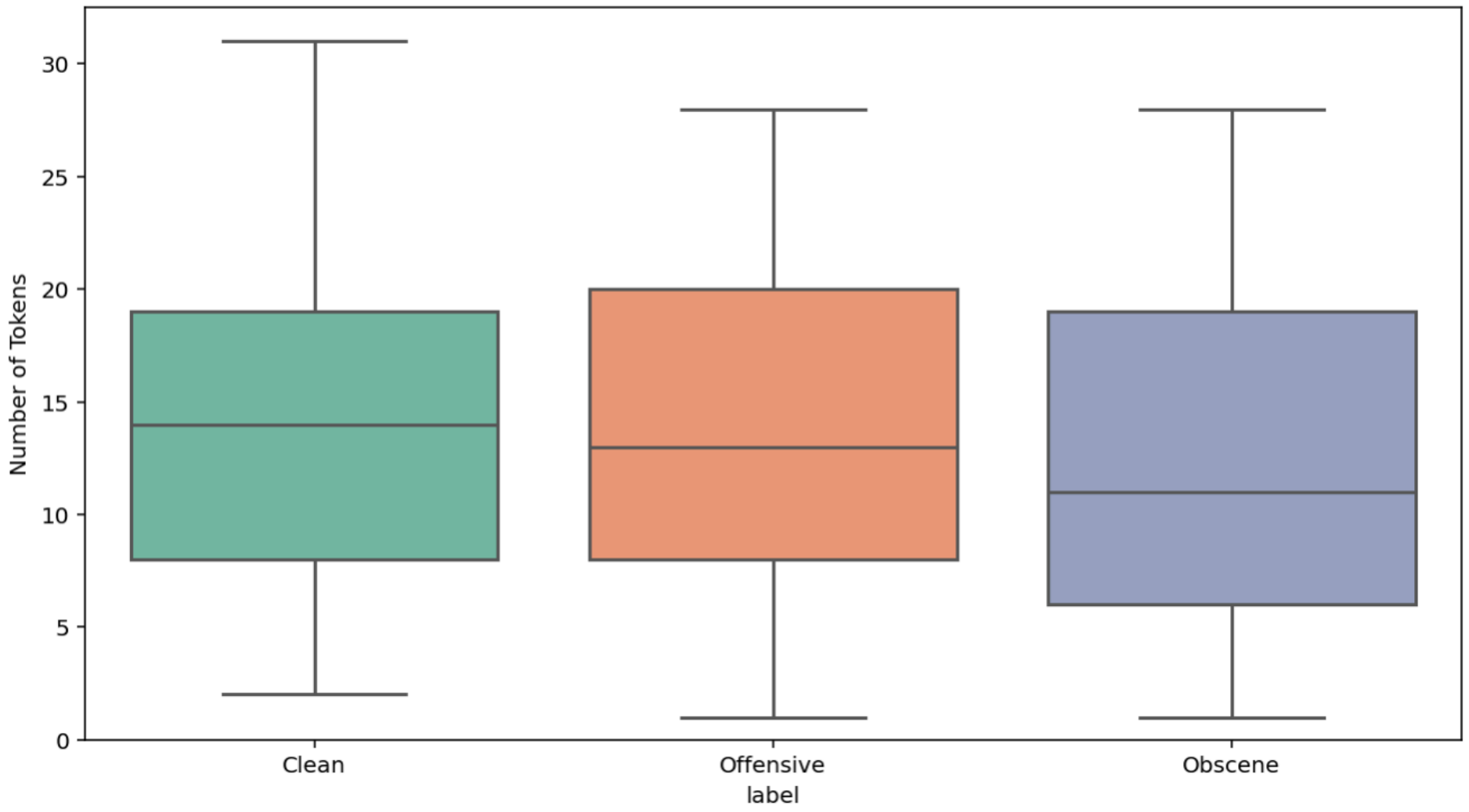

**Figure 20 Statistics of each label in the Egyptian Tweets dataset based on the number of tokens per tweet**

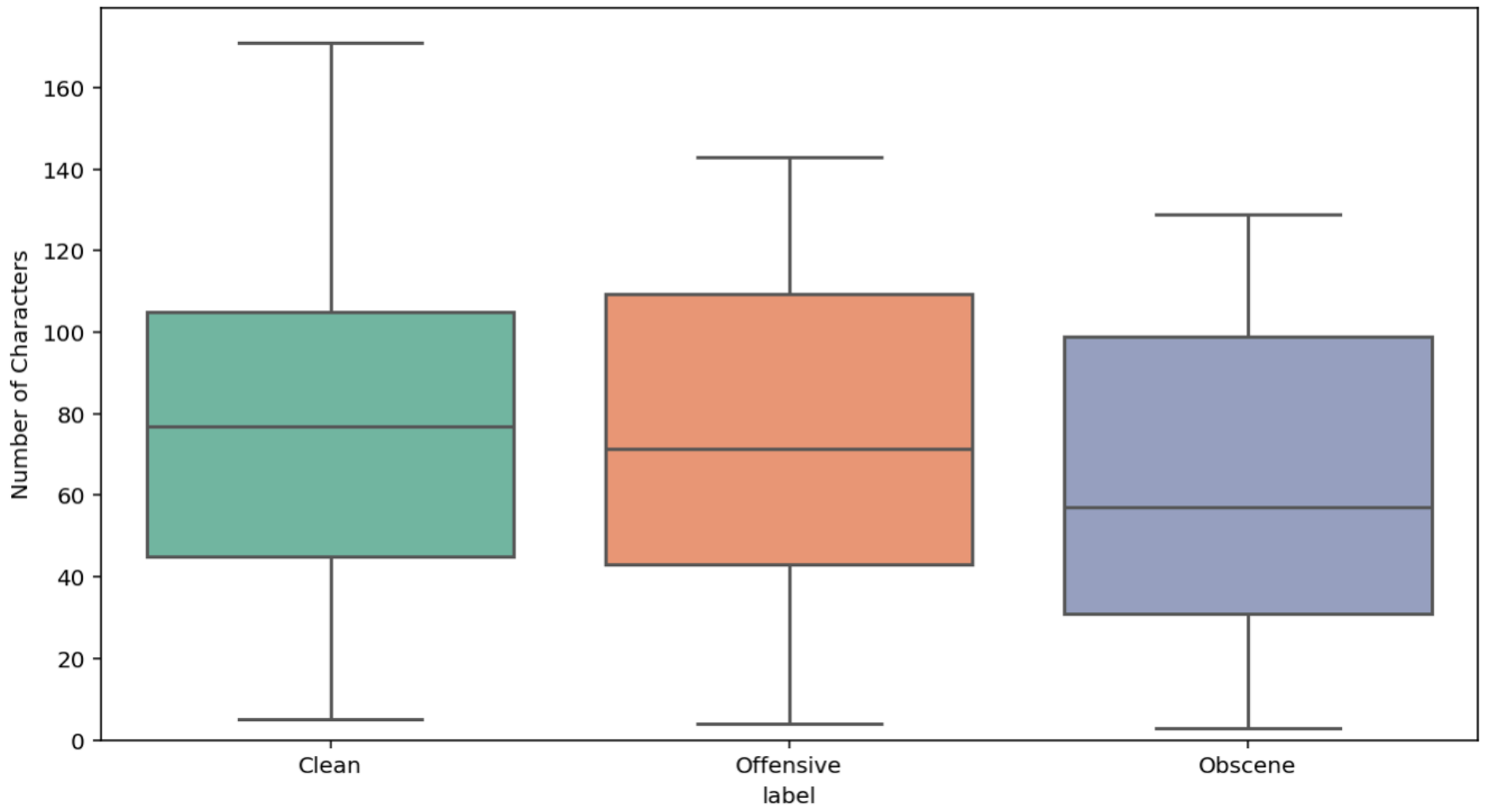

**Figure 21 Statistics of each label in the Egyptian Tweets dataset based on the number of characters per token**

Figures 22 to 24 indicate some similarities between obscene tweets and offensive tweets in term of using stop words. The particles “يا / you”, “من / from”, and “في / in” respectively, are the top three frequent stop words in both obscene tweets and offensive tweets, while clean tweets report “من / from”, “في / in”, and “و / and” respectively as the top frequent ones.

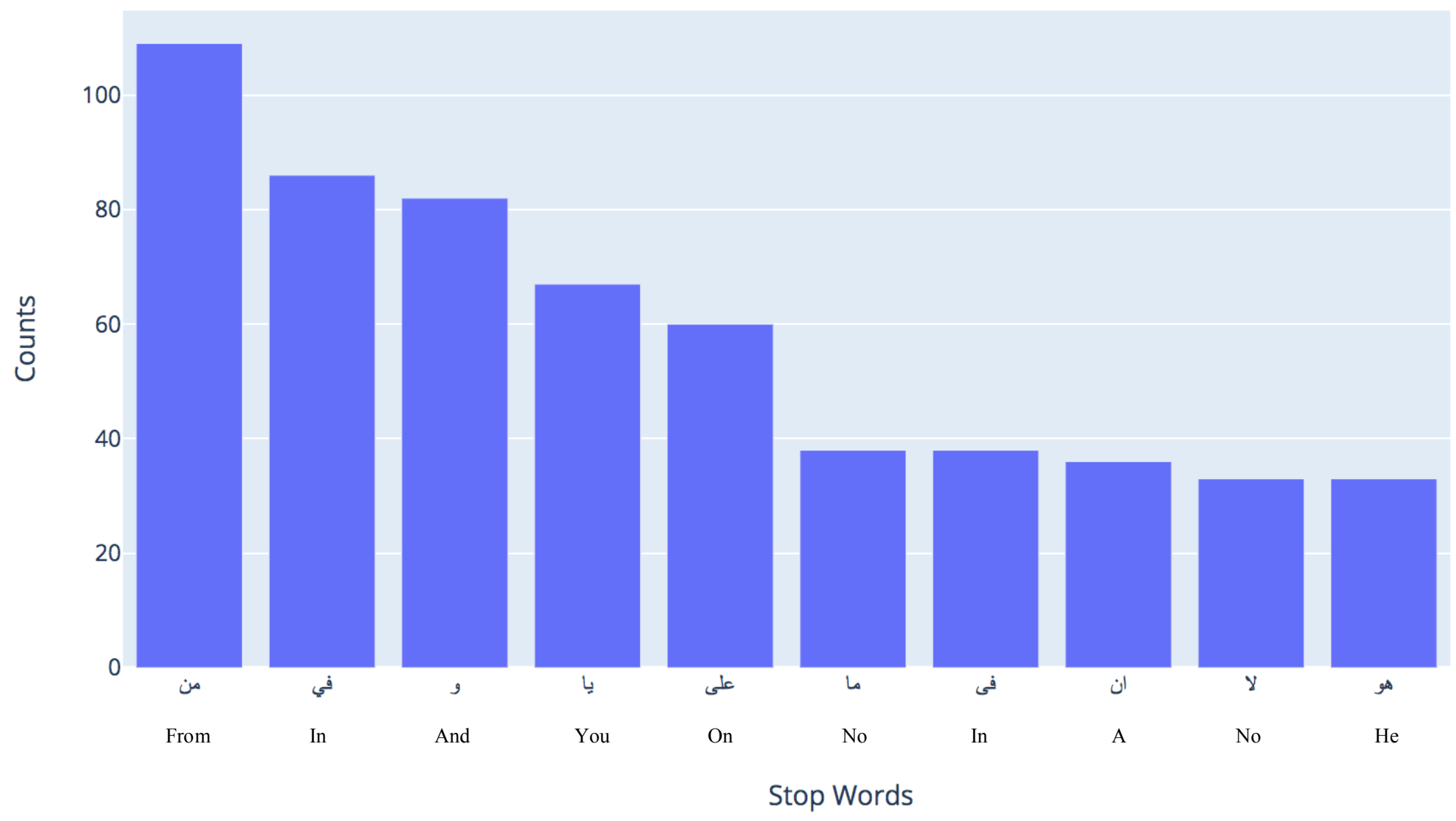

**Figure 22 Most common stop words in clean class from the Egyptian Tweets dataset**

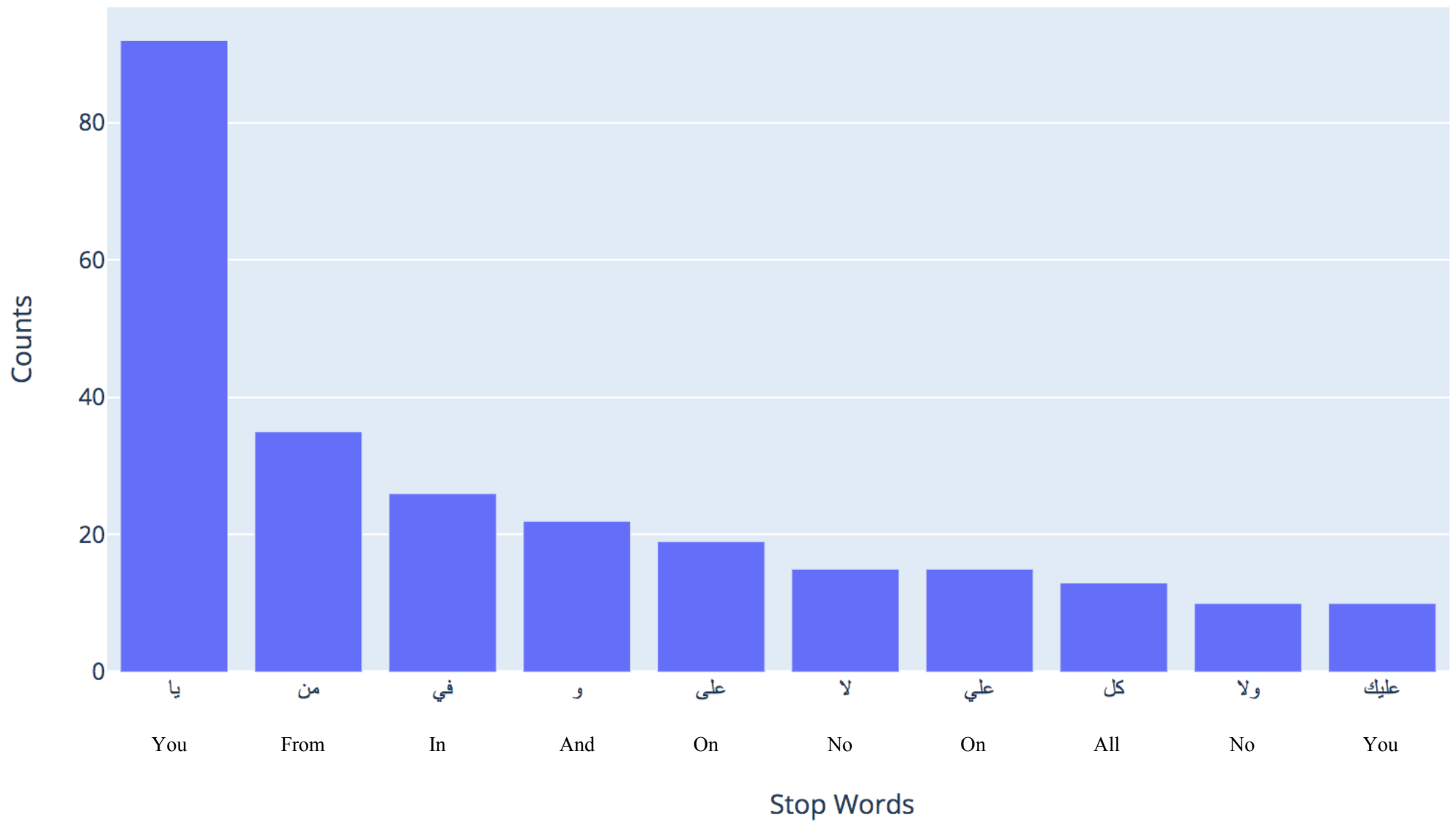

**Figure 23 Most common stop words in obscene class from the Egyptian Tweets dataset**

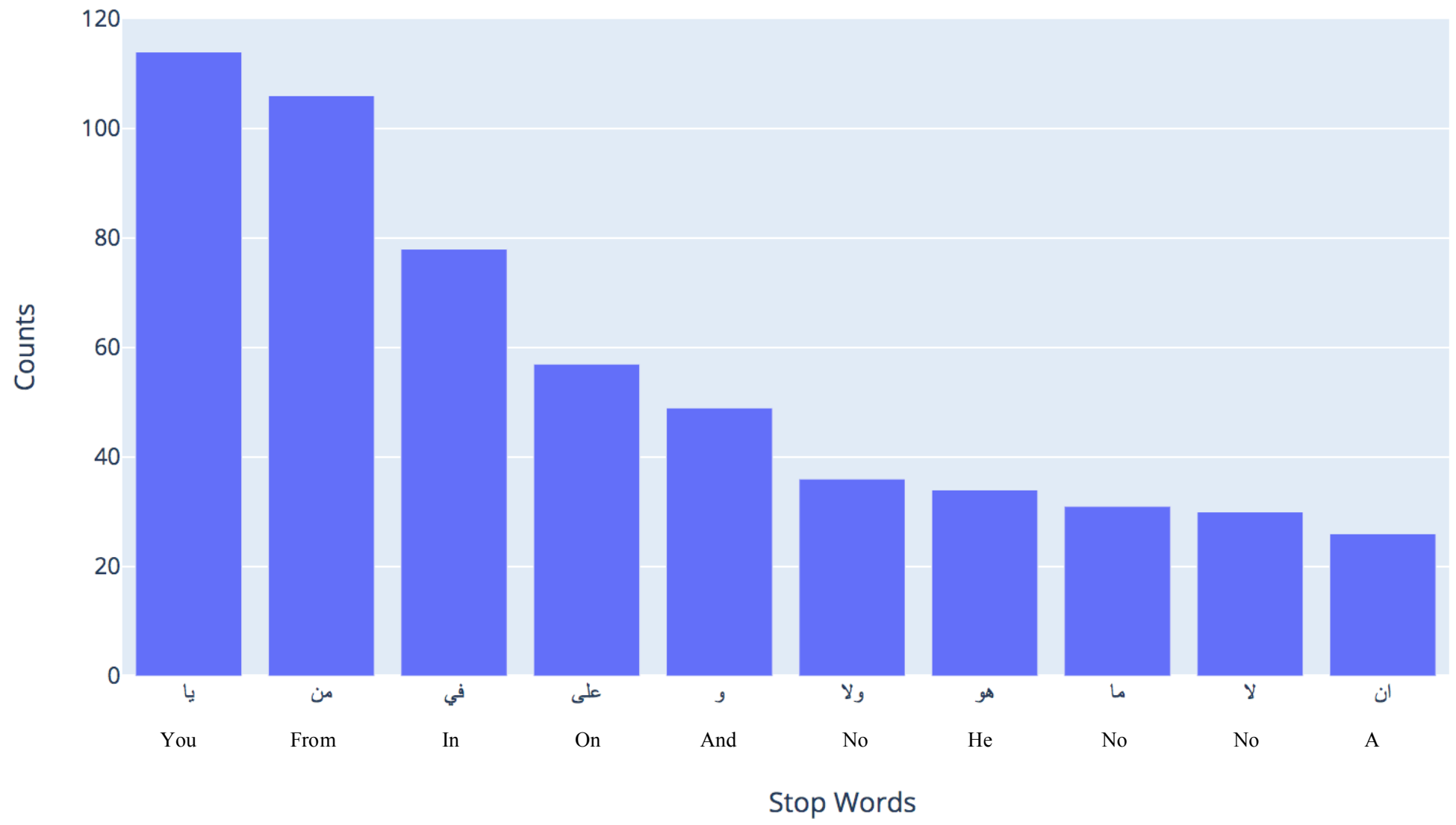

**Figure 24 Most common stop words in offensive class from the Egyptian Tweets dataset**

Sentiment analysis results highlight the overall relationship with negative sentiment in among all classes. Clean tweets have more positive sentiments than the others. Figure 25 shows more detailed about the sentiment analysis results.

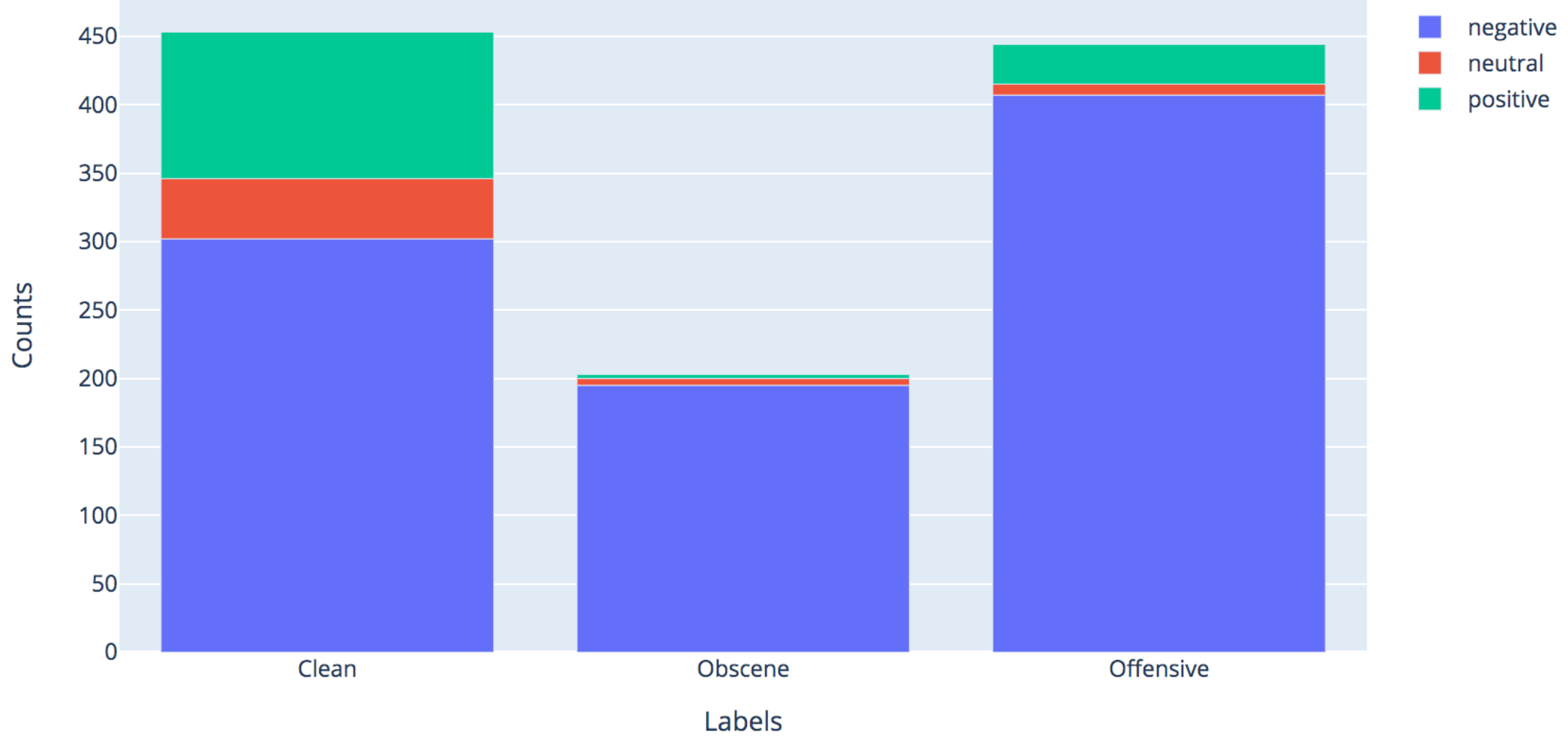

**Figure 25 Sentiment analysis based on labels for the Egyptian Tweets dataset**

Investigating the use of emoji within classes from Figures 26 to 28 reveals some similar patterns among all tweets, such as the high frequent use of the face with tears of joy emoji, "😂". It is also noticeable that obscene tweets have very limited use of emoji. Only three types of emoji are occurred within the entire obscene tweets.

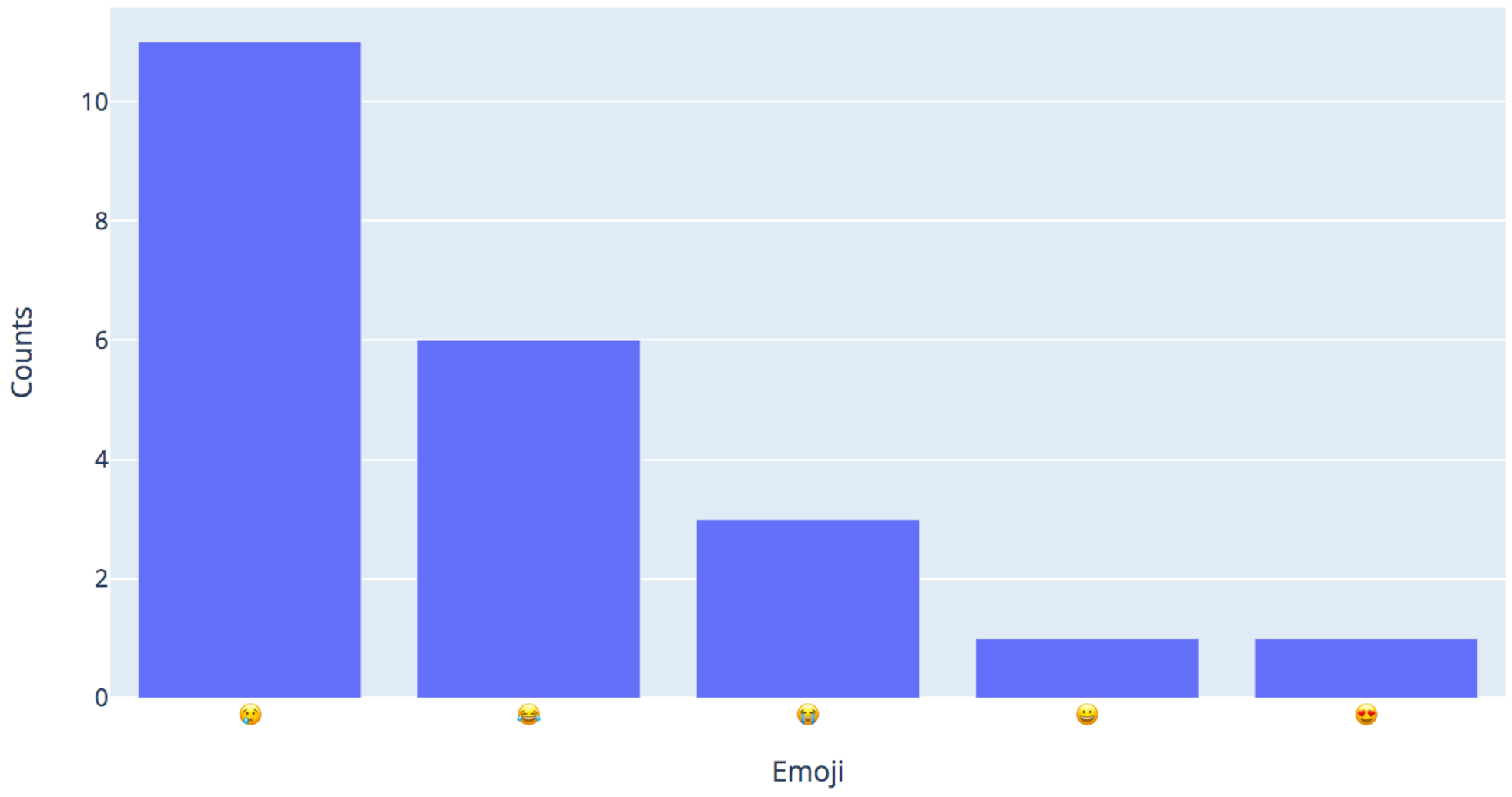

**Figure 26 Most common clean emoji in the Egyptian Tweets dataset**

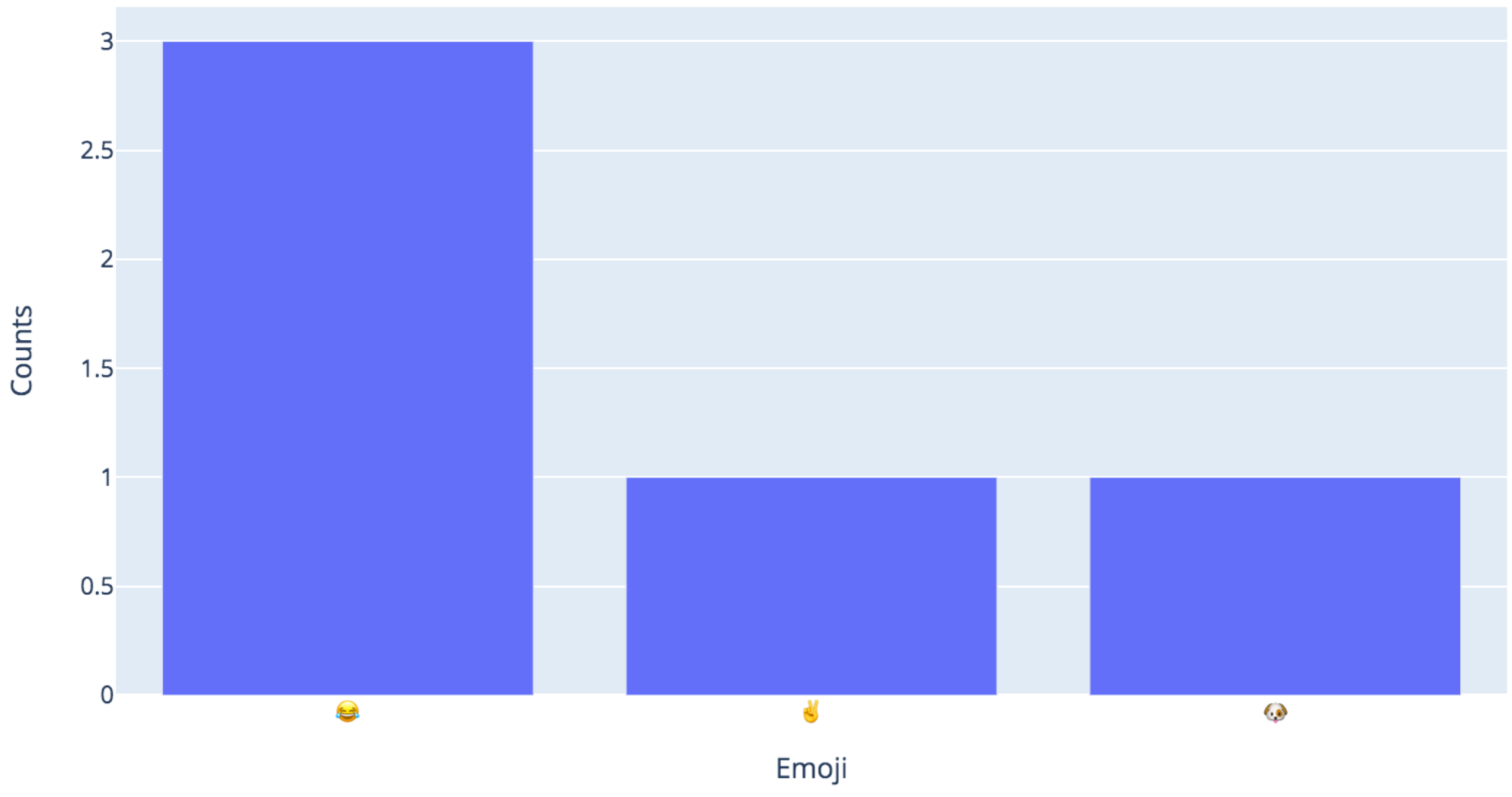

**Figure 27 Most common obscene emoji in the Egyptian Tweets dataset**

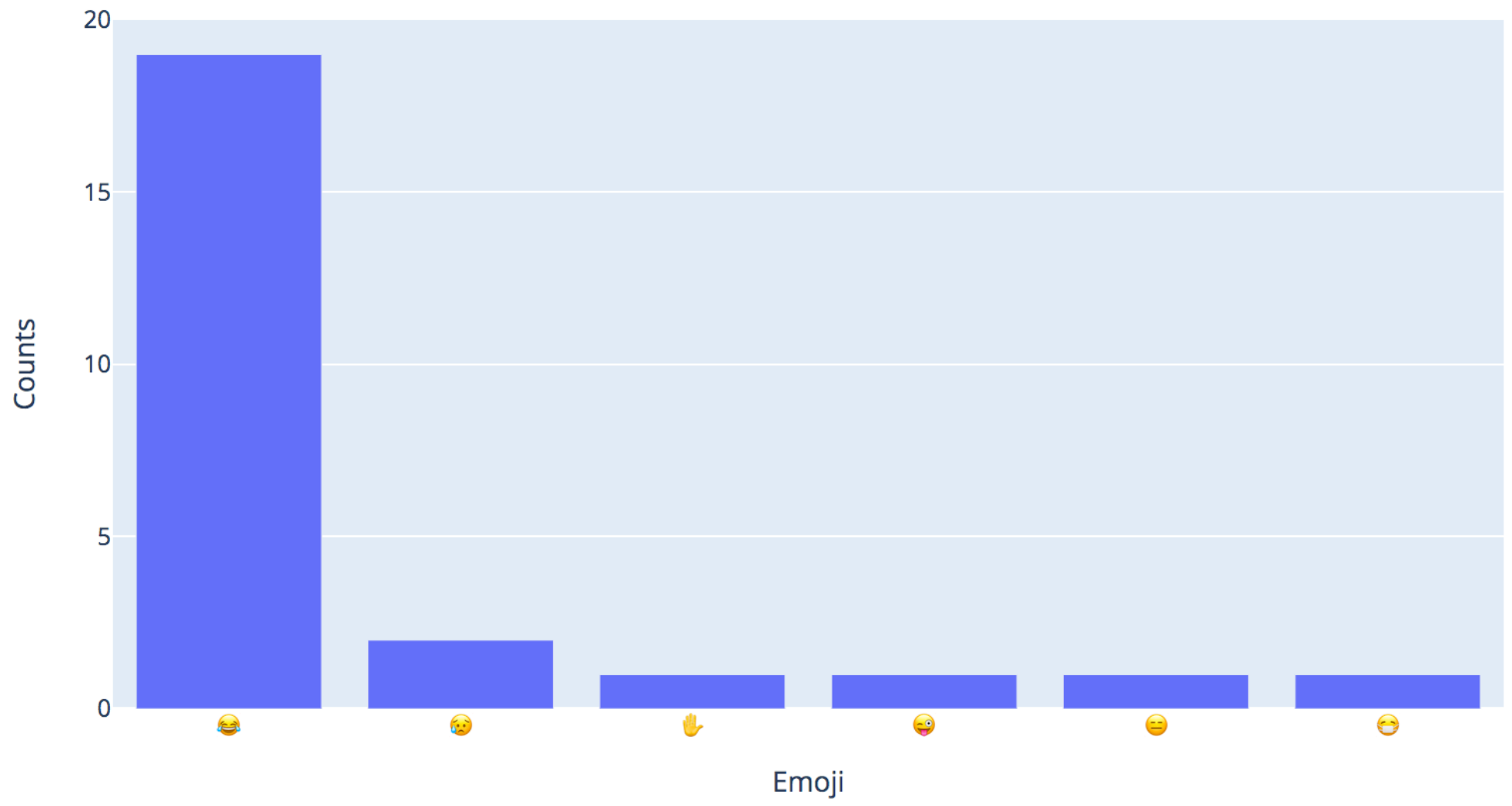

**Figure 28 Most common offensive emoji in the Egyptian Tweets dataset**

Similar to emoji results, the top frequent punctuation among the classes are very similar, with "." and """ as the first and second top respectively among all classes. Figures 29 to 31 cover more information for the use of punctuation among all classes.

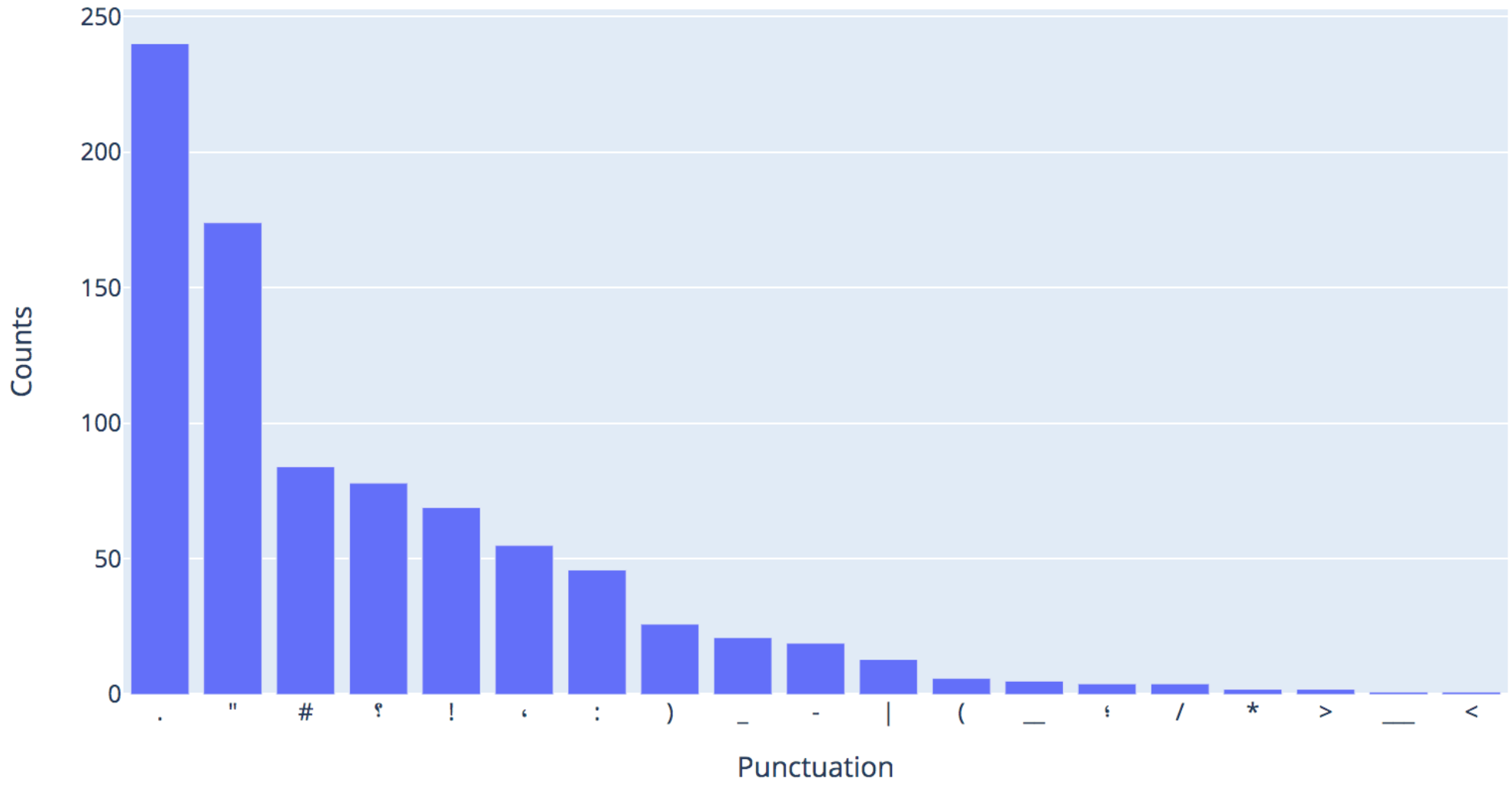

**Figure 29 Most common clean punctuation in the Egyptian Tweets dataset**

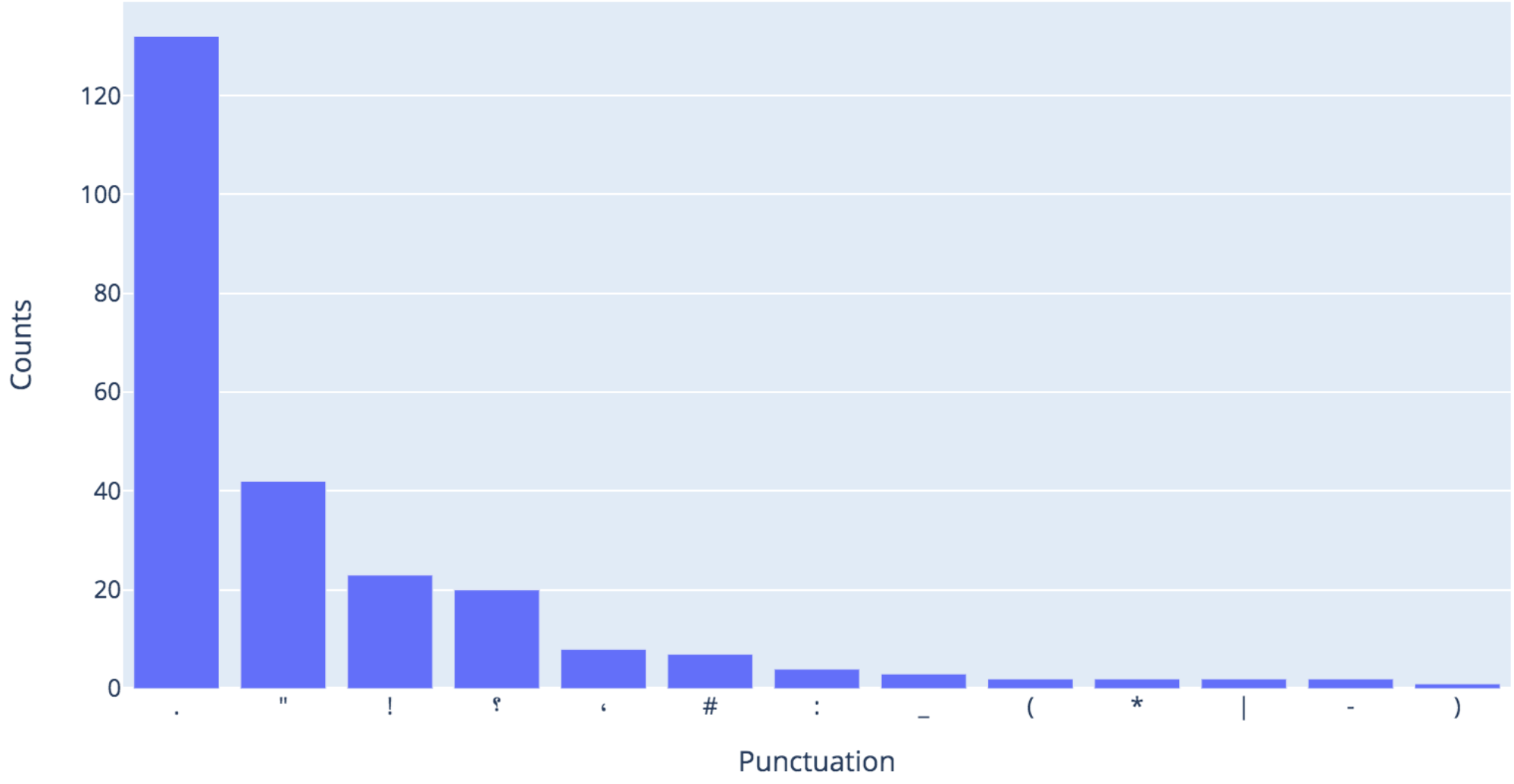

**Figure 30 Most common obscene punctuation in the Egyptian Tweets dataset**

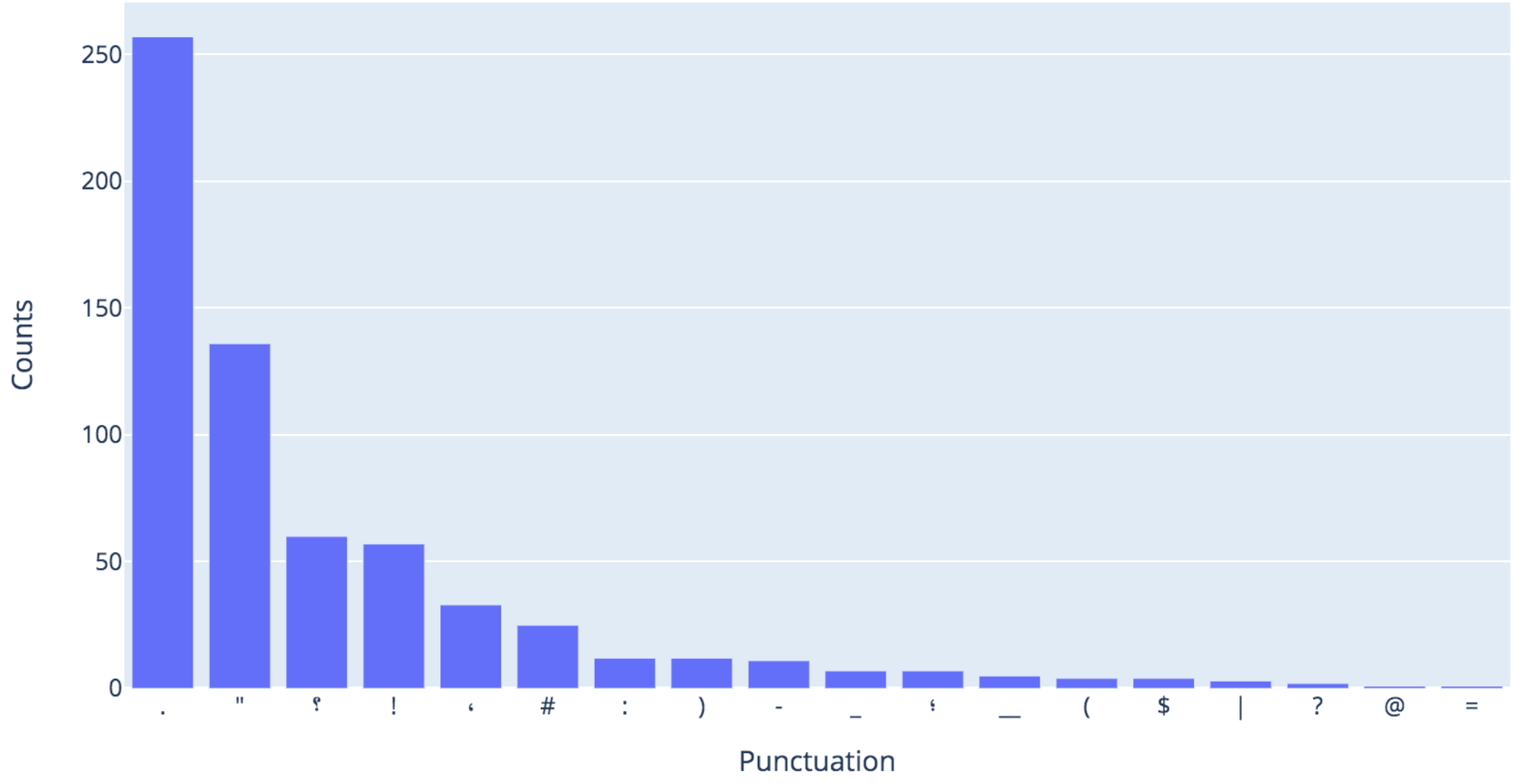

**Figure 31 Most common offensive punctuation in the Egyptian Tweets dataset**

3) Religious Hate Speech Dataset:

Albadi, Kurdi, and Mishra (2018) publish the Religious Hate Speech Dataset, which consists of 6,137 Arabic tweets; 2,762 hate, 3,375 not hate. Figure 32 shows the classes distribution. The dataset has multiple duplicates of the same tweets; 8 original tweets. The following is an example of a not hate tweet that has 13 instances in the dataset:

اللهم وحد أمة الإسلام و لاتفرقها

Translation: O' my God unite all Muslims and never disperse them

The following is another example from hate tweets that has been duplicated for 4 times in the dataset:

التّدين المغشوش قد يكون أنكى بالأمم من الإلحاد الصارخ !محمد الغزالي

Translation: Fake religiosity may have more sever effects on nations from Atheism! Muhammad al-Ghazali

Furthermore, some tweets are included multiple times and are classified as hate for some of their instances while classified as not hate for the others, an example of this type of tweet that is duplicated for 26 times; 21 not hate and 5 hate; is mentioned below:

اللهم ارحم من فارقونا الى القبور وانزل عليهم الضياء والنور والفسحة والسرور، رب اجعل بطون الالحاد خير منازلهم وفسيح جنانك هي ديارهم وقرارهم

Translation: O' my God have mercy upon the dead and lighten their graves, O' my lord let them rest in heaven

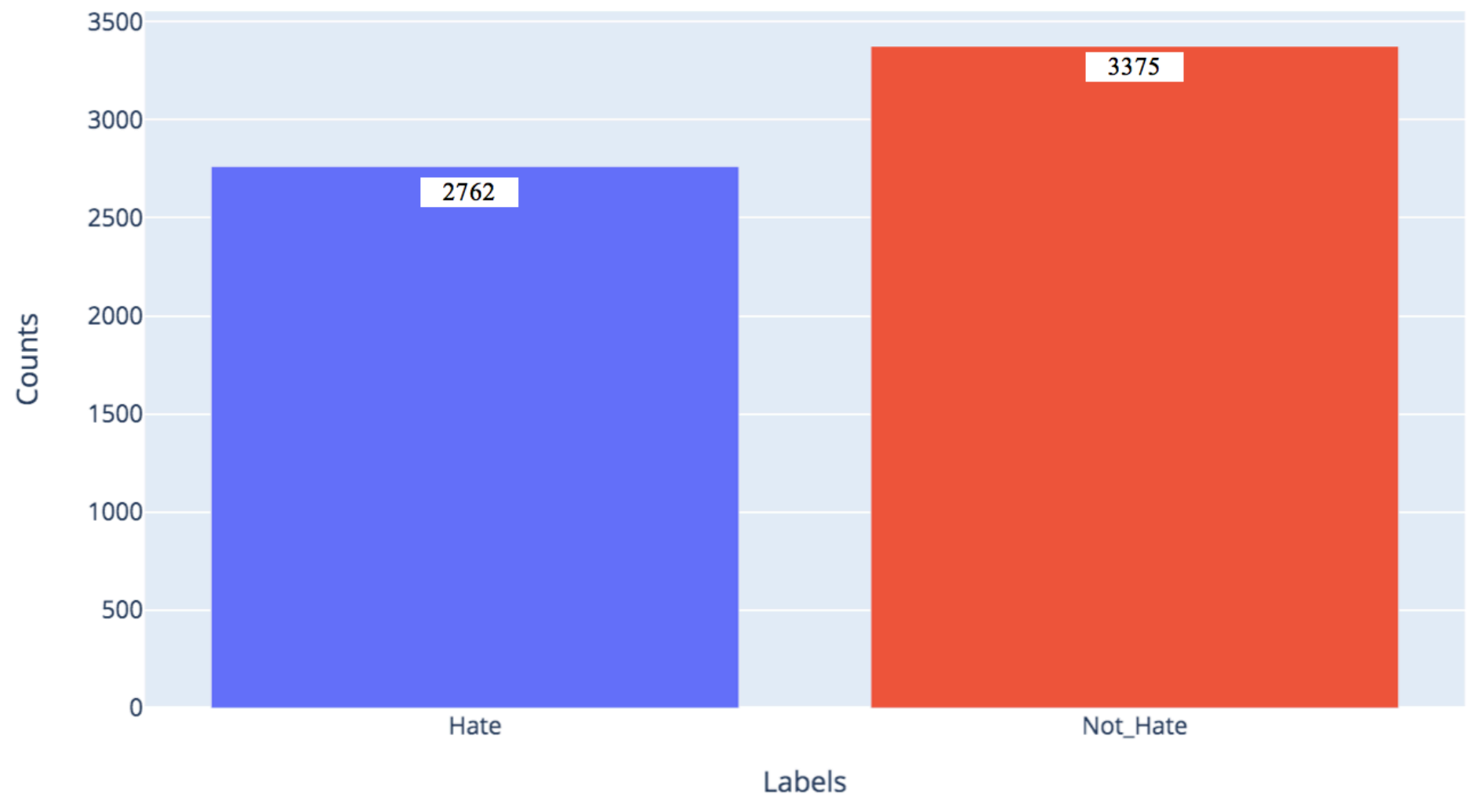

**Figure 32 Class distribution for the Religious Hate Speech dataset**

The word clouds in Figure 33 highlight the word “اليهود / the Jews” from hate tweets and the word “أهل / people” from not hate tweets.

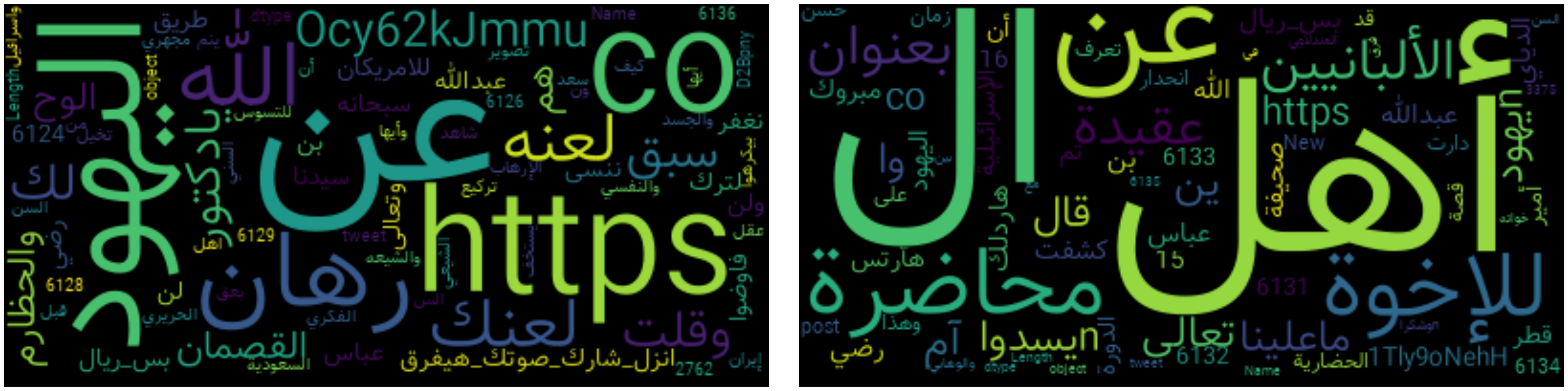

(a) Hate (b) Not hate

**Figure 33 The word cloud of the Religious Hate Speech dataset (a. hate, b. not hate)**

Figure 34 and 35 plot the count frequencies of the top tokens from each class. Both hate and not hate tweets report the word “الله / God” as the top frequent one. For hate tweets, the second most frequent token is “يهود / Jews” and the third is “شيعة / Shia”, while for not hate tweets, the second most frequent token is “لهم / for them” and the third is “مسلمين / Muslims”.

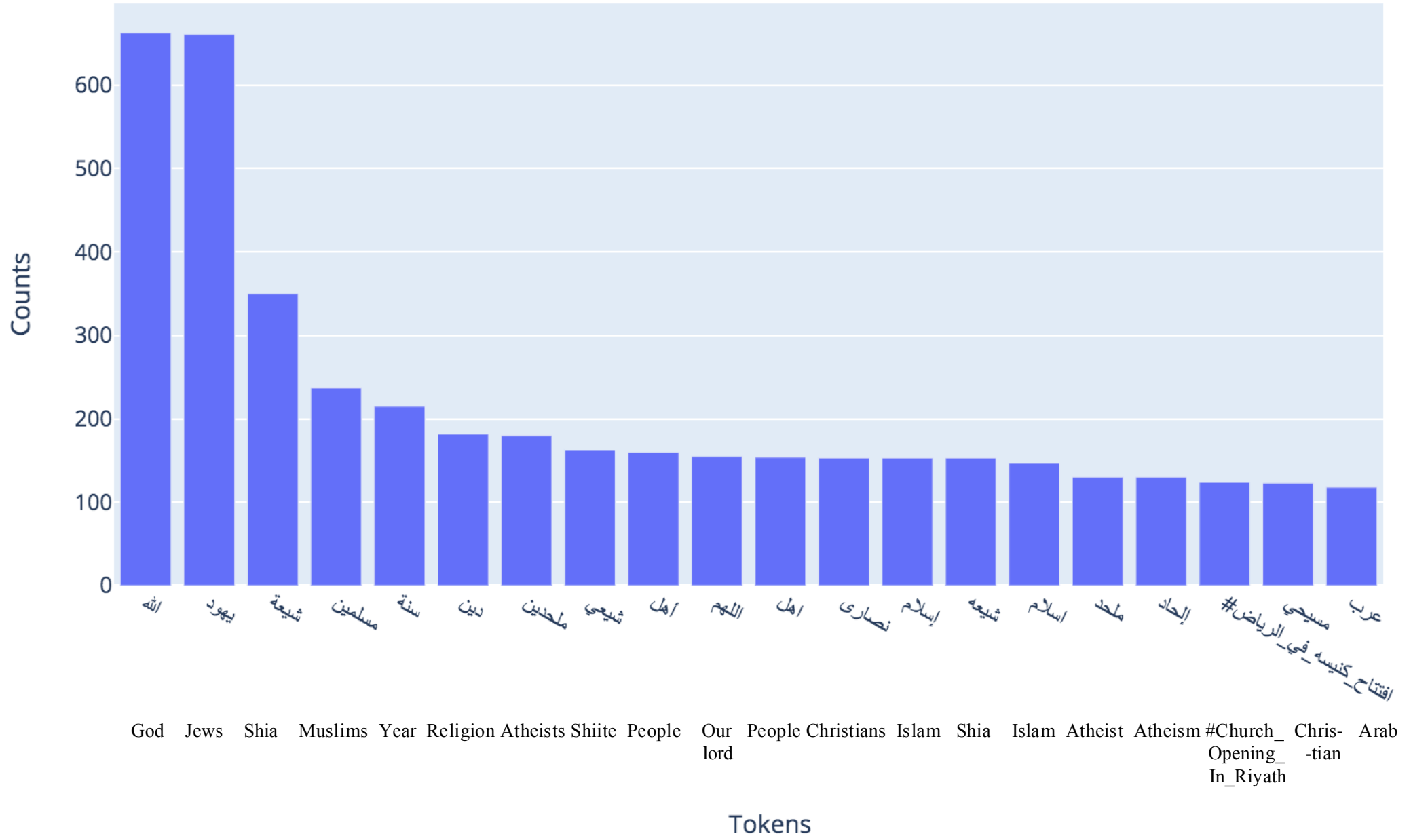

**Figure 34 Most common hate tokens in the Religious Hate Speech dataset**

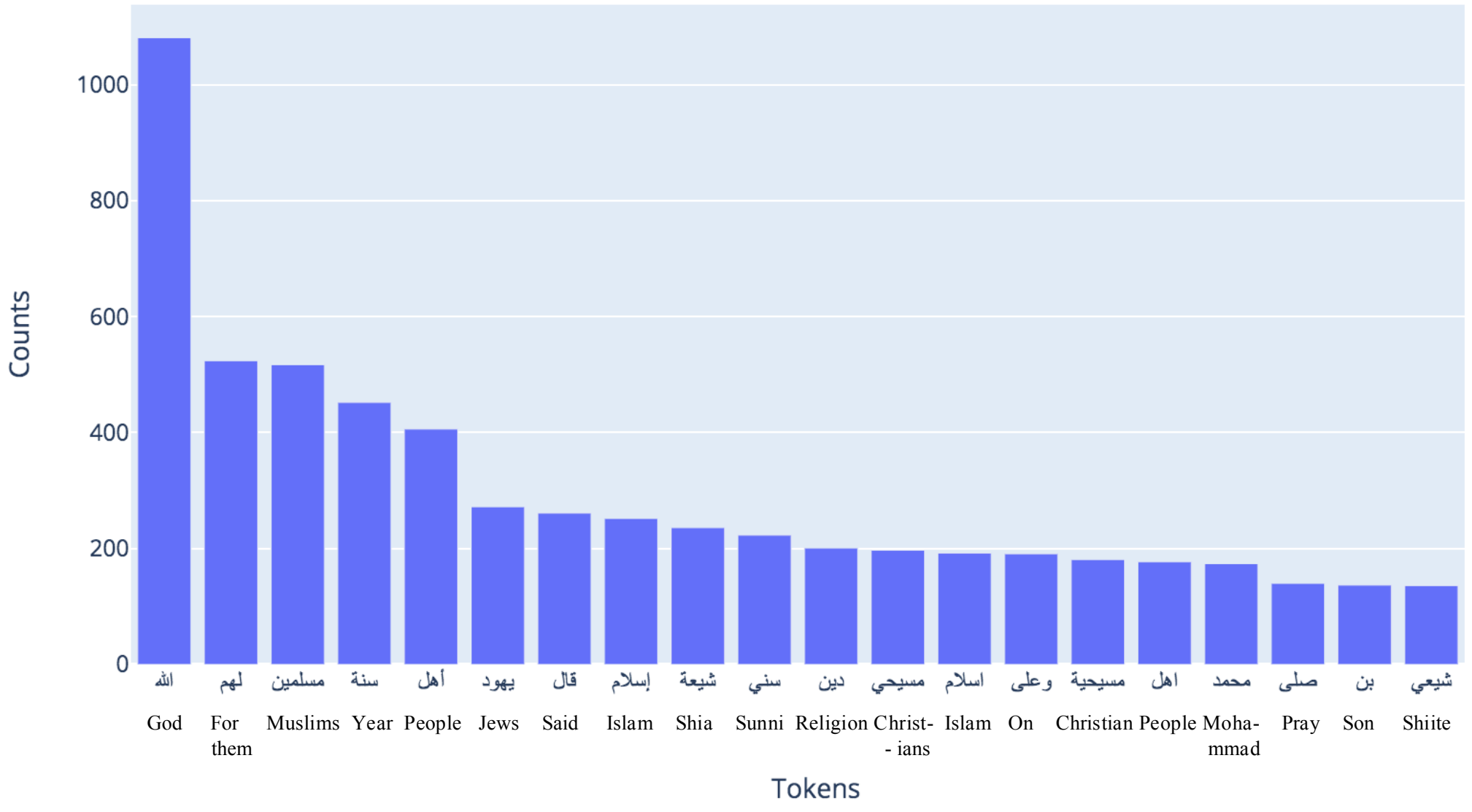

**Figure 35 Most common not hate tokens in the Religious Hate Speech dataset**

The followings are the top five distinctive words:

1. Hate: اللعنة / curse (32), العهر / immorality (20), متحدون / united (12), كلاب / dogs (12), جاهل / ignorant (11).
2. Not hate: فطرة / primitiveness (75), الإخلاص / the sincerity (64), إبراهيم / Ibrahim (61), وموتى / dead (52), ولوالدينا / our parents (50).

Very close patterns are noticed between hate and not hate classes for the results from the statistical analysis of the length of tweets and the length of tokens, except that hate tweets have more outliers for tweets length. See Figure 36 and 37 for more details.

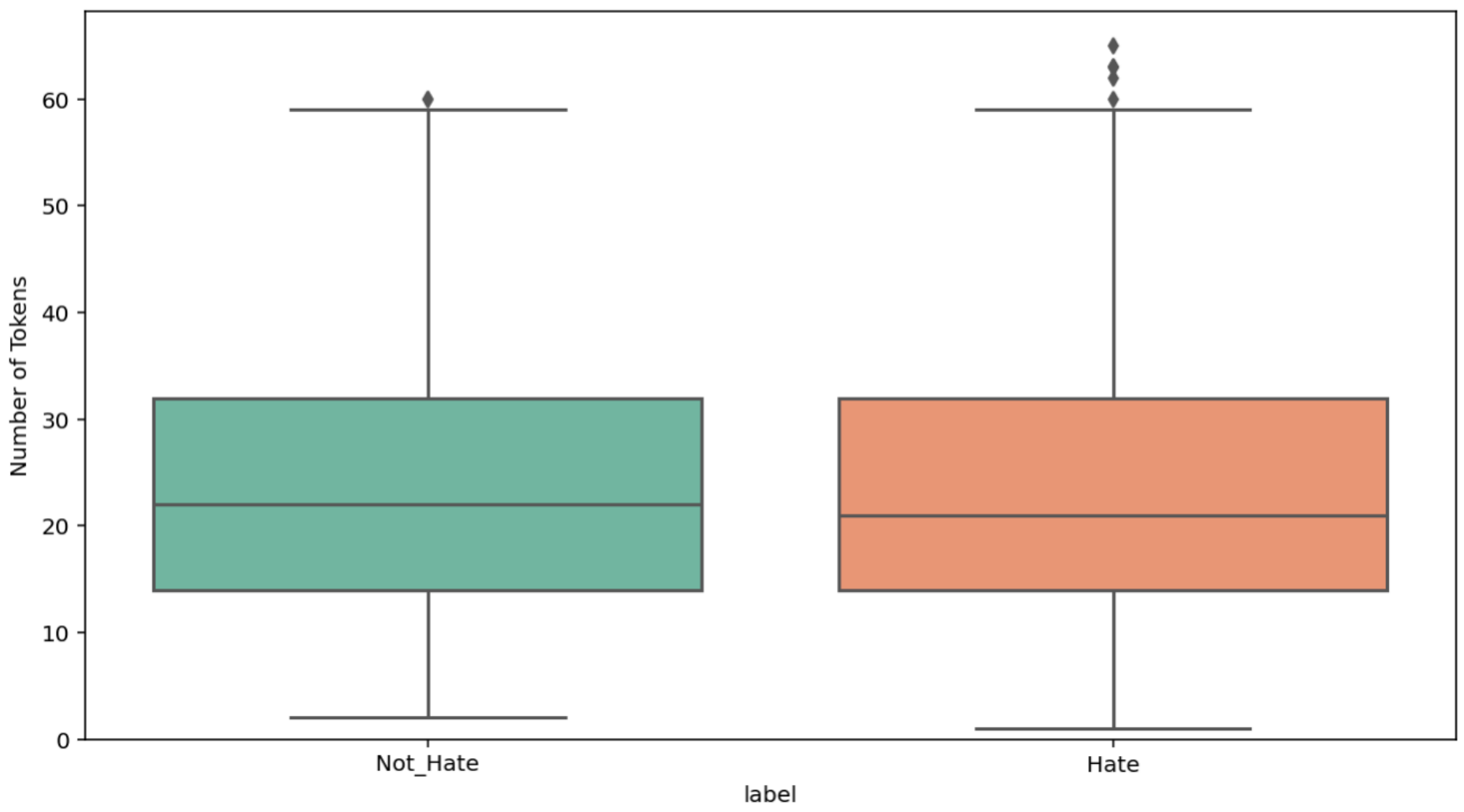

**Figure 36 Statistics of each label in the Religious Hate Speech dataset based on the number of tokens per tweet**

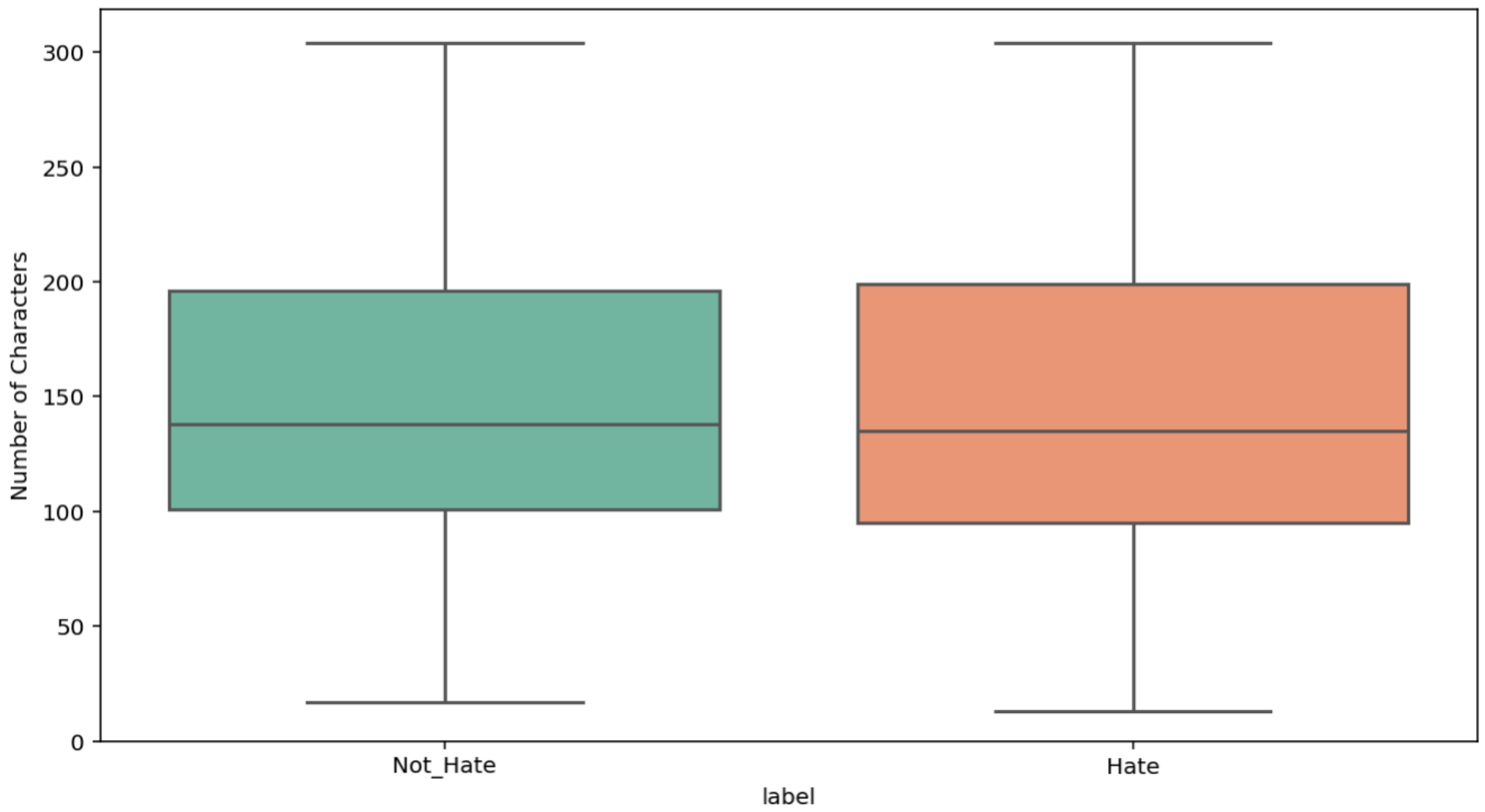

**Figure 37 Statistics of each label in the Religious Hate Speech dataset based on the number of characters per token**

Like results from statistical analysis of the length, frequencies of the top stop words report similar results for both classes (see Figures 38 and 39). The top three frequent stop words in both classes are “من / from”, “في / in”, and “على /on”, respectively.

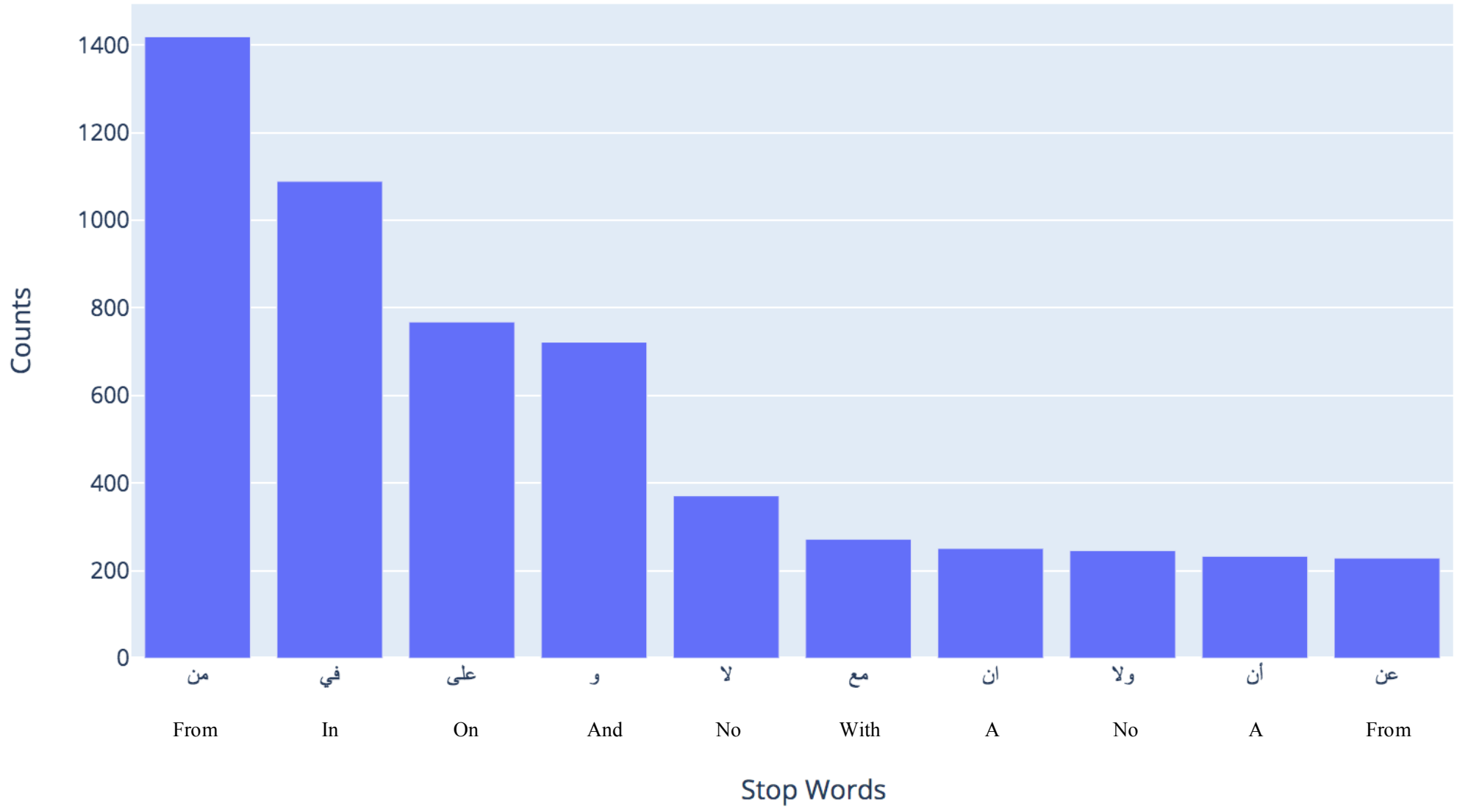

**Figure 38 Most common stop words in hate class from the Religious Hate Speech dataset**

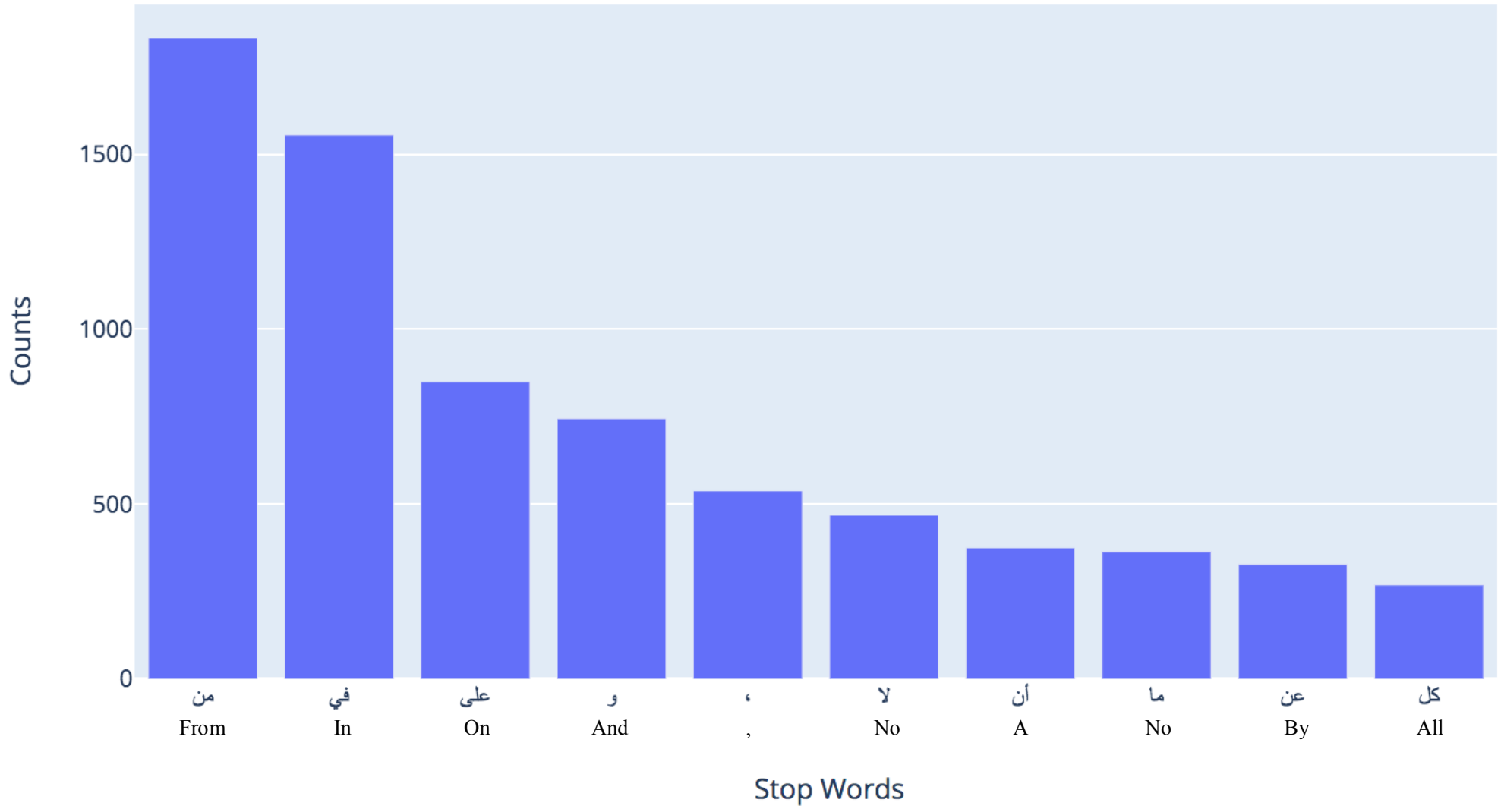

**Figure 39 Most common stop words in not hate class from the Religious Hate Speech dataset**

Results from the sentiment analysis in Figure 40 illustrate the wide spread of negative sentiment in both classes. However, hate tweets have similar number of tweets with neutral and positive sentiments. Not hate tweets contains a slightly larger number of positive sentiments than neutral sentiment.

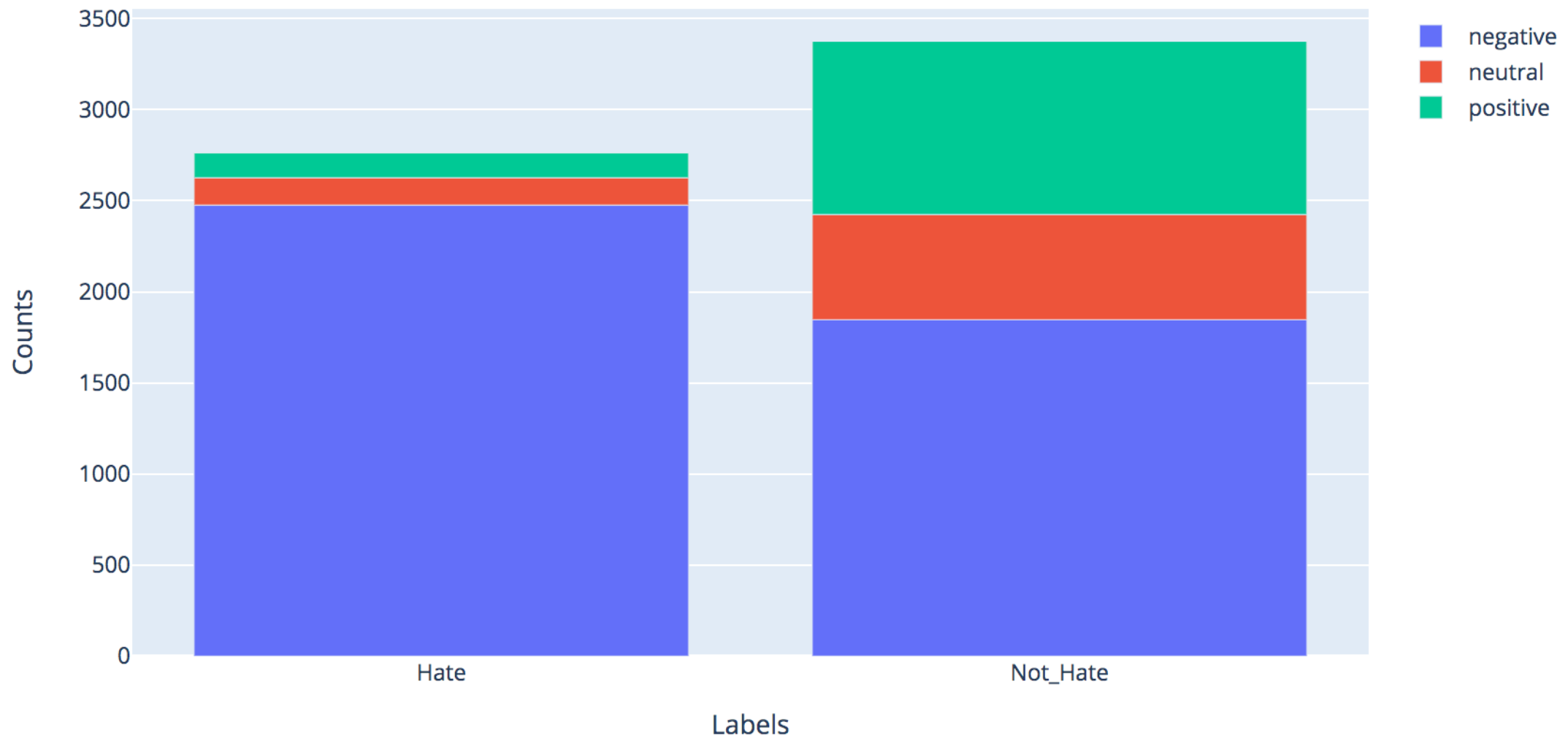

**Figure 40 Sentiment analysis based on labels for the Religious Hate Speech dataset**

Emojis analysis reports similar patterns to the previous dataset, Egyptian Tweets Dataset, the first top frequent emoji in both classes is the face with tears of joy, "😂". Hate tweets show Qatar flag emoji, "🇶🇦", as the second one and the backhand index pointing down, "👇", as the third one. Not hate tweets record the red rose emoji, "🌹", as the second one and the black heart, "🖤", as the third one. Figure 41 and 42 show more details about the use of emojis among the tweets in both classes.

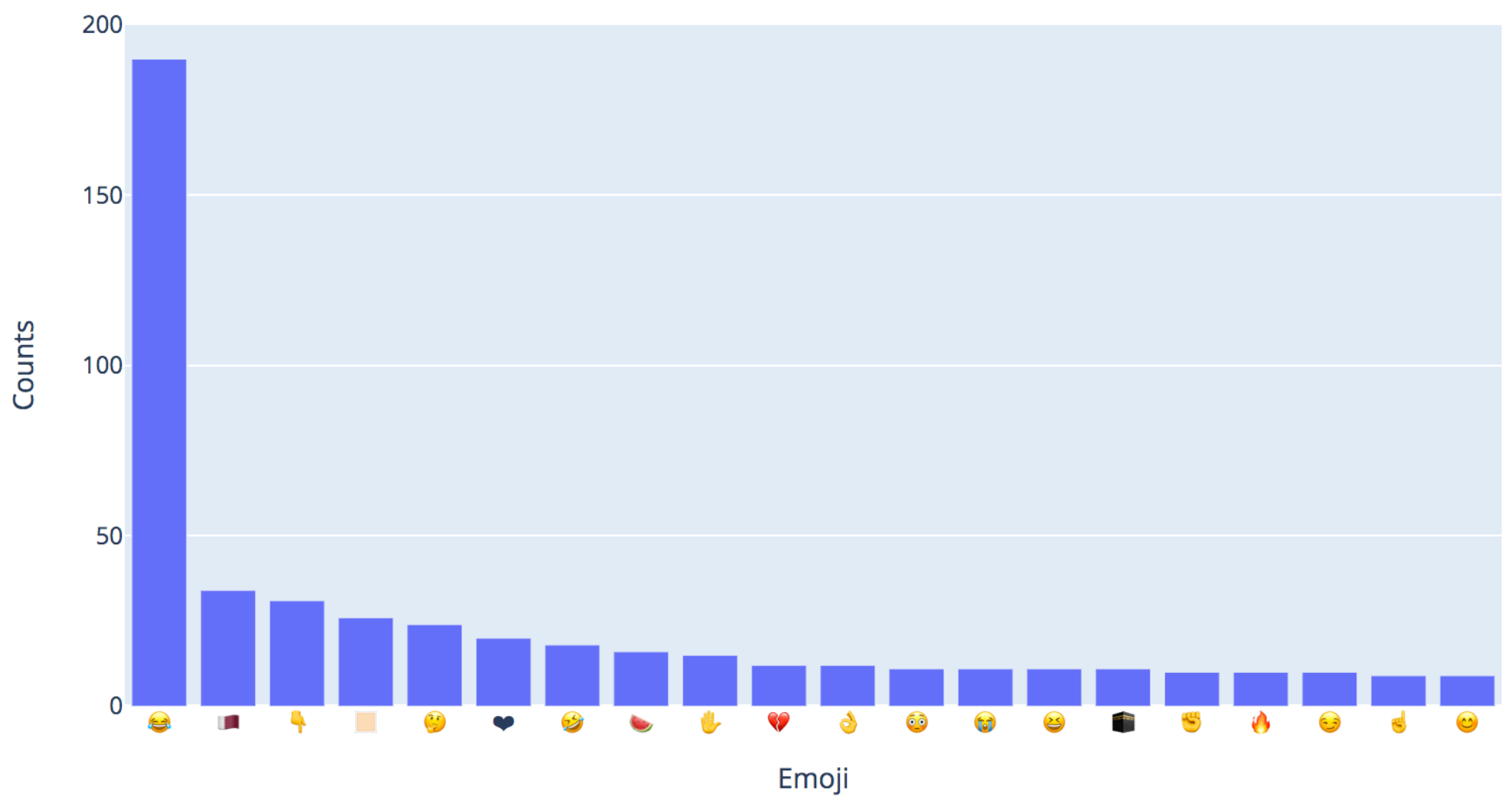

**Figure 41 Most common hate emojis in the Religious Hate Speech dataset**

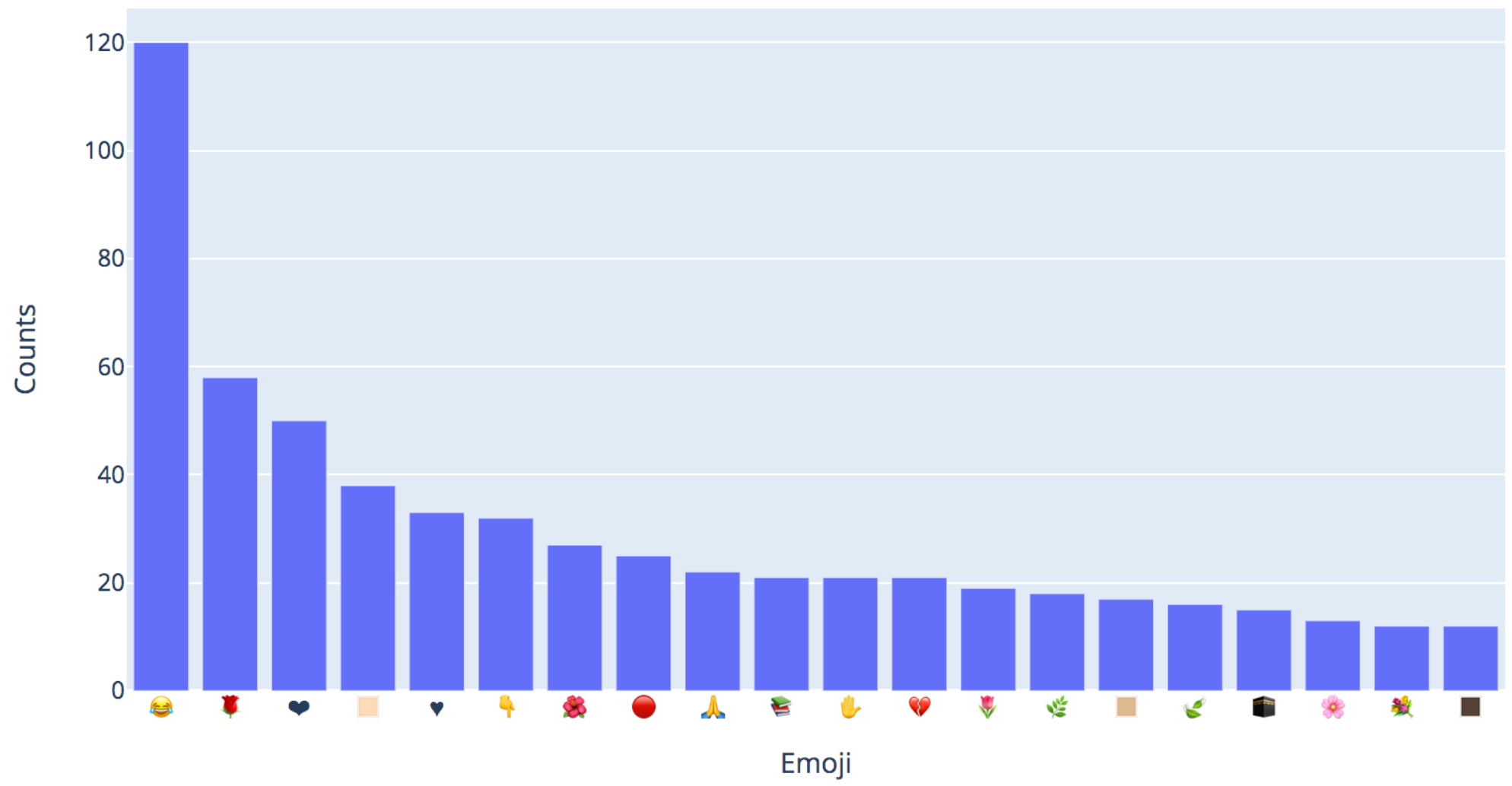

**Figure 42 Most common not hate emojis in the Religious Hate Speech dataset**

Figure 43 and 44 illustrate the similarities between the two classes in term of using punctuation. In both classes the top three frequently used punctuation are exactly the same; “.”, “/”, and “#”.

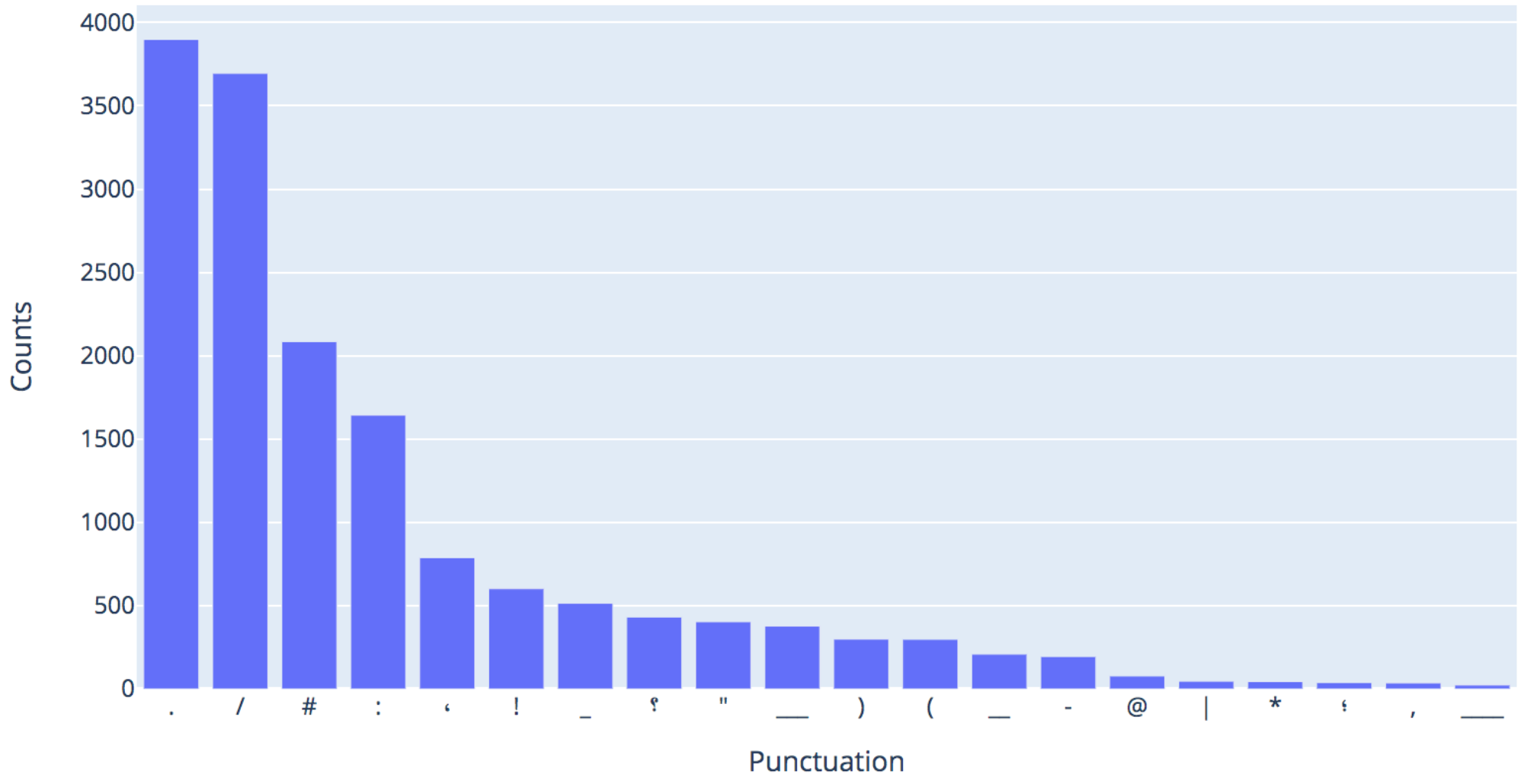

**Figure 43 Most common hate punctuation in the Religious Hate Speech dataset**

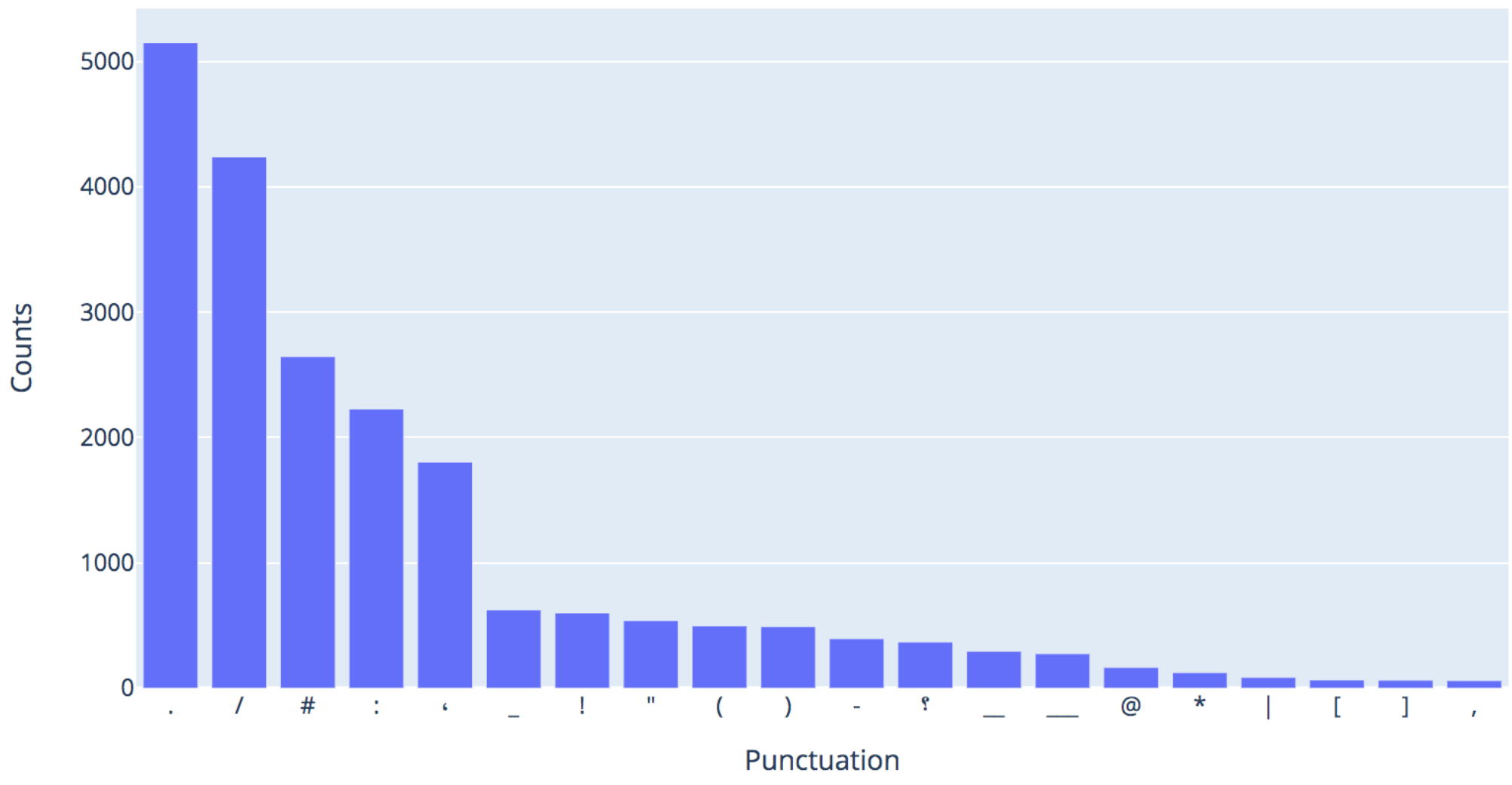

**Figure 44 Most common not hate punctuation in the Religious Hate Speech dataset**

4) YouTube Comment Dataset:

Alakrot, Murray, and Nikolov (2018) extract a comments dataset from a set of controversial YouTube channels that contains a total of 15,050 comments. From Figure 45, it can be seen that classes are imbalanced with 9,237 not offensive comments and 5,813 offensive comments.

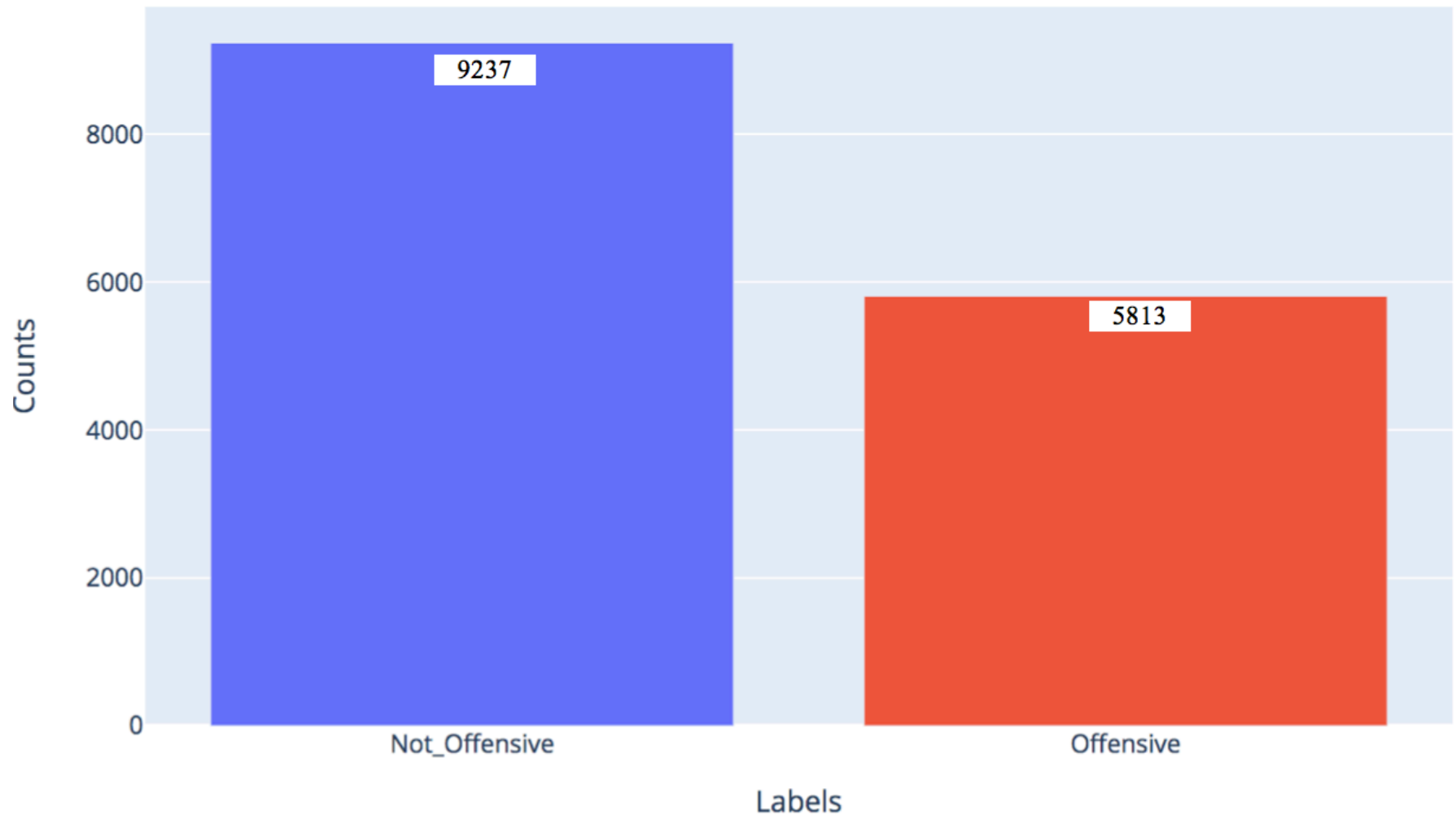

**Figure 45 Class distribution for the YouTube Comments dataset**

The dataset has two comments that have been repeated for 11 times. One of them is offensive and has only one word as the following:

عاهرة

Translation: Whore

The second comment is a not offensive as the following:

مضغوطين من القيصر

Translation: Forced by the Cesar

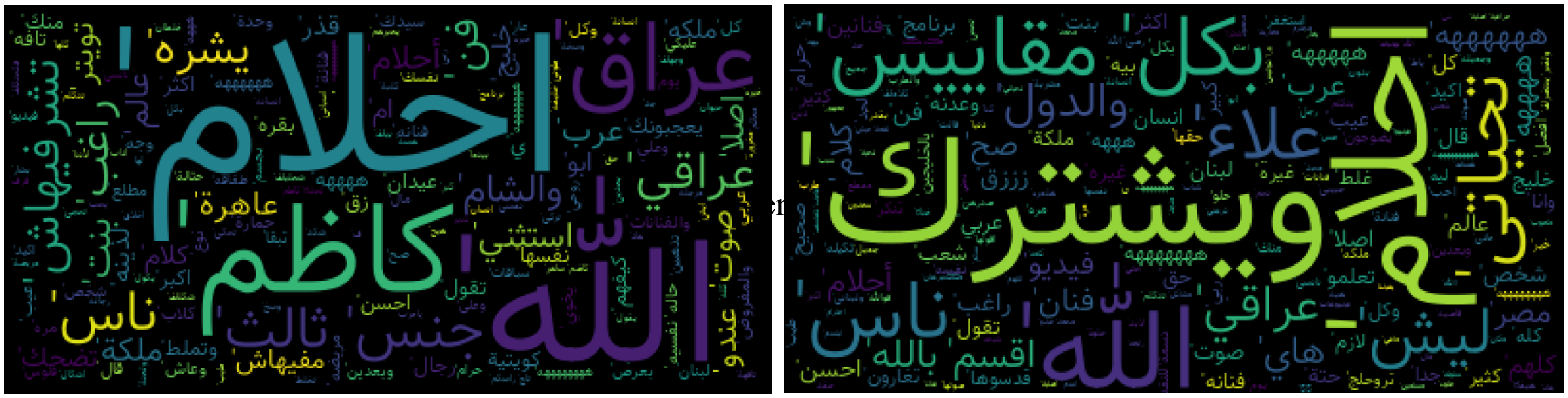

(a) Offensive (b) Not offensive

**Figure 46 The word cloud of the YouTube Comments dataset (a. offensive, b. not offensive)**

From Figure 46 for the word cloud, it can be seen that the word “احلام / Ahlam” and “الله / God” occurred in both classes, but they are more frequent within offensive comments than they are within not offensive comments. It can also be seen that the words “كاظم / Kadhim” and “عراق / Iraq” are related to offensive comments, while the words “مقاييس / scales” and “يشترك / share” are related to not offensive comments.

Most frequent tokens demonstrate the similarities between the two classes. The first three tokens from Figure 47 and 48 are the same but with different orders. These top three tokens are “كاظم / Kadhim”, “الله / God”, and “احلام / Ahlam”.

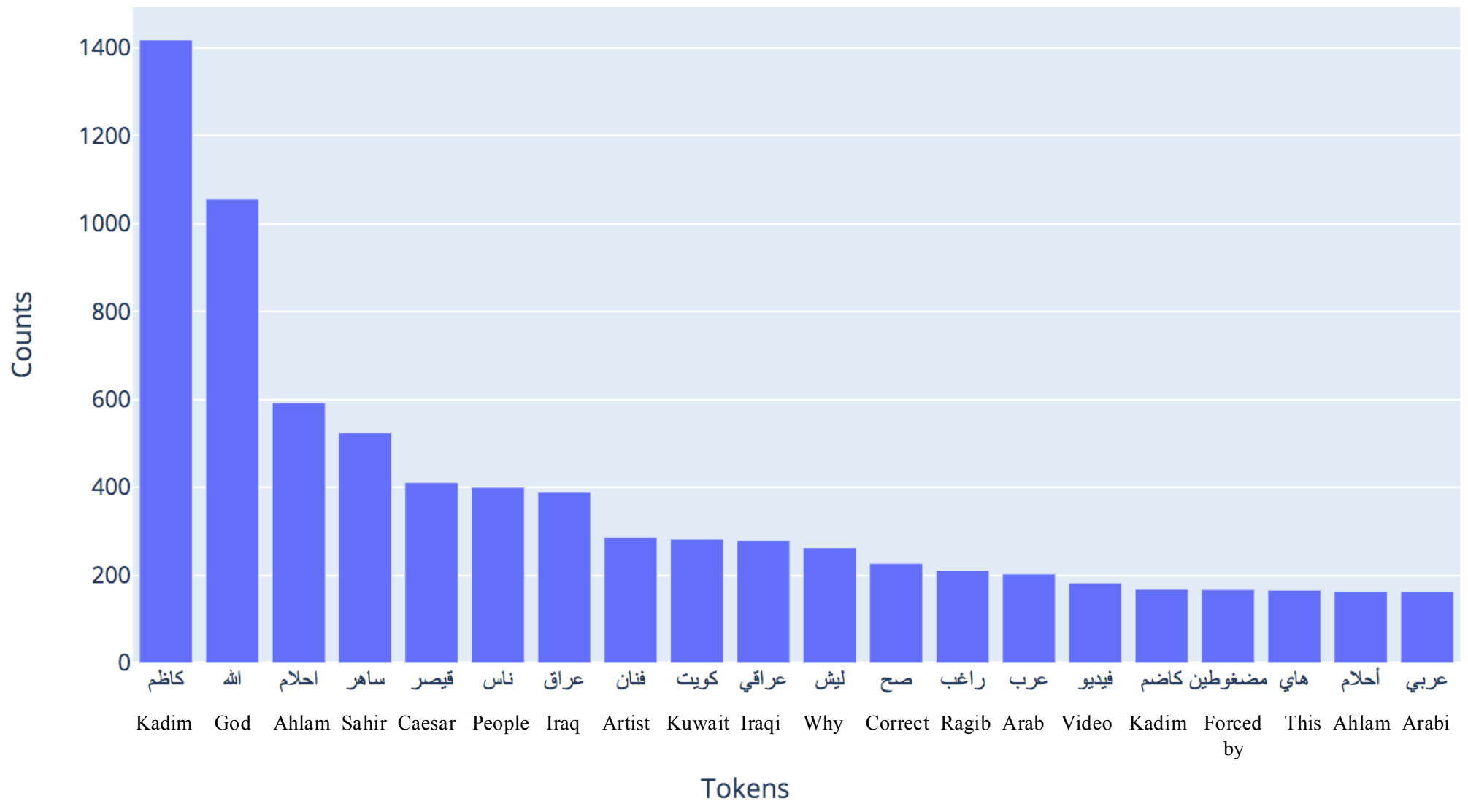

**Figure 47 Most common not offensive tokens in the YouTube Comments dataset**

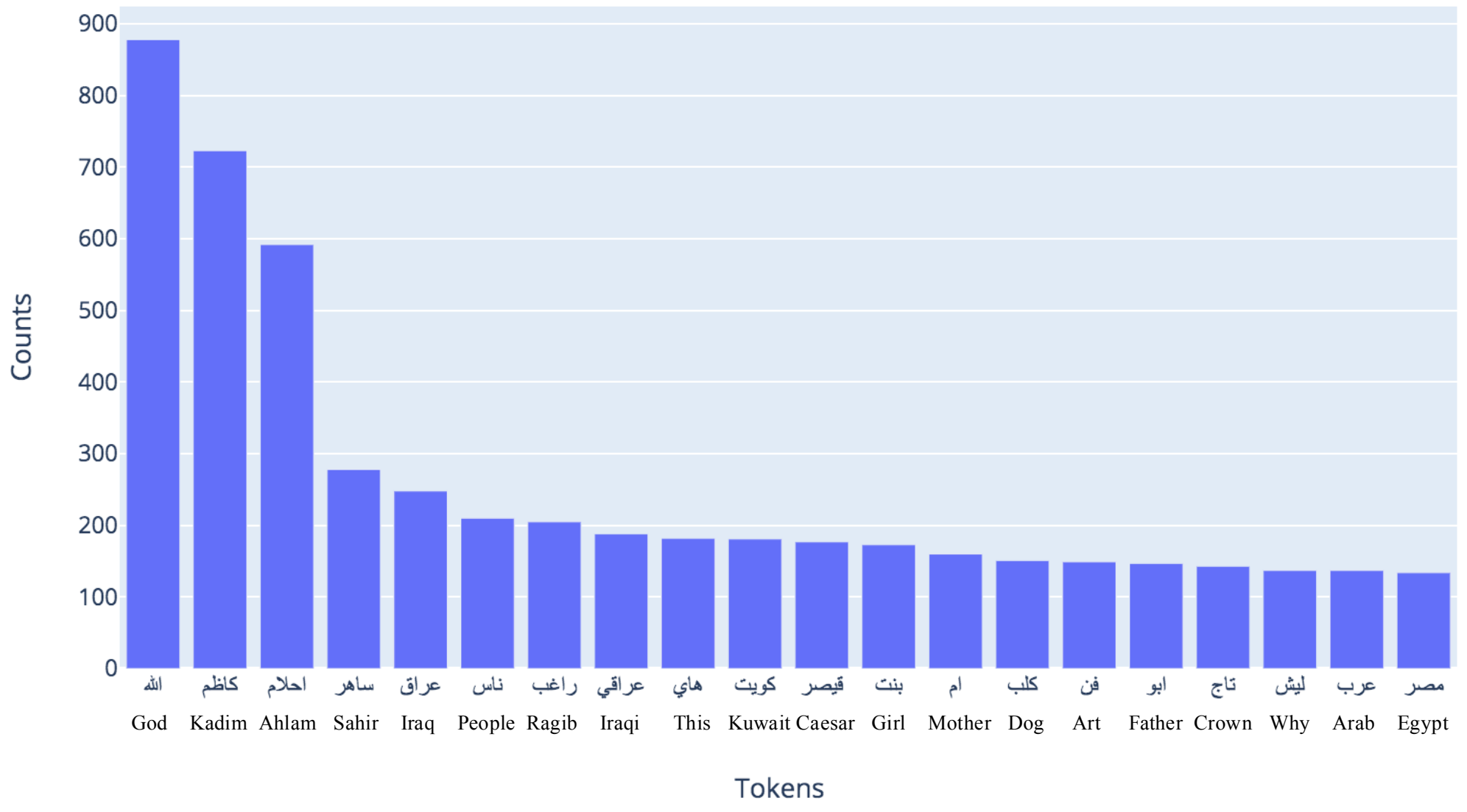

**Figure 48 Most common offensive tokens in the YouTube Comments dataset**

Distinctive words for each class are mentioned below:

1. Not offensive: بقناتي / in my channel (27), مها / Maha (23), يهديك / give you (20), يهدينا / give us (17), مسلمة / Muslim (17).
2. Offensive: زبالة / trash (70), زبالة / trash (62), خرة / shit (61), خره / shit (59), تفو / spit (55).

On general, the dataset contains very close statistical properties for both offensive and not offensive classes. Figures 49 and 50 illustrates the size of the dataset in more details. As can be noticed, there are several outliers in both figures; However, offensive comments have larger outlier values.

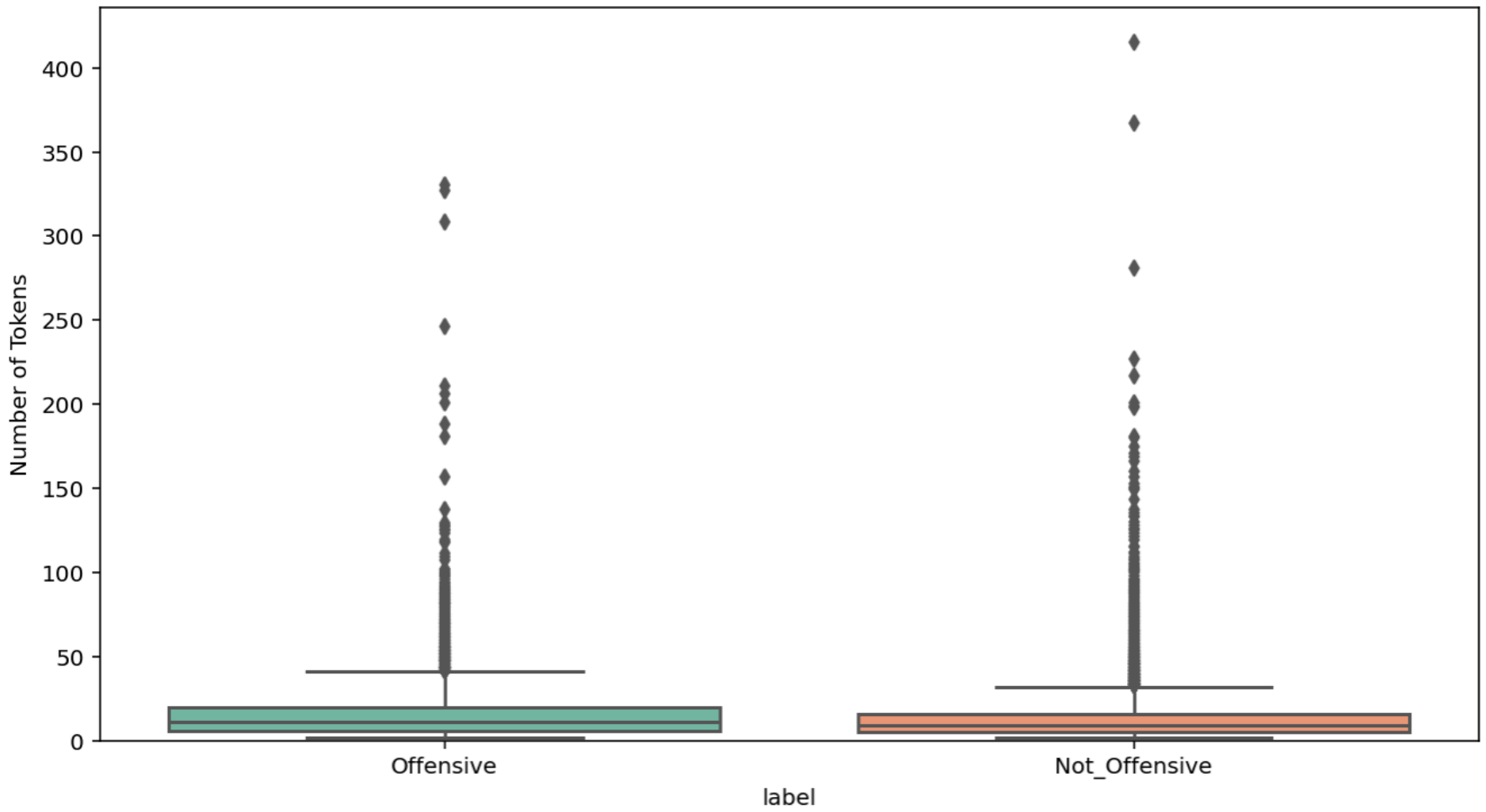

**Figure 49 Statistics of each label in the YouTube Comments dataset based on the number of tokens per tweet**

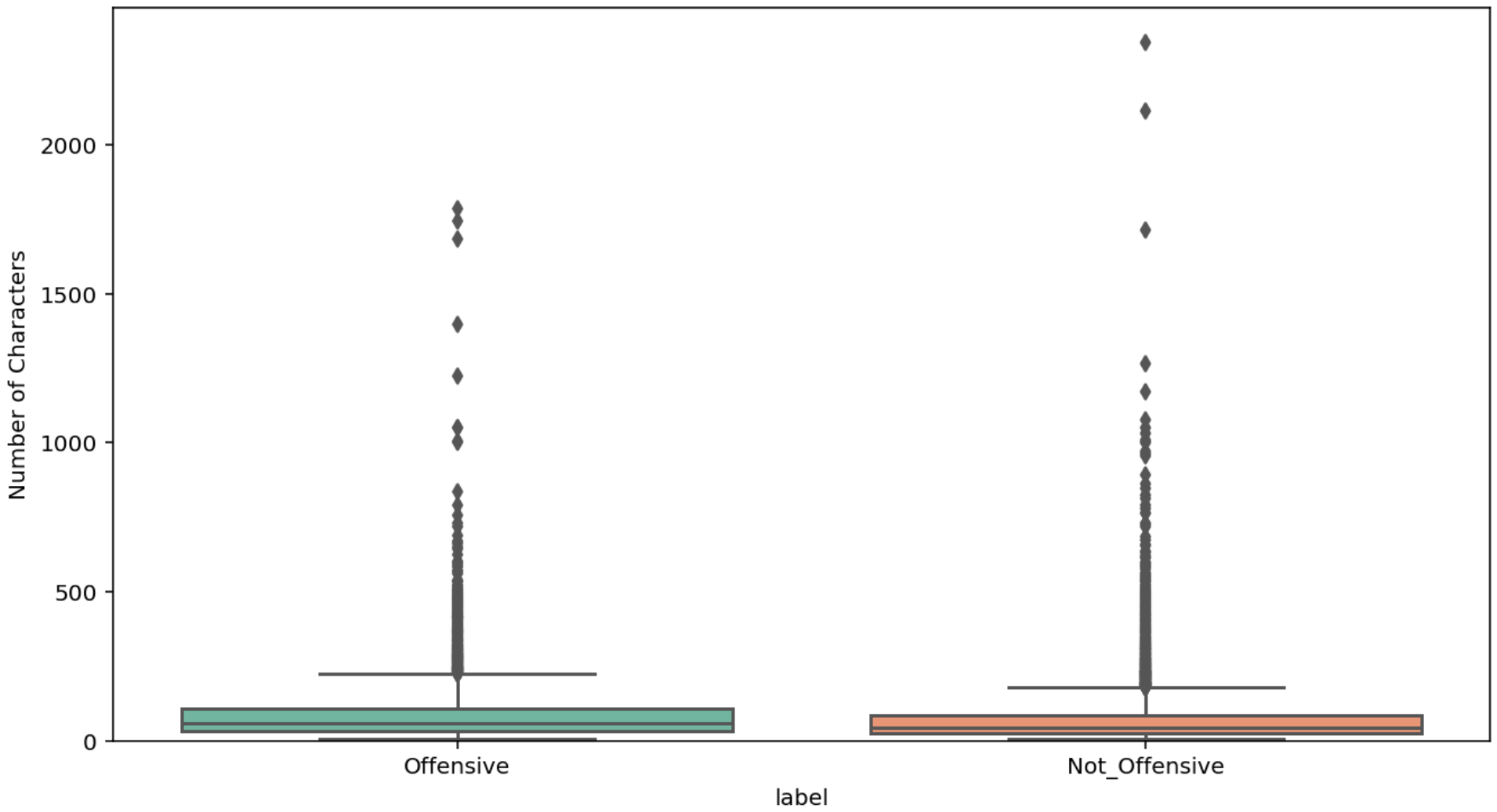

**Figure 50 Statistics of each label in the YouTube Comments dataset based on the number of characters per token**

The top frequent stop word is "من / from" in both classes. From Figure 51, the second stop word for not offensive comments is "و / and" and the third is "في / in". Figure 52 for offensive comments shows "يا / you" as the second stop word and "و / and" as the third one.

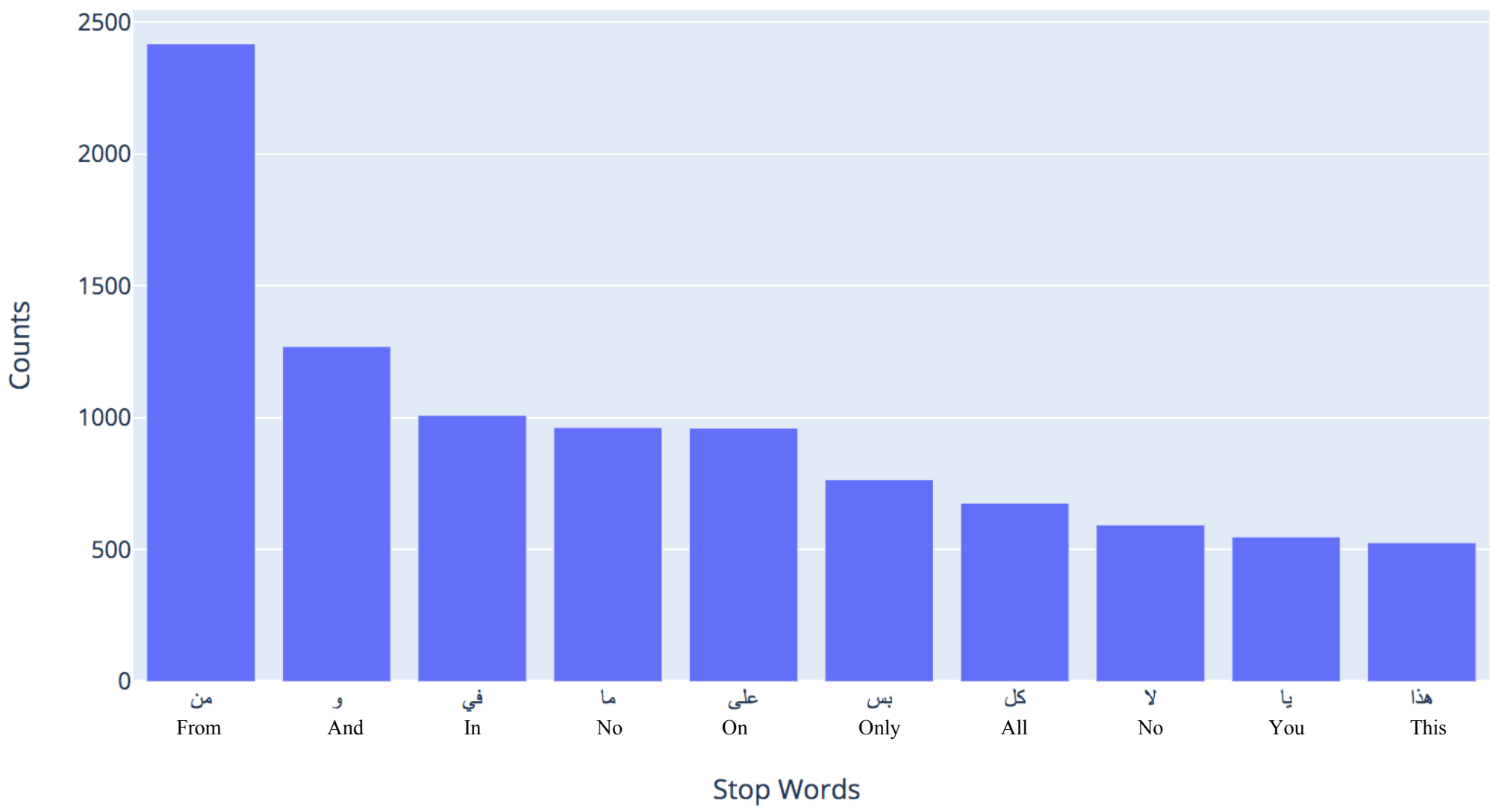

**Figure 51 Most common stop words in not offensive class from the YouTube Comments dataset**

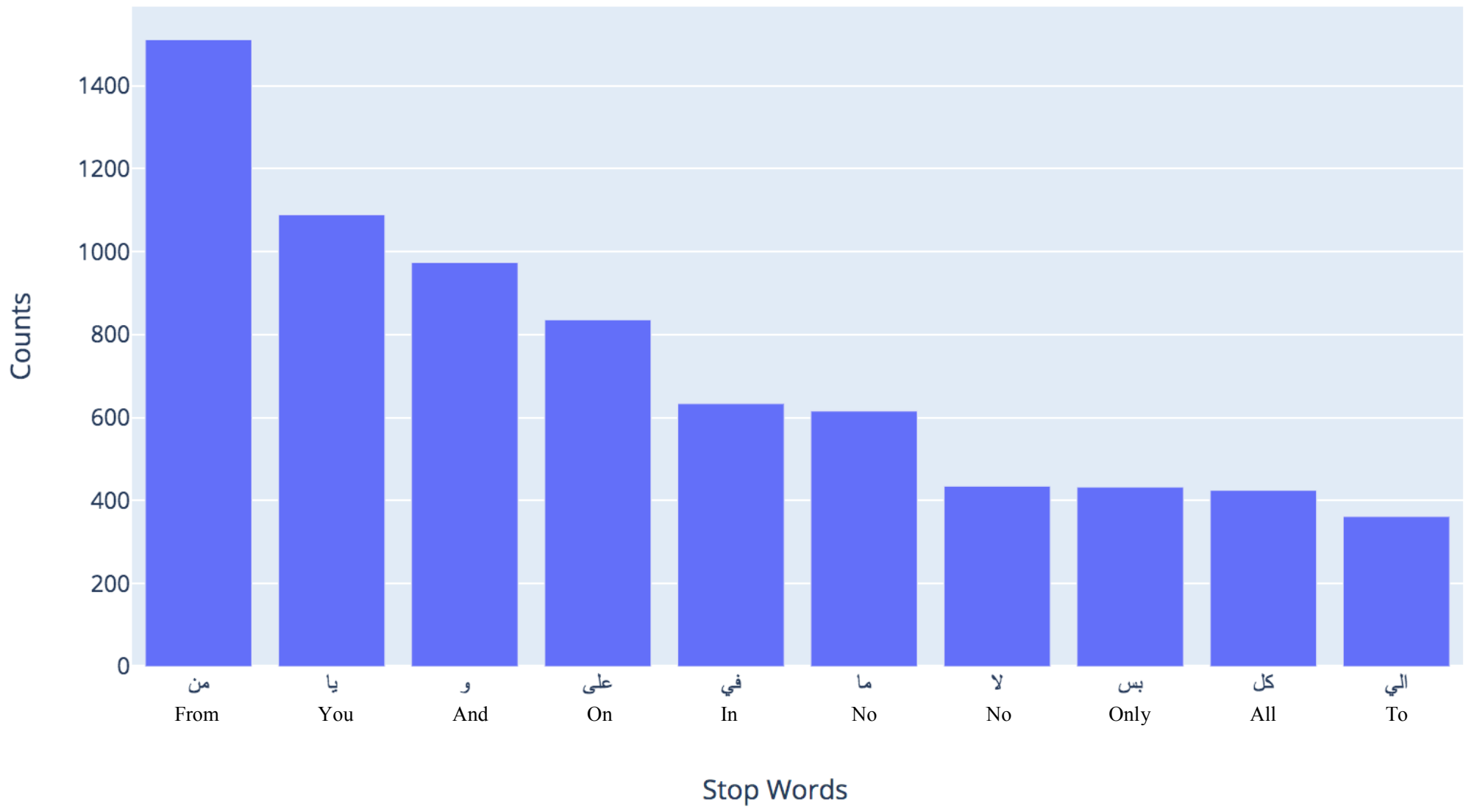

**Figure 52 Most common stop words in offensive class from the YouTube Comments dataset**

Sentiment analysis results illustrates distribution of sentiments among the classes. It can be noticed from Figure 53 that in both classes, the majority of comments are negative followed by positive, and the least occurred sentiment is the neutral.

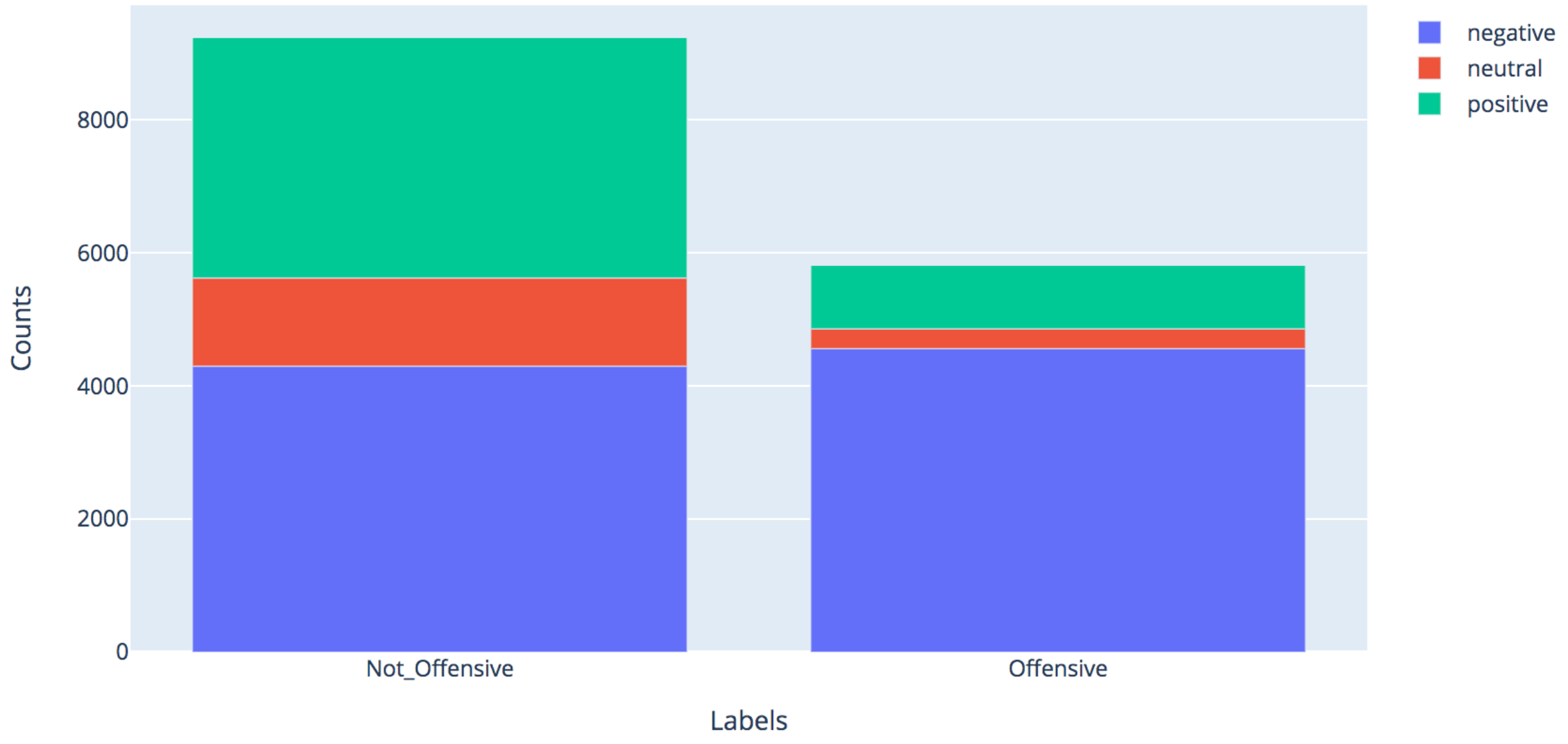

**Figure 53 Sentiment analysis based on labels for the YouTube Comments dataset**

This dataset also highlights the same top emojis of the previously discussed dataset in both classes, which is the face with tears of joy emoji, "😂". Not offensive comments at Figure 54 reports the black heart emoji, "🖤", second and the smiling face with heart-eyes emoji, "😍", third. Figure 55 shows thumbs down emoji, "👎", as the second frequent emoji among offensive comments and the pensive face emoji, "😔", as the third one.

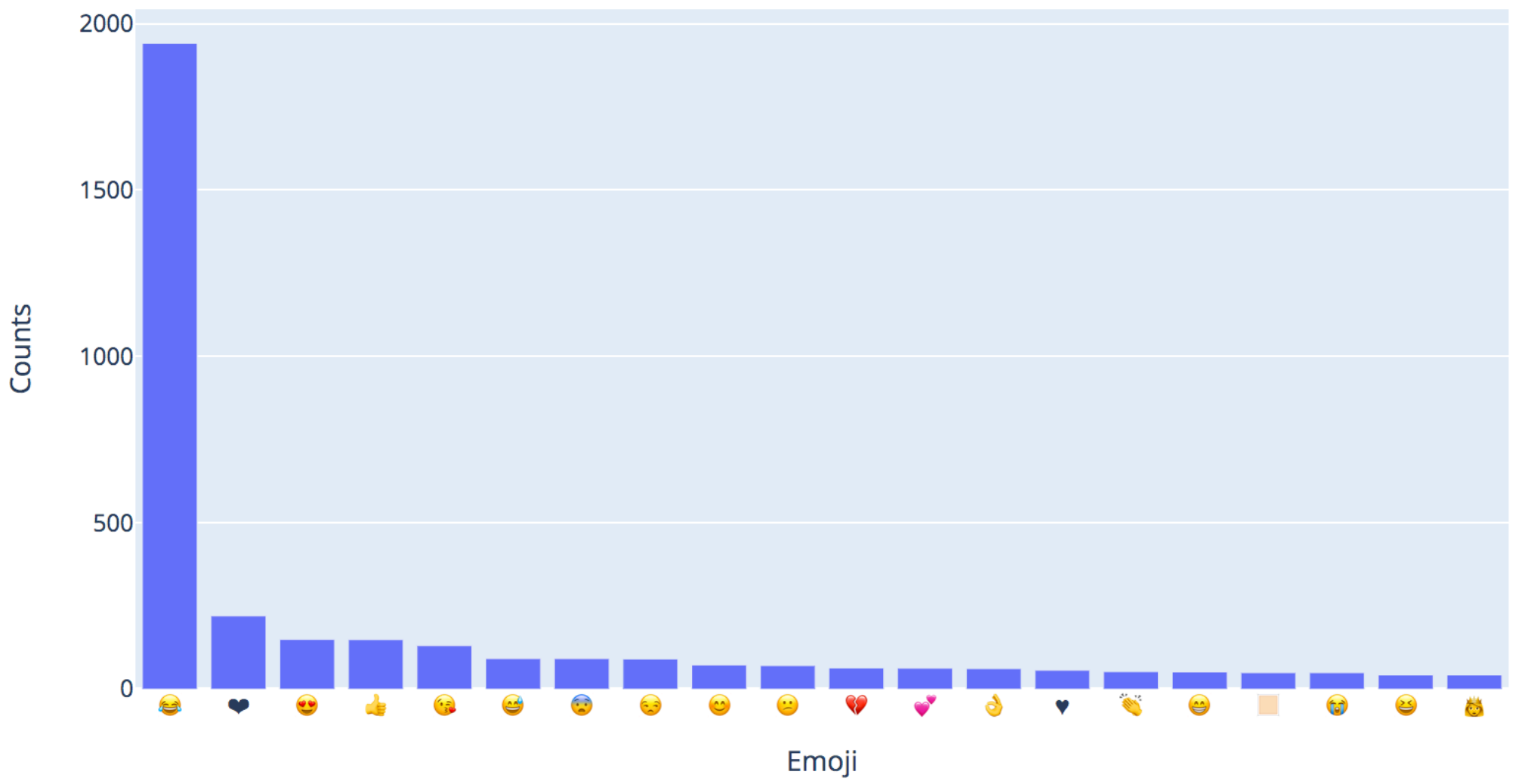

**Figure 54 Most common not offensive emojis in the YouTube Comments dataset**

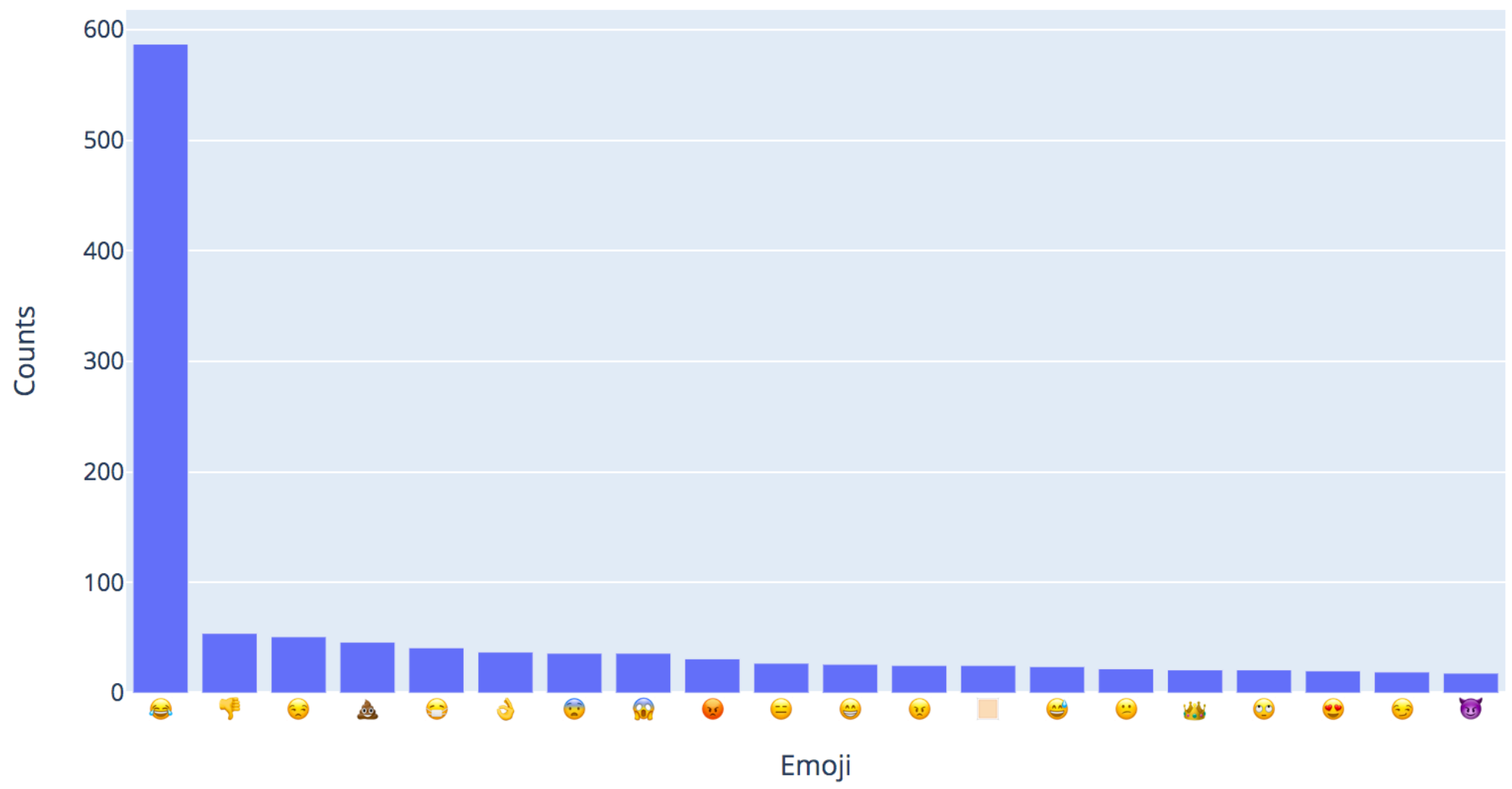

**Figure 55 Most common offensive emojis in the YouTube Comments dataset**

The bar charts in Figures 56 and 57 plot the same punctuation for the first and second top counts in both classes, which are “.” and “؟”. However, for not offensive comments, the third punctuation is “+”, while for offensive comment, it is “,”.

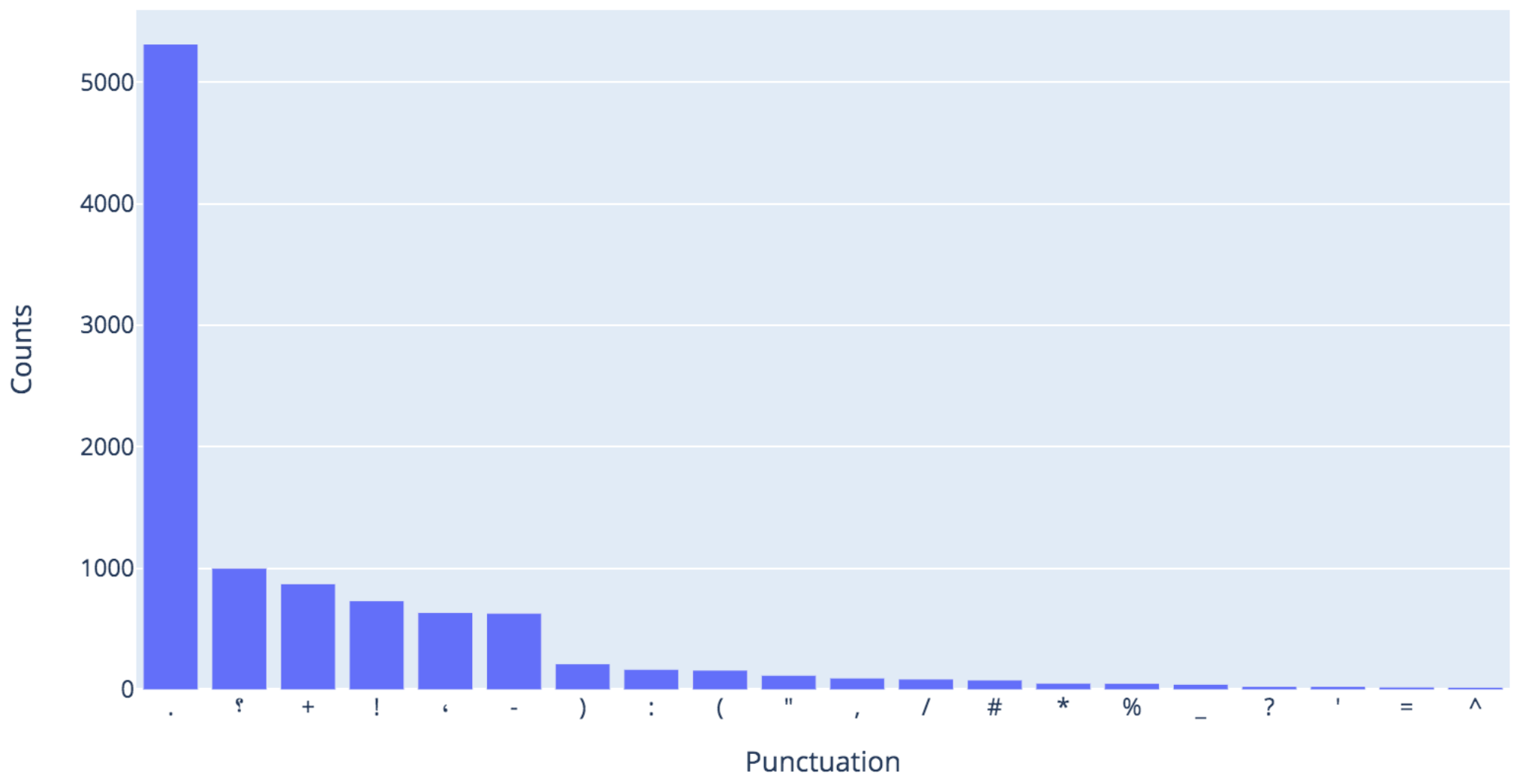

**Figure 56 Most common not offensive punctuation in the YouTube Comments dataset**

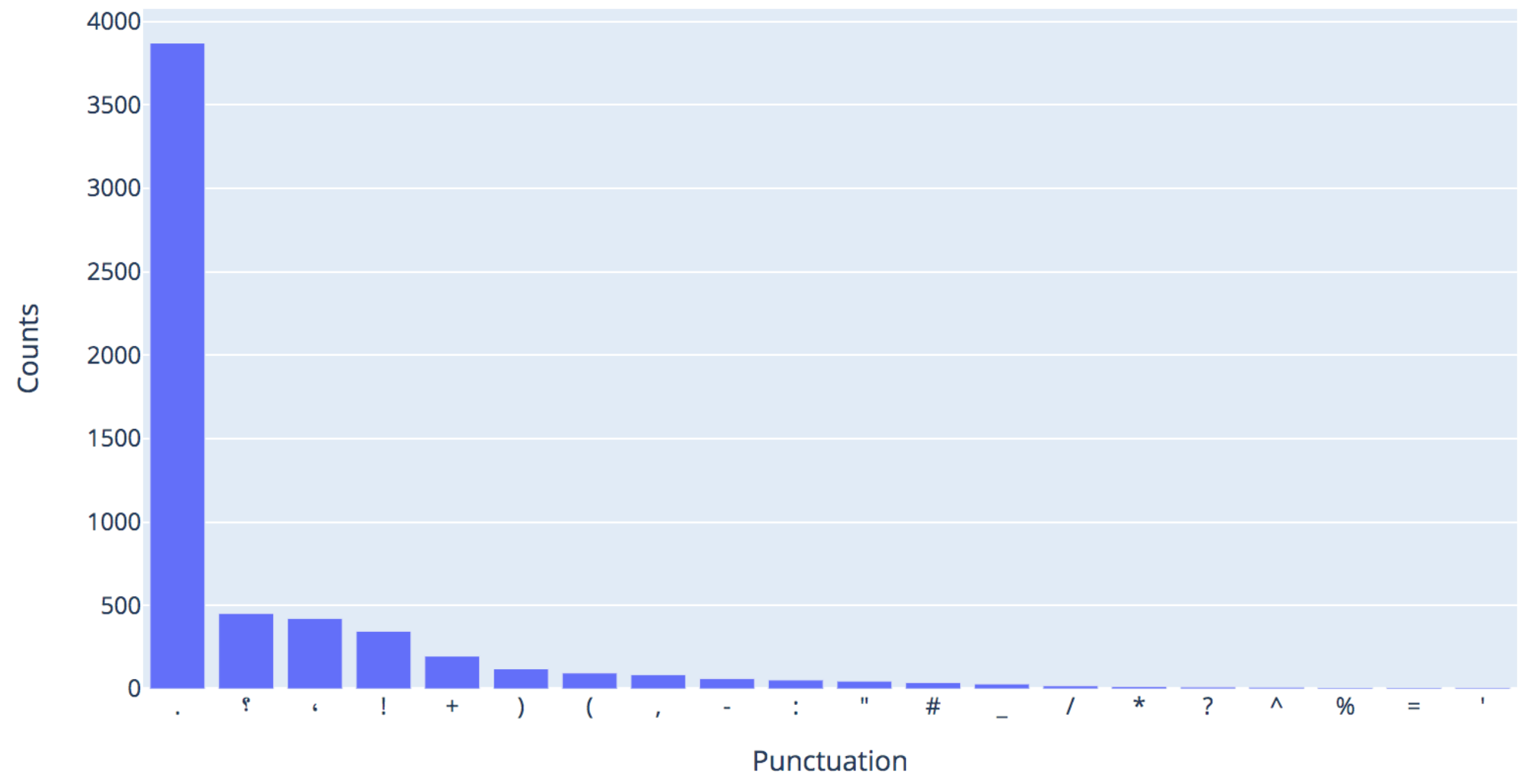

**Figure 57 Most common offensive punctuation in the YouTube Comments dataset**

5) Levantine Twitter Dataset for Hate Speech and Abusive Language (L-HSAB):

The L-HSAB dataset is a dialectic specific dataset that contains 5,846 Levantine tweets (Mulki et al., 2019). The class distribution is shown in Figure 58. It has three class as the following: hate = 468 tweets, abusive = 1,728 tweets, and normal = 3,650 tweets. Tweets were preprocessed to remove some characters, such as @, RT, and #. Multiple identical repetitions of tweets were found, for instance the following normal tweet is repeated for three times tweets:

من مقابلة جبران باسيل مع سيإنإن أمن إسرائيل هو حق

Translation: From Gebran Bassil's interview with CNN, Israel's security is a right

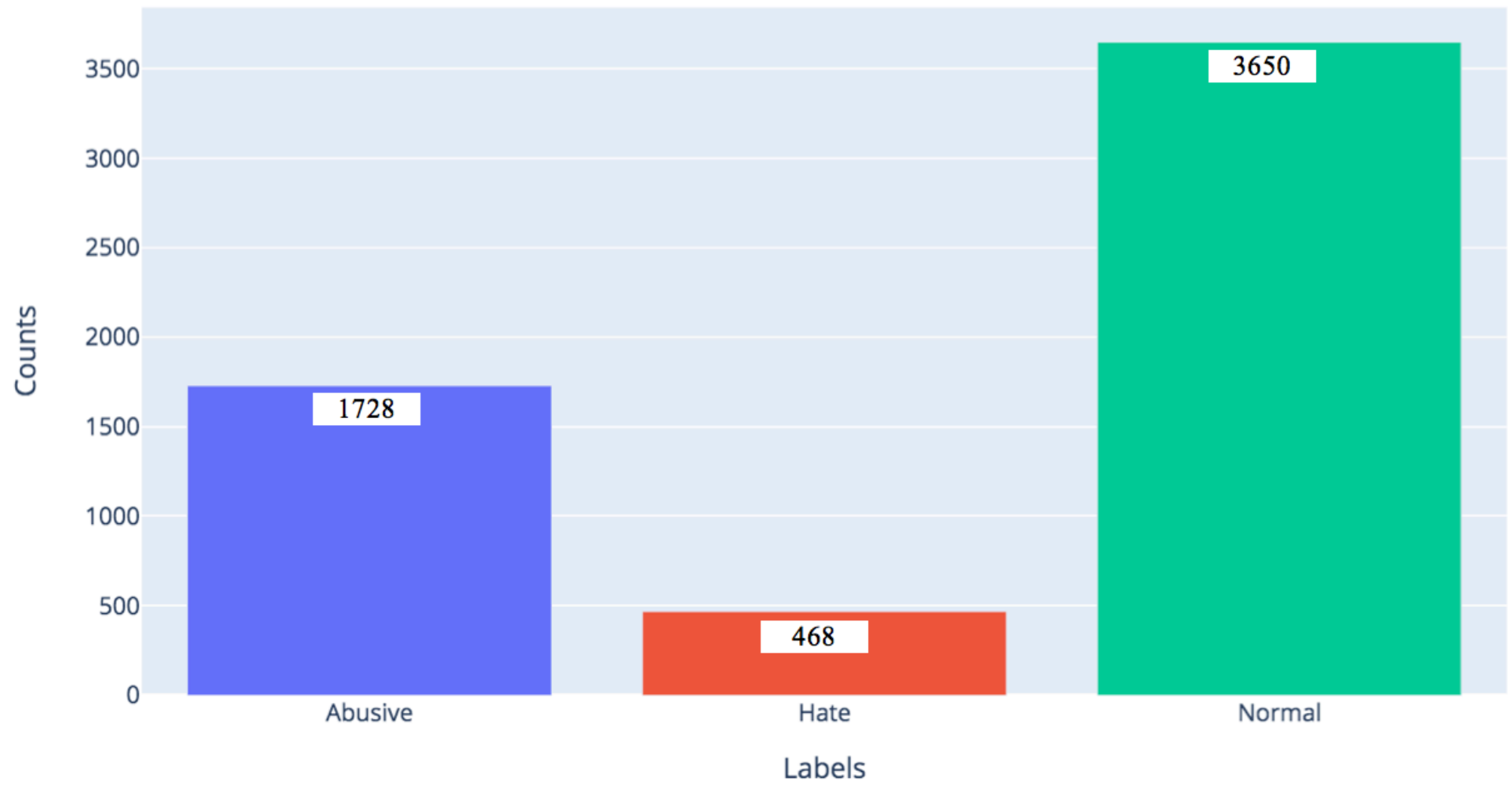

**Figure 58 Class distribution for the L-HSAB dataset**

Figure 59 show the word cloud graphs. The word cloud graphs demonstrate the importance of cleaning text to remove stop words, which show stop words as the most frequent words in the tweets such as “ كانت/was”, “ من/from”, “ على /above”. Moreover, the word cloud graphs reveal that the normal class has more diverse vocabulary than the others. This variation in the vocabulary size could be related to the variation in classes distribution in the dataset. The preposition “ يا / you” is occurring very frequent in abusive and hate tweets than in normal tweets.

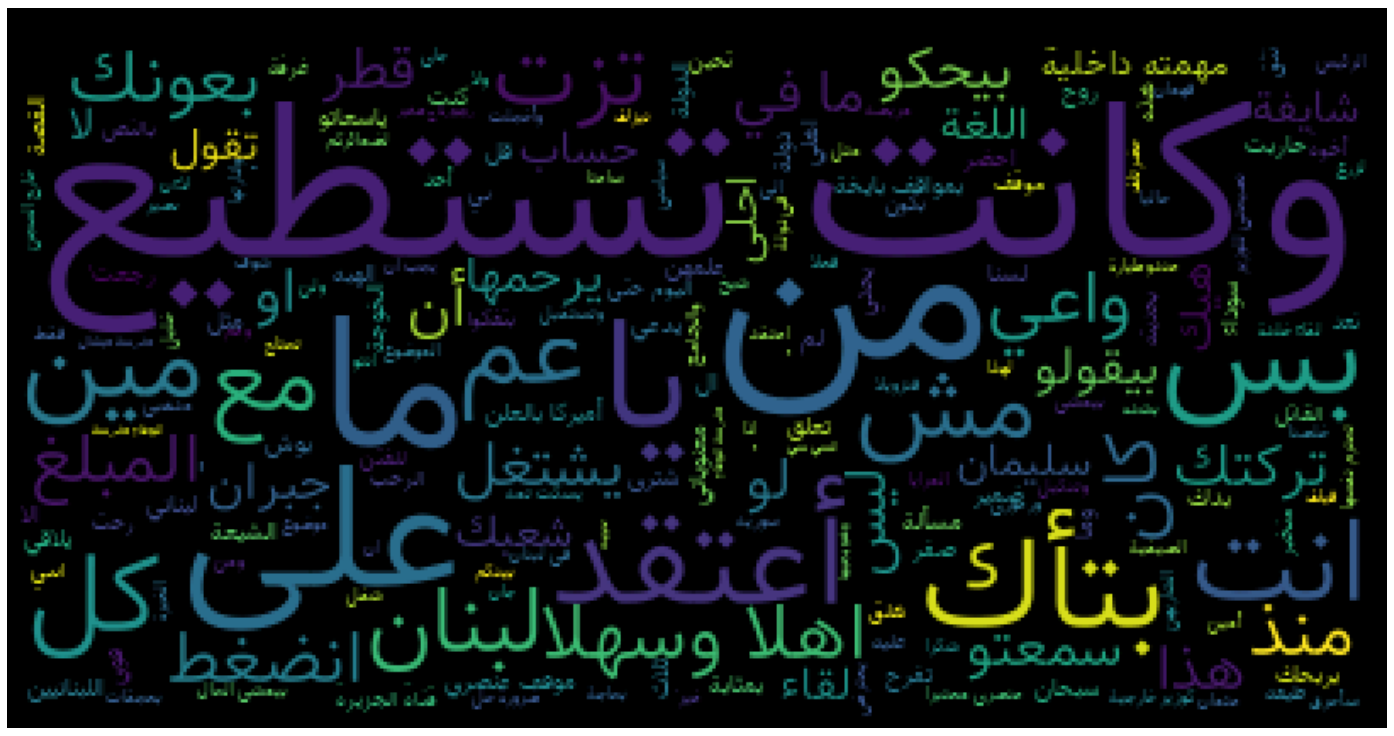

(b) Normal

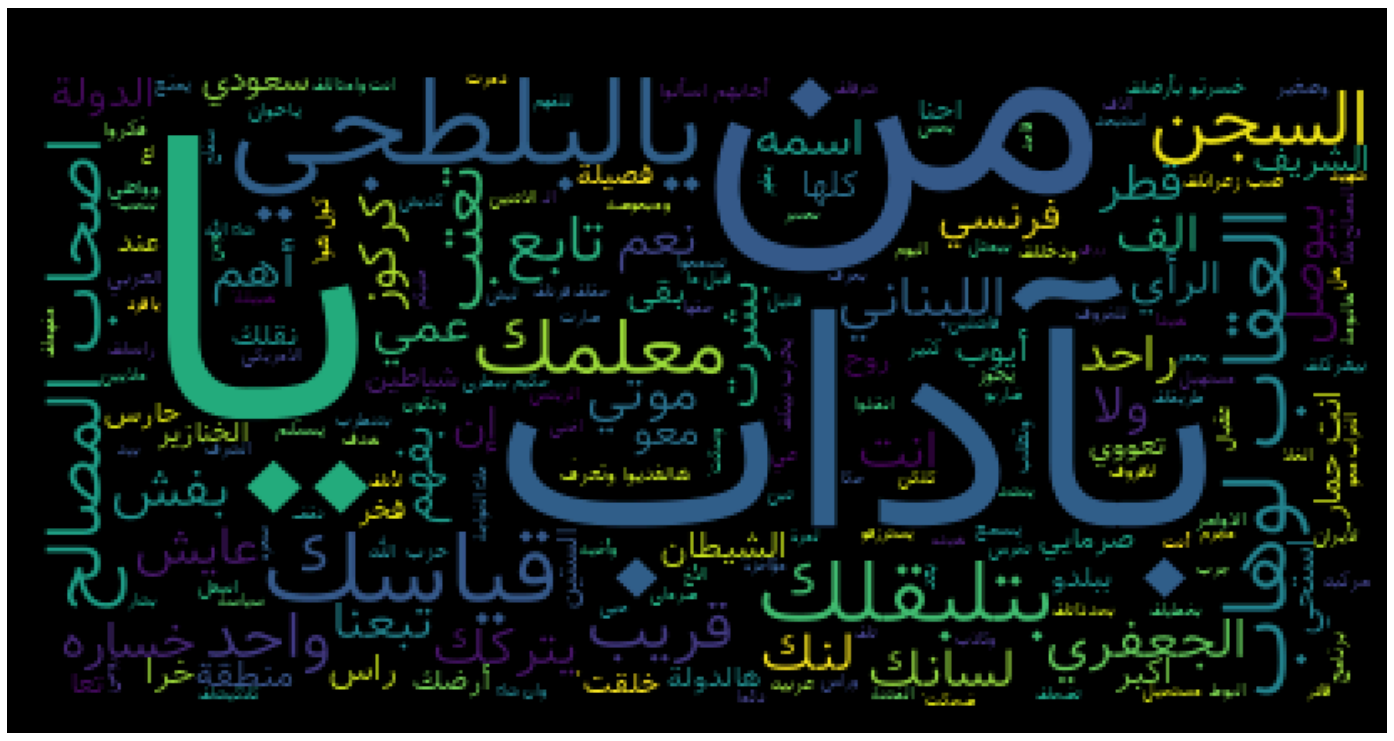

(a) Abusive

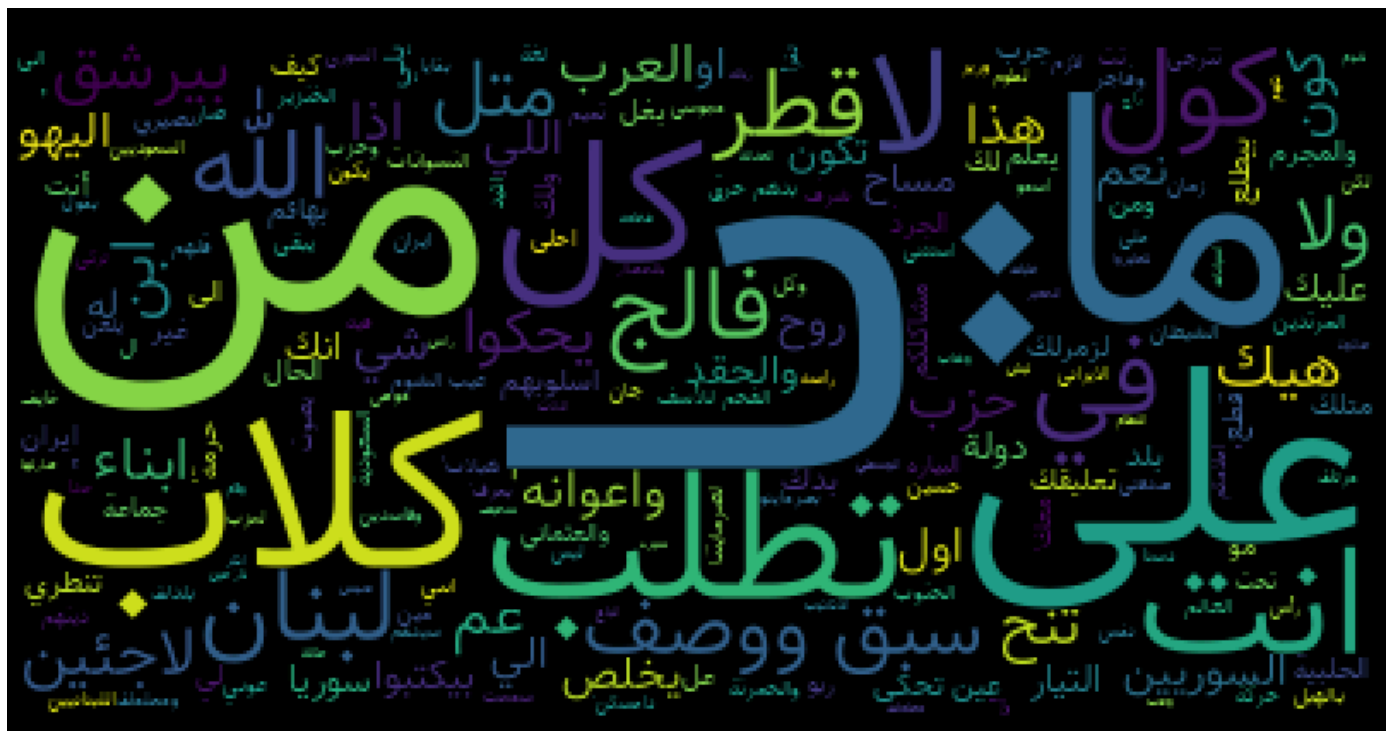

(c) Hate

**Figure 59 The word cloud of the L-HSAB dataset (a. normal, b. abusive, c. hate)**

Figures 60 to 62 show the most common 20 tokens or tokens in each label. The top frequent words differ among the classes. For the normal tweets, the name of the Lebanese politician, " باسيل جبران / Gebran Bassil", and the word "وزير/ minister" are the top ones. The term "كول هوا/ eat air" and the word "كلب/ dog" are commonly used in abusive tweets. Hate tweets also have the word "كلب/ dog" among the top frequent words, in addition to its plural term "كلاب/ dogs".

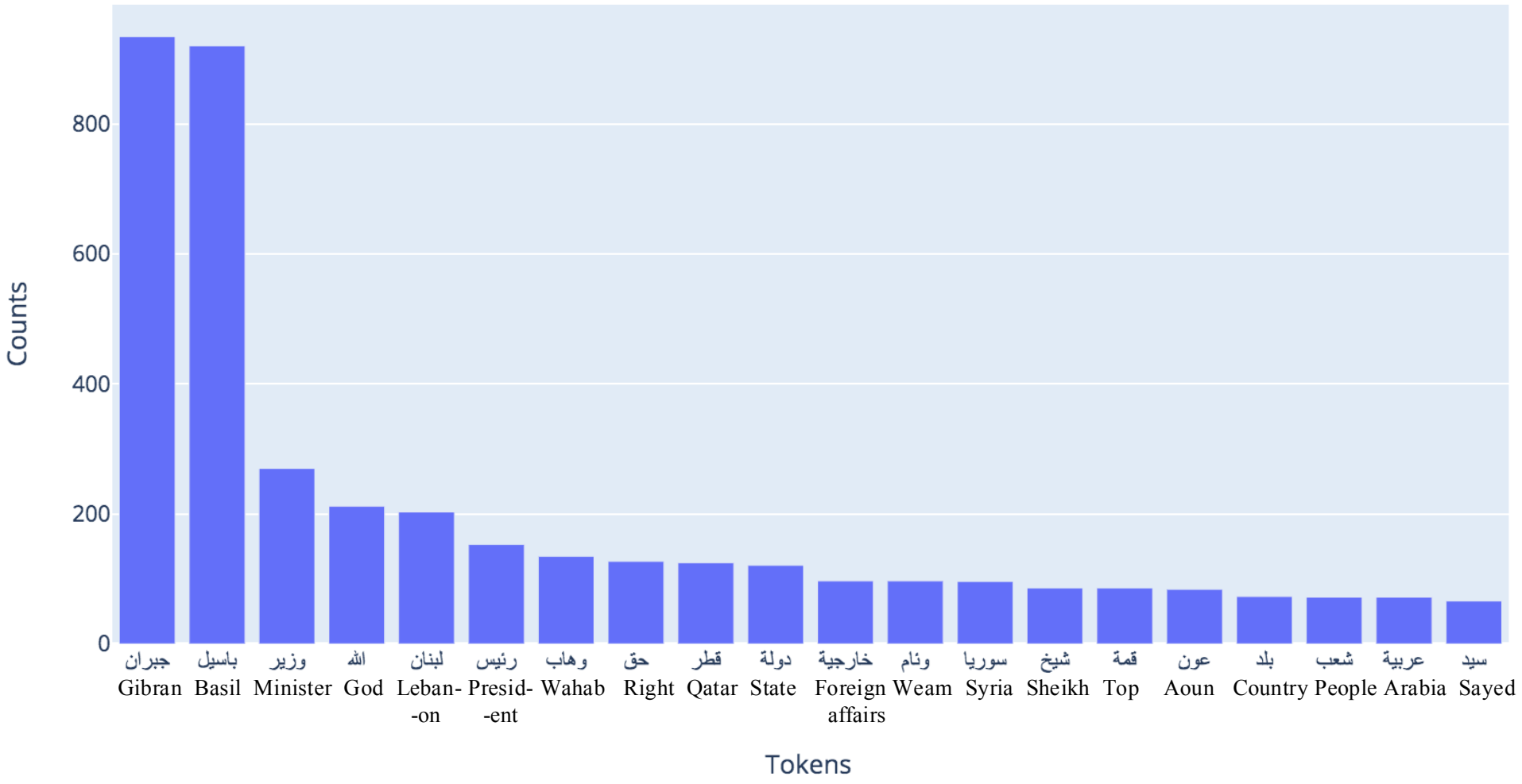

**Figure 60 Most common normal tokens in the L-HSAB dataset**

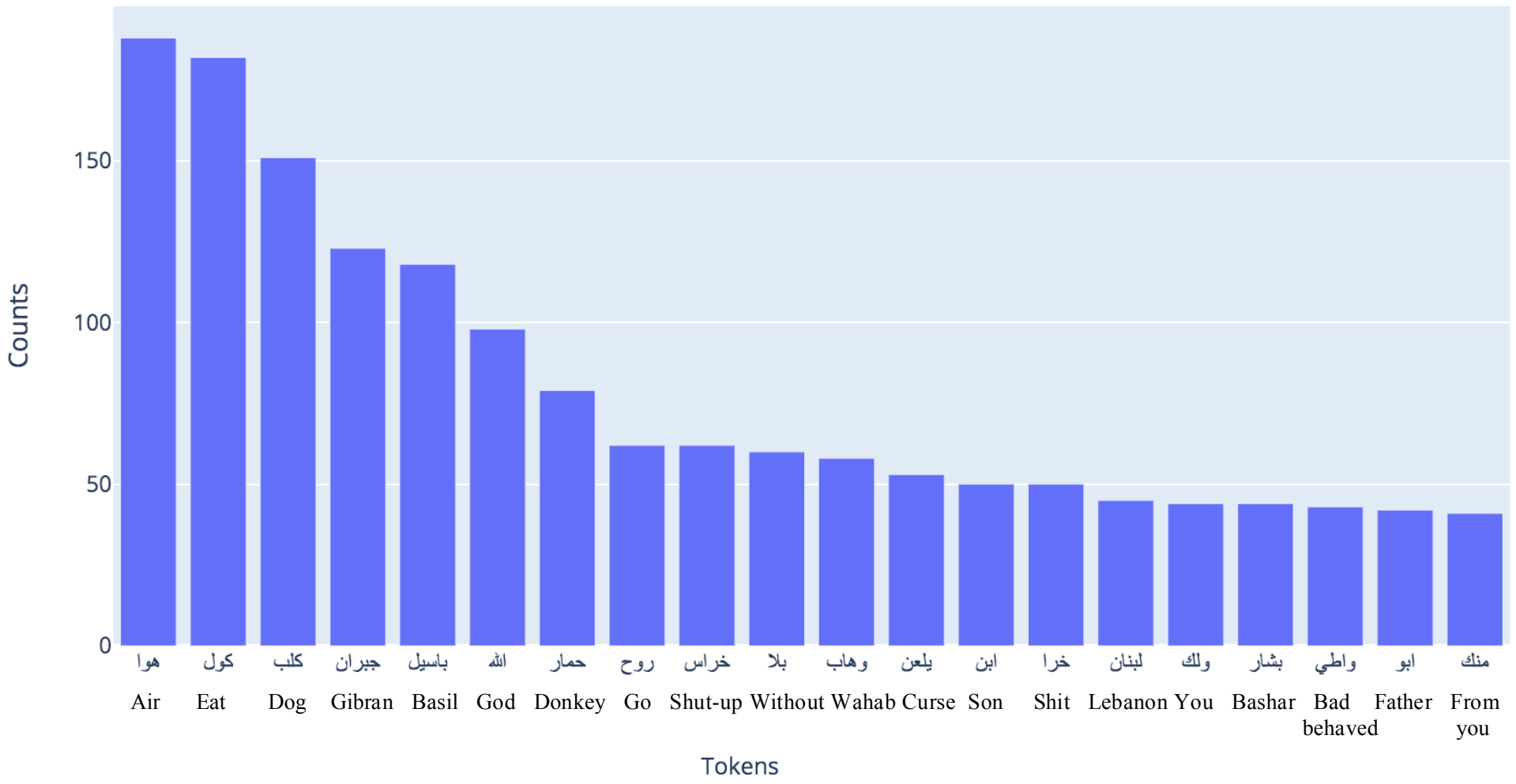

**Figure 61 Most common abusive tokens in the L-HSAB dataset**

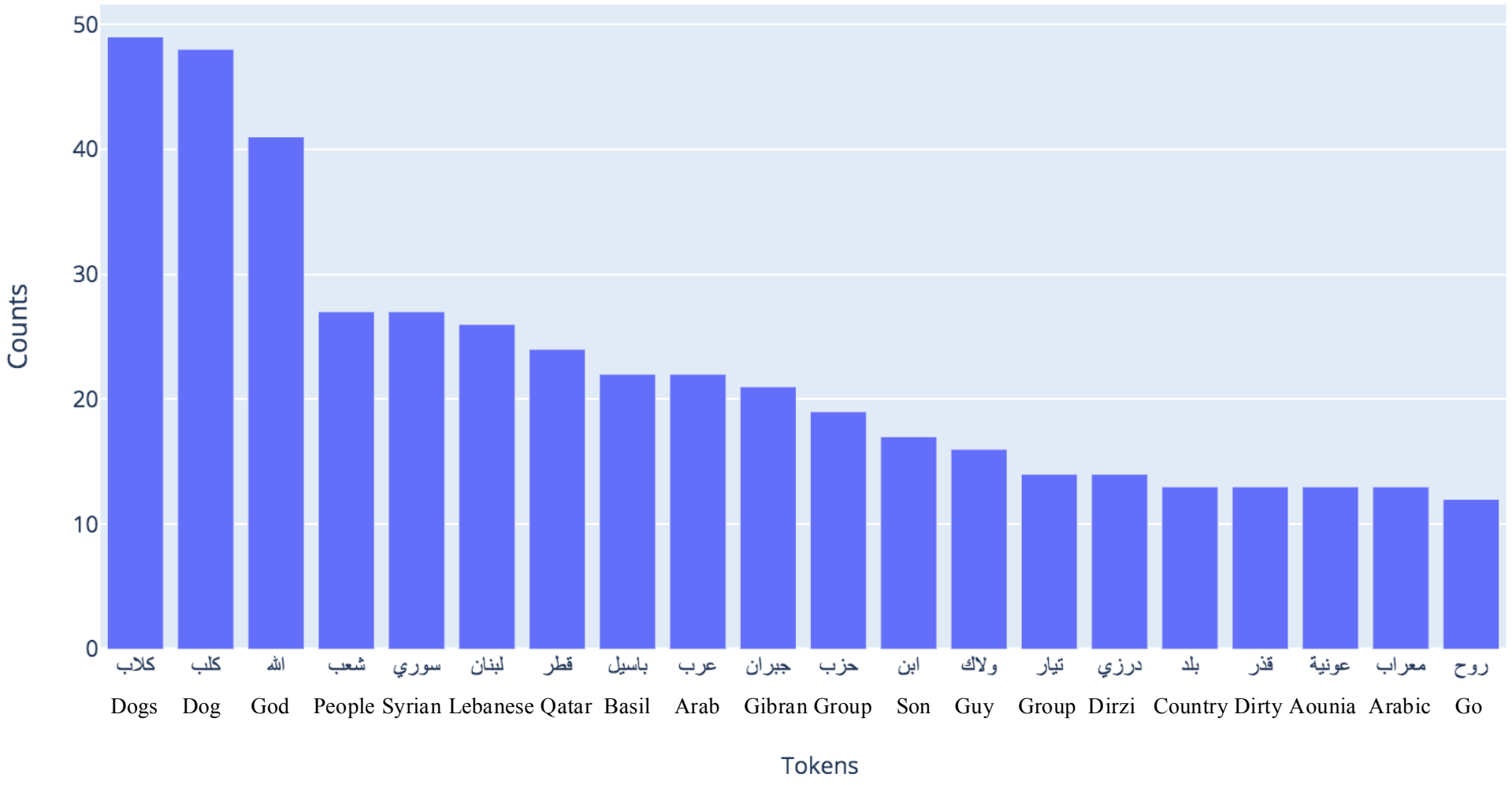

**Figure 62 Most common hate tokens in the L-HSAB dataset**

According to the Voyant tool, the top five most distinctive words of each class are as the following:

1. Normal: القمة / the top (73), الخارجية / the external (69), الوطني / the national (23), الاقتصادية / the economical (18), مقابلة / interview (17).
2. Abusive: خرا / shit (44), واطي / cheap man (43), شرفك / your honor (11), عاهرة / whore (10), طعمرك / for you (10).
3. Hate: عالمطبخ / to the kitchen (9), البنات / girls (6), ولاك / guys (16), لاجئ / refugee (5), فلاح / farmer (5).

For both measures of text length, on general, all classes have very similar measurements with multiple outliers. The hate tweets have slightly higher numbers followed by the normal then the abusive. Figure 63 plots the number of tokens per tweet and Figure 64 plots the number of characters per token.

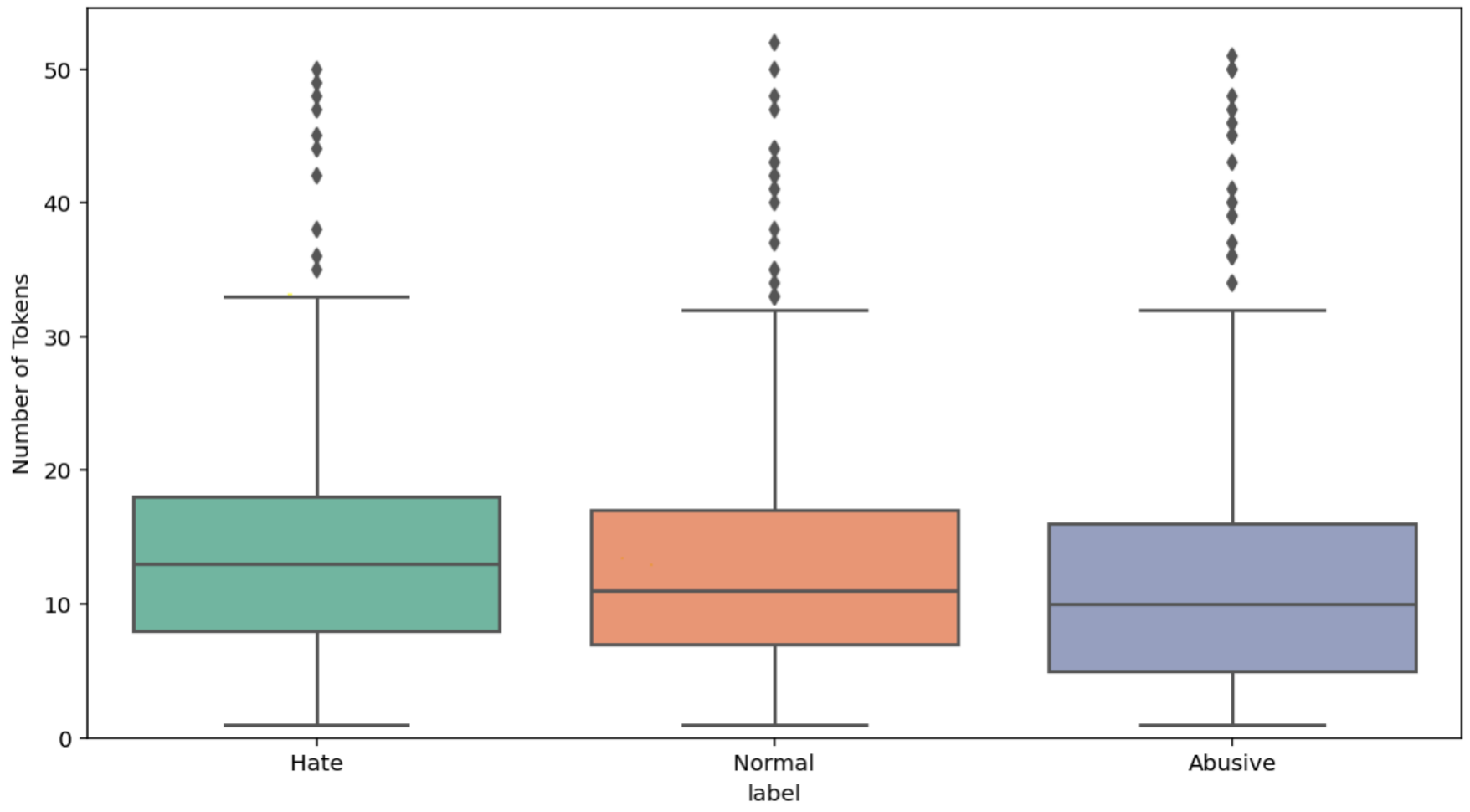

**Figure 63 Statistics of each label in the L-HSAB dataset based on the number of tokens per tweet**

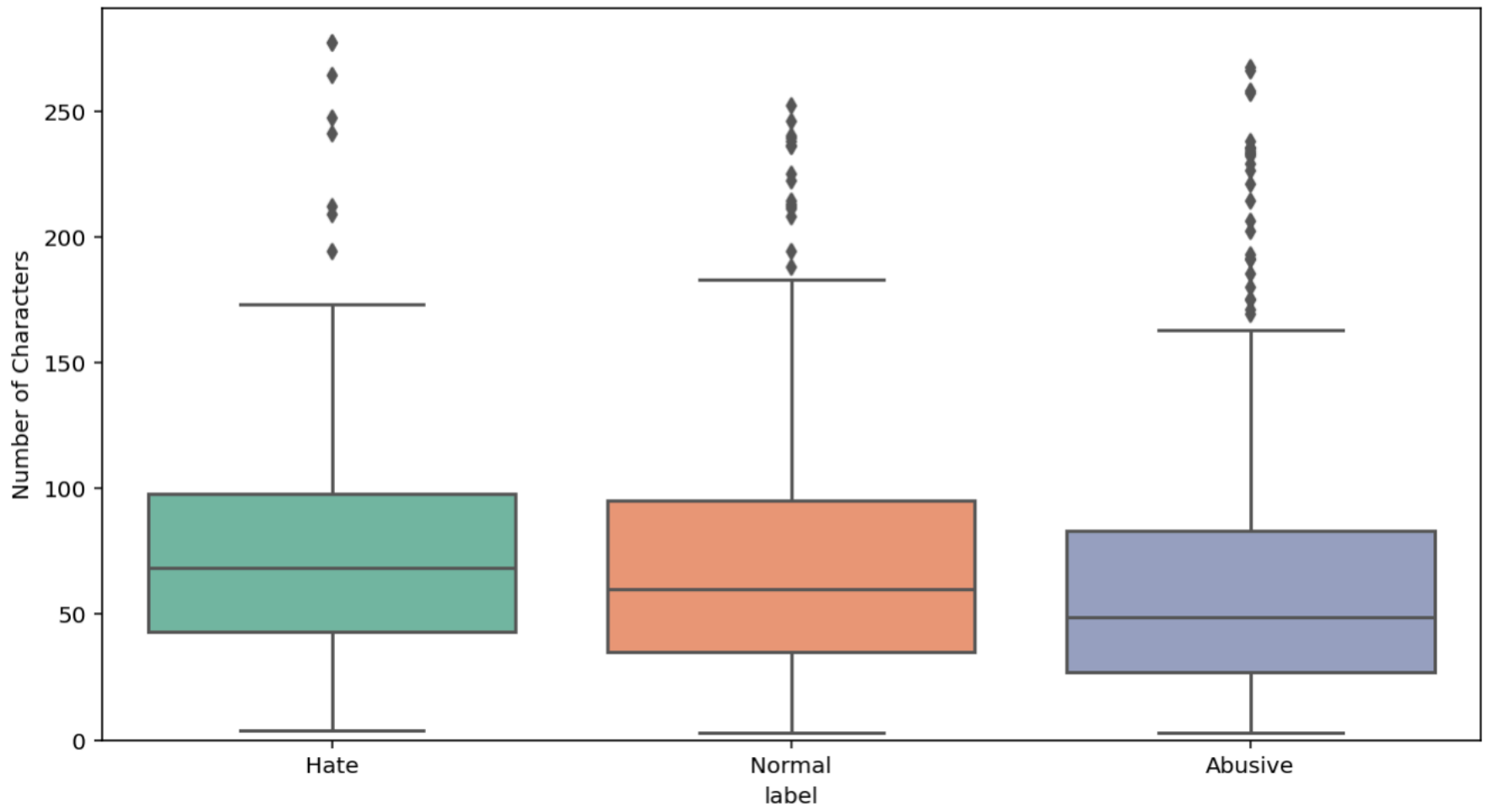

**Figure 64 Statistics of each label in the L-HSAB dataset based on the number of characters per token**

Figure 65 to 67 shows bar charts for the stop words used in the dataset based on the class label. The stop word "ع" just appears in hate tweets, and the preposition يا /you is the first top one for hate and abusive tweets, while for the normal tweets, it is the fourth top stop word.

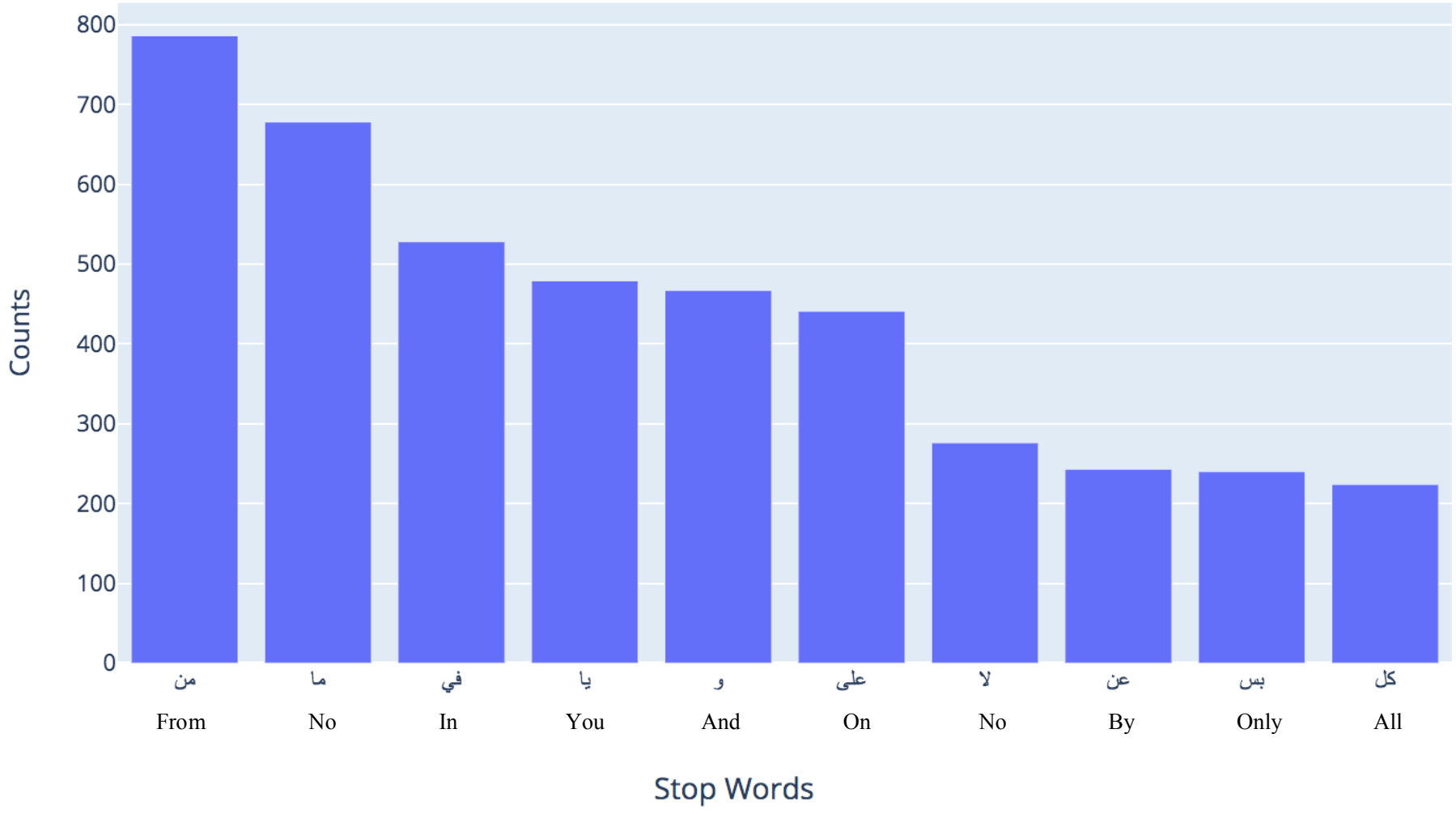

**Figure 65 Most common stop words in normal class from the L-HSAB dataset**

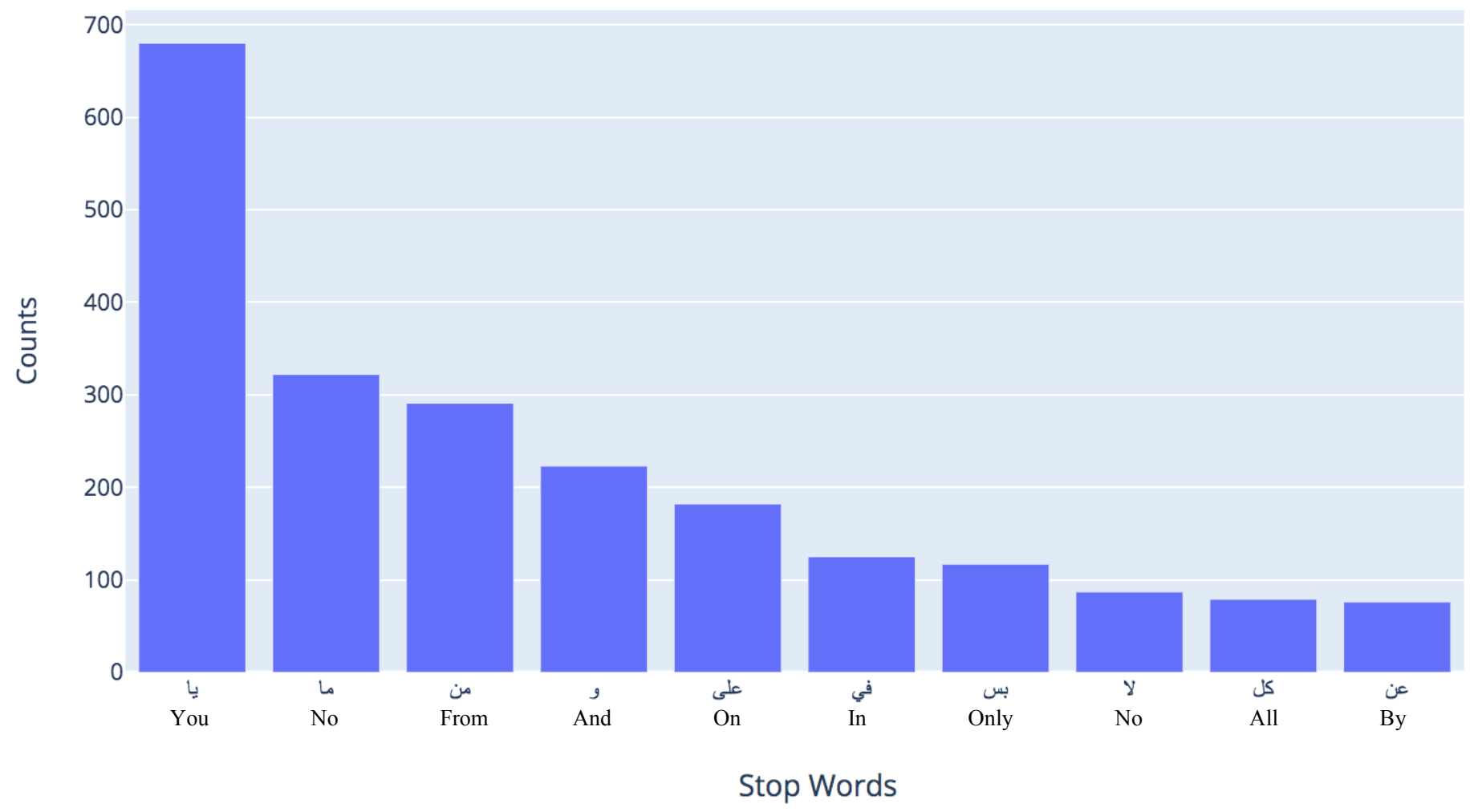

**Figure 66 Most common stop words in abusive class from the L-HSAB dataset**

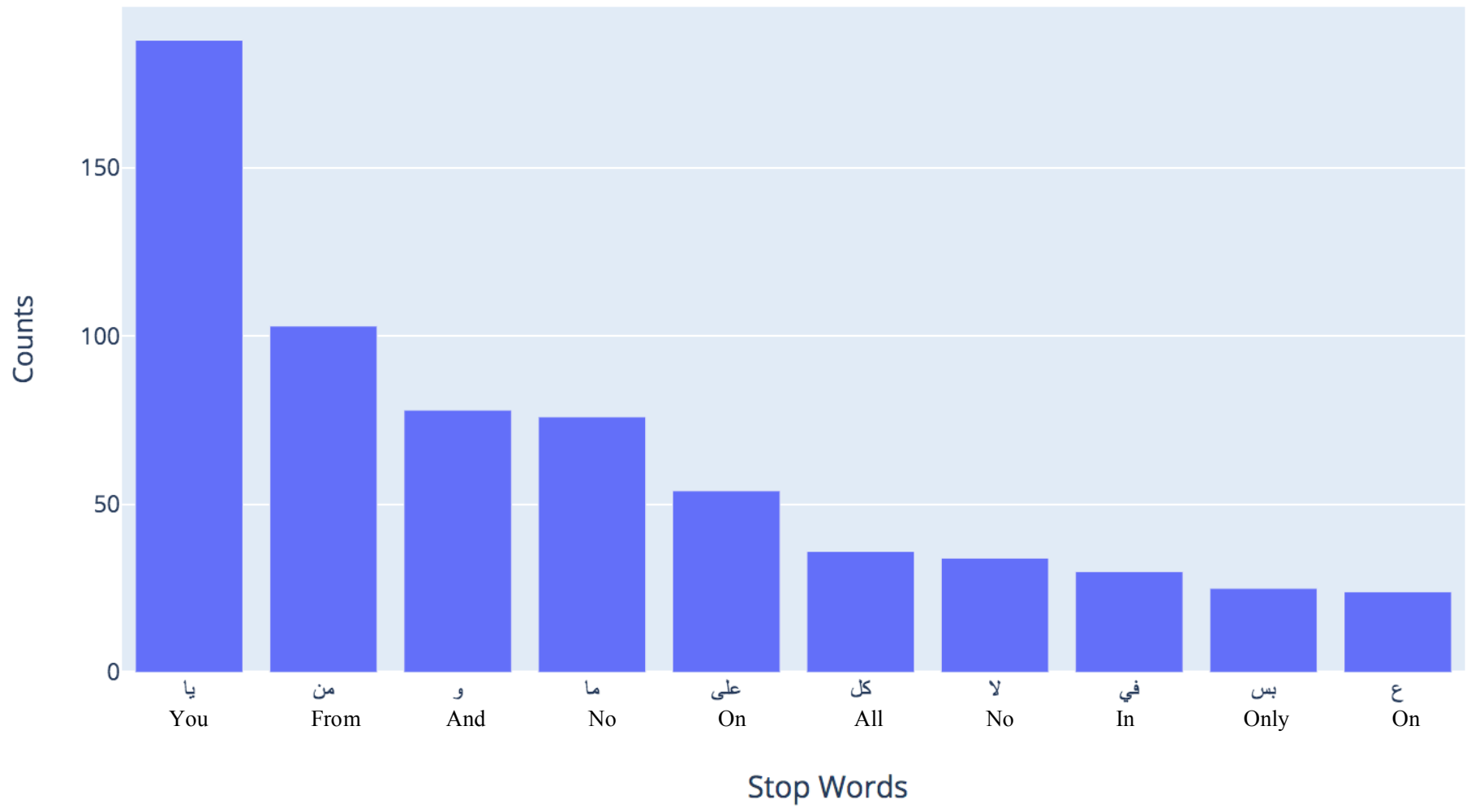

**Figure 67 Most common stop words in hate class from the L-HSAB dataset**

Figure 68 shows a bar chart for the distribution of the sentiment for each class. On general, tweets are mostly negative among all classes. The normal tweets have higher percentages of neutral and positive followed by the abusive tweets. While the hate tweets are dominated by the negative sentiments.

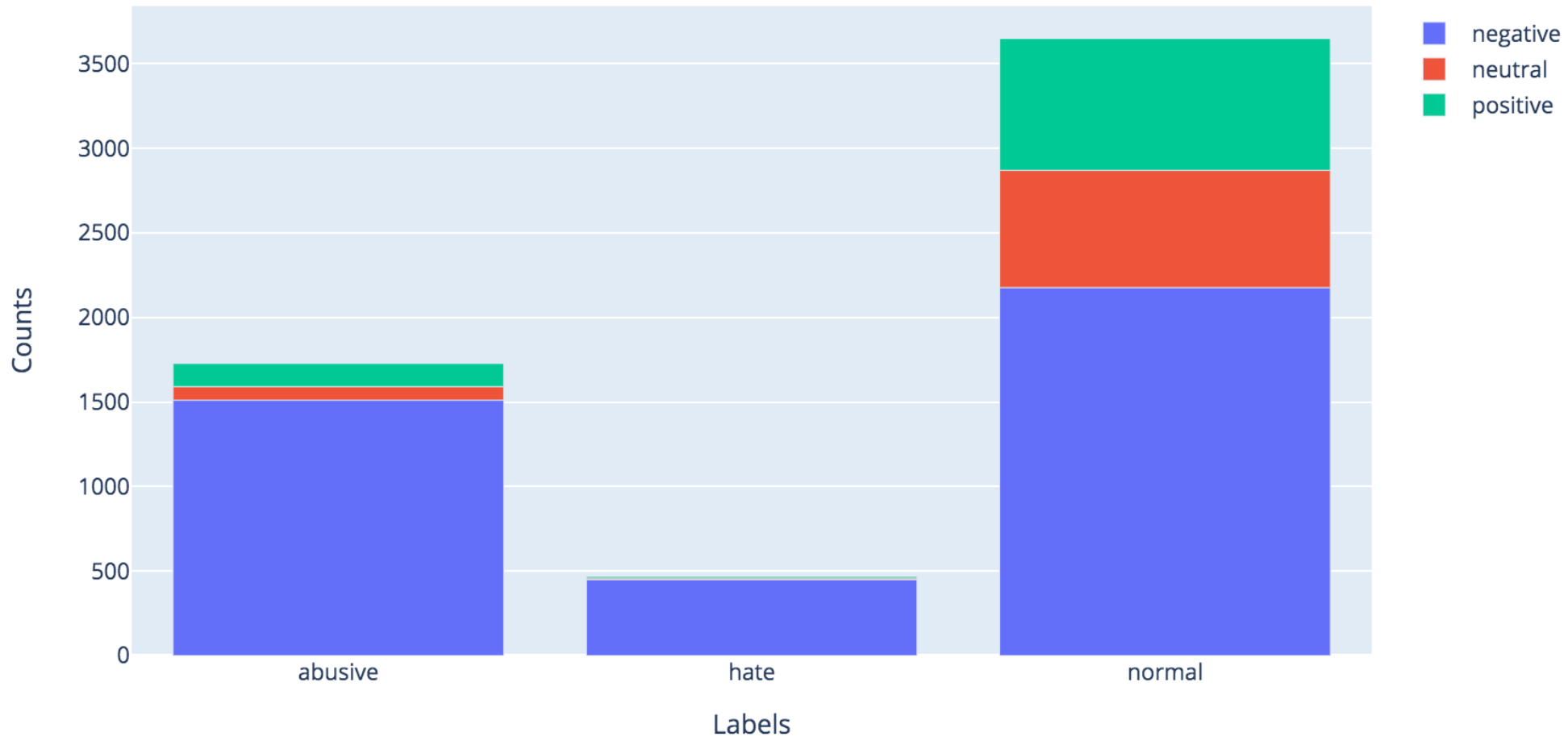

**Figure 68 Sentiment analysis based on labels for the L-HSAB dataset**

Emojis analysis results show the smiling face with sunglasses emoji, "😎", among the top first three emojis in all classes. Normal tweets also record the black heart emoji, "🖤", and the women symbol, "♀", among the top most frequent emoji. Abusive tweets show the face with tears of joy emoji, "😂", and the man shrugging emoji, "🤷‍♂️". Very few emojis were recorded by hate tweets. Only three emojis resulted from the analysis for hate tweets, including the smiling face with sunglasses emoji, "😎", the face with tears of joy emoji, "😂", and the man symbol, "♂". More frequent emojis are plotted in Figures 69 to 71.

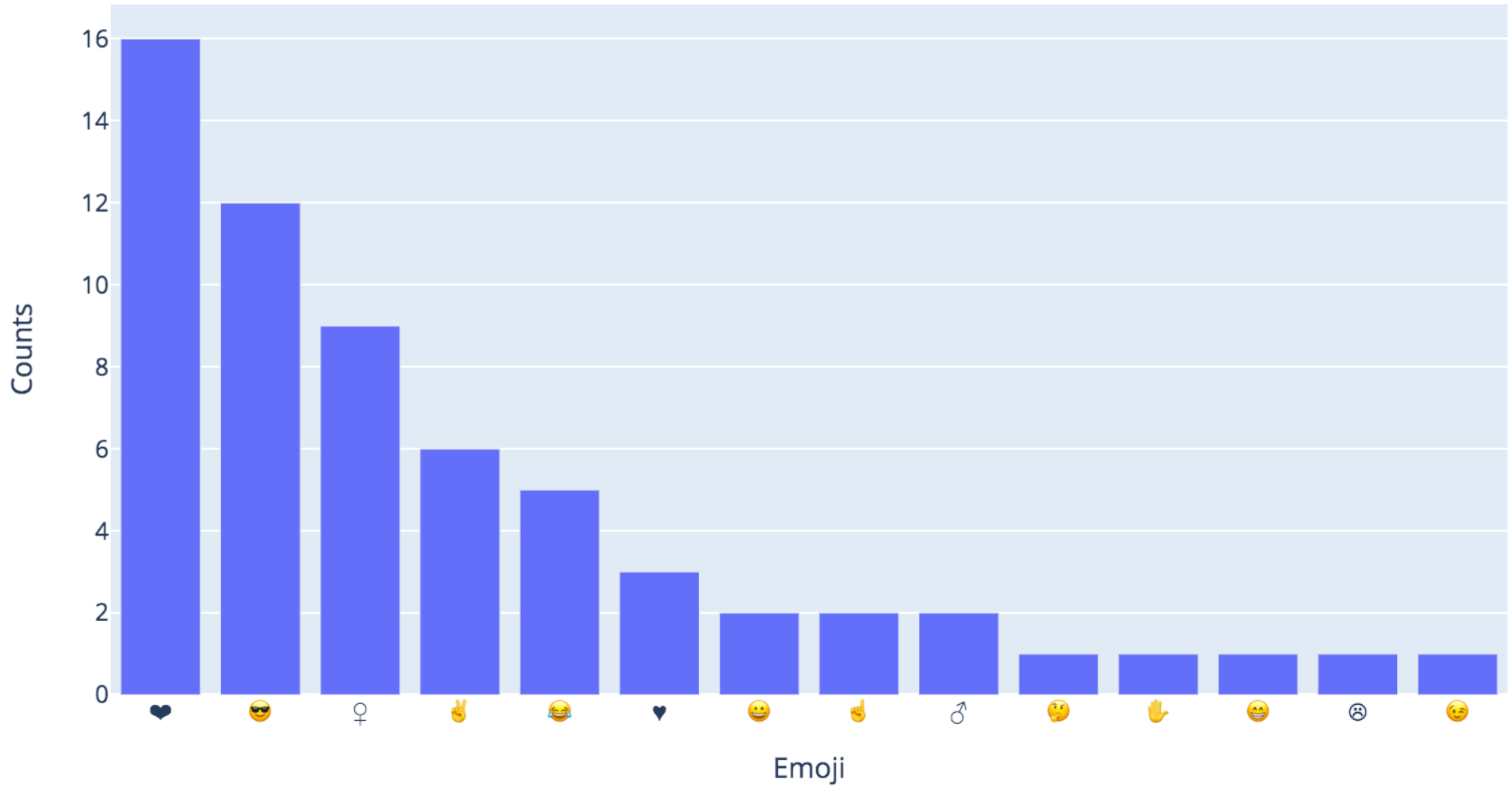

**Figure 69 Most common normal emojis in the L-HSAB dataset**

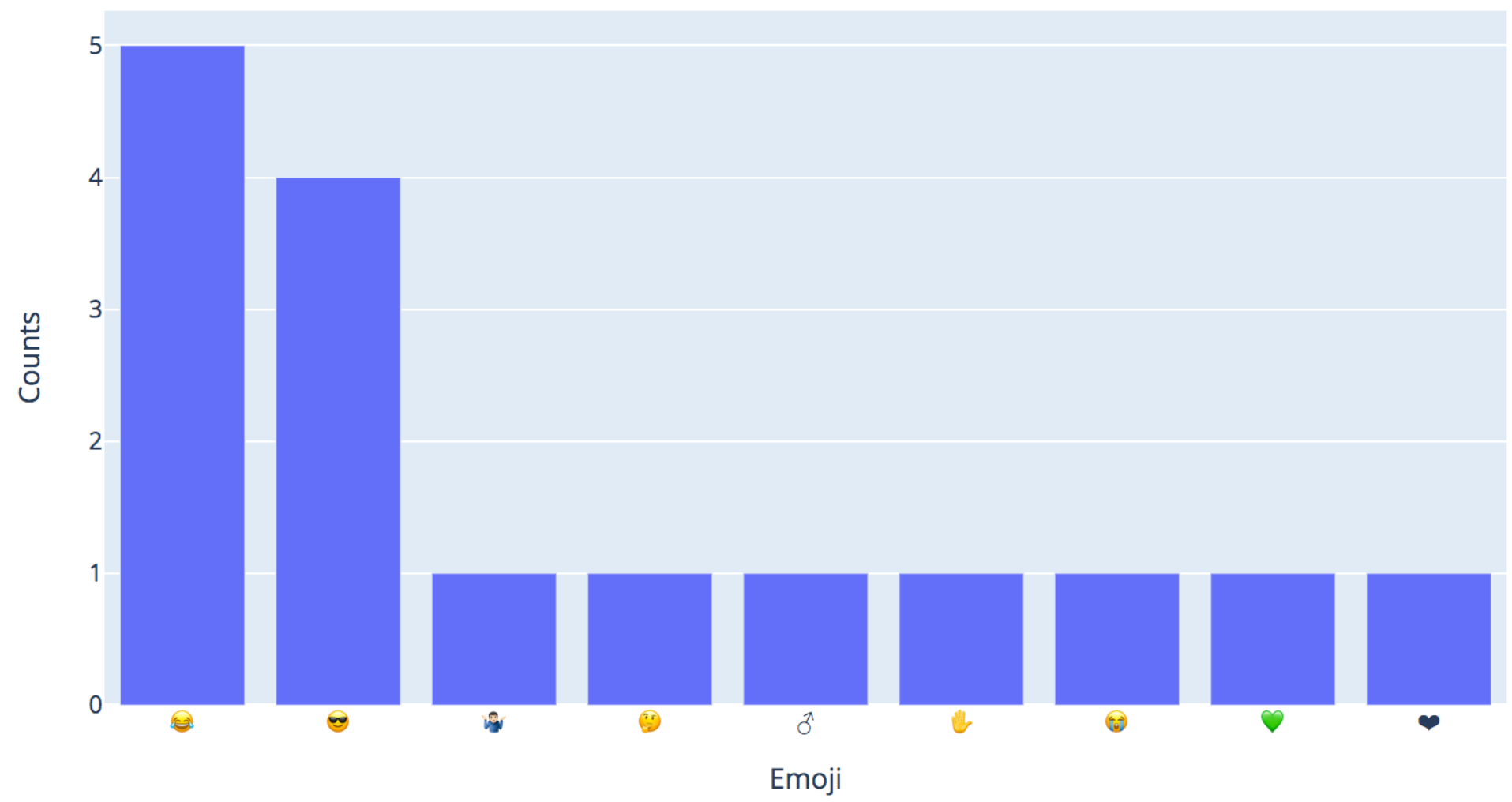

**Figure 70 Most common abusive emojis in the L-HSAB dataset**

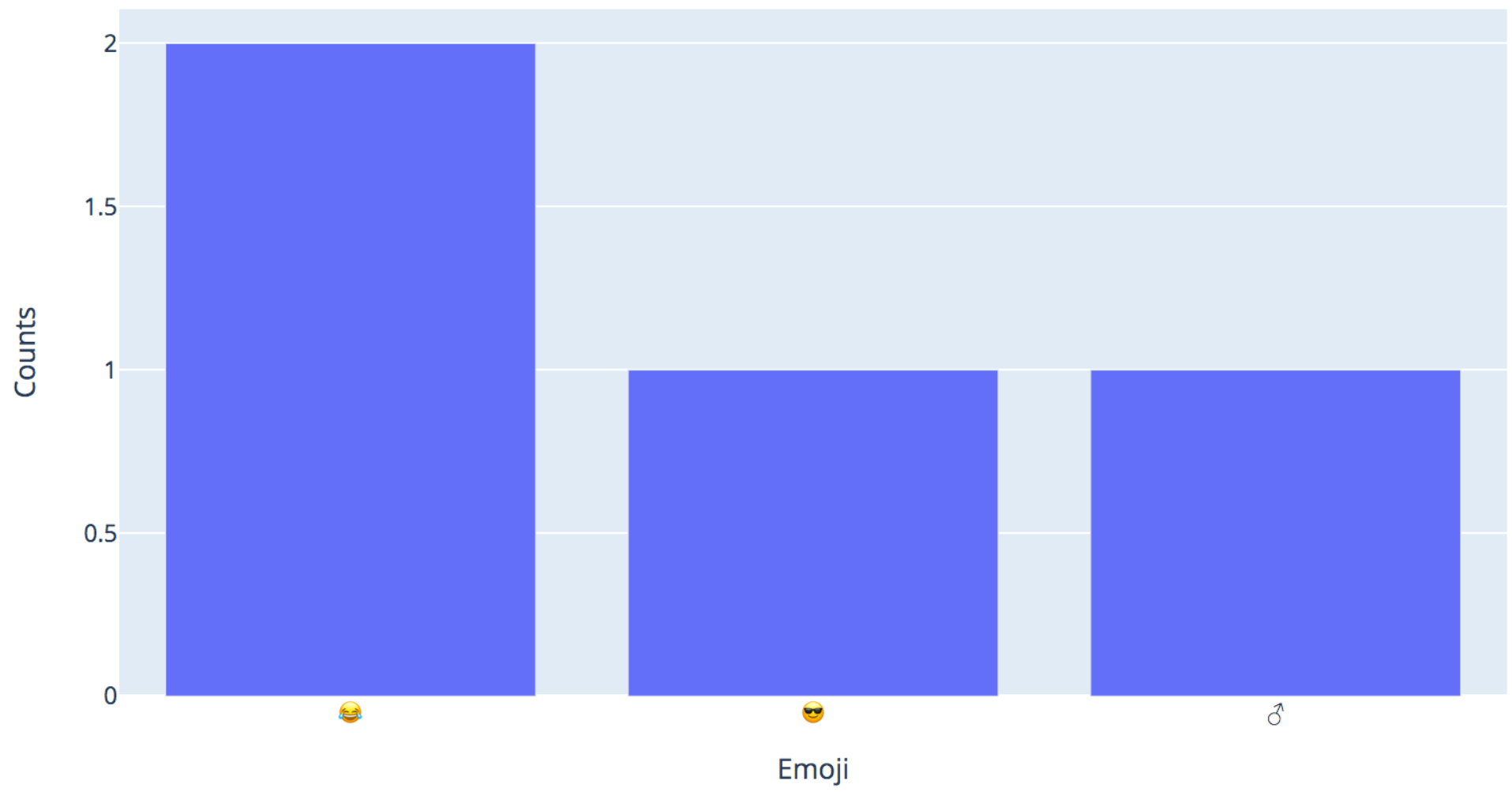

**Figure 71 Most common hate emojis in the L-HSAB dataset**

Figures 72 and 74 show the top punctuation. All classes report "," and "؟" among the top used punctuation. Thus, there is not particular pattern between punctuation used and offensive content.

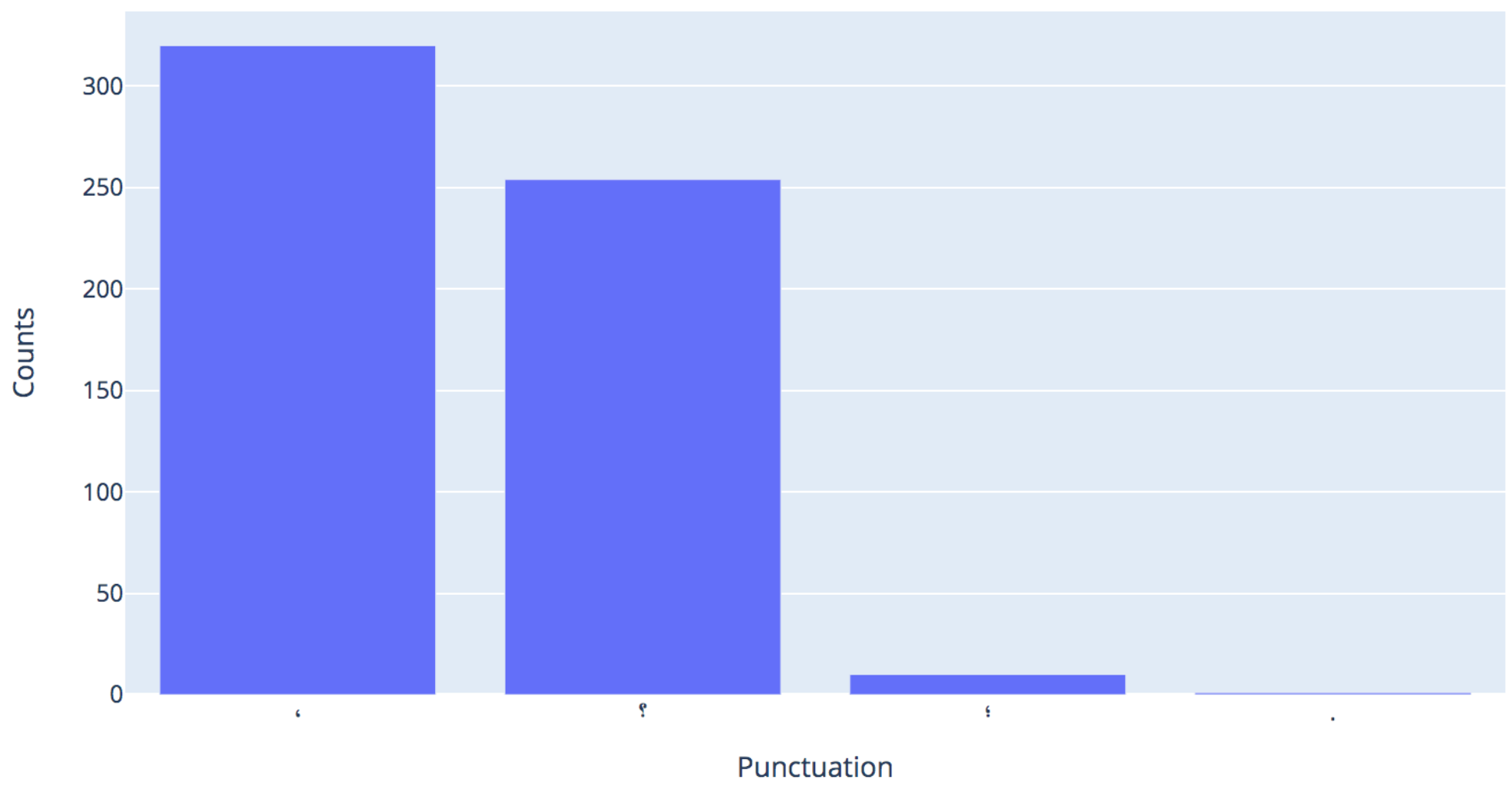

**Figure 72 Most common normal punctuation in the L-HSAB dataset**

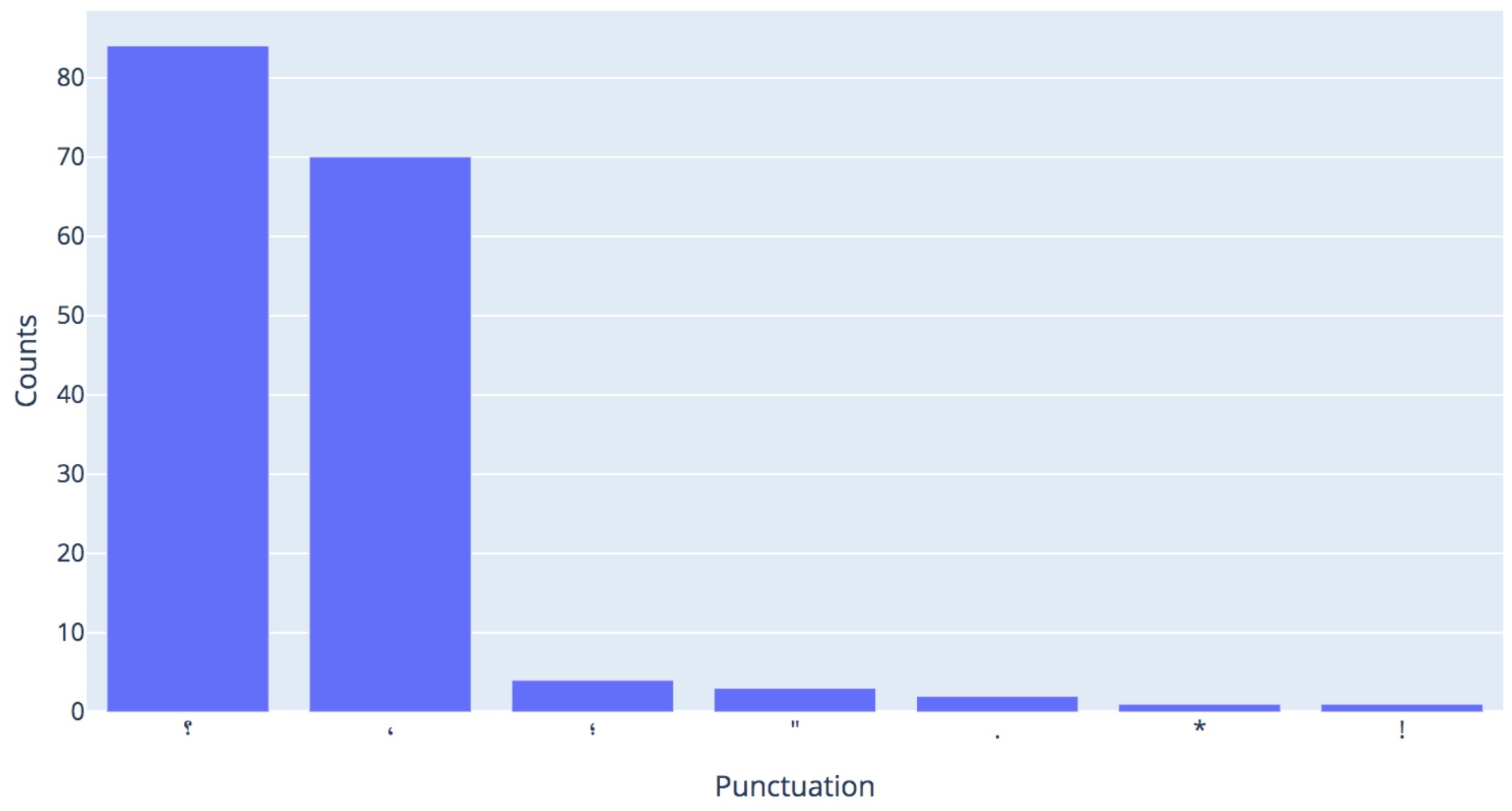

**Figure 73 Most common abusive punctuation in the L-HSAB dataset**

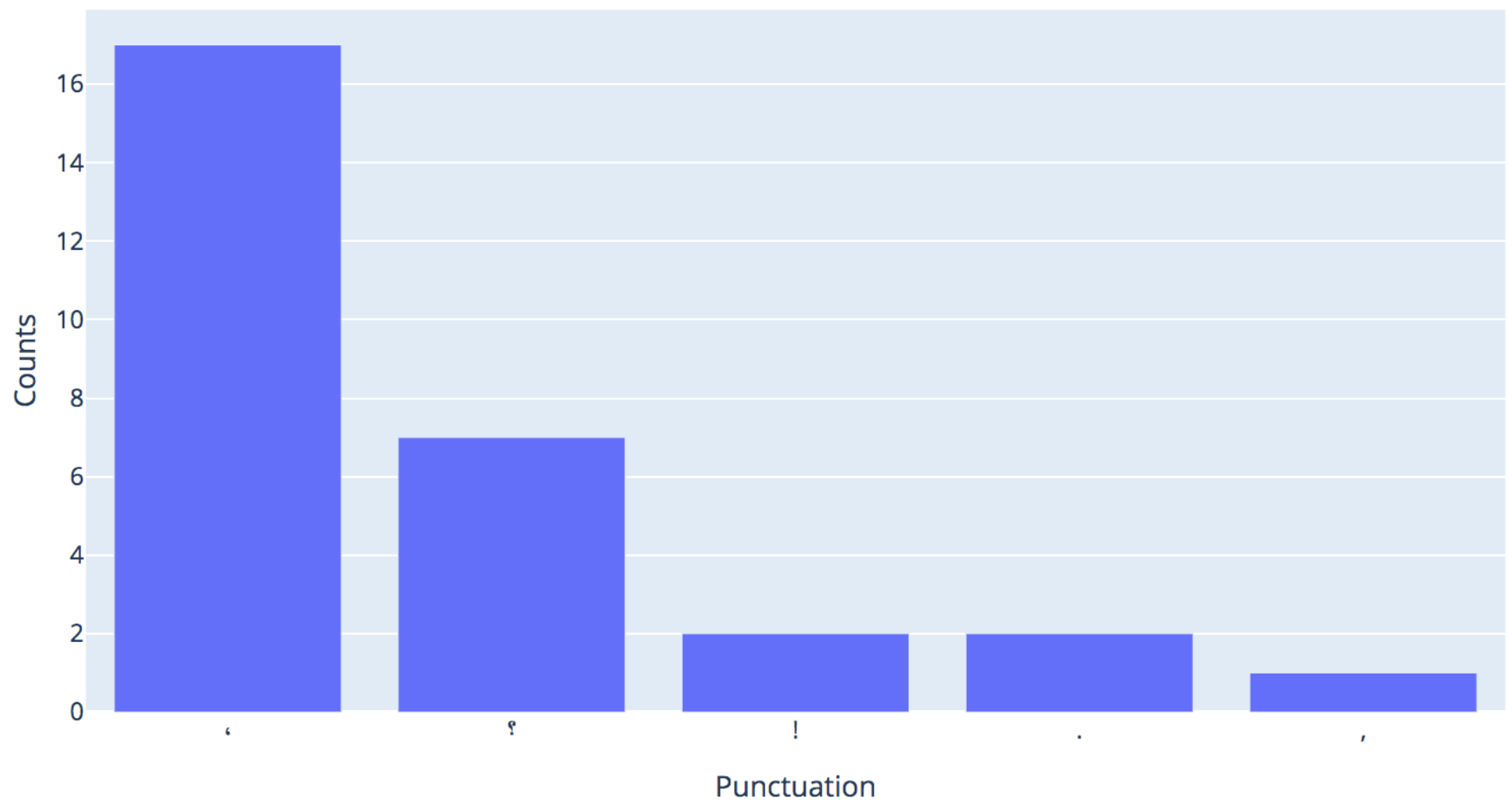

**Figure 74 Most common hate punctuation in the L-HSAB dataset**

5) The Tunisian Hate and Abusive speech dataset (T-HSAB):

Haddad et al. (2019) develop T-HSAB dataset, which has a total of 6,075 comments; 3,834 normal, 1,127 abusive, and 1,078 hate (see Figure 75). Text cleaned to remove platform specific symbols; RT, usermention (@), and hashtags (#), and preprocessed by removing emojis, digits, and all non-Arabic characters. Duplicate comments were removed.

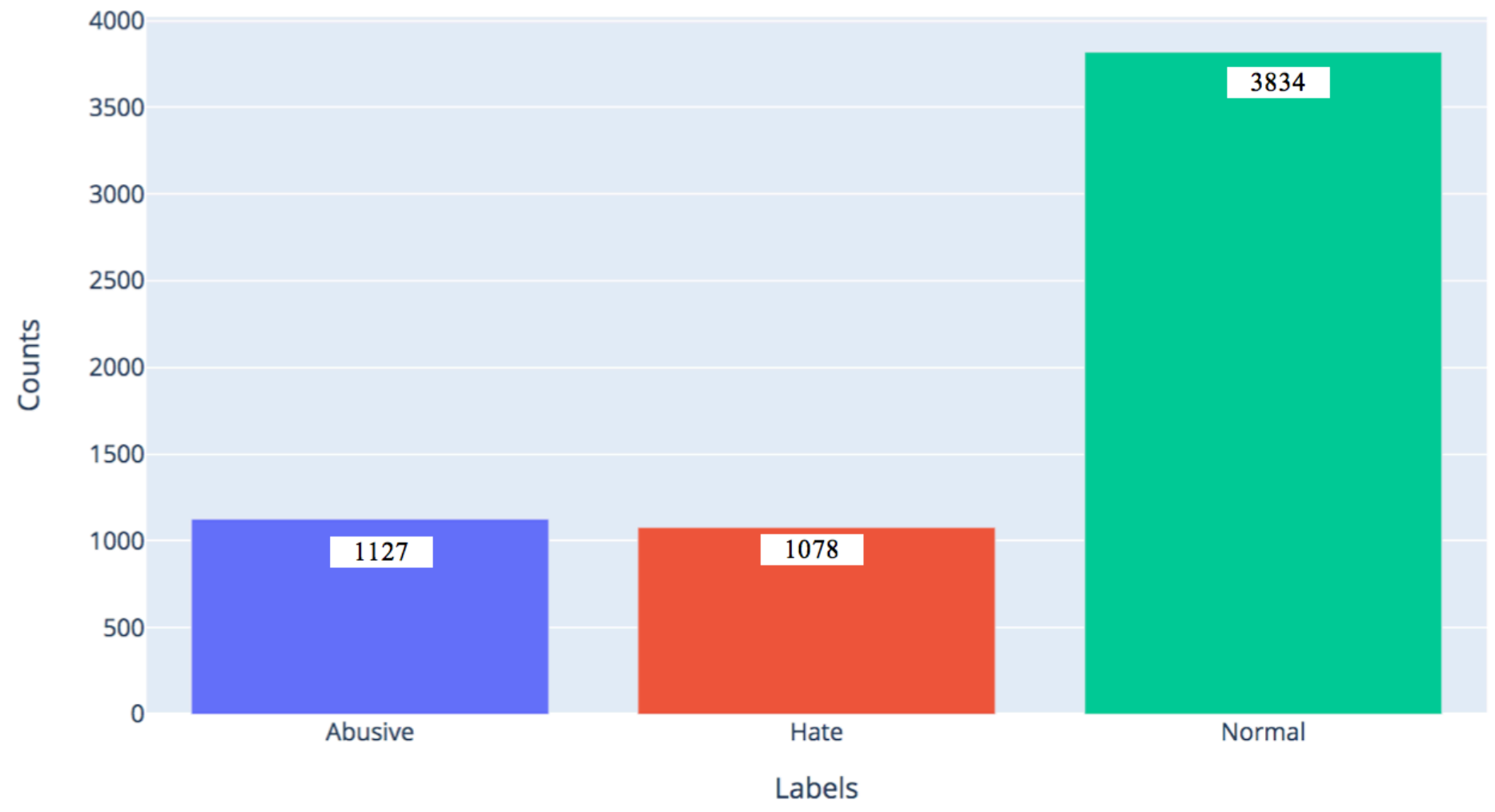

**Figure 75 Class distribution for the T-HSAB dataset**

Figure 76 shows the word cloud for each class separately, as it observed, all classes include similar words, such as "تونس / Tunisia", "الاسلام / the Islam", and "شعب / people". It is hard to identify specific patterns from the word clouds.

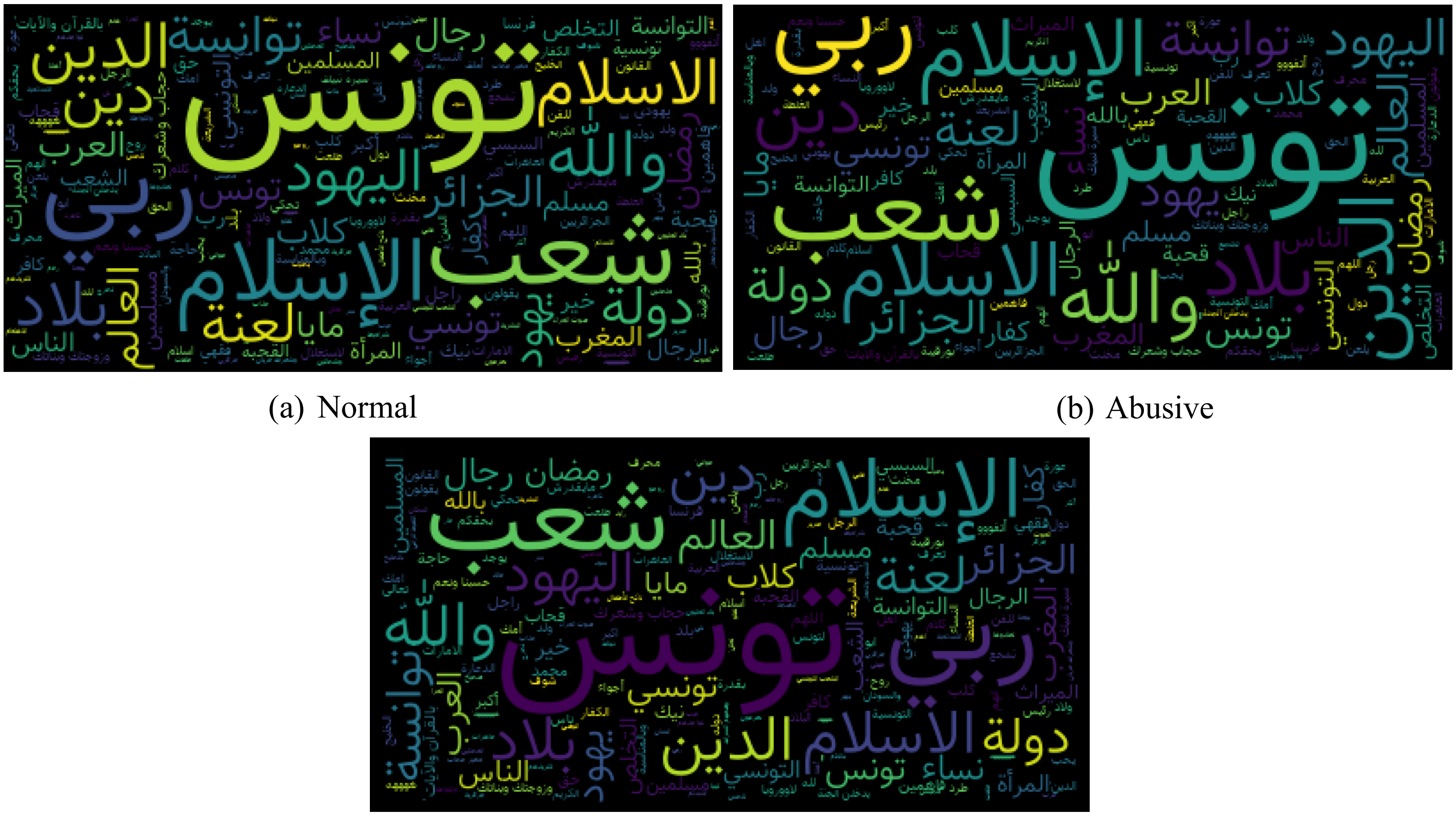

(a) Normal (b) Abusive

(c) Hate

**Figure 76 The word cloud of the T-HSAB dataset (a. normal, b. abusive, c. hate)**

From the frequencies bar charts for the top ten tokens from each class, all figures plot “تونس / Tunisia” among the top first three tokens. Figure 77 for normal class, the second most frequent token is “ربي / my lord” and the third is “لطفي / Lutfy”. In Figure 78, before “تونس / Tunisia” comes “قحبة / whore” and after it comes “نيك / fuck” within abusive comments. For the hate class in Figure 79, the words “شعب / people” and “دين / religion” are also very frequent.

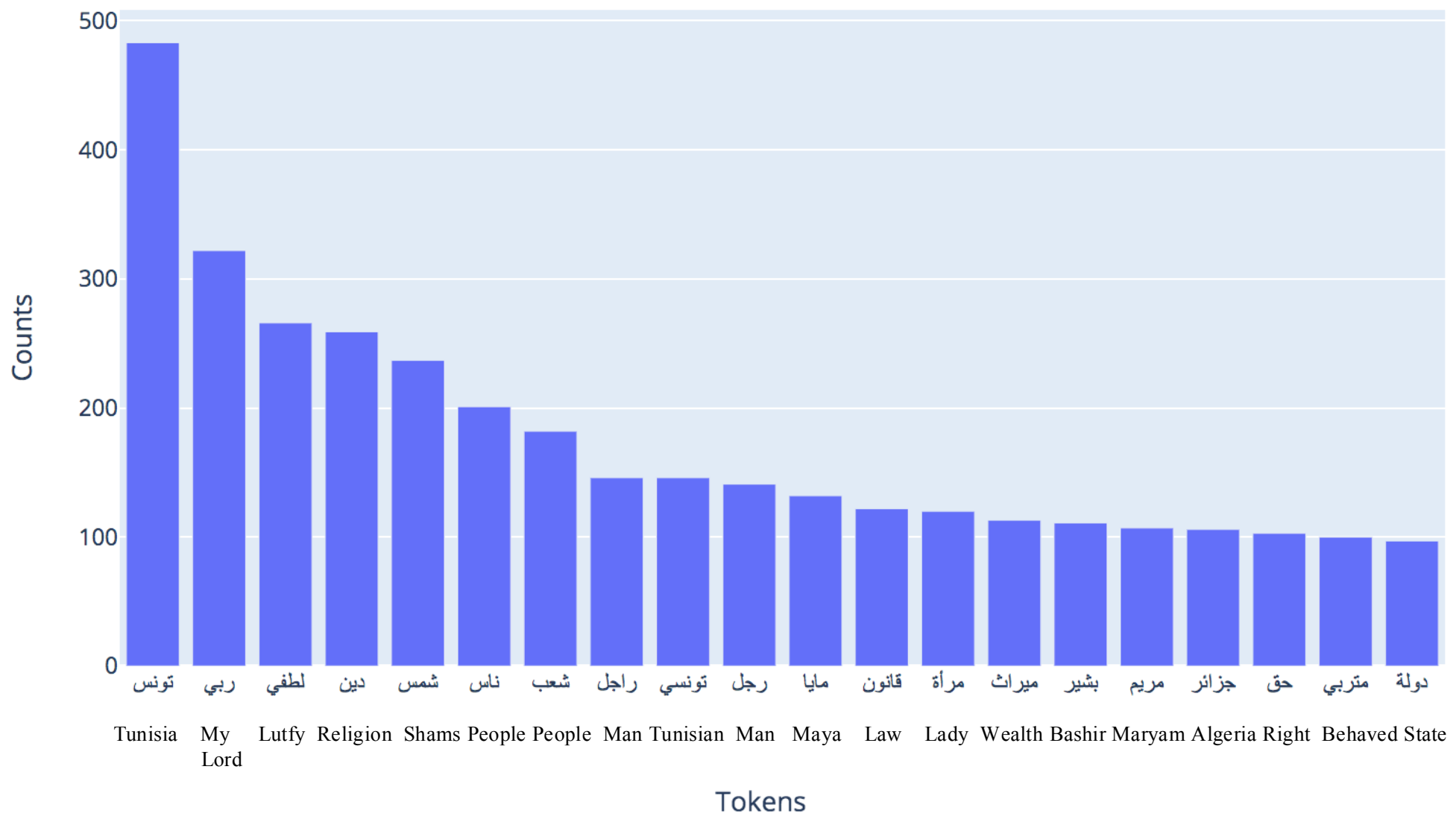

**Figure 77 Most common normal tokens in the T-HSAB dataset**

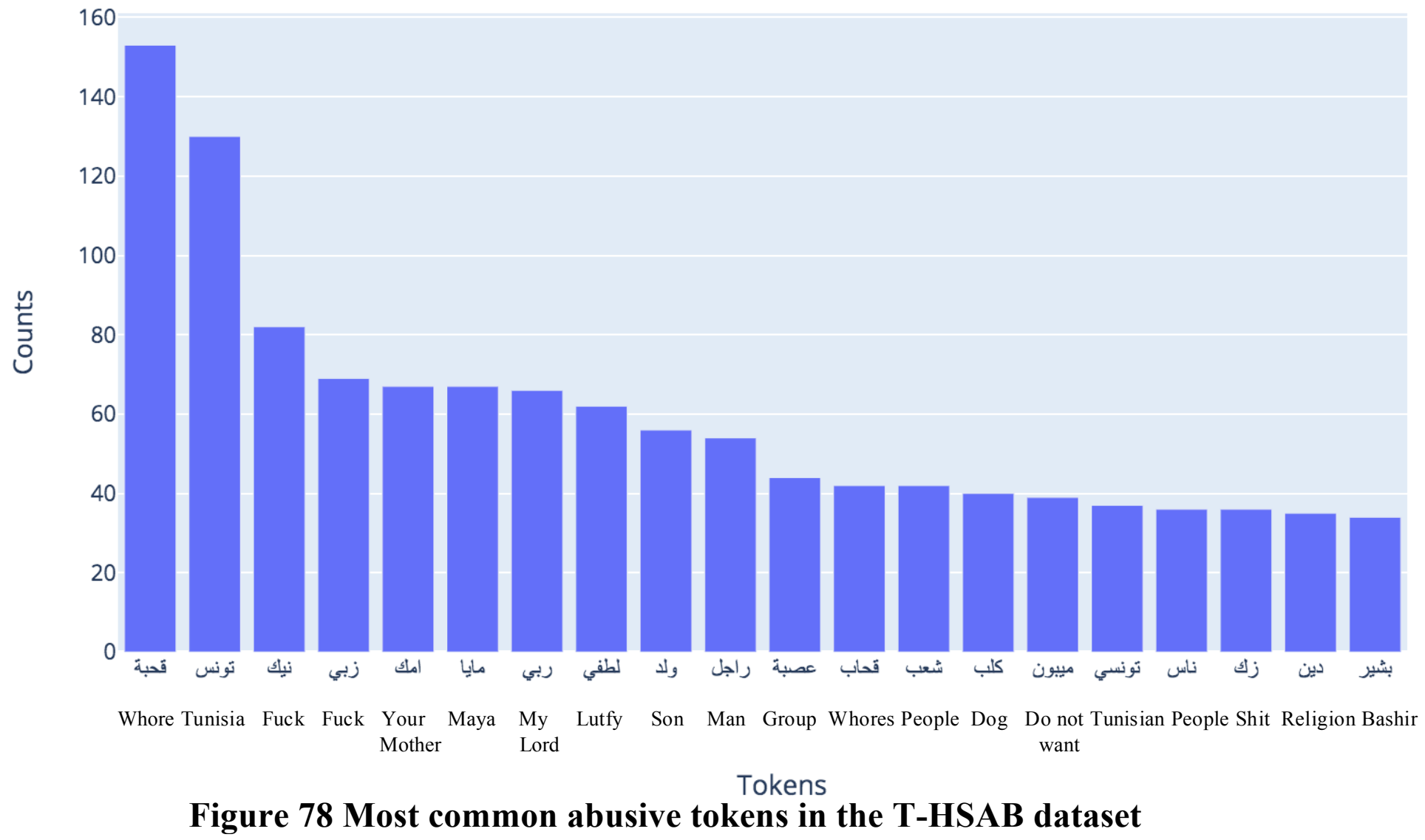

**Figure 78 Most common abusive tokens in the T-HSAB dataset**

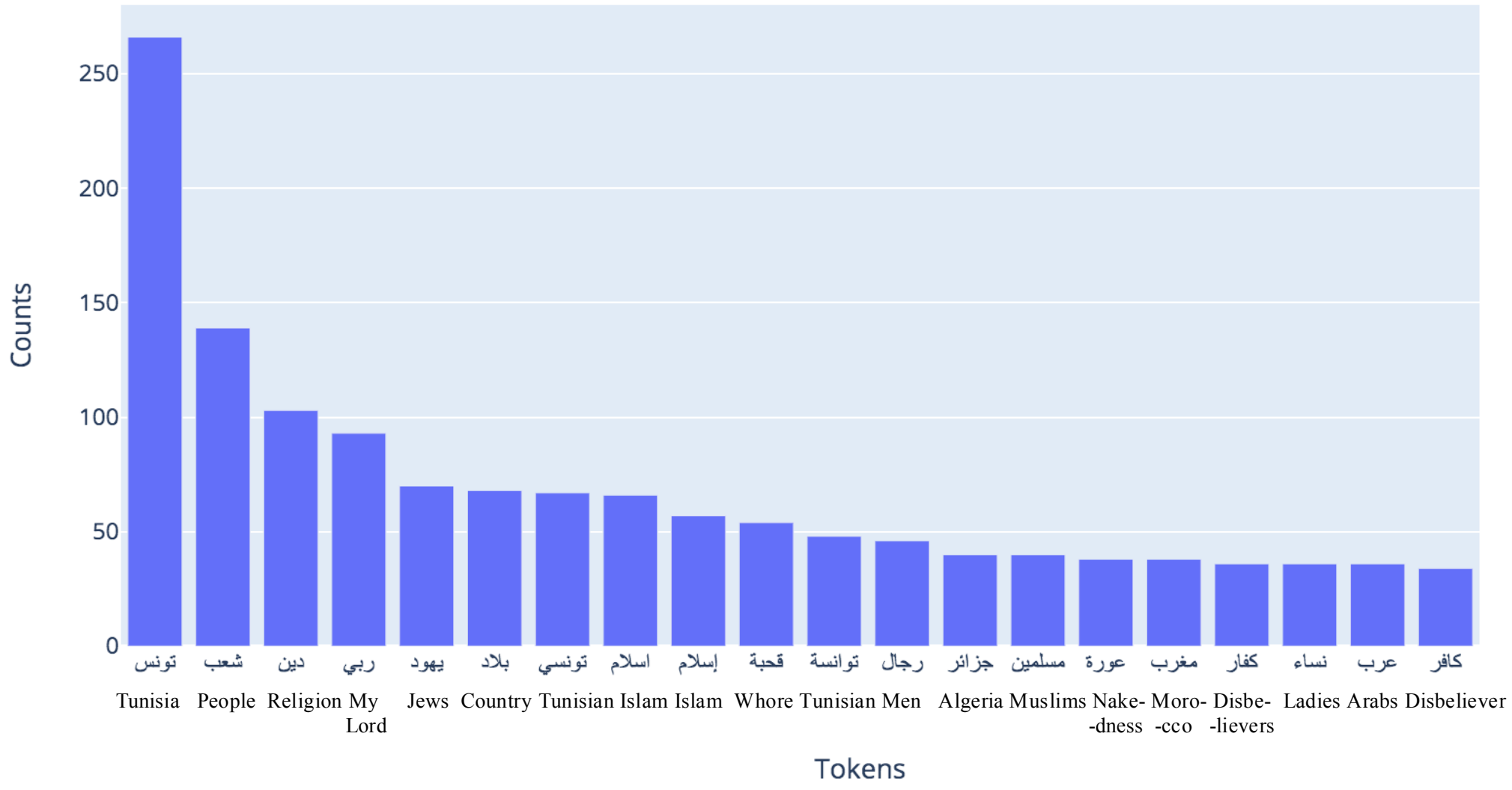

**Figure 79 Most common hate tokens in the T-HSAB dataset**

The followings are the top five most distinctive words:

1. Abusive: نيك / fuck (80), قحبة / whore (78), القحبة / the whore (75), عصبة/ band (38), زك / shit (35).
2. Hate: القحبة / the whore (33), ودين / religion (8), قحبة / whore (21), ناكح / fucked (7), لوط / Lot (7).
3. Normal: عواطف / emotions (60), الخطاب / Al-Khattab (19), صلاح / Salah (17), شكرا / thanks (17), الدباغ / Al-Dabagh (45).

Investigating the lengths of comments and tokens illustrates the very short and limited content of the dataset in general. Figure 80 shows that most of the comments regardless of their classes are about 9 tokens length. Tokens are also slightly short among all classes as can be seen from Figure 81.

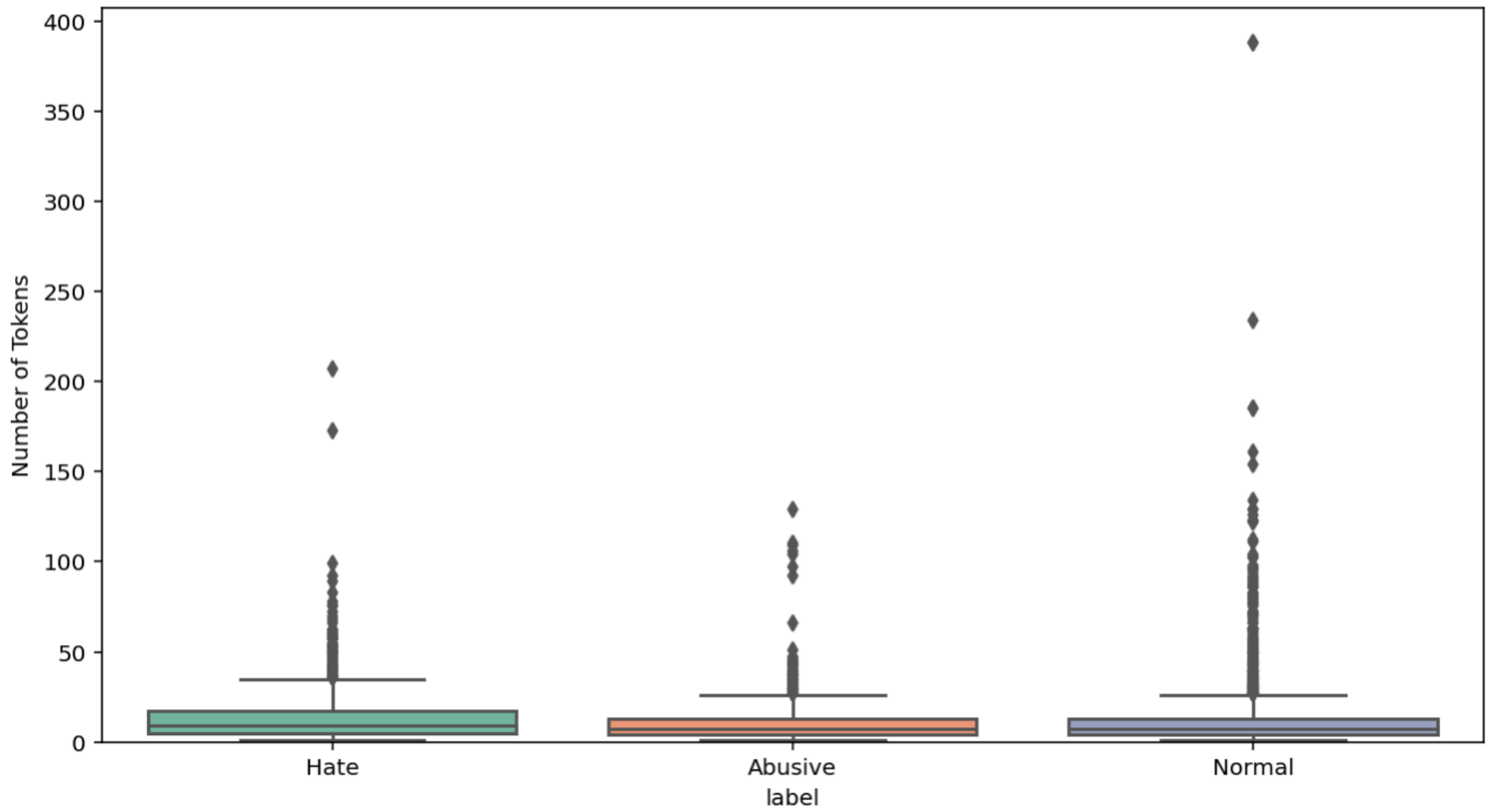

**Figure 80 Statistics of each label in the T-HSAB dataset based on the number of tokens per tweet**

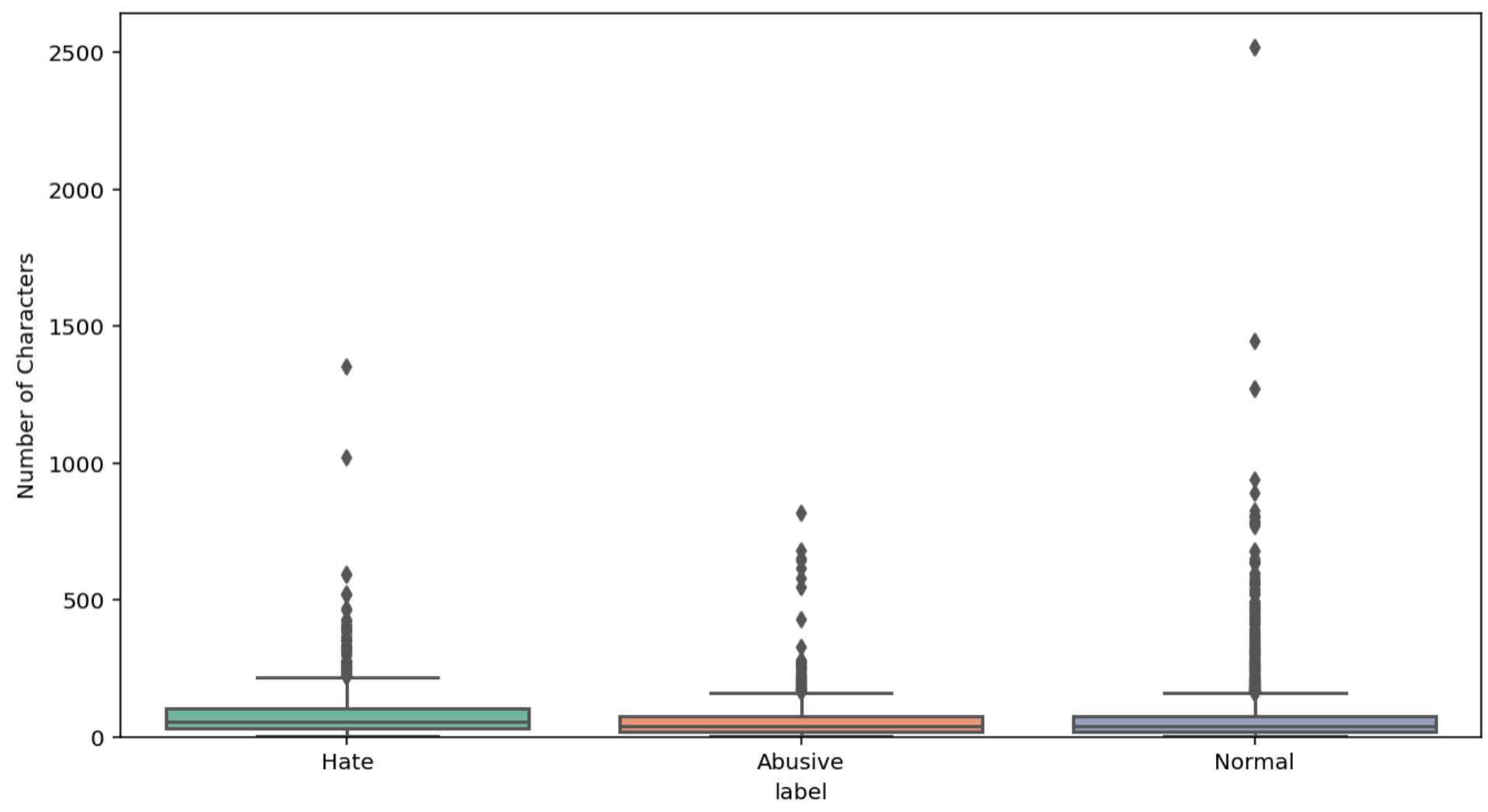

**Figure 81 Statistics of each label in the T-HSAB dataset based on the number of characters per token**

Stop words are very similar in all classes as illustrated in Figures 82 to 84. The first and second stop words in all classes are "ه / ha" and "أكثر / more". Normal comments also contain "هكذا / this is" very frequently, abusive comments show "بكم / by you", and hate comments report "الآن / now".

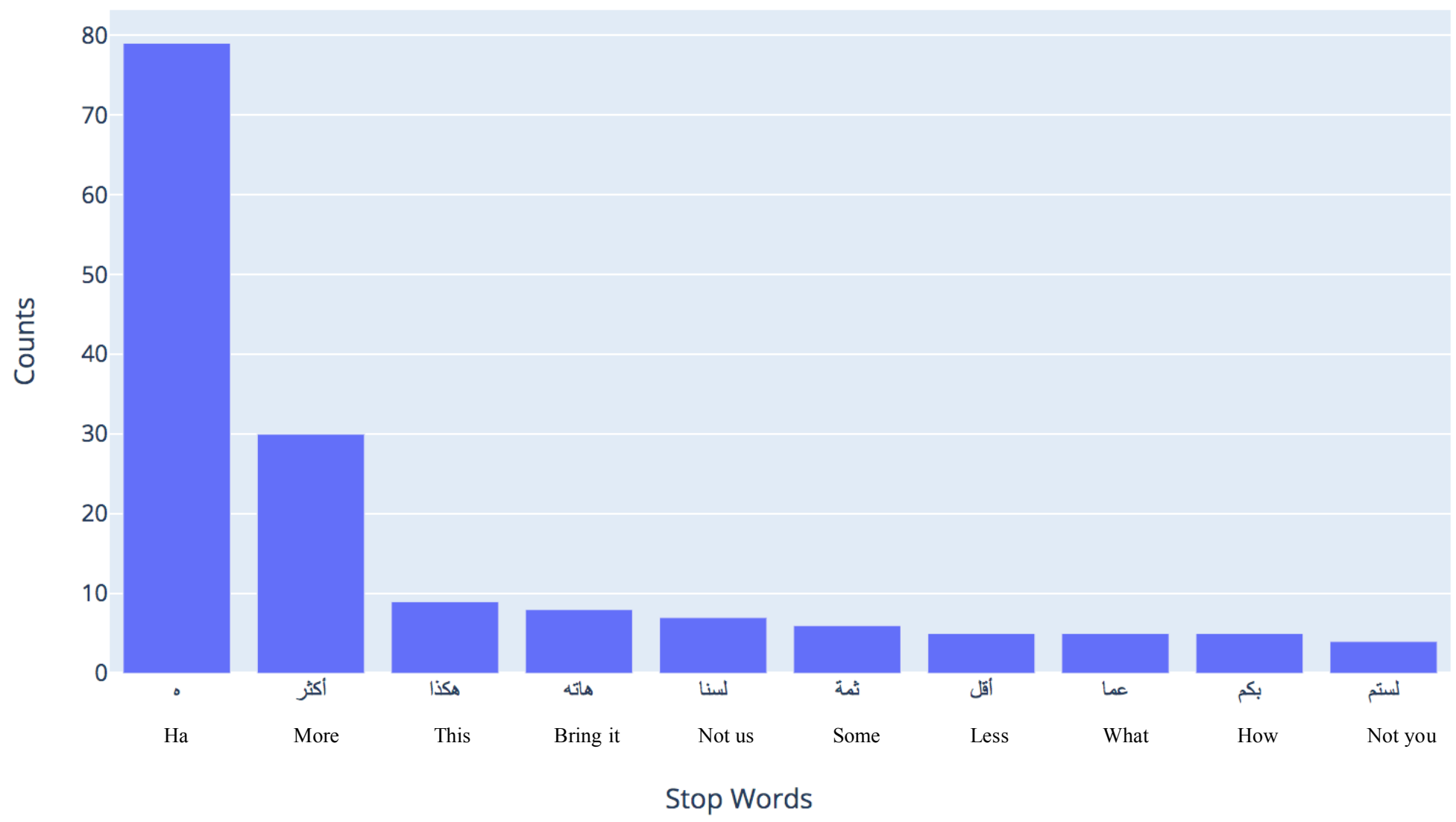

**Figure 82 Most common stop words in normal class from the T-HSAB dataset**

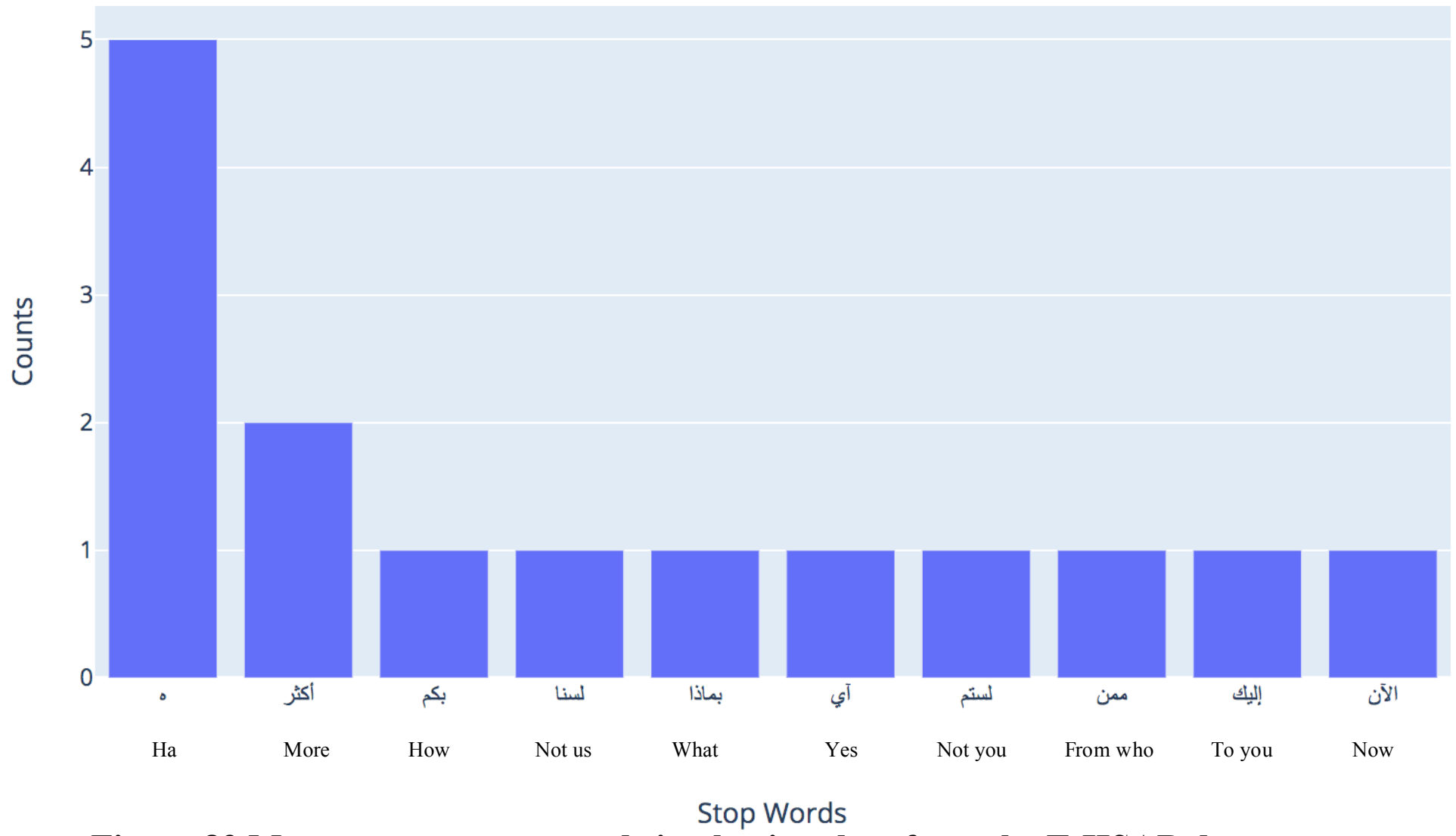

**Figure 83 Most common stop words in abusive class from the T-HSAB dataset**

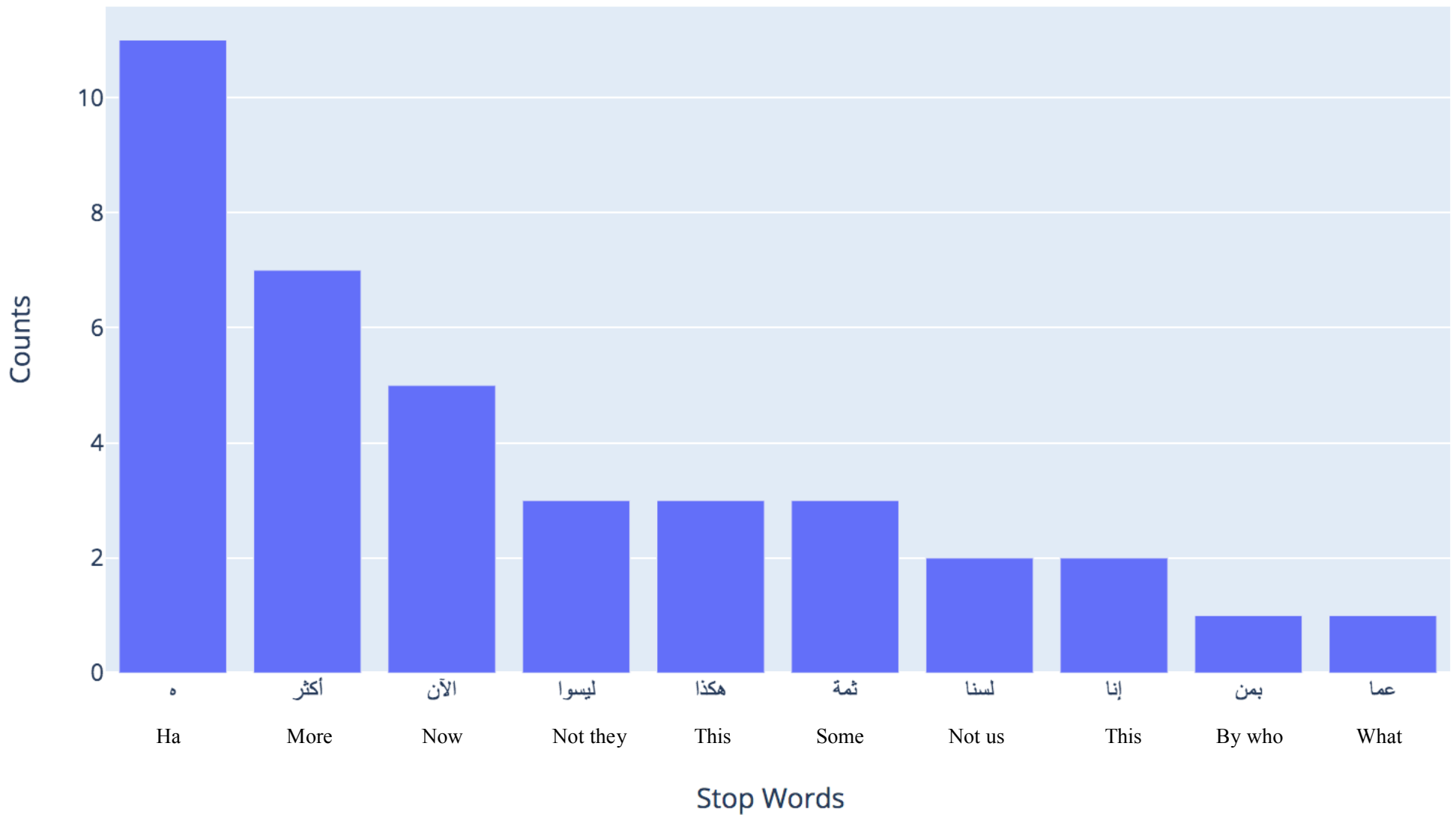

**Figure 84 Most common stop words in hate class from the T-HSAB dataset**

For all classes, the sentiment analysis in Figure 85 highlight the same pattern. Majority of comments are negative, followed by positive and neutral.

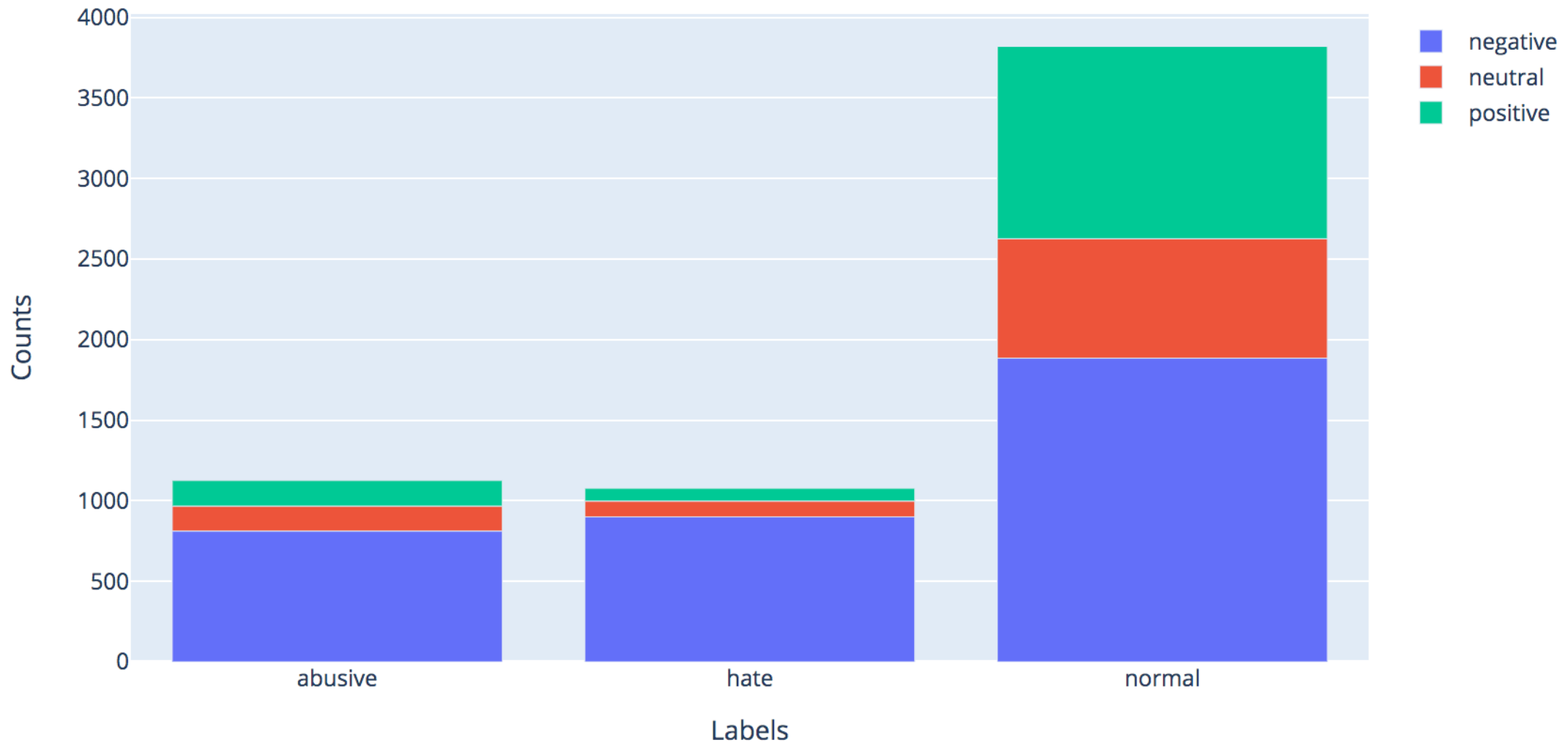

**Figure 85 Sentiment analysis based on labels for the T-HSAB dataset**

6) The Multi-Platform Offensive Language Dataset (MPOLD):

Chowdhury et al. (2020) develop the MPOLD dataset[3]. Figure 86 and 87 plots the classes distribution, the MPOLD dataset consists of 4,000 comments; 3,325 are not offensive comments and 675 are offensive comments. Offensive comments are further classified to vulgar, hate, and other as shown in Figure 87. For the purpose of this research, we just focus on the first labeling hierarchy.

---

[3] https://github.com/shammur/Arabic-Offensive-Multi-Platform-SocialMedia-Comment-Dataset/tree/master/data

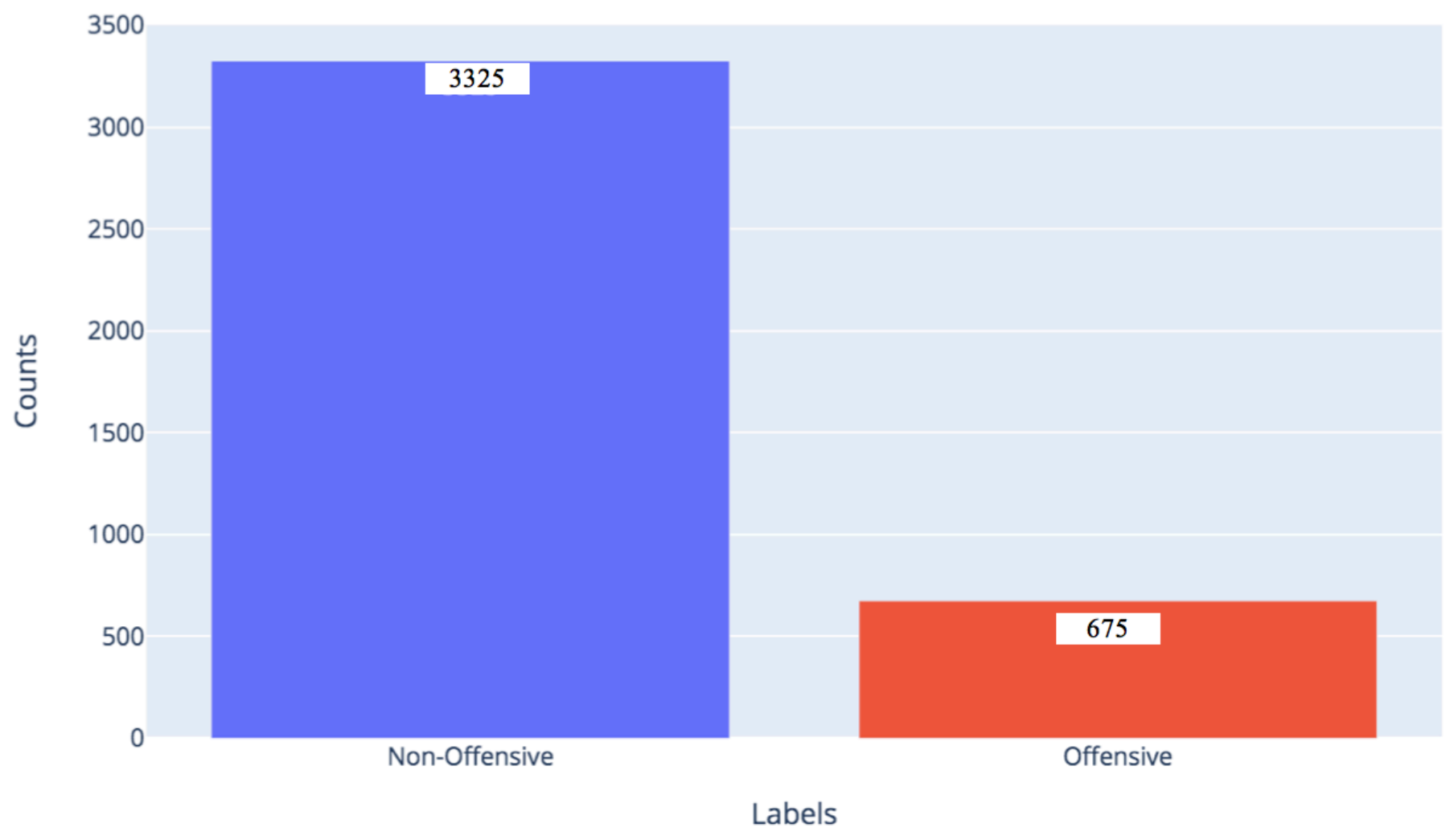

**Figure 86 The first level of labeling distribution for the MPOLD dataset**

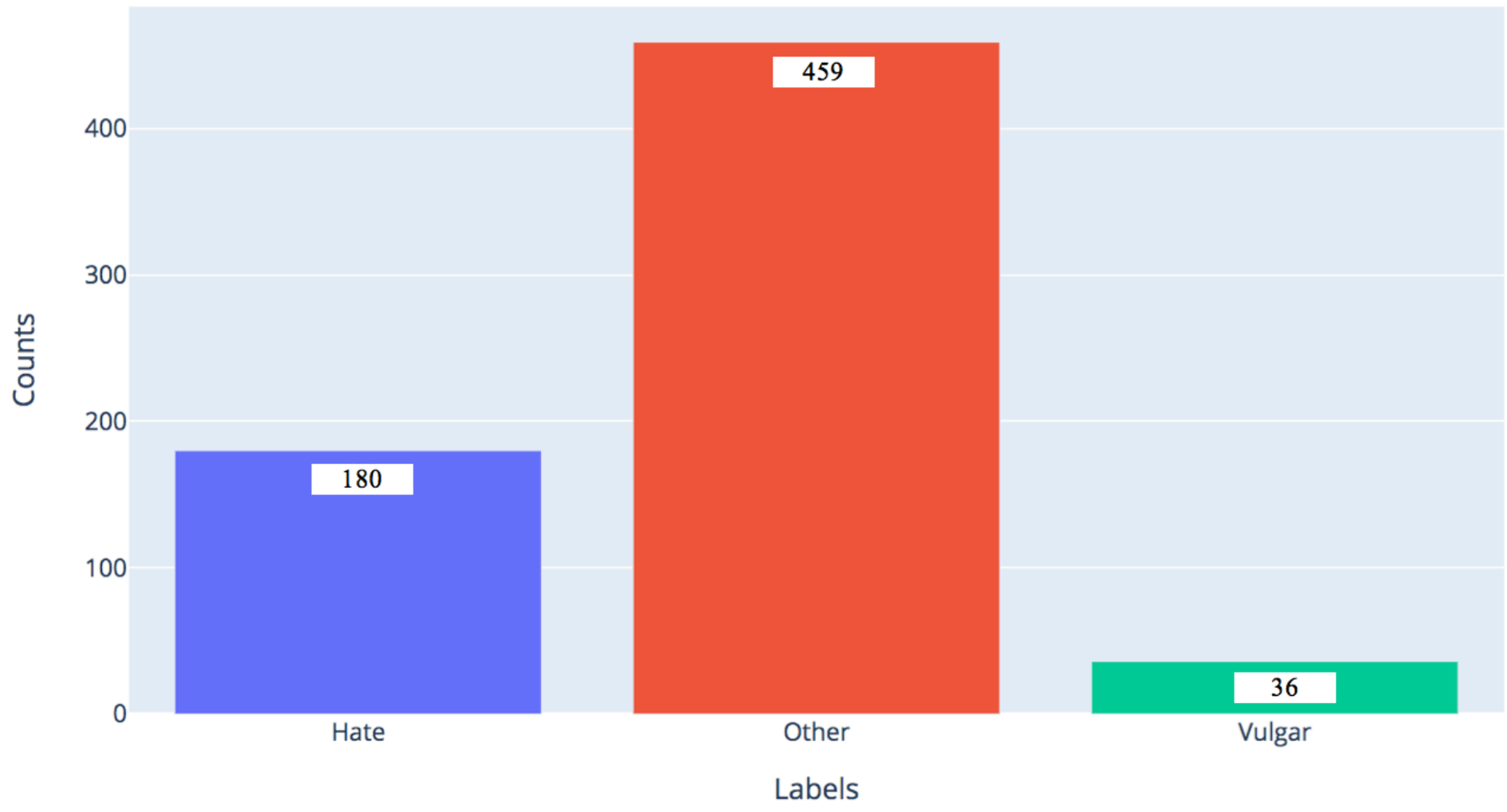

**Figure 87 The second level of labeling distribution for the MPOLD dataset**

Figure 88 shows the word cloud for each class separately. Offensive comments have more user mentions than not offensive comments. Not offensive comments include large numbers of stop words, such as "من / from", "على / on", "في / in", and "ليه / why".

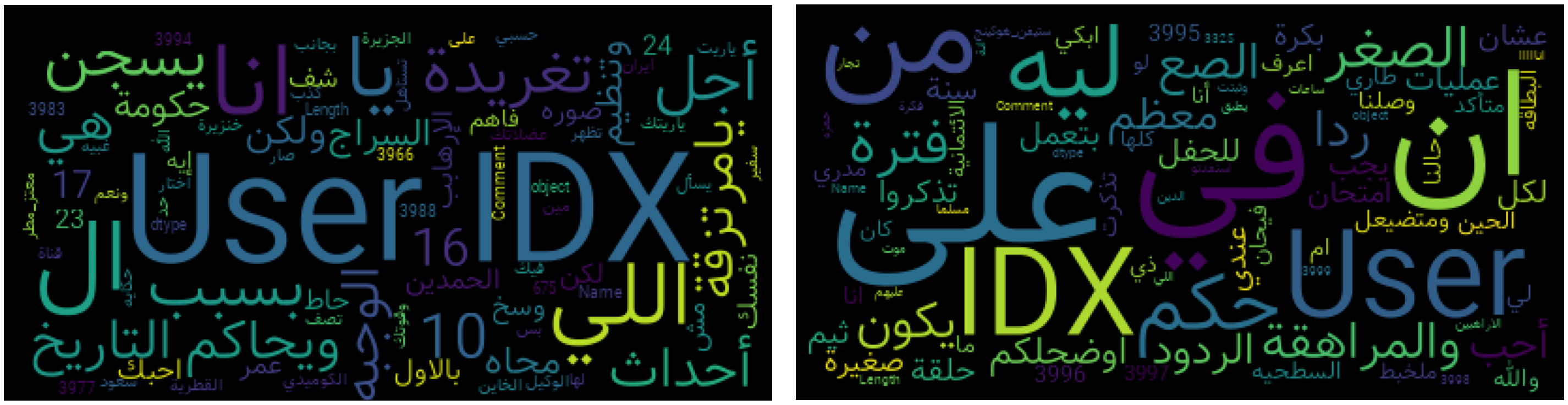

(a) Offensive (b) Not offensive

**Figure 88 The word cloud of the MPOLD dataset (a. offensive, b. not offensive)**

The bar charts in Figure 89 and 90 are consistent with the word cloud, in both classes the most frequent token is "@User.IDX", which is used to anonymize username mentions. Moreover, both figures have the exact same first four top frequent tokens.

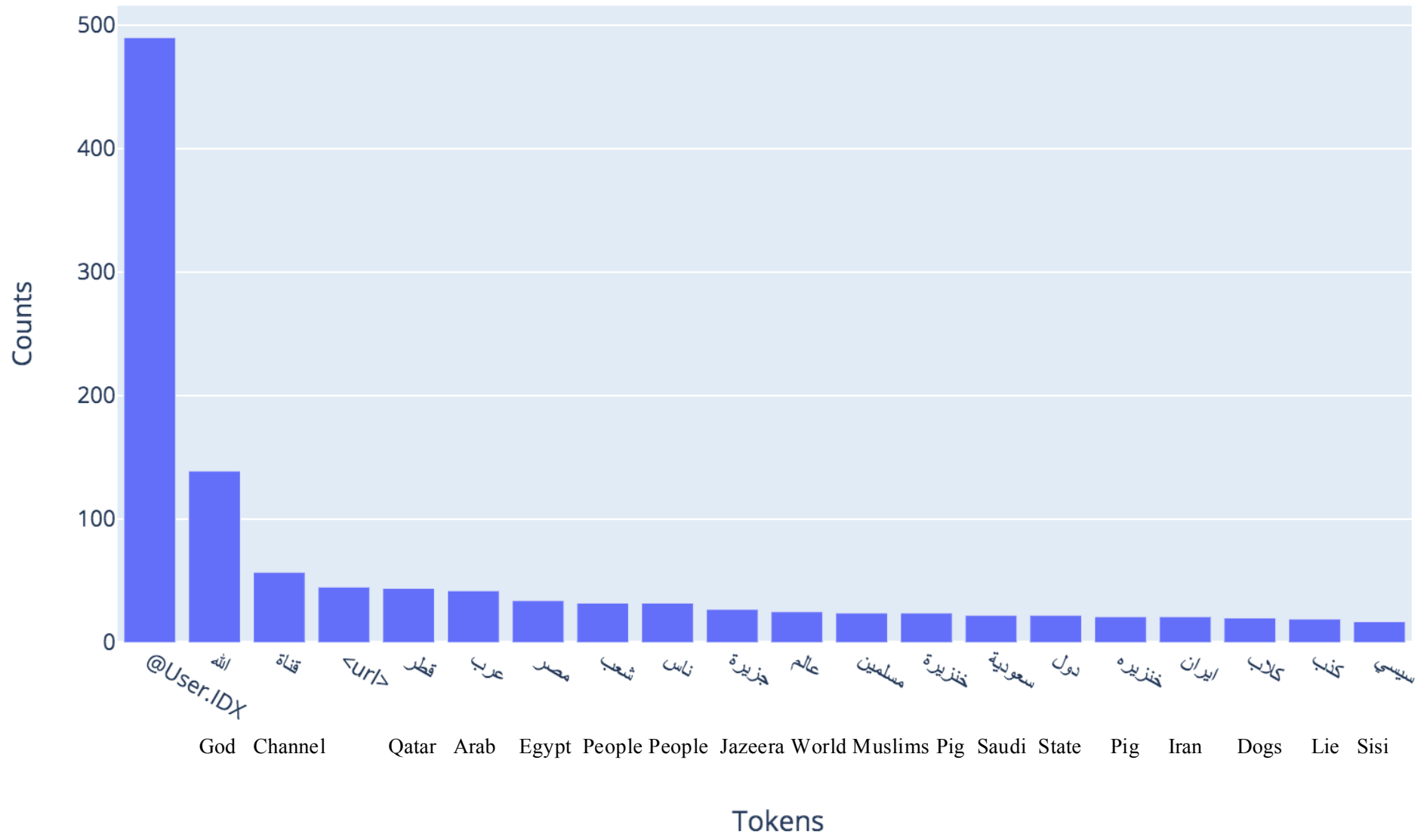

**Figure 89 : Most common not offensive tokens in the MPOLD dataset**

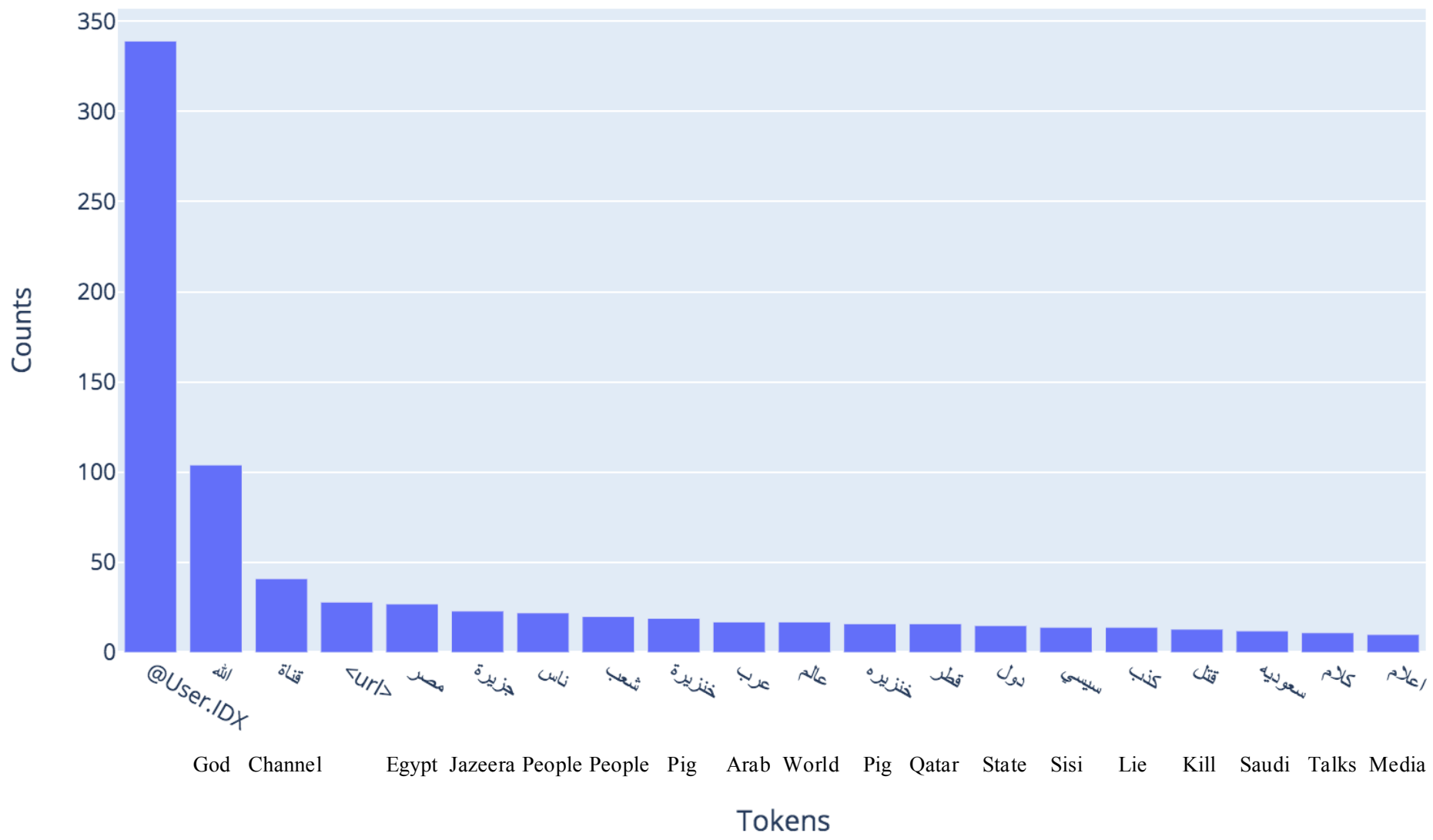

**Figure 90 : Most common offensive tokens in the MPOLD dataset**

Distinctive words are as the followings:

1. Not offensive: شكرا / thanks (46), برنامج / program (22), افضل, better (22), 1 (22), دون / without (21).
2. Offensive: الخنزيره / pig (21), الحمدين / the two Hamad (10), قذر / dirty (8), الكلب / the dog (8), المرتزقة / mercenaries (7).

Results from the statistical analysis in Figures 91 and 92 for the length of comments and tokens report an overall limited content for the entire dataset with more outliers among the not offensive comments.

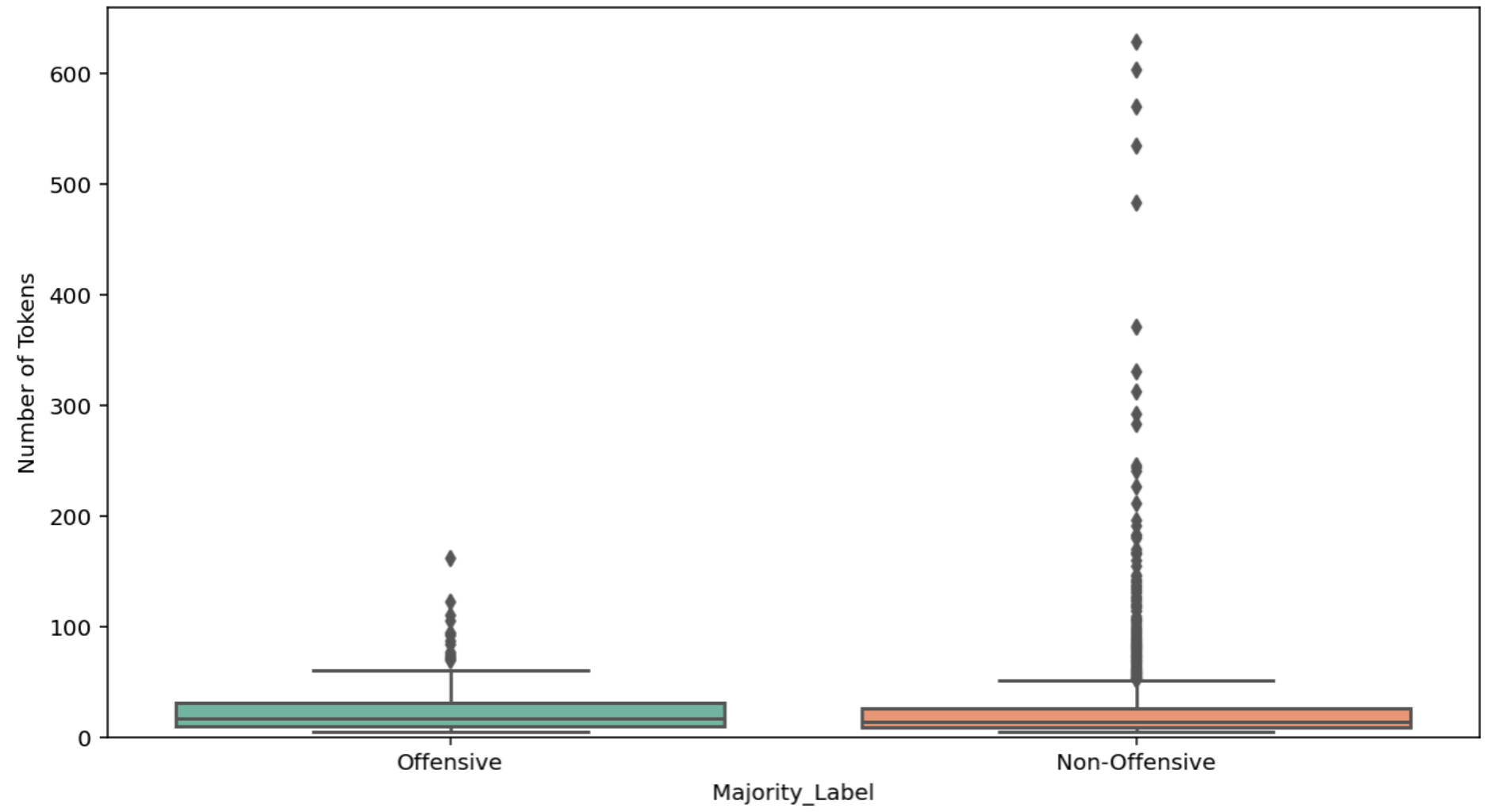

**Figure 91 : Statistics of each label in the MPOLD dataset based on the number of tokens per comment**

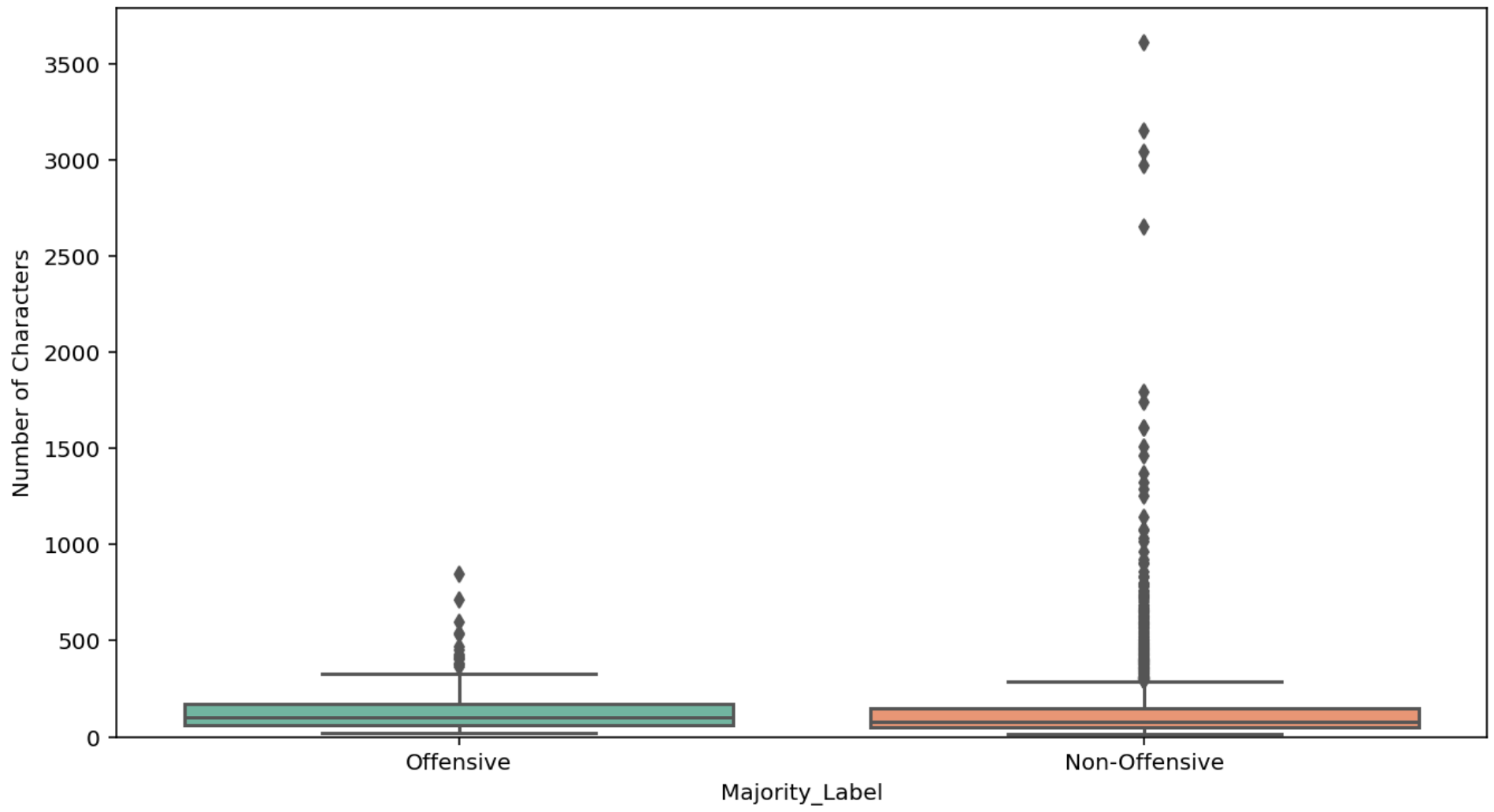

**Figure 92 : Statistics of each label in the MPOLD dataset based on the number of characters per token**

Very similar pattern of using stop words is observed in Figures 93 and 94 for the two classes with “من / from”, “في / in”, and “و / and” are the used stop words.

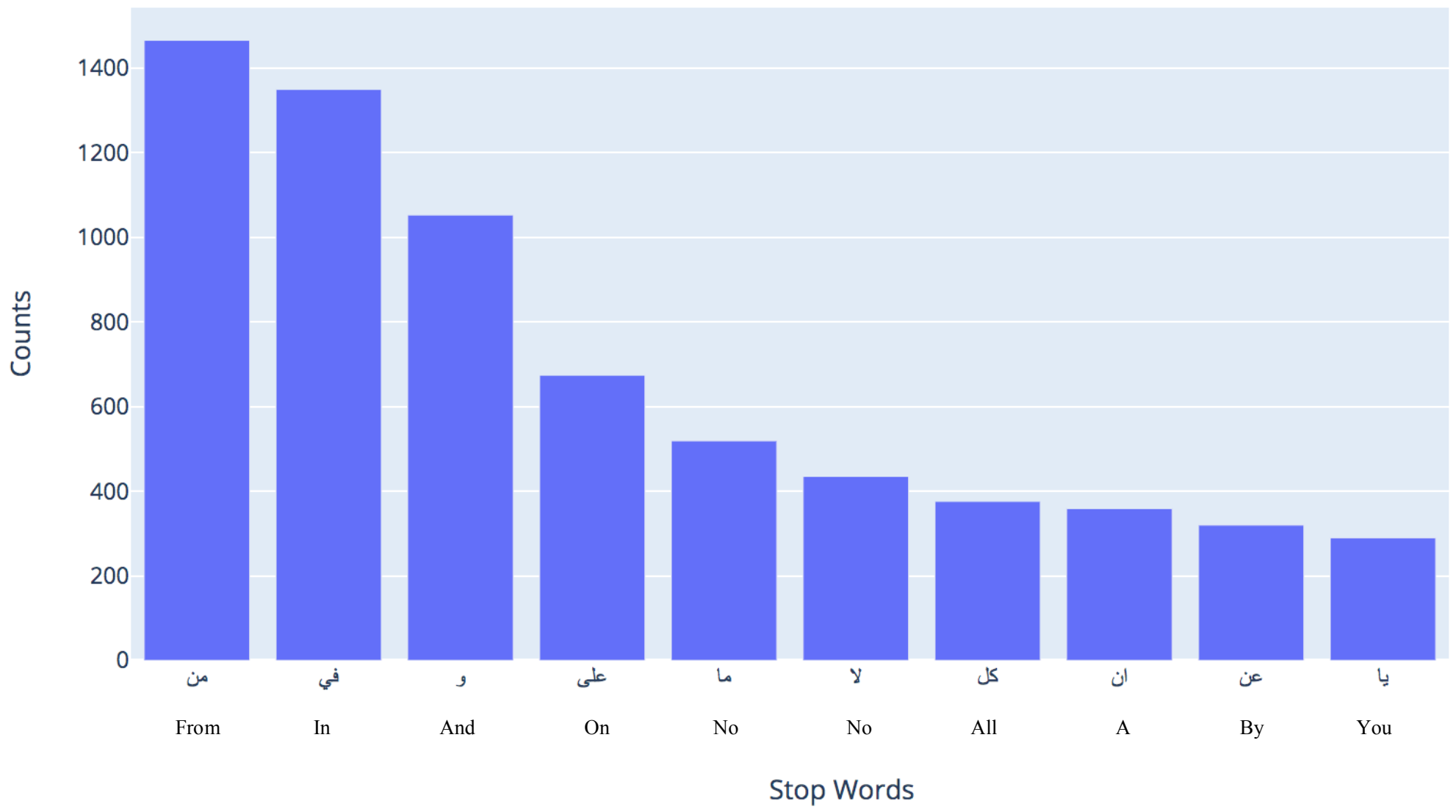

**Figure 93 : Most common stop words in the not offensive class from the MPOLD dataset**

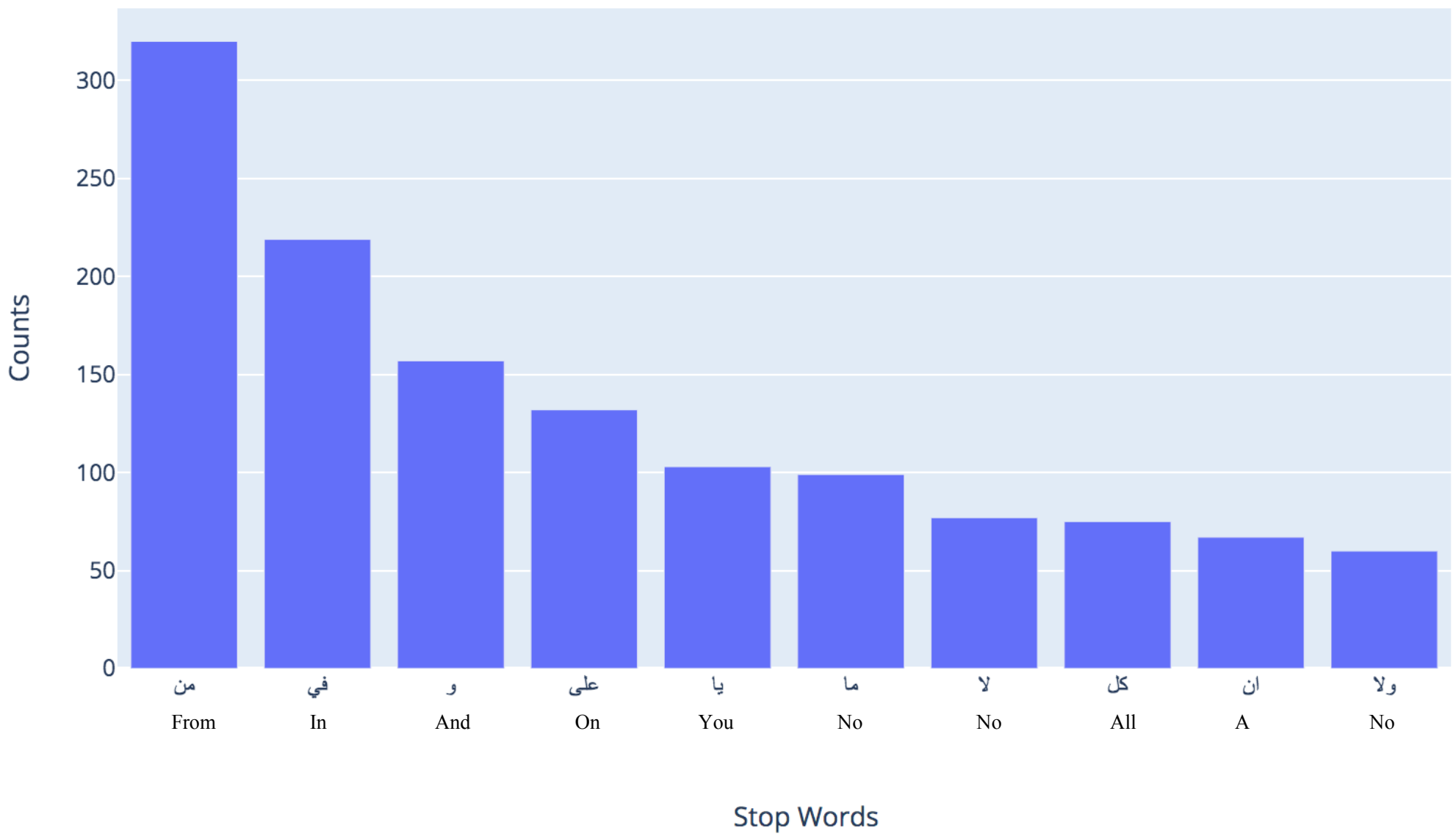

**Figure 94 : Most common stop words in the offensive class from the MPOLD dataset**

Both classes have majority of negative sentiment as shown in Figure 95. Positive sentiment is more than neutral sentiment in both classes.

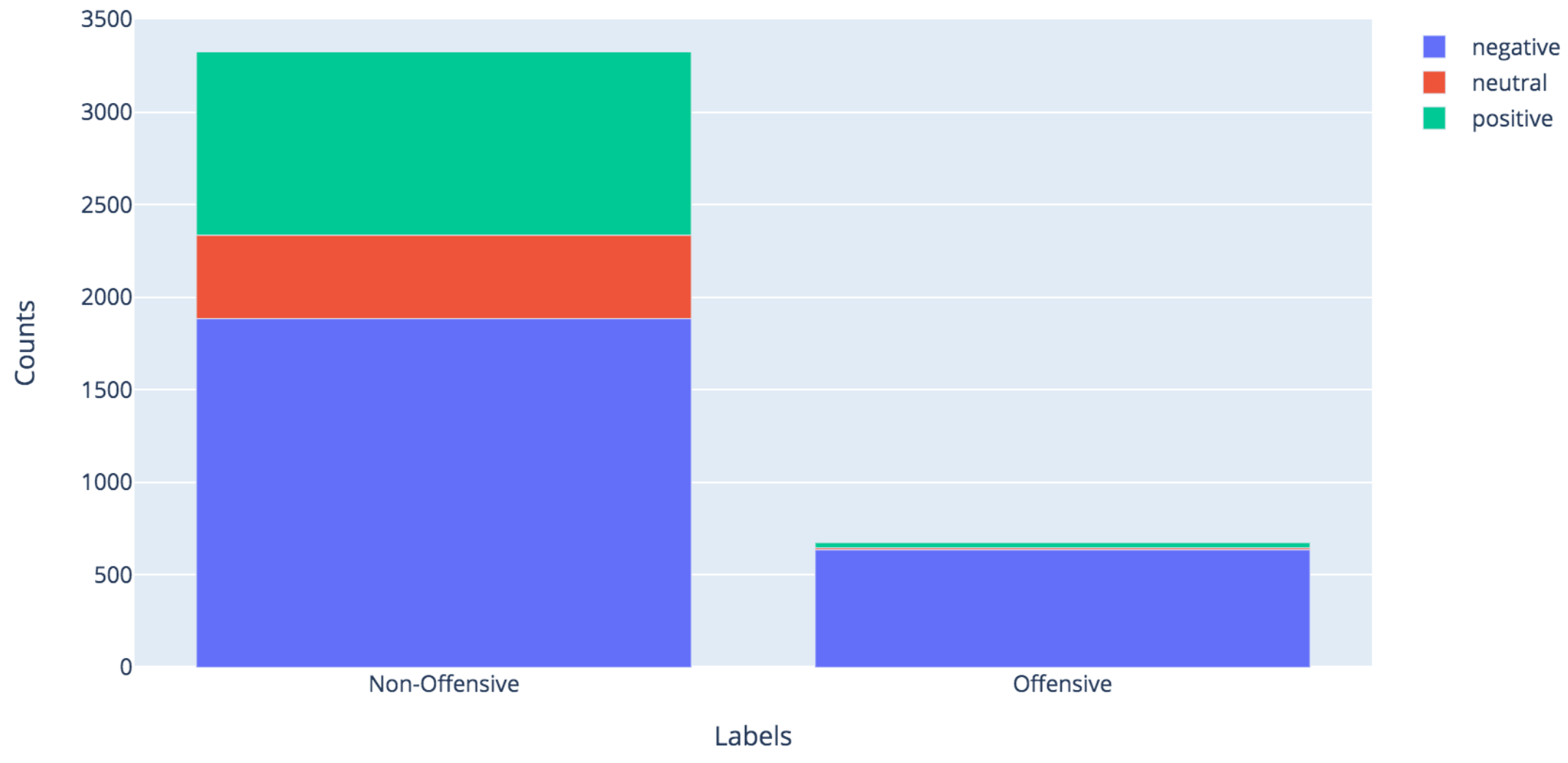

**Figure 95 : Sentiment analysis based on labels for the MPOLD dataset**

Figure 96 and 97 contain bar charts for the top ten used emojis. From both classes, only one emoji is used sharply more than others in both classes. This most commonly used emoji is the face with tears of joy emoji, "😂", which is also consistent with the results from other datasets mentioned previously. The black heart emoji, "🖤", and the thumps up emoji, "👍" also ranked among the top frequently used emoji for not offensive comments, while the backhand index pointing down emoji, "👇", and the rolling on the floor laughing emoji, "🤣", for the offensive comments.

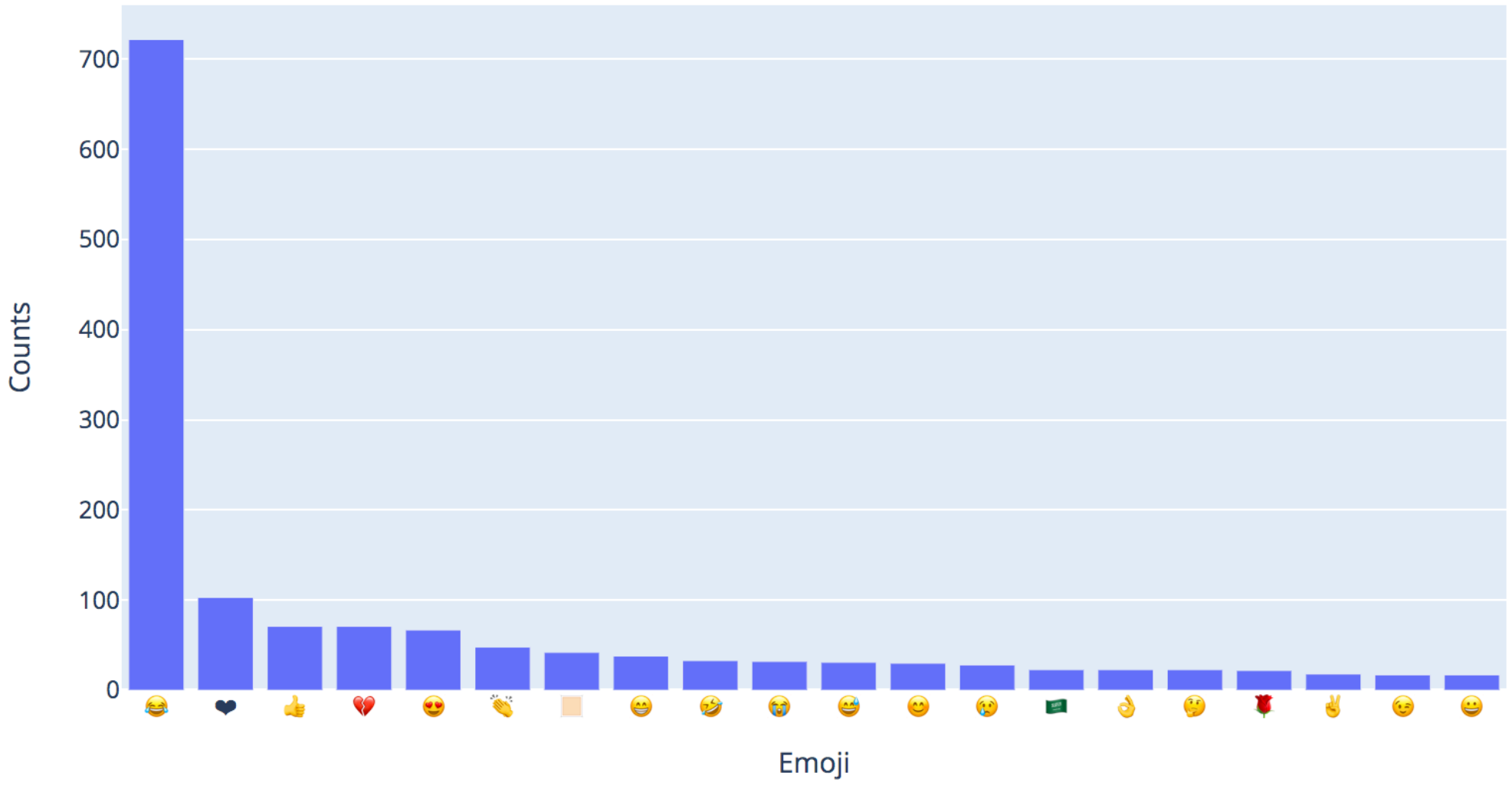

**Figure 96 : Most common not offensive emojis in the MPOLD dataset**

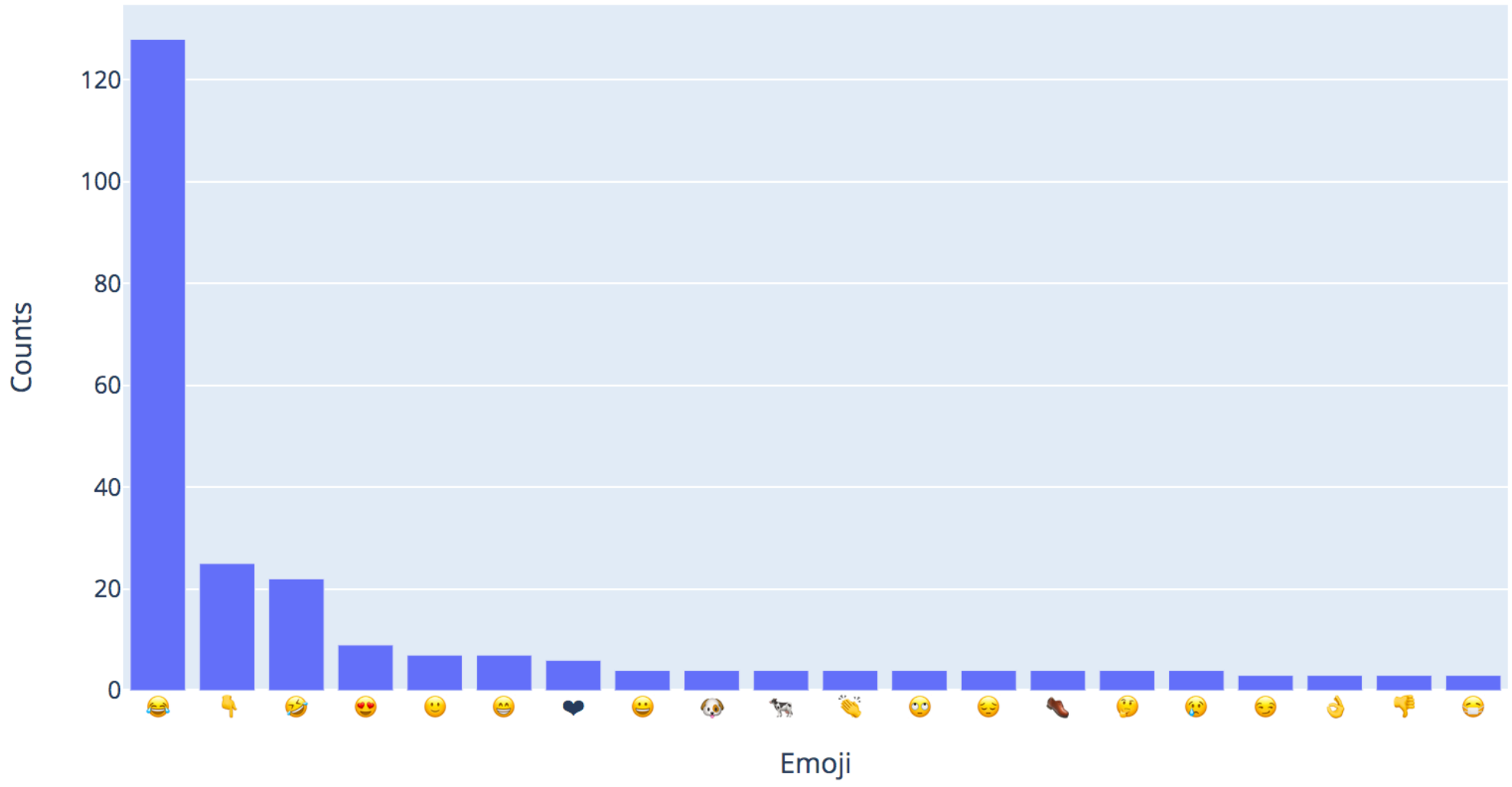

**Figure 97 : Most common offensive emojis in the MPOLD dataset**

The use of punctuations among the offensive and not offensive comments is illustrated in Figures 98 and 99. From both Figures, the "." is used sharply more than the rest of the punctuation. Secondly top used punctuation is "@" in both classes, followed by "،" in not offensive comments and "!" in offensive comments.

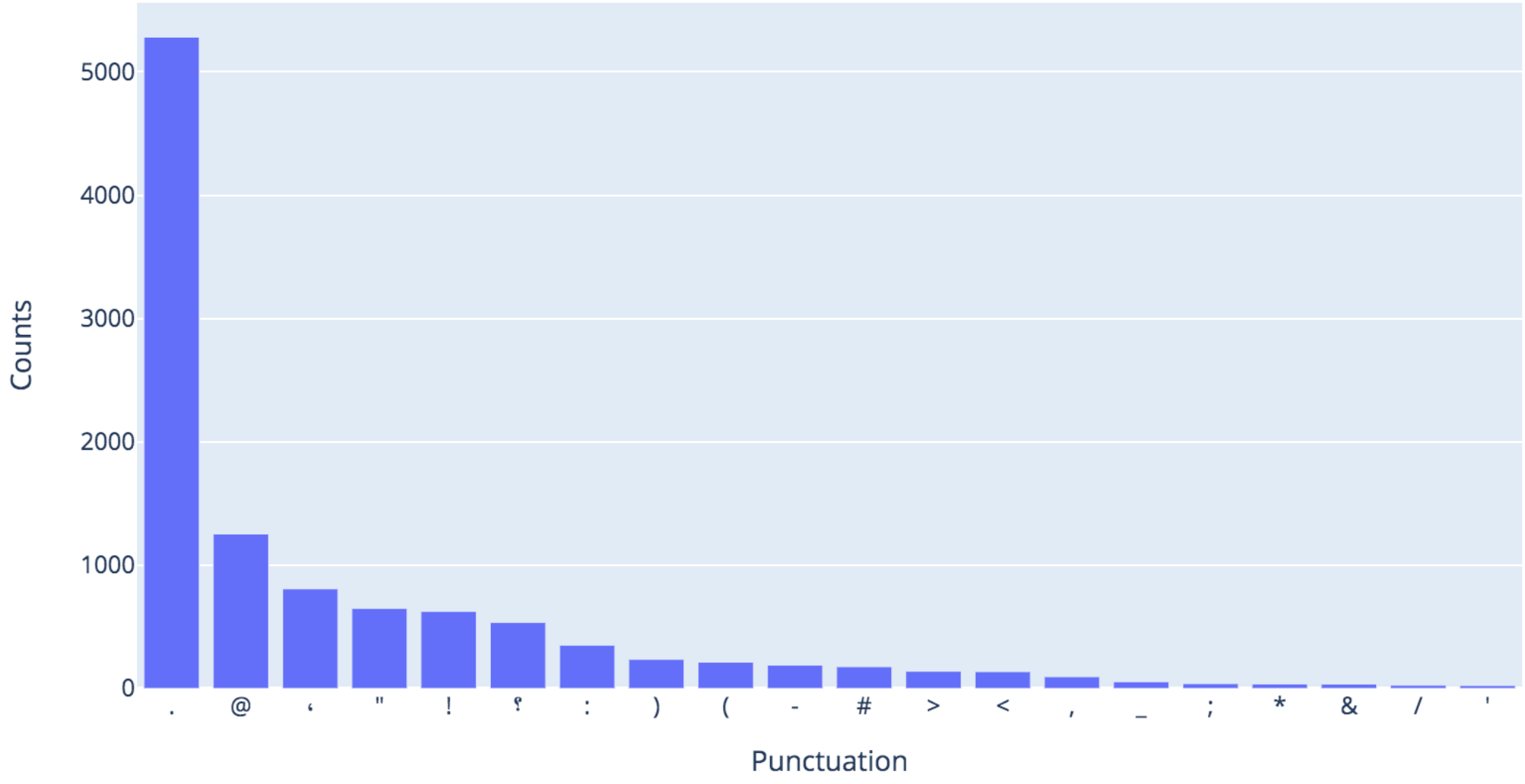

**Figure 98 : Most common not offensive punctuation in the MPLOT dataset**

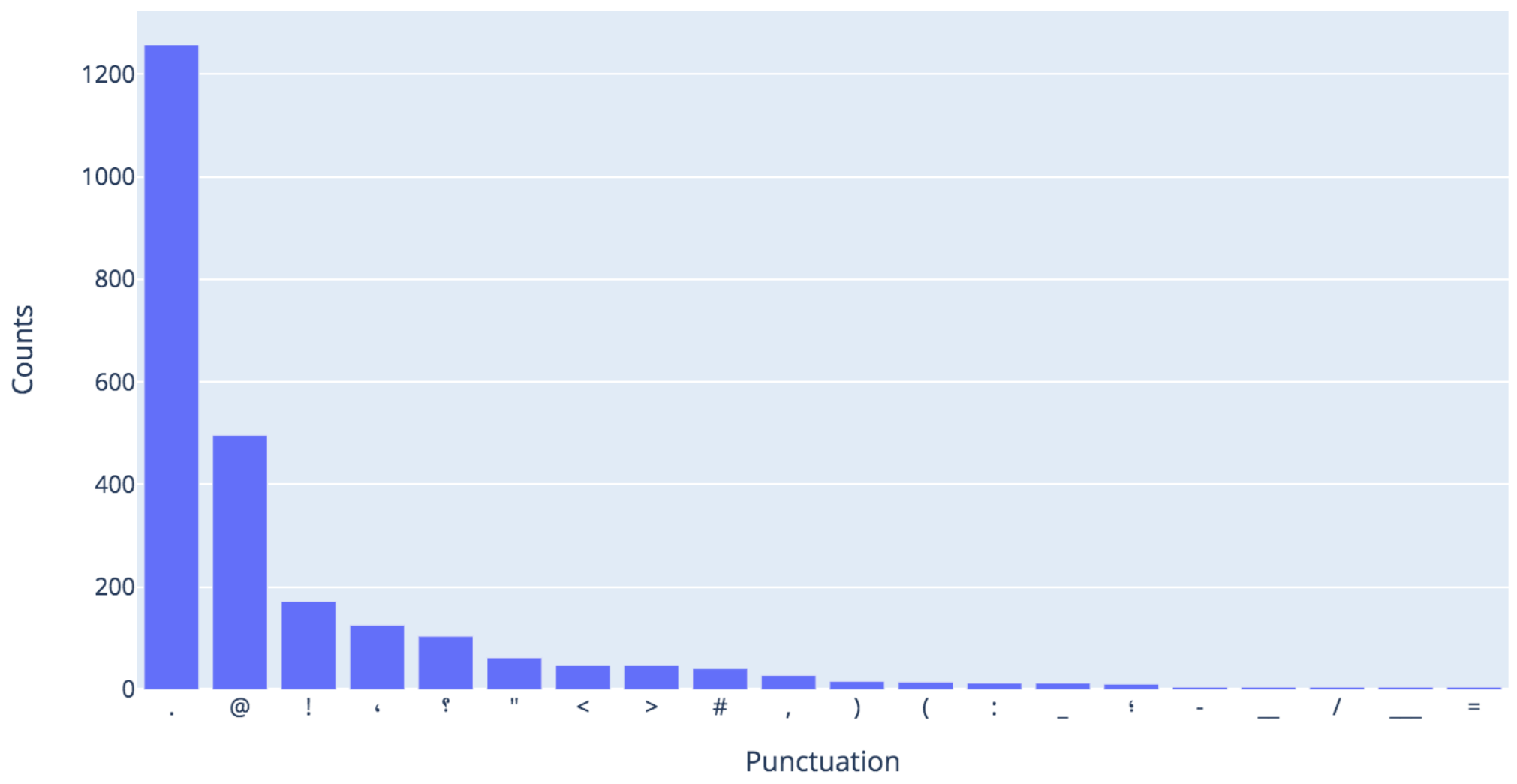

**Figure 99 : Most common offensive punctuation in the MPLOT dataset**

7) The Fourth Workshop on Open-Source Arabic Corpora and Corpora Processing Tools Dataset (OSACT4):

The OSACT4 dataset is released by Mubarak et al. (2020). Each tweet has two labels; the first label used to classify the tweet into offensive or not offensive and the second one to classify it into hate speech or not hate speech. As can be noticed from Figure 100 and 101, the dataset is imbalanced. The total number of tweets is 10,000; 1,900 are offensive tweets and out of these offensive tweets, only 500 are hate speech. For the purpose of this research, we consider only the first level of labeling in the analysis. Tweets are preprocessed to replace user mentions with @USER, URLs with URL, and empty lines with <LF>. The same dataset is used for OffenEval 2020 Arabic shard task but with the binary offensive or not offensive labels only. The dataset has multiple duplicated tweets, for example, the following one is one of the duplicated not offensive tweets from the dataset:

RT @USER: يا زمالك يا مدرسة لعب وفن وهندسة ، الف مبروك صعود نادى الزمالك 👏👏👏👏

Translation: RT @USER Zamalik is a school for playing, art, and engineering, congratulation for the outperformance of Zamalik team 👏👏👏👏

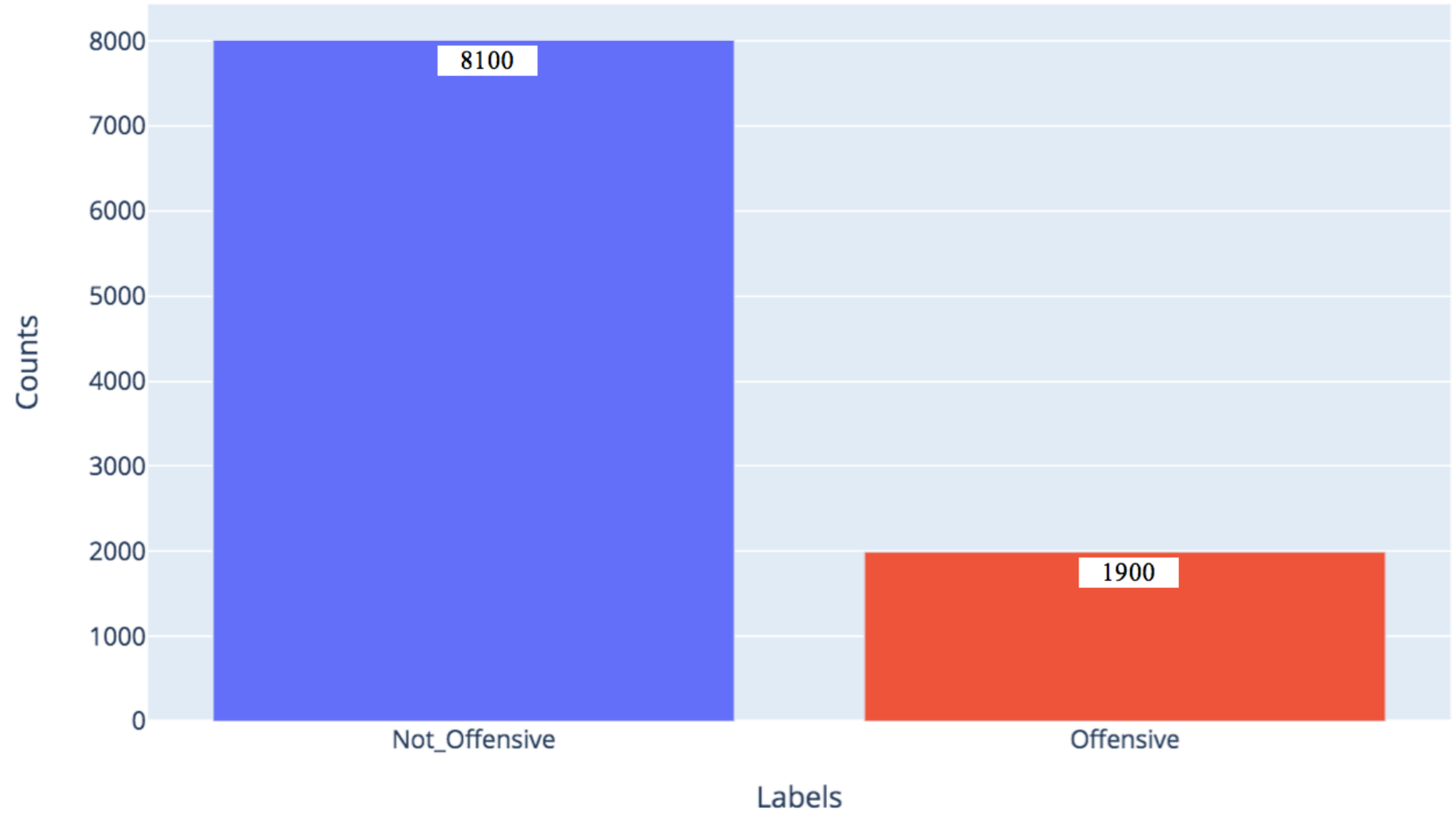

**Figure 100 : The first label distribution for the OSACT4/OffensEval 2020 Arabic dataset**

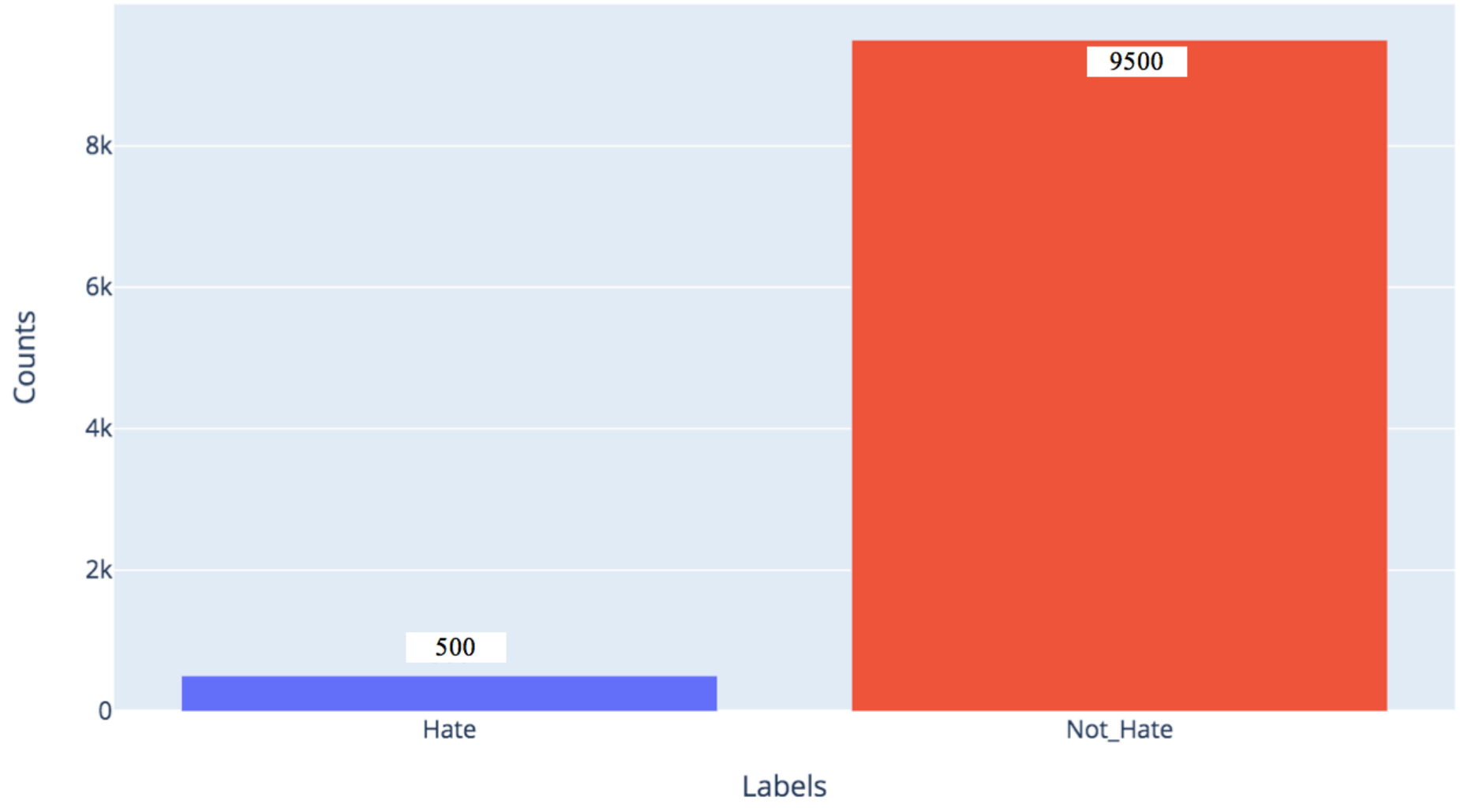

**Figure 101 : The second label distribution for the OSACT4/OffensEval 2020 Arabic dataset**

The word clouds from Figure 102 highlight a common criterion in both class, which is the high frequency of username mentions. Another common feature from both word cloud is the large occurrence of the particle “يا / you”.

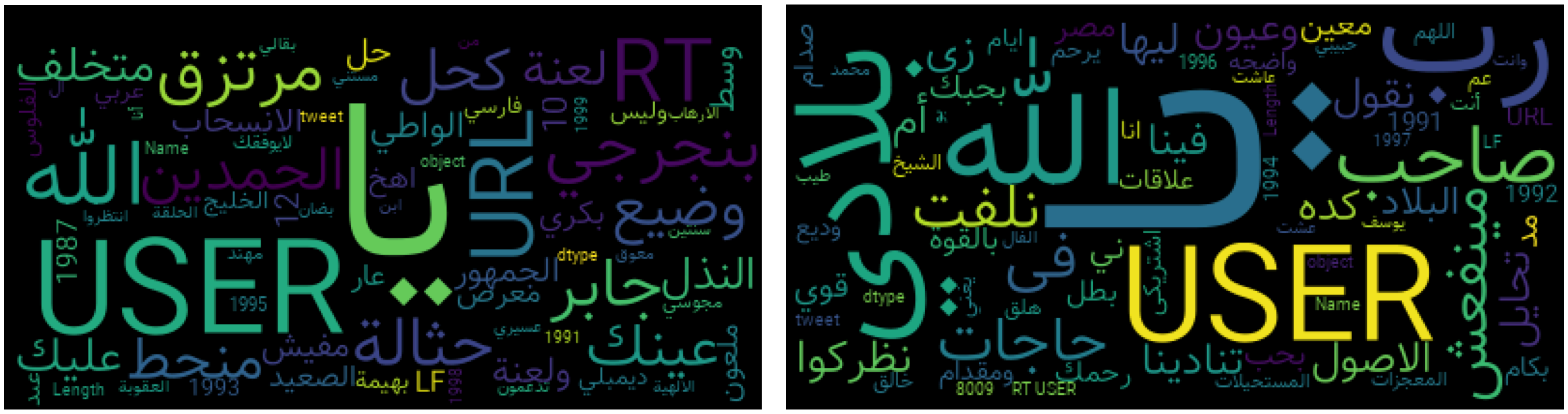

(a) Offensive (b) Not offensive

**Figure 102 : The word cloud of the OSACT4/OffensEval 2020 Arabic dataset (a. offensive, b. not offensive)**

The bar charts in Figure 103 and 104 plot the same most frequent token in both classes, “user”, which refers to username mentions. Offensive tweets also show “الله / God” as the second most frequent token, while not offensive tweets show “RT” secondly, which illustrate the large retweeted content. Both classes have the same token for the third most frequent one, which is “URL”.

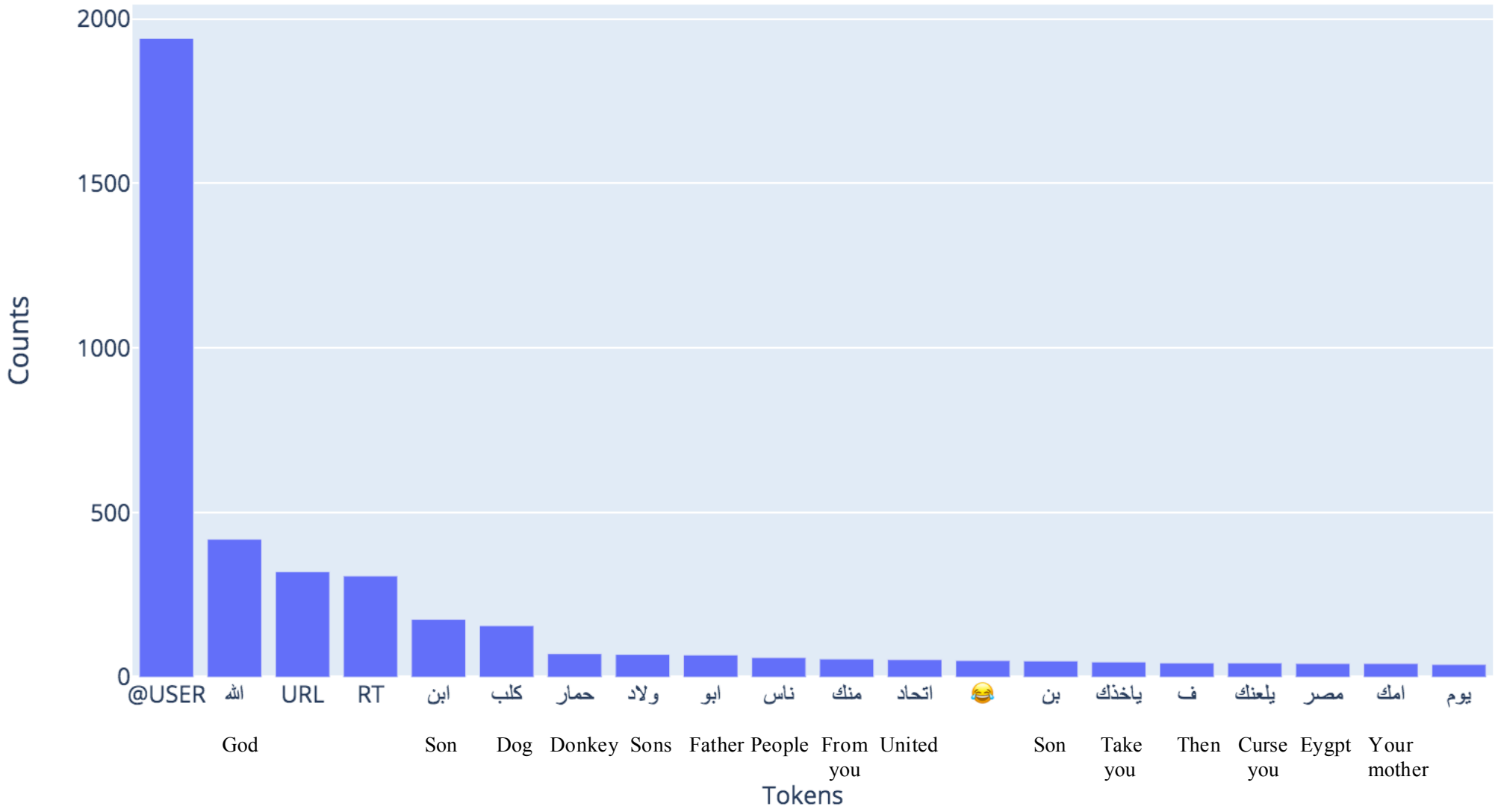

**Figure 103 : Most common offensive tokens in the OSACT4/OffensEval 2020 Arabic dataset**

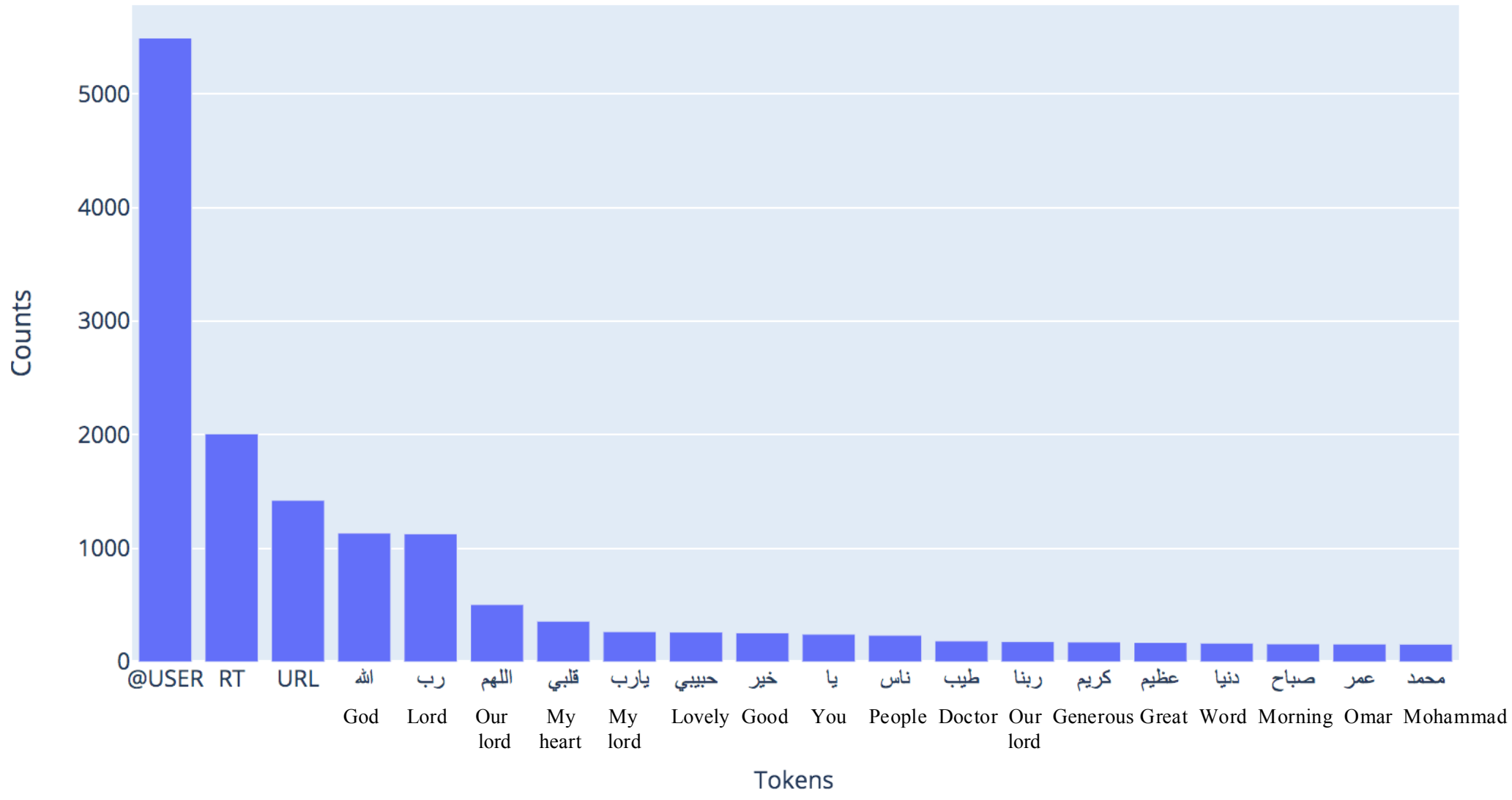

**Figure 104 : Most common not offensive tokens in the OSACT4/OffensEval 2020 Arabic dataset**

Results of distinctive words list the following top five words:

1. Offensive: فسواي / shit (20), كلووووب / small dog (12), كلبه / female dog (11), الغيري (10), بجم / how much (9).
2. Not offensive: قيوم / everlasting (111), مجيب / answerer (91), رحمن / merciful (88), الراحمين / the merciful (84), برحمتك / by your mercy (70).

The length of not offensive tweets is more scattered than that of the offensive tweets, but with smaller average length. Similar length characteristics applied to the length of the token for each class. Figures 105 and 106 provide more detailed information.

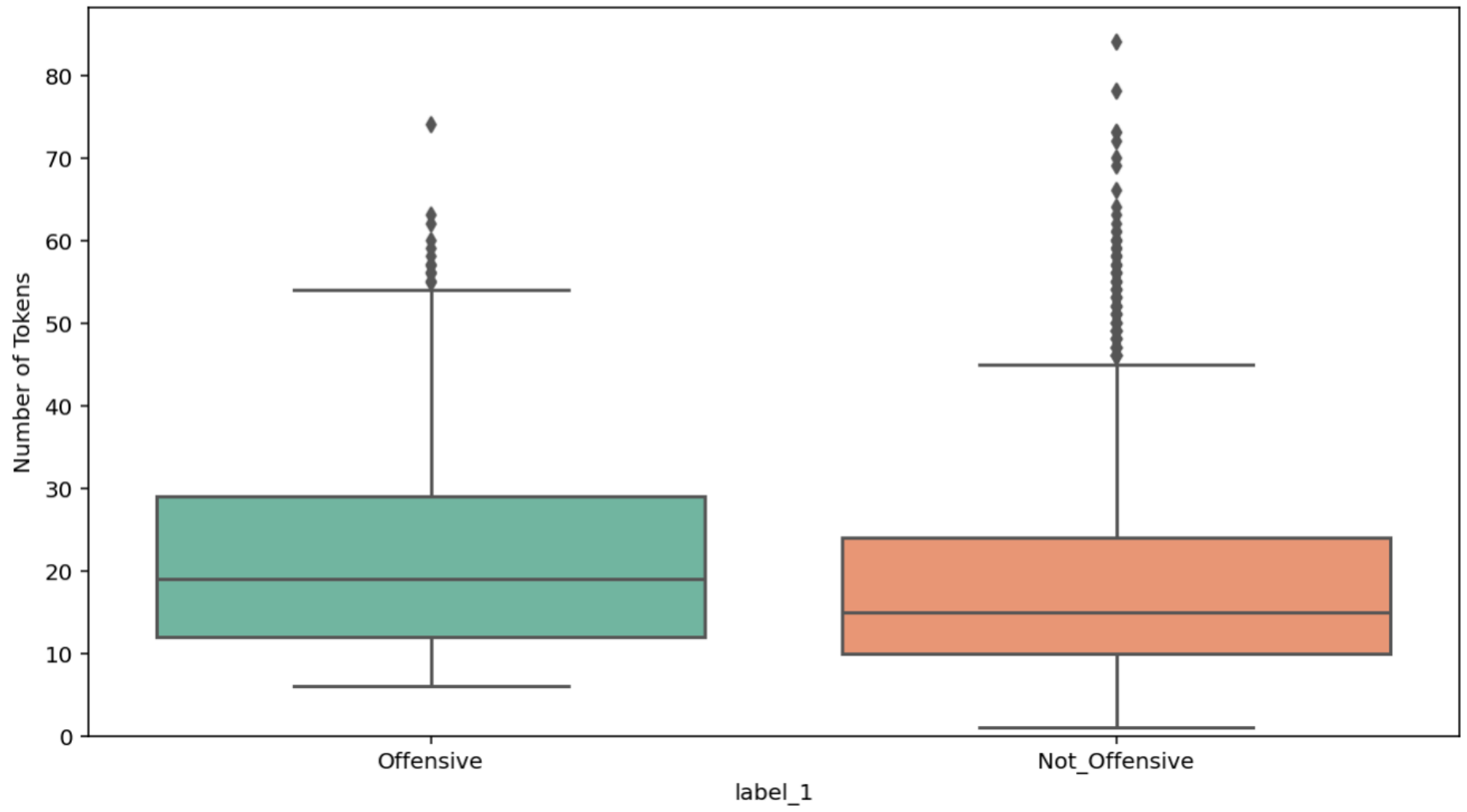

**Figure 105 : Statistics of each label in the OSACT4/OffensEval 2020 Arabic dataset based on the number of tokens per tweet**

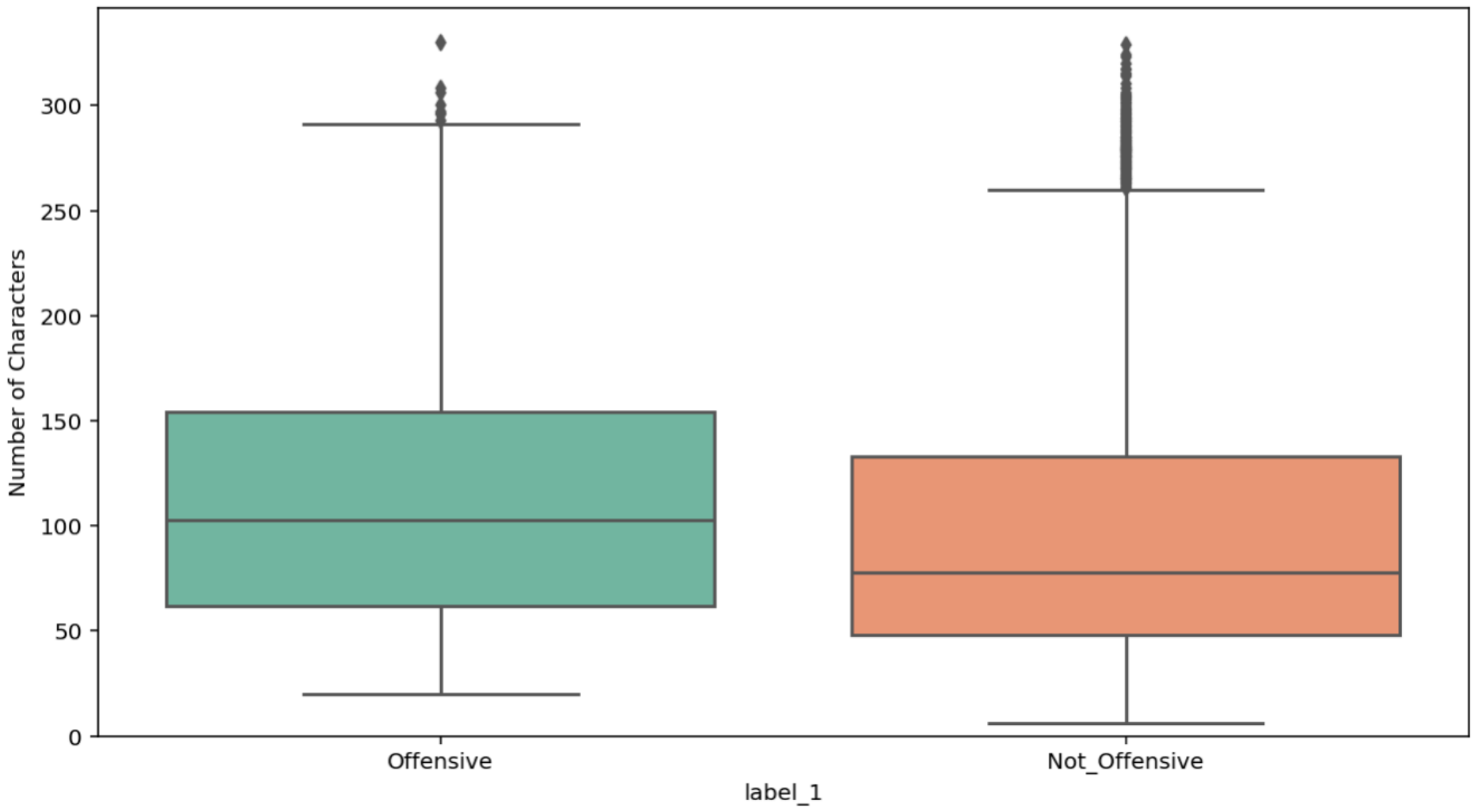

**Figure 106 : Statistics of each label in the OSACT4/OffensEval 2020 Arabic dataset based on the number of characters per token**

The distributions of stop words are very similar for both offensive and not offensive tweets. Only the particle "يا / you" shows a very high frequency in both Figures 107 and 108, and the other top frequent stop words are "من / from" and "و / and".

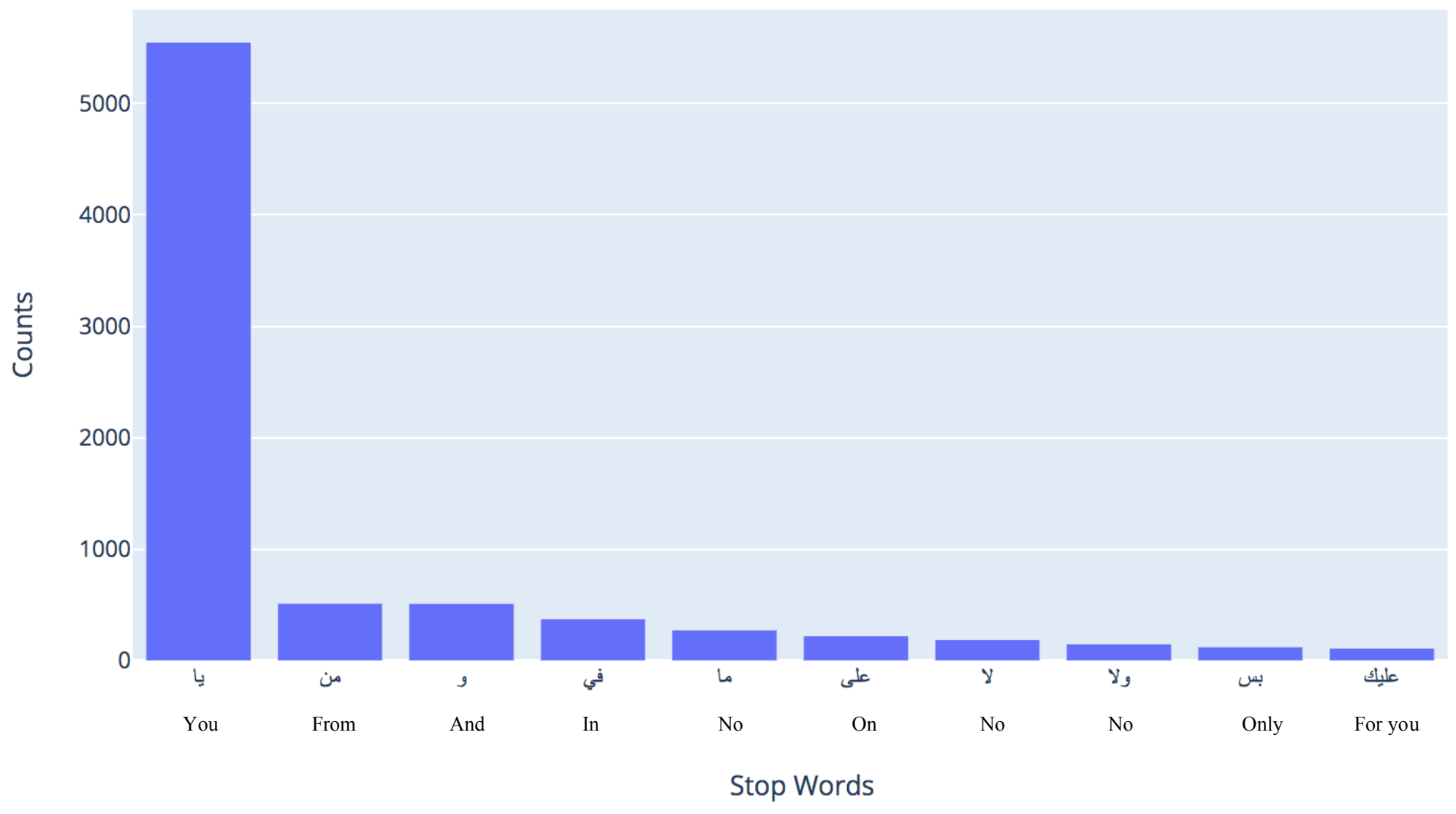

**Figure 107 : Most common stop words in offensive class from the OSACT4/OffensEval 2020 Arabic dataset**

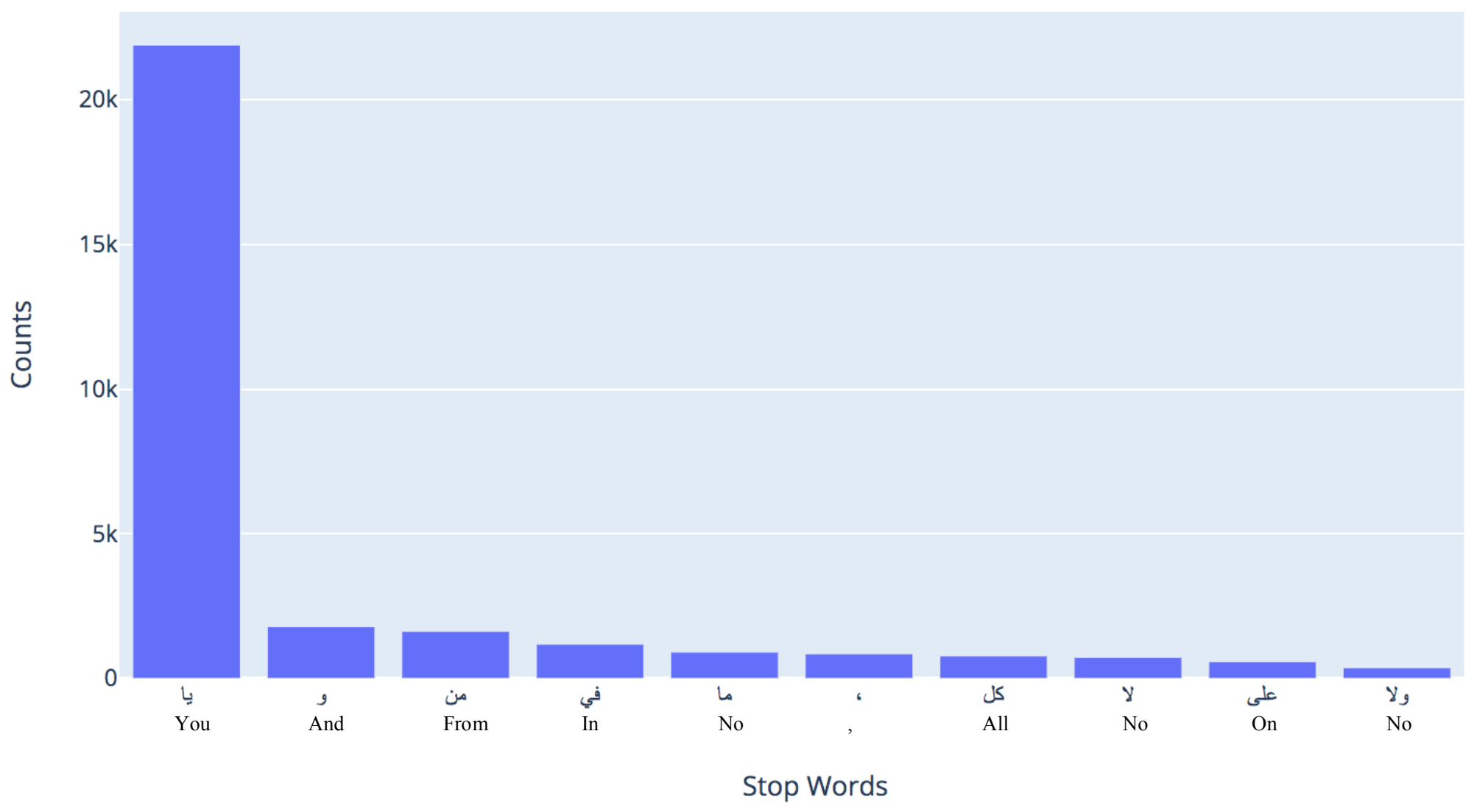

**Figure 108 : Most common stop words in not offensive class from the OSACT4/OffensEval 2020 Arabic dataset**

From the sentiment analysis bar chart in Figure 109, not offensive tweets are mostly positive, while the offensive tweets are mostly negative. Both classes have very small number of neutral sentiment tweets.

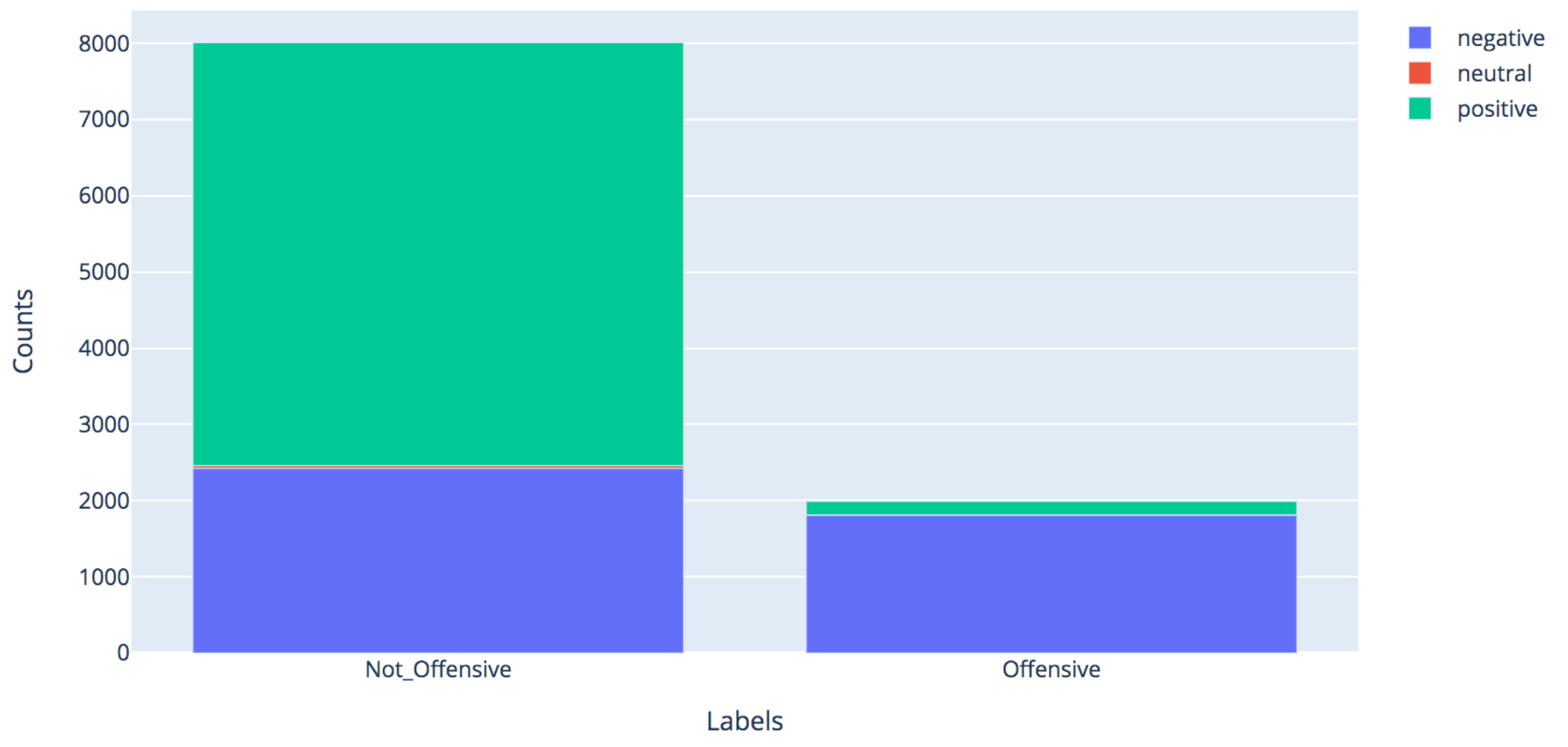

**Figure 109 : Sentiment analysis based on labels for the OSACT4/OffensEval 2020 Arabic dataset**

Like the previous datasets, the face with tears of joy emoji, "😂", is sharply more frequent than the other emoji for all classes. Results of offensive tweets secondly record the floor laughing emoji, "🤣", while not offensive tweets record the black heart emoji, "🖤". The same emoji, the blue heart emoji "💙", shows as the top third most frequent emoji for all classes. Figures 110 and 111 present more detailed information about the use of emoji.

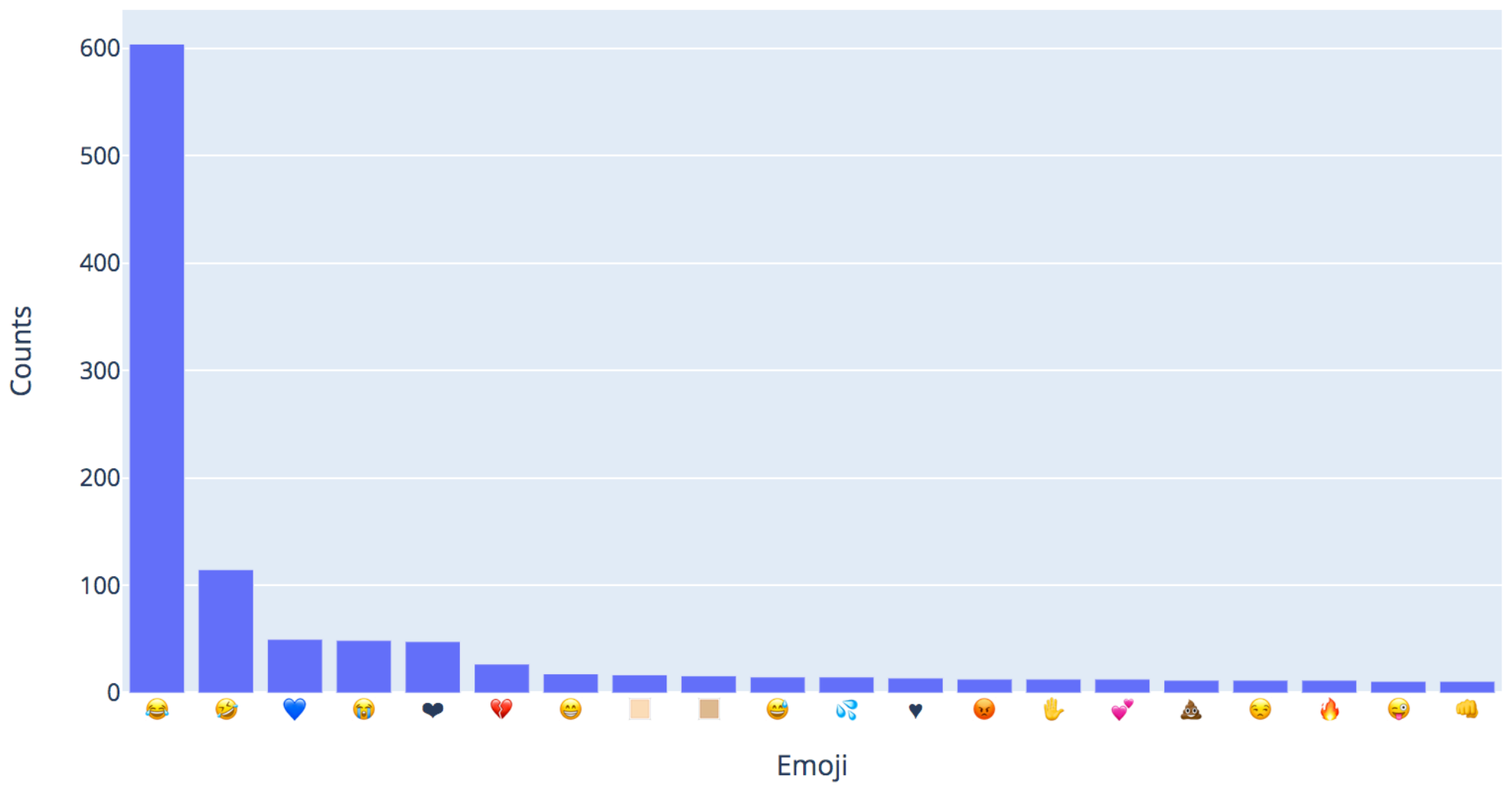

**Figure 110 : Most common offensive emojis in the OSACT4/OffensEval 2020 Arabic dataset**

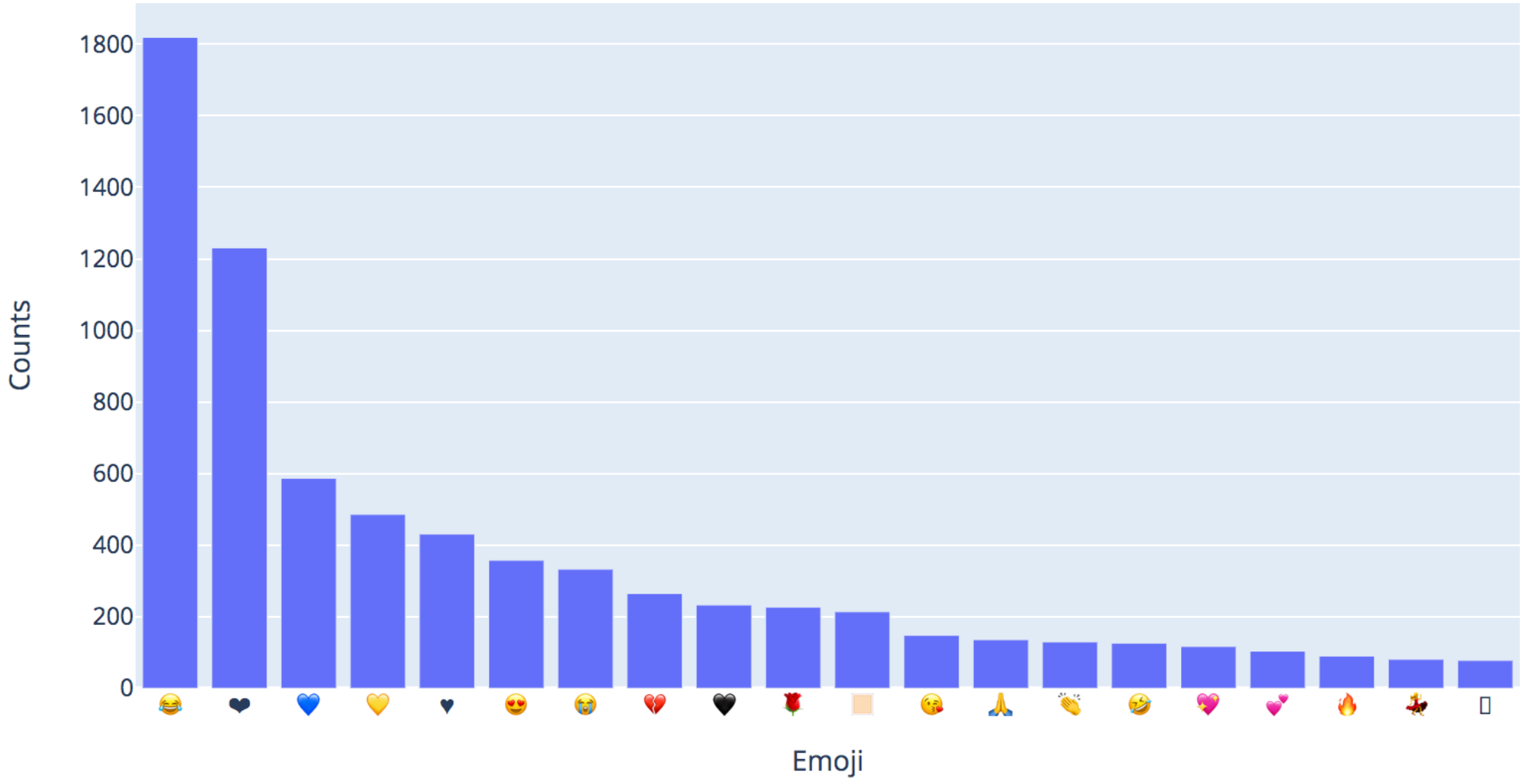

**Figure 111 : Most common not offensive emojis in the OSACT4/OffensEval 2020 Arabic dataset**

Punctuation frequencies do not highlight pattern that could be related to offensive content. The most commonly used punctuations among both classes are “@”, “.”, “<”, and “>”. The results of punctuation analysis can be checked from Figures 112 and 113.

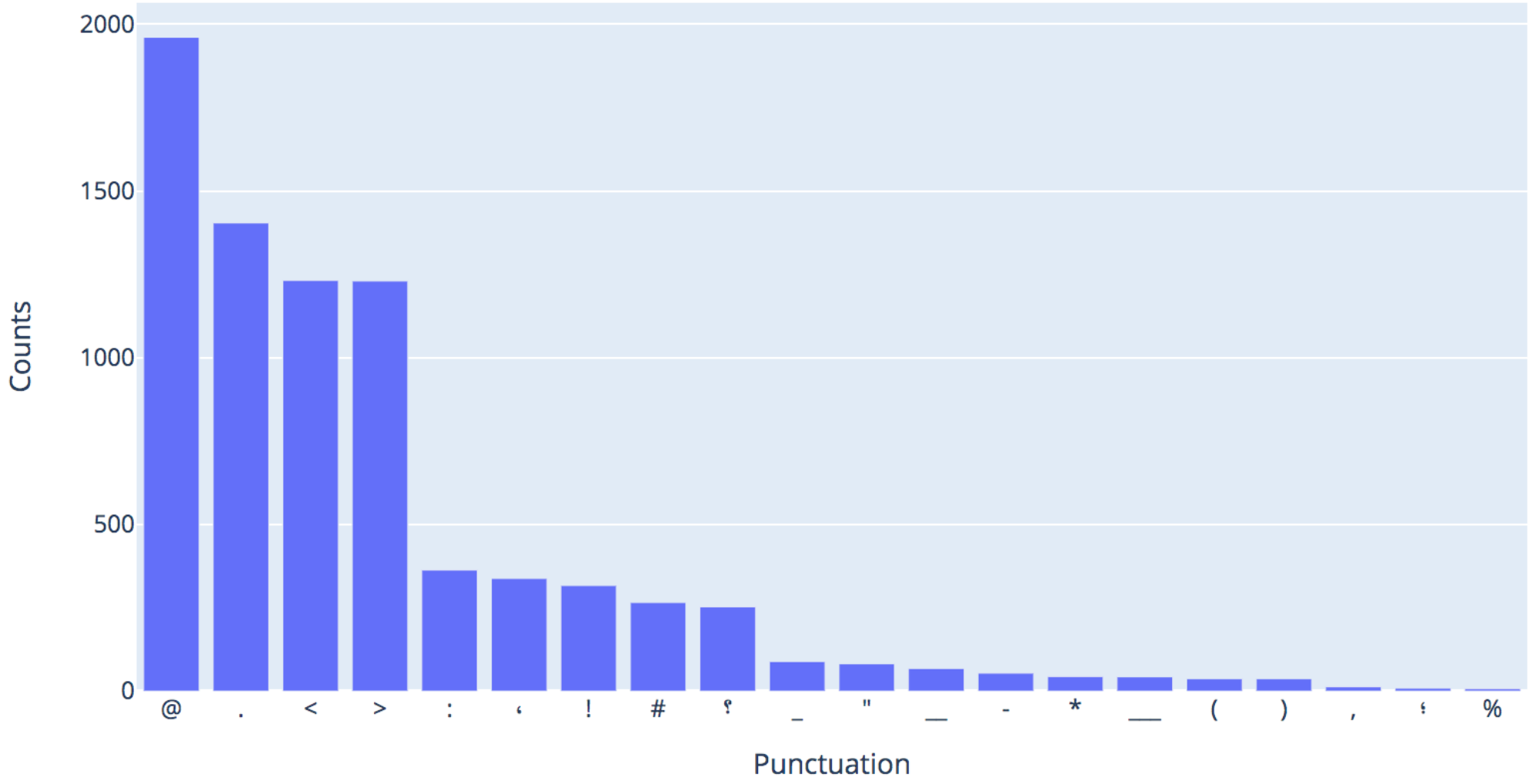

**Figure 112 : Most common offensive punctuation in the OSACT4/OffensEval 2020 Arabic dataset**

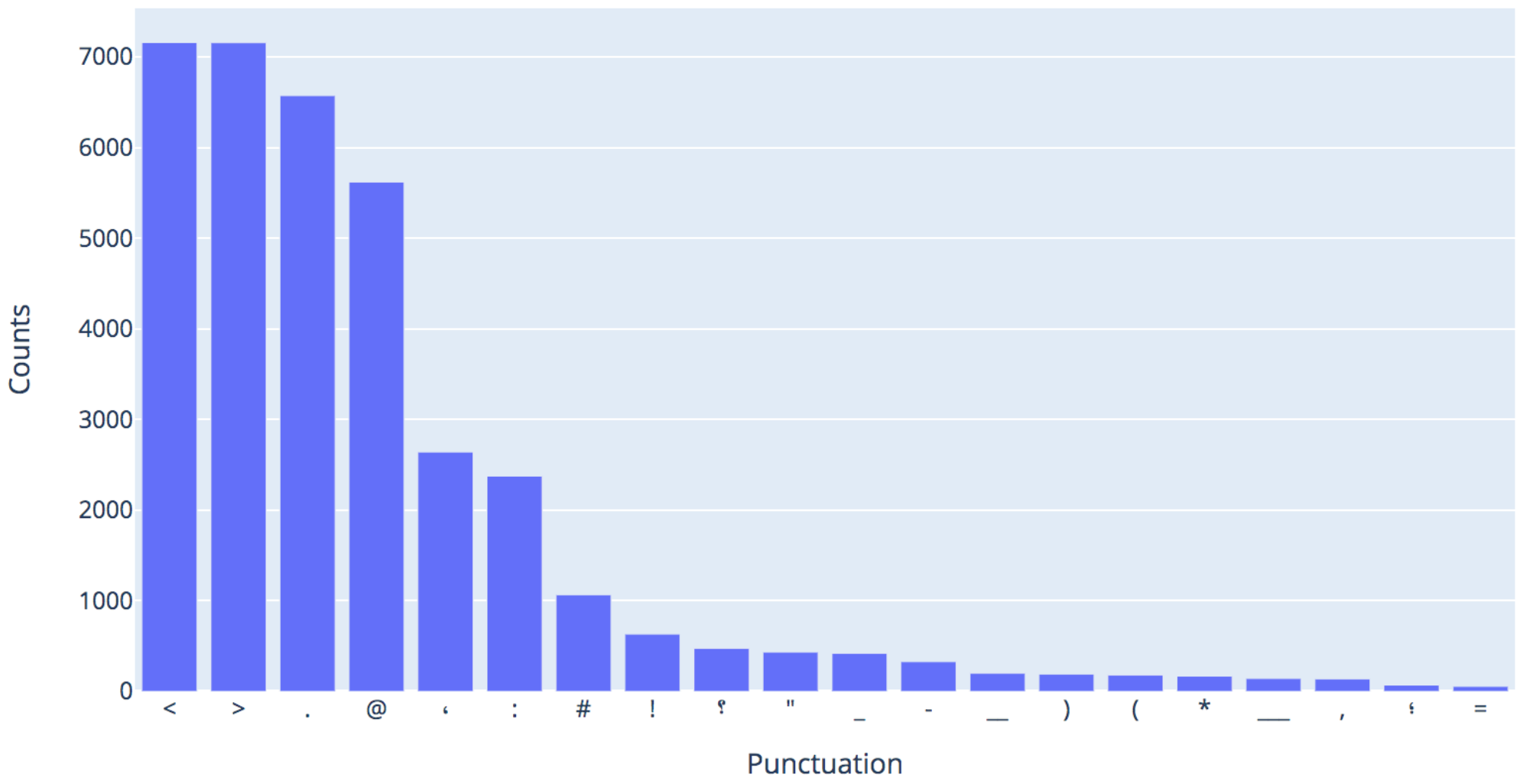

**Figure 113 : Most common not offensive punctuation in the OSACT4/OffensEval 2020 Arabic dataset**

8) The Multi-Platform Hate Speech Dataset:

Omar, Mahmoud, and Abd El-Hafeez (2020) release the first multi-platform dataset for Arabic hate speech detection. Comments were collected from four social media platforms; Facebook, Twitter, YouTube, and Instagram. From Figure 114, the dataset is balanced with 10,000 hate comments and 10,000 not hate comments. Content of the comments were preprocessed to remove non-Arabic characters, emoji, URLs, and posts with less than 2 words were also deleted.

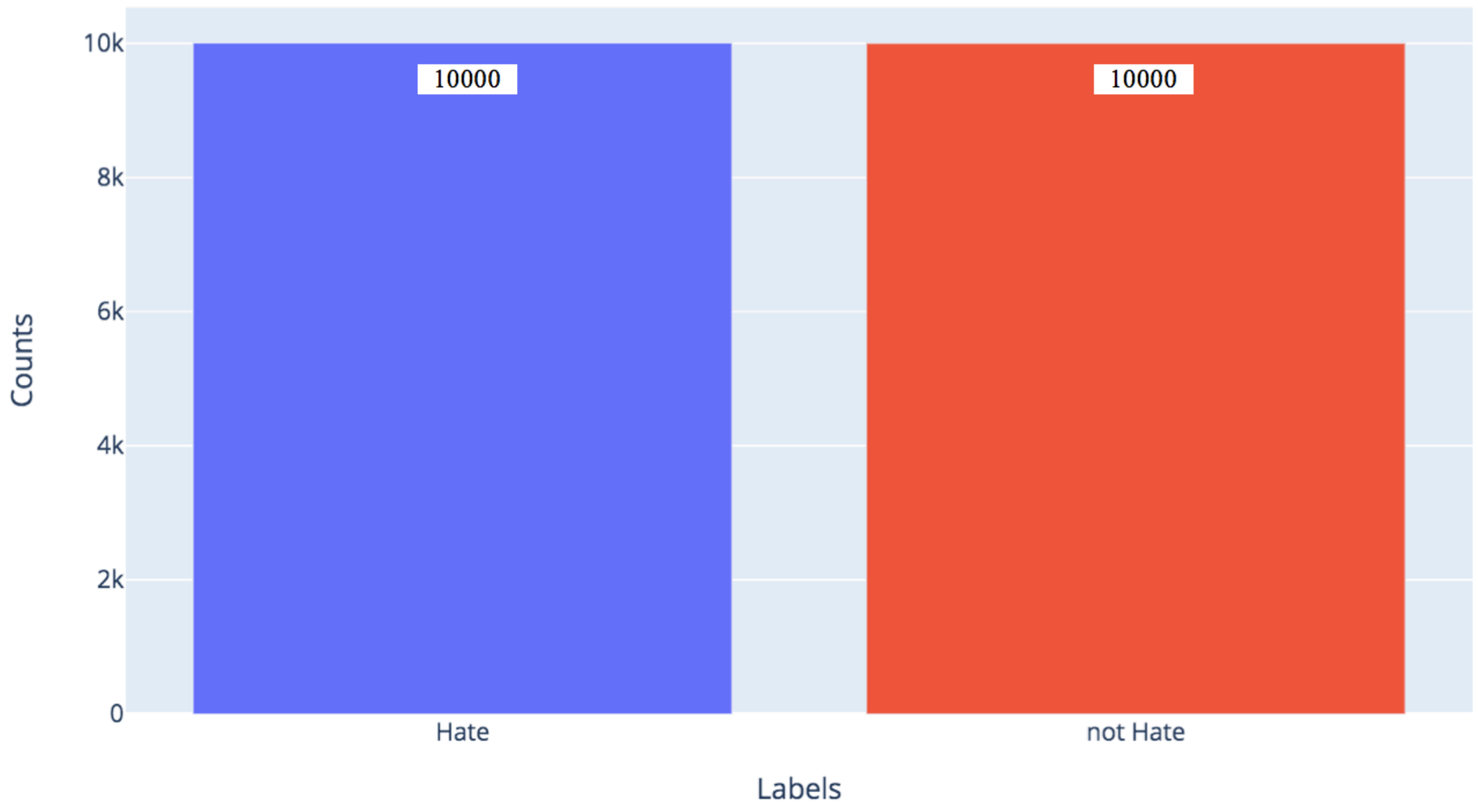

**Figure 114 : Class distribution for the Multi-Platform Hate Speech dataset**

Only two Duplicated comments were included, and both are classified as hate, the following is one of them:

أحمد موسى عار على الإعلام

Translation: Ahmad Mousa is a shame to the media

The word cloud graphs in Figure 115 provide an overview for the content from each class, for example, the stop word "على /on" is very frequent on both classes, while "عروس / pride" and "الحيوان / the animal" appear only in hate comments. The words "رئيس / president", "ممنهجة / well planned", and "البرلمان / the parliament" appear only in not hate word cloud.

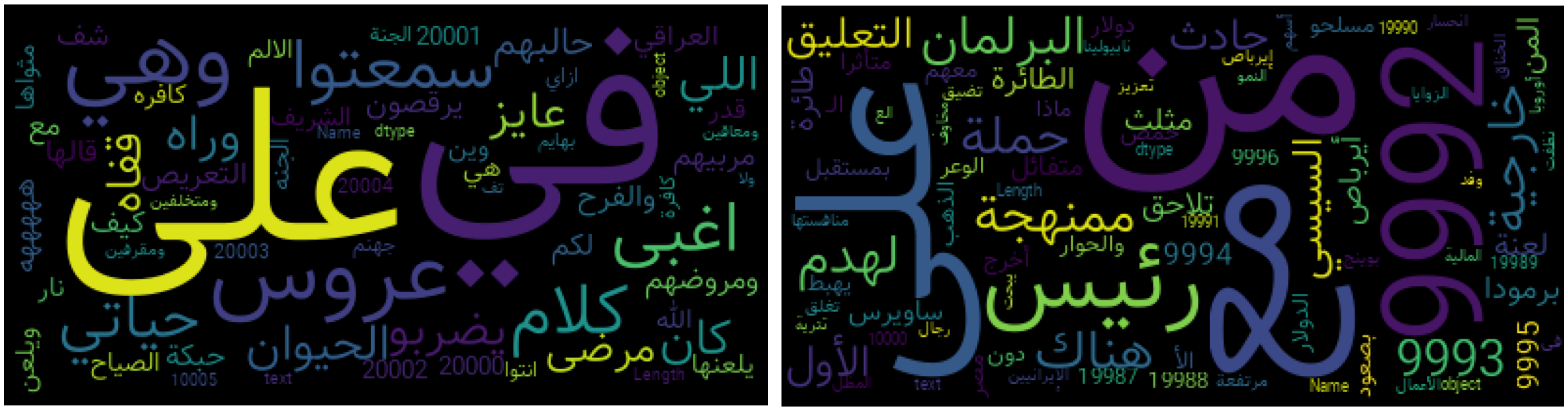

(a) Hate (b) Not hate

**Figure 115 : The word cloud of the Multi-Platform Hate Speech dataset (a. hate, b. not hate)**

The top three frequent tokens for not hate class are "اللهم / our God", "الله / God", and "تقنية / technology" as shown in Figure 116, while the top three ones for hate class are "الله / God", "ام / mother", and "كس / pussy" as shown in Figure 117.

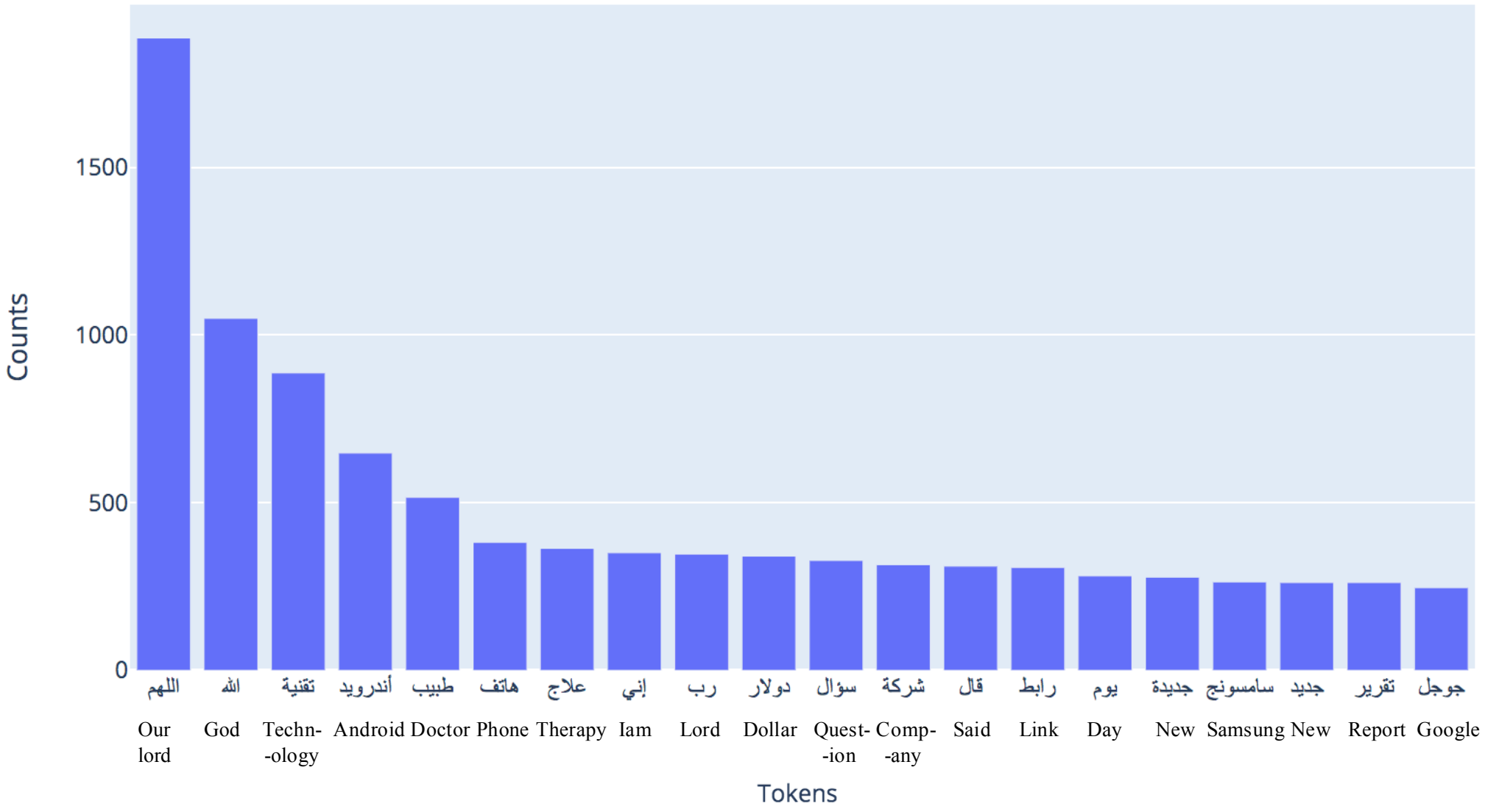

**Figure 116 : Most common not hate tokens in the Multi-Platform Hate Speech dataset**

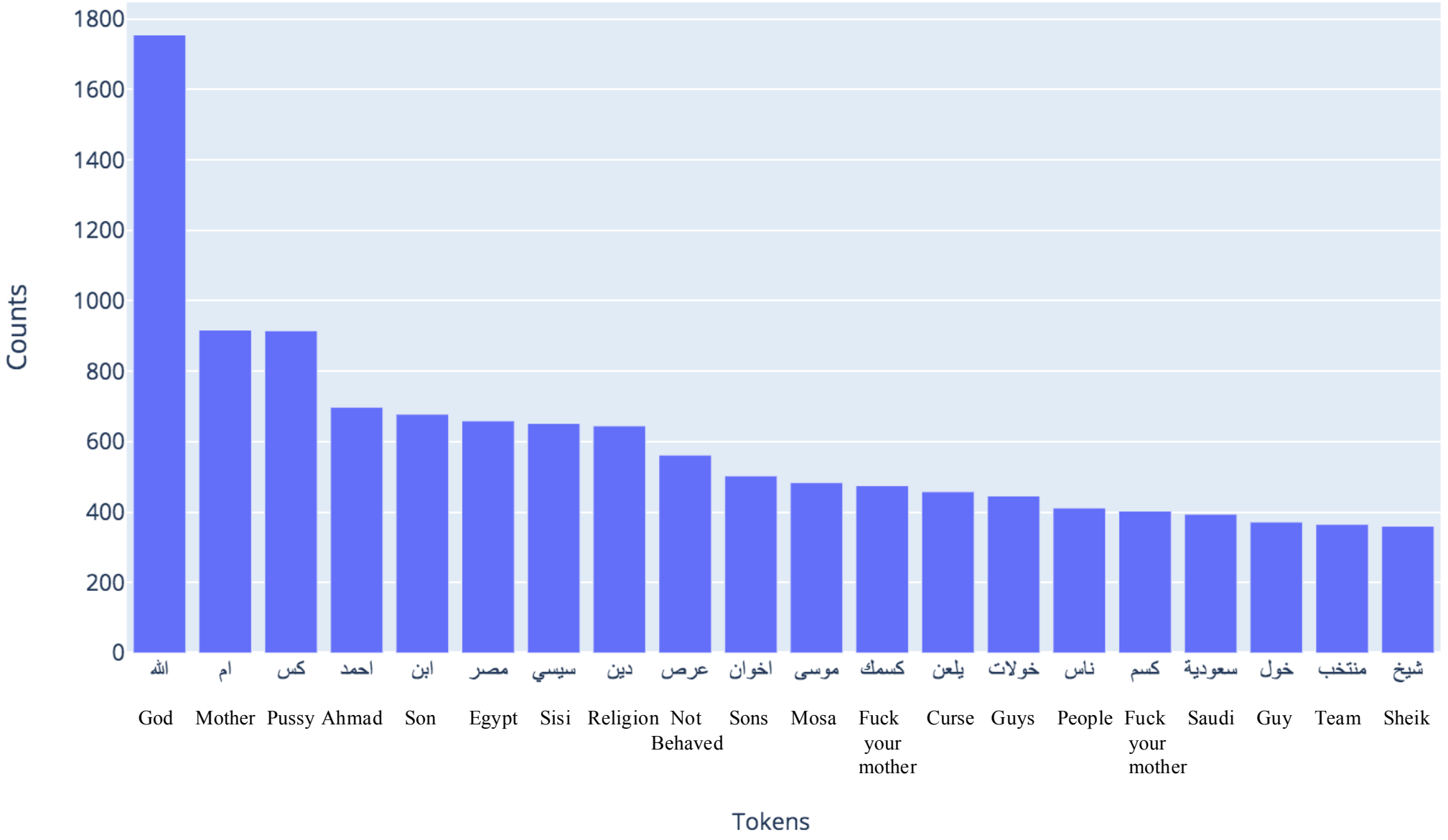

**Figure 117 : Most common hate tokens in the Multi-Platform Hate Speech dataset**

The followings are the top most distinctive words:

1. Hate: كس / pussy (952), كسمك / your mother's pussy (512), عرص / bad behaved (496), يلعن / curse (459), خولات / gays (445).
2. Not hate: تقنية / technology (847), أندرويد / Android (636), الطبيب / doctor (467), الرابط / link (306), سامسونج / Samsung (265)

Results of the statistical measurements for the length of comments and the length of tokens report very similar patterns among the two classes and among the two lengths measurements as shown in Figures 119 and 120. However, hate comments have larger outliers' values than those of not hate comments.

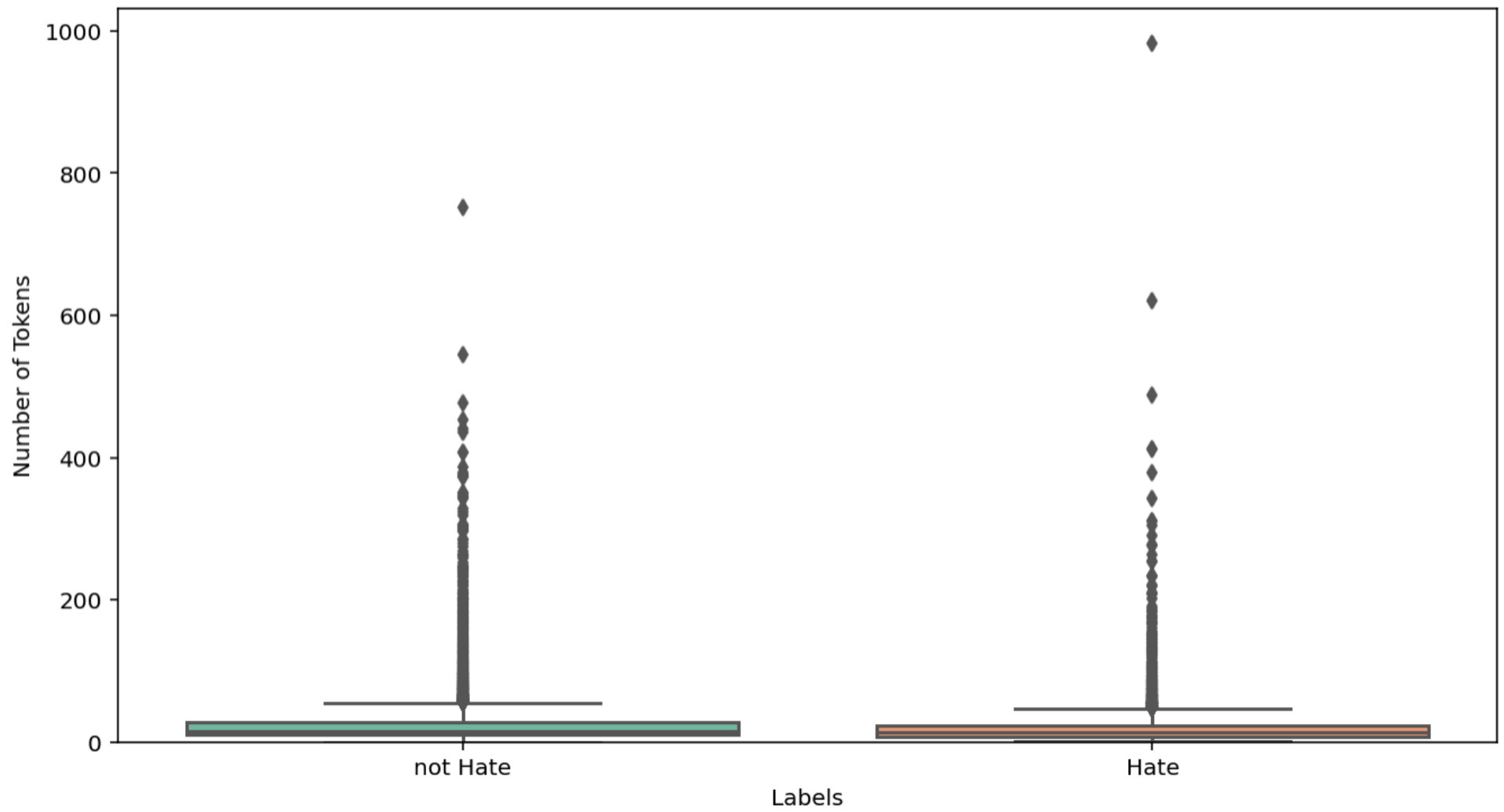

**Figure 118 : Statistics of each label in the Multi-Platform Hate Speech dataset based on the number of tokens per tweet**

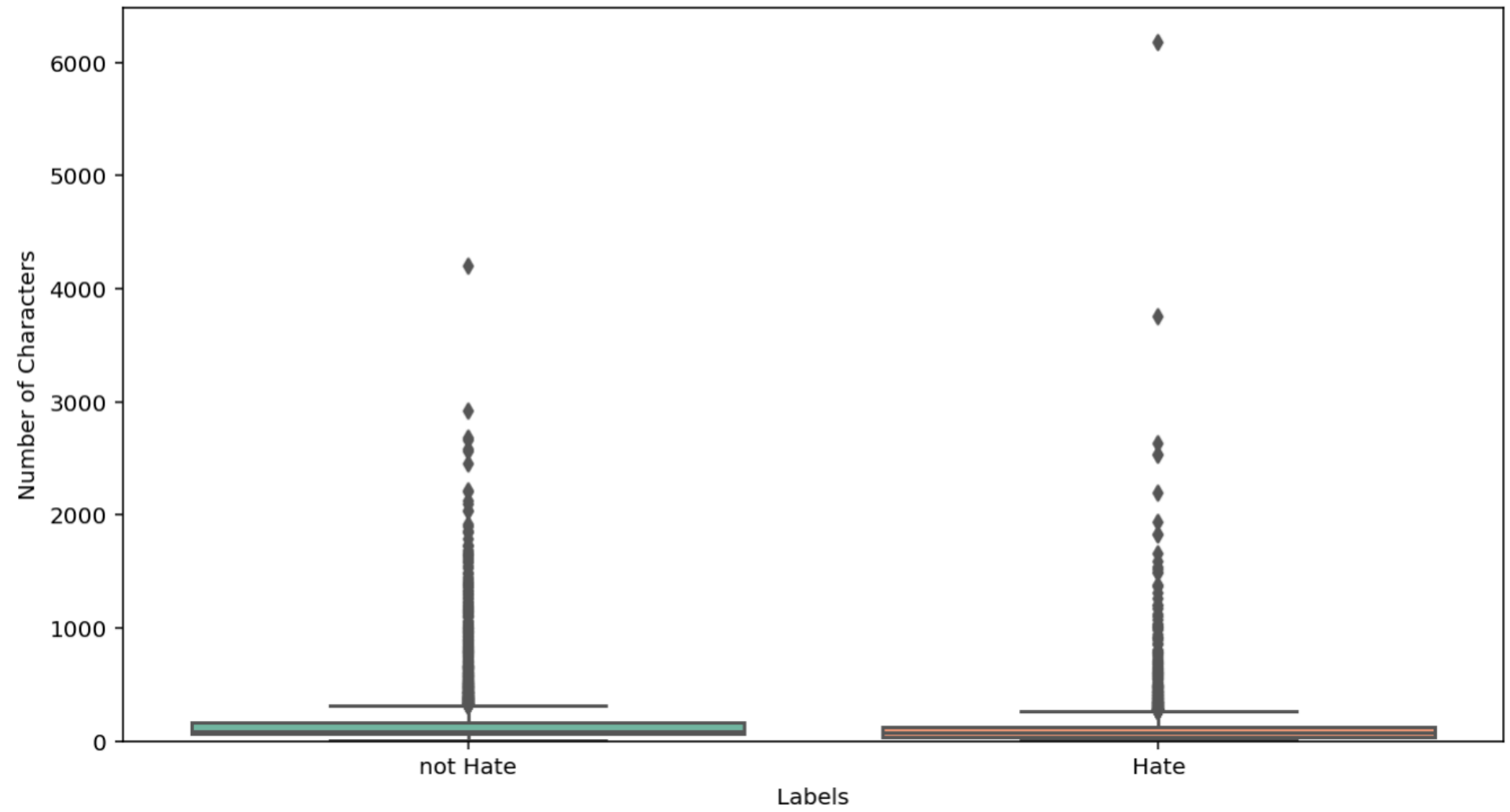

**Figure 119 : Statistics of each label in the Multi-Platform Hate Speech dataset based on the number of characters per token**

The appearance of stop words is very similar in both hate and not hate comments, "من/ from" is the top shown stop word in both Figures 120 and 121. Moreover, the stop word "في / in" is ranked the second for not hate comments and the fourth for hate comments and "على / on" is ranked the third for not hate comments and the fifth for hate comments. For hate class, “يا / you” is very frequent that it comes on the second top used stop word and “و / and” ranks the third top used stop word.

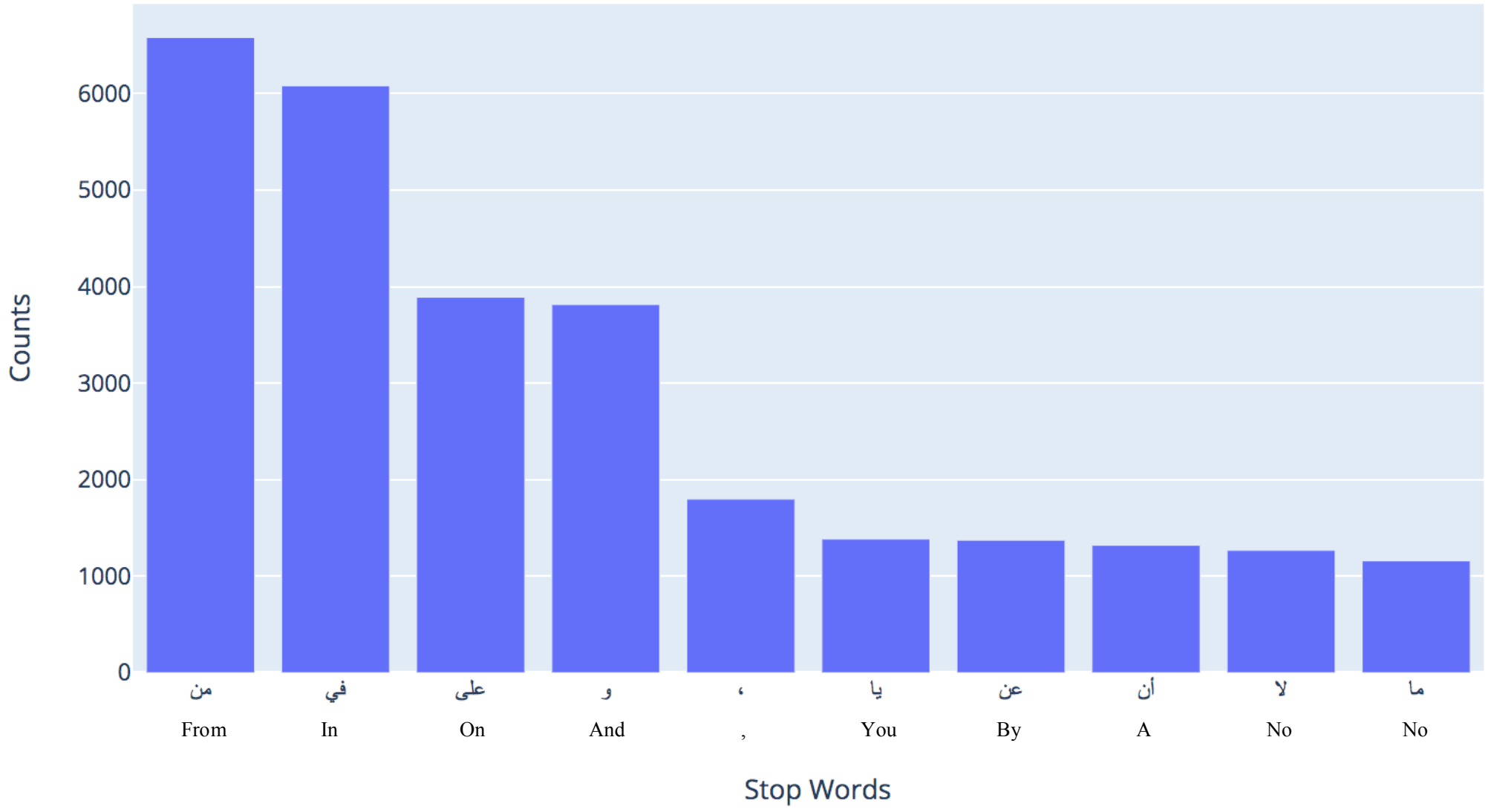

**Figure 120 : Most common stop words in not hate class from the Multi-Platform Hate Speech dataset**

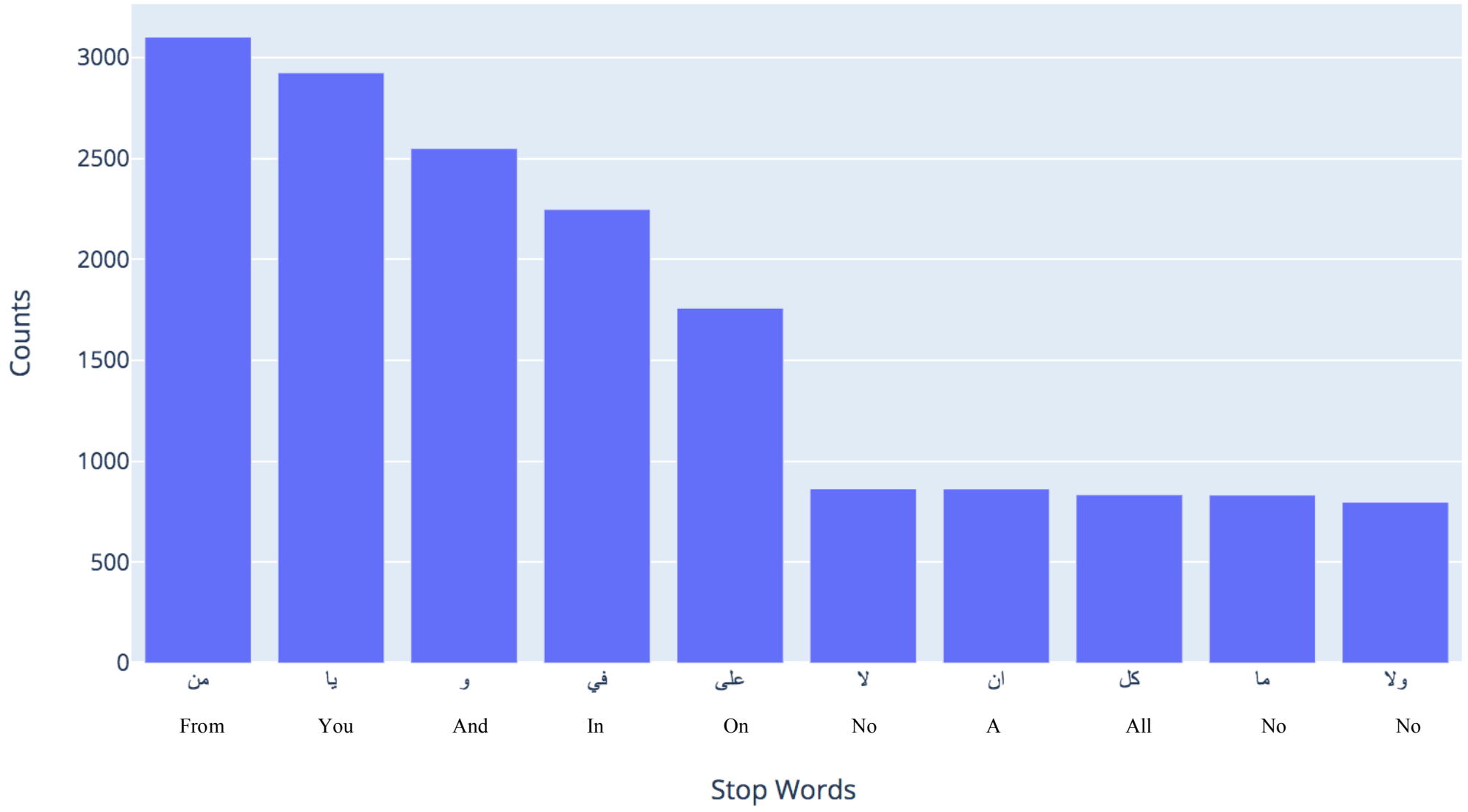

**Figure 121 : Most common stop words in hate class from the Multi-Platform Hate Speech dataset**

The sentiment analysis bar chart in Figure 122 show a relationship between neutral sentiment and not hate comments and a relationship between negative sentiment and hate comments.

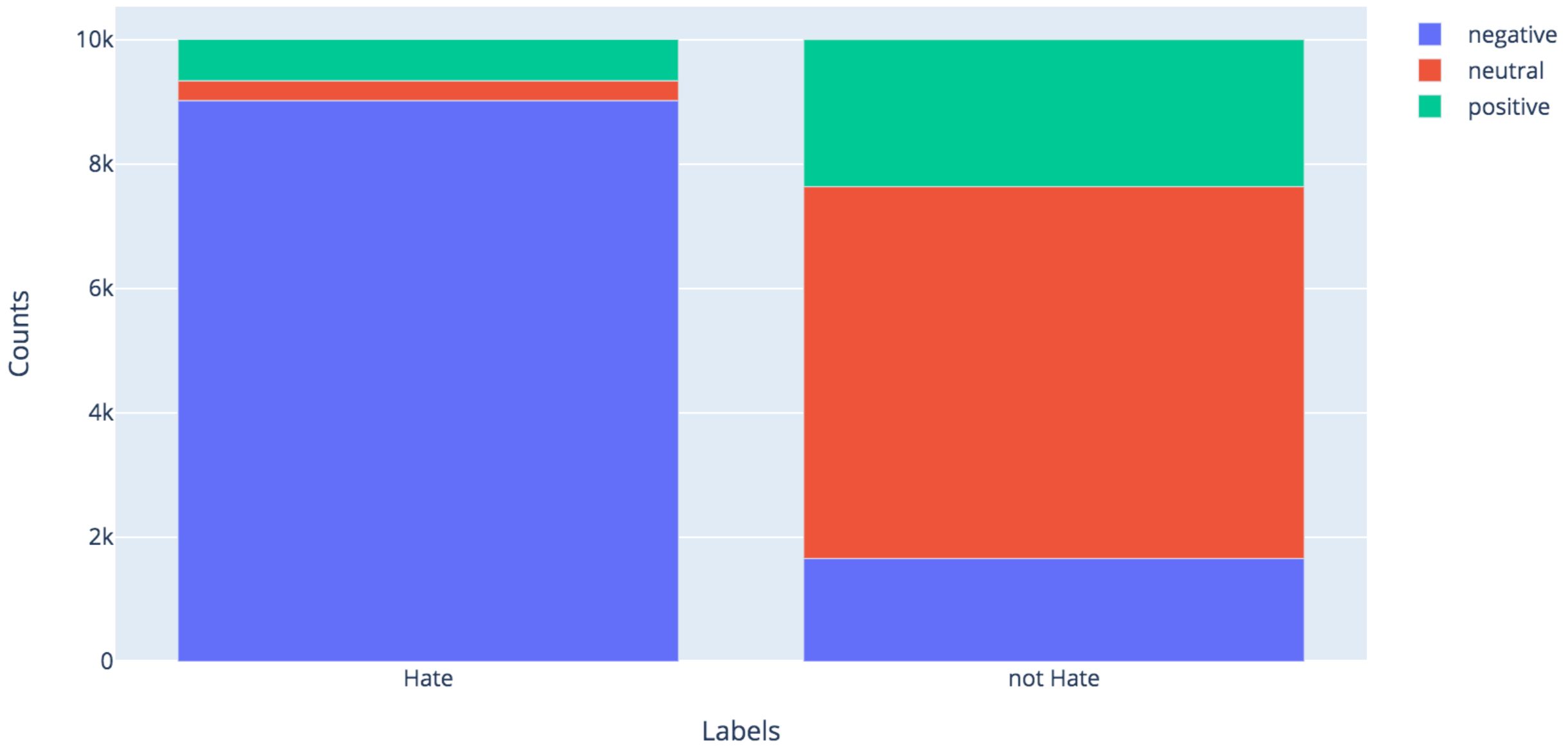

**Figure 122 : Sentiment analysis based on labels for the Multi-Platform Hate Speech dataset**

In both punctuation frequency counts Figures; Figure 123 for not hate class and figure 124 for hate class; "." is the top used one. The "‘" is the second most appeared punctuation within not hate comments and the third within hate comments. For not hate class, ":" is ranked the third top used punctuation and on the same time ranked the seventh for hate comments, while for hate class, "#" is ranked the second punctuation and does not appear within the top ten punctuation for not hate class.

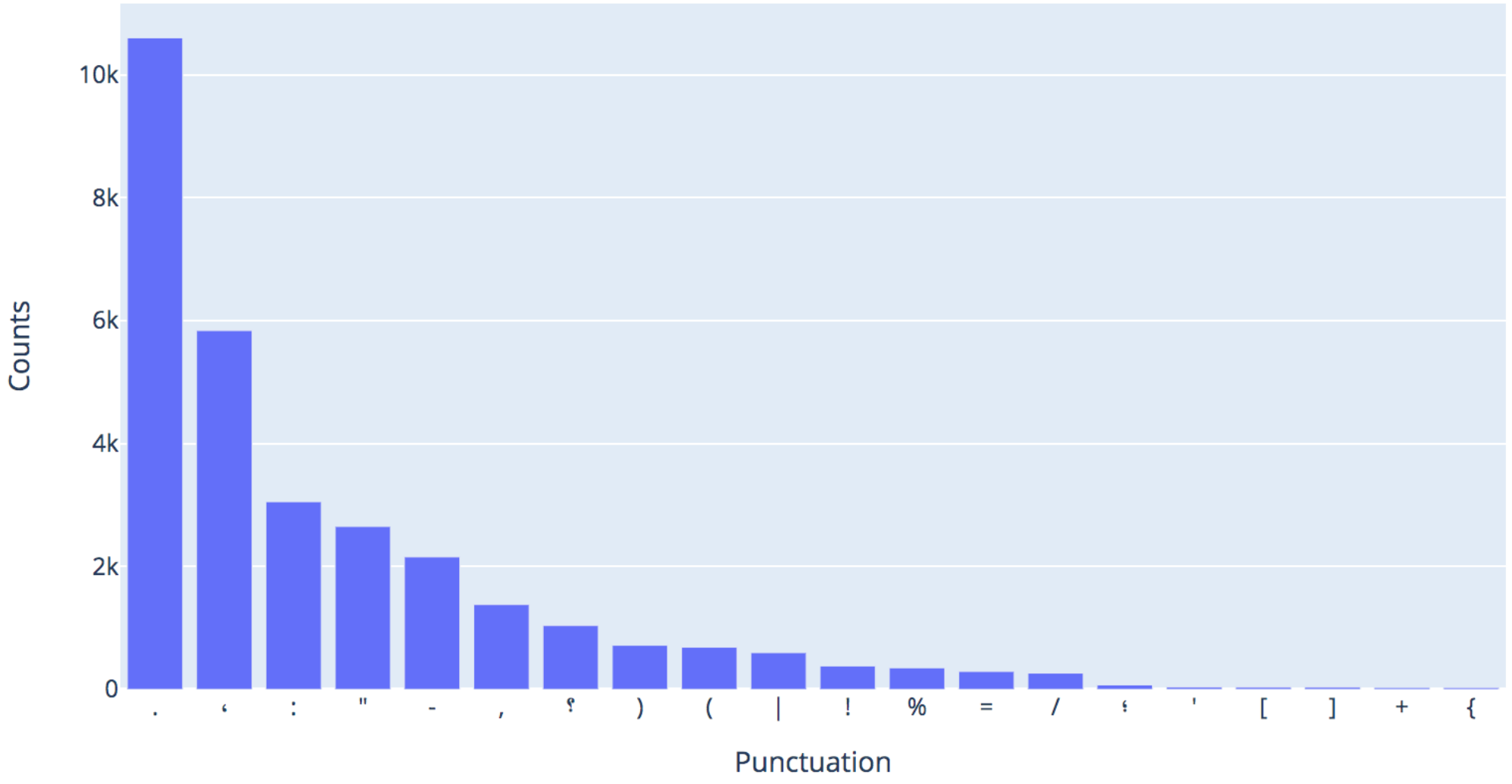

**Figure 123 : Most common not hate punctuation in the Multi-Platform Hate Speech dataset**

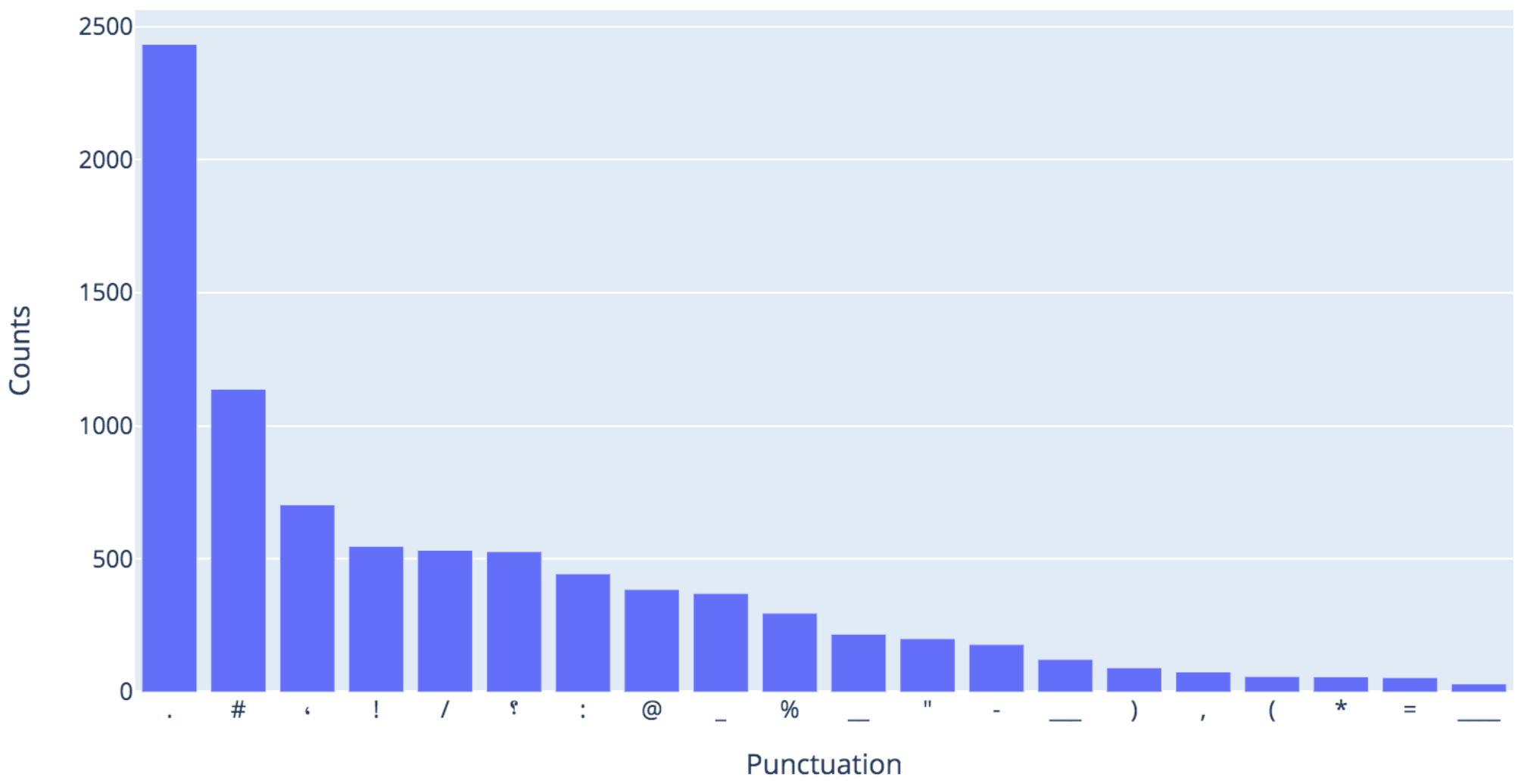

**Figure 124 : Most common hate punctuation in the Multi-Platform Hate Speech dataset**

## Summarizing Results

The following table summarize some features of the datasets that could be helpful to measure the quality of the datasets. The main quality related attributes include the size of the dataset and whether the classes are close in size, the number of the samples that have been included in the dataset for several times, the rounded average length of samples (comments) based on the number of tokens, the rounded average length of tokens based on the number of characters, the inclusion of emojis and punctuation, and whether the content is preprocessed for privacy and security reasons.

**Table 1 Quality features of the datasets**

| Dataset | Size per label | No. of Duplicates | Avg. length of comment | Avg. length of token | Emojis | Punctuation | Anonymized |
|---|---|---|---|---|---|---|---|
| The Aljazeera.net Deleted Comments | 31,692 (5,653 clean, 533 obscene, 25,506 offensive) | 7 original comments | 15 tokens | 80 characters | Not available | Available | Not available |
| Egyptian Tweets | 1,100 (453 clean, 203 obscene, and 444 offensive) | 2 original tweets | 13 tokens | 79 characters | Available | Available | Not available |
| Religious Hate Speech | 6,137 (2,762 hate and 3,375 not hate) | 8 original tweets | 22 tokens | 140 characters | Available | Available | Not available |
| YouTube Comments | 15,050 (9,237 not offensive and 5,813 offensive) | 2 original comments | 10 tokens | 100 characters | Available | Available | Not available |
| L-HSAB | 5,846 (468 hate, 1,728 abusive, and 3,650 normal) | No duplicates | 10 tokens | 70 characters | Available | Available | Not available |
| T-HSAB | 6,075 (3,834 normal, 1,127 abusive, and 1,078 hate) | No duplicates | 9 tokens | 80 characters | Not available | Not available | Not available |
| MPOLD | 4,000 (3,325 not offensive and 675 offensive) | No duplicates | 10 tokens | 80 characters | Available | Available | Available |
| OSACT4/OffensEval | 10,000 (8,100 not offensive and 1,900 offensive) | 3 original tweets | 19 tokens | 95 characters | Available | Available | Available |
| Multi-Platform Hate Speech | 20,000 (10,000 hate and 10,000 not hate) | 2 original comments | 30 tokens | 150 characters | Not available | Available | Not available |

The content of Table 2 summarizes the main findings for each dataset based on the class. This summarization supports the comparison of each type of offensive content (e.g. hate speech, offensive, obscene) among the entire Arabic offensive resources, which can provide a holistic picture of the available resources.

**Table 2 Label specific attributes summary**

| **Dataset** | **Label** | **Top three tokens** | **Top three distinctive words** | **Top three stop words** | **Major sentiment** | **Top three emoji** | **Top three punctuation** |
|---|---|---|---|---|---|---|---|
| The Aljazeera.net Deleted Comments | Clean | الله /God, دولة /state, جزيره /Jazeera | الخبر /the news, قال /said, ووفقكم /bless you | من /from, و /and, في /in | Negative | Not available | ., !, ؟ |
| | Obscene | الله /God, ابن /son, كس /pussy | المتناك / fucked, كسم / fuck your mother, كس / pussy | من /from, يا / you, في /in | Negative | Not available | ., !, ؟ |
| | Offensive | الله /God, دولة /state, مصر /Egypt | قال /said, الاسد /Al-Asad, السوري /Syrian | من /from, و /and, في /in | Negative | Not available | ., !, ؟ |
| Egyptian Tweets | Clean | الله /God, مصر / Egypt, شعب /people | الحل /the solution, الرئيس /the president, علاقة / relationship | من /from, في /in, و /and | Negative | 😢, 😂, 😭 | ., “, # |
| | Obscene | امك /your mother, عرص /bad behaved, كلب /dog | العرص /the bastard, امك /your mother, عرص /bastard | يا / you, من /from, في /in | Negative | 😂, ✌️, 🐶 | ., “,! |
| | Offensive | مصر / Egypt, الله /God, منك/ from you | البرادعى /El-Baradei, بطل /hero, اهبل / stupid | يا / you, من /from, في /in | Negative | 😂, 😥, ✋ | ., “,؟ |
| Religious Hate Speech | Not hate | الله /God, يهود /Jews, شيعة /Shia | فطرة/ primitiveness, الإخلاص/the sincerity, إبراهيم/ Ibrahim | من /from, في /in, على /on | Negative | 😂, 🌹, 🖤 | ., /, # |
| | Hate | الله /God, لهم /for them, مسلمين /Muslims | اللعنة /curse, العهر /immorality, متحدون /united | من /from, في /in, على /on | Negative | 😂,🇶🇦, 👇 | ., /, # |
| YouTube Comments | Not offensive | كاظم/ Kadhim, الله/ God, احلام/ Ahlam | بقناتي / in my channel, مها / Maha, يهديك / give you | من /from, و/ and, في /in | Negative | 😂, 🖤, 😍 | .,؟ , + |
| | Offensive | الله /God, كاظم/ | زباله /trash, زبالة /trash, | من /from, | Negative | 😂, 👎, 😒 | ., ؟, ‘ |
| | | Kadhim, احلام/ Ahlam | خرة /shit | يا / you, و / and | | | |
| L-HSAB | Normal | جبران باسيل/ Gebran Bassil, وزير/ minister, الله/ God | القمة /the top, الخارجية /the external, الوطني /the national | من /from, ما / what, في /in | Negative | 🖤, 😎, ♀ | ،, ؟, ؛ |
| | Abusive | كول هوا/ eat air, كلب/ dog | خرا /shit, واطي /cheap man, شرفك /your honor | يا / you, , ما / what, في /in | Negative | 😂, 😎, 🤷‍♂️ | ؟, ، , ؛ |
| | Hate | عالمطبخ /to the kitchen, البنات /girls, ولاك /guys | عالمطبخ /to the kitchen, البنات /girls, ولاك /guys | يا / you, من /from, و/ and | Negative | 😂, 😎, ♂ | ؟, ،, ! |
| T-HSAB | Normal | تونس / Tunisia, "ربي /my lord, لطفي /Lutfy | عواطف /emotions, الخطاب / Al-Khattab, صلاح / Salah | ه /ha, أكثر / more, هكذا /this is | Negative | Not available | Not available |
| | Abusive | قحبة /whore, تونس / Tunisia, نيك /fuck | نيك /fuck, قحبة /whore, القحبة / the whore | ه /ha, أكثر / more, بكم / by you | Negative | Not available | Not available |
| | Hate | تونس / Tunisia, شعب /people, دين /religion | القحبة / the whore, ودين/ religion, قحبة /whore | ه /ha, أكثر / more, الآن /now | Negative | Not available | Not available |
| MPOLD | Not offensive | @User.IDX, الله/ God, قناة /channel | شكرا / thanks, برنامج /program, افضل/ better | من/ from, في/ in, و / and | Negative | 😂, 🖤, 👍 | ., @, ، |
| | Offensive | @User.IDX, الله/ God, قناة /channel | الخنزيره /pig, الحمدين /the two Hamad, قذر /dirty | من/ from, في/ in, و / and | Negative | 😂, 👇, 🤣 | ., @, ! |
| OSACT4/OffensEval | Not offensive | @USER, RT, URL | قيوم /everlasting, مجيب /answerer, رحمن /merciful | يا / you, من /from, و/ and | Positive | 😂, 🖤, 💙 | <, >, . |
| | Offensive | @USER, الله/ God, URL | فسواي /shit, كلووووب /small dog, كلبه /female dog | يا / you, و/ and,من /from | Negative | 😂, 🤣, 💙 | @, ., < |
| Multi-Platform Hate Speech | Not hate | اللهم /our God, الله / God, تقنية /technology | تقنية /technology, أندرويد /Android, الطبيب /doctor | من /from, في/ in, على /on | Neutral | Not available | ., ‘, : |
| | Hate | الله / God, ام / mother, كس / pussy | كس /pussy, كسمك /your mother's pussy, عرص /bad behaved | من /from, يا / you, و/ and | Negative | Not available | ., #, ‘ |

**Synthesizing Results**

The problem of online offensive language is a very complex one. Our analysis shows that most surveyed datasets have average length of comment less than or equal to 15 tokens, which make it difficult to accurately identify the context and differentiate between offensive content and other similar content such as sarcasm content. This problem gets more difficult when the comment does not include non-textual elements, such as emojis or pictures, that might otherwise add more insight into the textual content. In addition, majority of the surveyed datasets are imbalanced with very small percentage of offensive content, which make it very difficult to depend on one dataset to develop an offensive language detection system with sufficient training instances and obtain accurate results.

Offensive words differ among the datasets regardless of the offensive type. From the analysis, the first three top frequent hate speech tokens in L-HSAB dataset are "عالمطبخ /to the kitchen ", "البنات /girls", and "ولاك /guys", while for T-HSAB dataset, they are "تونس / Tunisia", "شعب /people", and "دين /religion". The results show very limited possibility of overlapping among the top frequent tokens from the datasets, which could be a result for the variations of Arabic dialects and Arabic sub-cultures. Overall, Arabic users of online social media are commonly use the face with tears of joy emoji, "😂", in their conversations as it appears as the top frequent emoji in most surveyed datasets for offensive and not offensive samples. The use of punctuation does not demonstrate any specific patterns among the datasets. Stop word analysis highlights the relationship of "يا / you" with offensive content as it appears among the top frequent stop words in multiple offensive types from multiple datasets. Some of the top frequent tokens are part of a name of famous figures; for example, "كاظم/ Kadhim", and "احلام/ Ahlam" from YouTube Comments dataset; thus, it would be better to consider studying the relationship between names of famous personal and offensive content.

**Design Considerations**

The available Arabic offensive language datasets that we discuss in this paper cover several offensive contents from different platforms with different Arabic dialects that can provide valuable insights into the problem of online offensive content. Accordingly, researchers in this domain of research need to develop an advanced method to extract collective knowledge from all offensive language dataset in Arabic language, which can help researchers from Arabic offensive language detection to understand the problem and develop offensive language detection system that can look to the problem from several dimensions with a holistic view.

**Conclusions**

In this paper, we investigate the content of 9 Arabic offensive language datasets to provide in-depth analysis of their content. This research aims to guide researchers in Arabic offensive language to select the appropriate dataset based on their content, and in creating new Arabic offensive language resources to support and complement the available ones. Results demonstrate the limited content of the surveyed dataset in terms of the offensive sample size and the length of the samples. Results also report variations of the offensive content among the datasets. Thus, it is very important to consider developing an innovative method to extract valuable insight about Arabic offensive content from the available datasets collectively to apply that knowledge into an Arabic offensive language detection system.